%% file: main.tex
\documentclass[11pt, a4paper, oneside]{Thesis} 
\usepackage{floatrow}
\usepackage{textcomp}
\usepackage{lscape}
\usepackage{adjustbox}
\usepackage{rotating}
\usepackage{graphicx}
\usepackage{caption}
\usepackage{amsmath}
\usepackage{upgreek}
\usepackage{gensymb}
\usepackage{csquotes}
\usepackage{pdfpages}
\usepackage{multirow}
\usepackage{wrapfig}
 \usepackage{booktabs}
\usepackage{float}
\DeclareUnicodeCharacter{0301}{\'{}}

\newcommand\crule[3][black]{\textcolor{#1}{\rule{#2}{#3}}}
\definecolor{ao}{rgb}{0.0, 0.5, 0.0}
\definecolor{dark}{rgb}{0.20784313725,0.10980392156,0.45882352941}
\definecolor{med}{rgb}{0.40392156862,0.30588235294,0.65490196078}
\definecolor{light}{rgb}{0.70588235294,0.65490196078,0.83921568627}
\definecolor{light_green}{rgb}{0.85098039215,0.91764705882,0.82745098039}
\definecolor{special_red}{rgb}{0.59607843137,0.59607843137,0.59607843137}
\definecolor{g1}{rgb}{0.36078431372,0.65098039215,0.2862745098}
\definecolor{g2}{rgb}{0.26666666666,0.50588235294,0.55294117647}
\definecolor{lightg}{rgb}{0.8509803922,	0.9215686275,	0.831372549}
\definecolor{lorg}{rgb}{0.9882352941,	0.8980392157,	0.8117647059}
\definecolor{lpink}{rgb}{0.9215686275,	0.8156862745,	0.862745098}
\definecolor{lvio}{rgb}{0.8509803922,	0.8196078431,	0.9098039216}
\definecolor{special_red}{rgb}{0.65098039215,0.30196078431,0.47450980392}
\usepackage{acronym}
\usepackage{array}
\newcolumntype{L}{>{\centering\arraybackslash}m{3cm}}

\makeatletter
\AtBeginDocument{%
  \renewcommand*{\AC@hyperlink}[2]{%
    \begingroup
      \hypersetup{hidelinks}%
      \hyperlink{#1}{#2}%
    \endgroup
  }%
}
\makeatother

\graphicspath{{images/}} 
\usepackage[square]{natbib} 
\renewcommand\cite{\citep}	
\newcommand\shortcite{\citeyearpar}
\newcommand\newcite{\citet}

\hypersetup{urlcolor=black, colorlinks=true} 
\title{\ttitle} 

\begin{document}
\makeatletter
\renewcommand*{\NAT@nmfmt}[1]{\textsc{#1}}
\makeatother


\frontmatter 

\setstretch{1.6} 

\fancyhead{} 
\rhead{\thepage} 
\lhead{} 

\pagestyle{fancy} 

\newcommand{\HRule}{\rule{\linewidth}{0.5mm}} 

\hypersetup{pdftitle={\ttitle}}
\hypersetup{pdfsubject=\subjectname}
\hypersetup{pdfauthor=\authornames}
\hypersetup{pdfkeywords=\keywordnames}


\begin{titlepage}
\begin{center}

\HRule \\[0.4cm] 
{\huge \bfseries \ttitle}\\[0.4cm] 
\HRule \\[1.5cm] 
 
\large \textit{A thesis submitted in fulfilment of the requirements\\ for the degree of \degreename}\\[0.3cm] 
\textit{by}\\[0.4cm]
\authornames \\
Roll No: 18104278
\\

\vfill
\graphicspath{ {./Figures/} }
\begin{figure}[hb]
  \centering
  \includegraphics[width=0.35\linewidth]{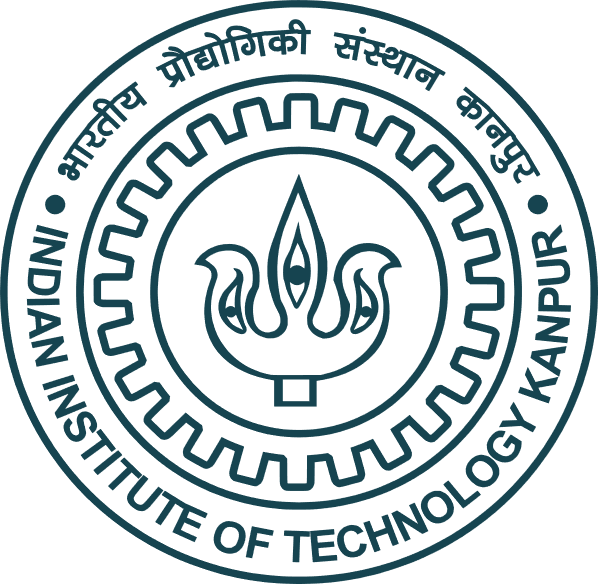}
\end{figure}

\DEPTNAME\\ 
\textsc{ \UNIVNAME}\\[1.5cm] 
\large \today\\[2cm] 

\end{center}

\end{titlepage}


\Declaration{\addtocontents{toc}{\vspace{1em}}} 
\setcounter{page}{2}

\begin{minipage}{\textwidth}
    
    It is certified that the work contained in this thesis entitled \textbf{\enquote{\ttitle}} by \textbf{\authornames} has been carried out under my supervision and that it has not been submitted elsewhere for a degree.
        
\end{minipage}

\vspace{20.00mm}

\begin{minipage}{0.49\textwidth}
		\begin{flushleft}
			{ 
  \includegraphics[width=0.35\linewidth]{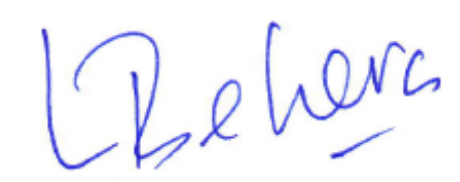}\\
            \supnameA\\ 
			Professor\\
			\deptname\\
			\univname}
		\end{flushleft}
\end{minipage}
\begin{minipage}{0.49\textwidth}
	\begin{flushleft} \large
			{\includegraphics[width=0.55\linewidth]{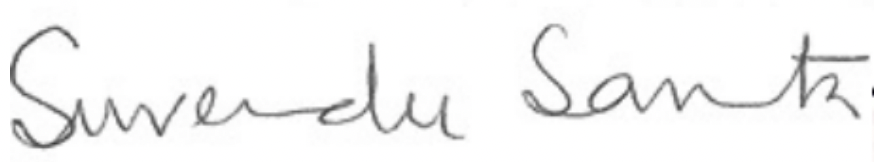}\\
   \supnameB\\
			Assistant Professor\\
			\deptname\\
			\univname}
	\end{flushleft}		
\end{minipage}
\vfill
\clearpage 
\StudentDeclaration{\addtocontents{toc}{\vspace{1em}}} 

This is to certify that the thesis titled \textbf{``\ttitle''} has been authored by me. It presents the research conducted by me under the supervision of \textbf{\supnameA} and \textbf{\supnameB}.\par

To the best of my knowledge, it is an original work, both in terms of research content and narrative, and has not been submitted elsewhere, in part or in full, for a degree. Further, due credit has been attributed to the relevant state-of-the-art and collaborations with appropriate citations and acknowledgments, in line with established norms and practices.\\ [2cm]
\begin{minipage}{.5\textwidth}
		\begin{flushleft}
			{\includegraphics[width=0.35\linewidth]{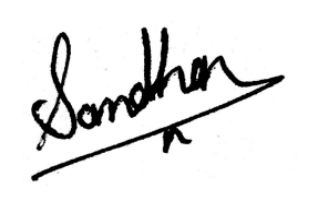}\\
   \authornames\\ Roll No. 18104278 \\
			\normalsize{\href{http://www.iitk.ac.in/ee}{EE Department}\\
			\univname}}
		\end{flushleft}
\end{minipage}
\vfill

\clearpage 

\lhead{\emph{Synopsis}}

\abstract{\addtocontents{toc}{\vspace{1em}} 
The primary focus of this thesis is to make Sanskrit manuscripts more accessible to the end-users through natural language technologies. The morphological richness, compounding, free word orderliness, and low-resource nature of Sanskrit pose significant challenges for developing deep learning solutions. We identify four fundamental tasks, which are crucial for developing a robust NLP technology for Sanskrit: word segmentation, dependency parsing, compound type identification, and poetry analysis. The first task, Sanskrit Word Segmentation (SWS), is a fundamental text processing task for any other downstream applications. However, it is challenging due to the \textit{sandhi} phenomenon that modifies characters at word boundaries. Similarly, the existing dependency parsing approaches struggle with morphologically rich and low-resource languages like Sanskrit. Compound type identification is also challenging for Sanskrit due to the context-sensitive semantic relation between components. All these challenges result in sub-optimal performance in NLP applications like question answering and machine translation. Finally, Sanskrit poetry has not been extensively studied in computational linguistics.

While addressing these challenges, this thesis makes various contributions, First, the thesis proposes \textit{linguistically-informed neural architectures} for these tasks. For the SWS task, our system accommodates language-specific \textit{sandhi} phenomenon as an inductive bias. We also propose a novel linguistically-informed context-sensitive multi-task learning architecture for compound type identification, with morphological tagging and dependency parsing auxiliary tasks. Additionally, we propose a linguistically-informed ensemble architecture for dependency parsing that augments several strategies. Finally, we propose to build an interpretable framework consisting of 7 linguistically-informed modules to analyze and classify Sanskrit poetry into levels of the best composition.

Second, we showcase the \textit{interpretability and multilingual extension} of the proposed systems. We conduct probing analyses on attention modules in word segmentation and compound type identification tasks. Our findings highlight the importance of attention modules and auxiliary tasks in NLP tasks and emphasize the need for explainable AI systems. We also showcase the efficacy of our systems on multiple languages for compound type identification and dependency parsing tasks.

Third, our proposed systems report \textit{state-of-the-art performance}. For instance, our system shows an average gain of 7.2 points over the current state-of-the-art system in word segmentation. The ensemble system used for dependency parsing outperforms the state-of-the-art system by 1.2 points for Sanskrit, and the context-sensitive compound type identification task reports a gain of 7.71 points compared to the current state-of-the-art.

Finally, we present a \textit{neural toolkit named SanskritShala}, a web-based application that provides real-time analysis of input for various NLP tasks. We release the source codes, 7 word embedding models trained on publicly available Sanskrit corpora and a multilingual code-mixed ST dataset for 25 languages. Additionally, we provide annotations for {\sl \'Sik\d{s}\={a}\d{s}\d{t}aka} poetry and a web application to illustrate its analysis and annotations. Overall, this thesis contributes to making Sanskrit manuscripts more accessible by developing robust NLP technology and releasing various resources, datasets, and web-based toolkit.
}

\newpage

\ListofPublications{\addtocontents{toc}{\vspace{1em}} 
\noindent\textbf{Author's Biography:} 
\begin{itemize}
\itemsep-0.7em 
\item \textbf{Name}: Mr. Jivnesh Sandhan; Ph.D. at IIT Kanpur
\item \textbf{Education}: Dual degree in Math's and Scientific Computing, IIT Kanpur, 2018. 
\item \textbf{Scholarship}: Recipient of the TCS Research Fellowship.
\item \textbf{Broad research area}: Deep learning for Sanskrit Computational Linguistics.
\item \textbf{Research supervisor}: Dr. Laxmidhar Behera and Dr. Suvendu Samanta.
\item \textbf{Goal}: Provide access to Sanskrit literature for pedagogical and annotation purposes.
\item \textbf{Current research}: Aesthetics of Sanskrit poetry from computational perspective.
\end{itemize}

\begin{enumerate}
  \item \textbf{Jivnesh Sandhan}, Amruta Barbadikar, Malay Maity, Tushar Sandhan, Pawan Goyal, Laxmidhar Behera. \emph{Aesthetics of Sanskrit Poetry from the Perspective of Computational Linguistics: A Case Study Analysis on  \'siks\={a}\d{s}\d{t}aka Poetry.} ACM Transactions on Asian and Low-Resource Language Information Processing, \textbf{TALLIP 2023*}.   \href{}{[\underline{Paper}]}\href{}{ } | \href{}{[\underline{Interface}]}\href{https://sanskritshala.github.io/shikshastakam/}{}  (Communicated).

  \item \textbf{Jivnesh Sandhan}, Anshul Agarwal, Laxmidhar Behera, Tushar Sandhan, Pawan Goyal. \emph{SanskritShala: A Neural Sanskrit NLP Toolkit with Web-Based Interface for Pedagogical and Annotation Purposes.} Proceedings of the Annual Meeting of the Association for Computational Linguistics: System Demonstrations, \textbf{ACL 2023*}, Toronto, Canada.  \href{https://arxiv.org/abs/2302.09527}{[\underline{Paper}]}\href{https://arxiv.org/abs/2302.09527}{ } | \href{https://github.com/Jivnesh/sanskritshala}{[\underline{Code}]}\href{https://github.com/Jivnesh/sanskritshala}{ }
  
    \item \textbf{Jivnesh Sandhan}, Laxmidhar Behera, Pawan Goyal. \emph{Systematic Investigation of Strategies Tailored for Low-Resource Settings for Low-Resource Dependency Parsing.} Proceedings of the European Chapter of the Association for Computational Linguistics, \textbf{EACL 2023}, Dubrovnik, Croatia.  \href{https://arxiv.org/abs/2201.11374}{[\underline{Paper}]}\href{https://arxiv.org/abs/2201.11374}{ } | \href{https://github.com/Jivnesh/SanDP}{[\underline{Code}]}\href{https://github.com/Jivnesh/SanDP}{ }
    
    \item \textbf{Jivnesh Sandhan}, Rathin Singha, Narein Rao, Suvendu Samanta, Laxmidhar Behera, Pawan Goyal. \emph{TransLIST: A Transformer-Based Linguistically Informed Sanskrit Tokenizer.} Proceedings of the Conference on Empirical Methods in Natural Language Processing, \textbf{EMNLP (Findings) 2022}, Abu Dhabi. \href{https://arxiv.org/abs/2210.11753}{[\underline{Paper}]}\href{https://arxiv.org/abs/2210.11753}{ } | \href{https://github.com/rsingha108/translist}{[\underline{Code}]}\href{https://github.com/rsingha108/translist}{ }

    \item \textbf{Jivnesh Sandhan}, Ashish Gupta, Hrishikesh Terdalkar, Tushar Sandhan, Suvendu Samanta, Laxmidhar Behera and Pawan Goyal. \emph{A Novel Multi-Task Learning Approach for Context-Sensitive Compound Type Identification in Sanskrit.} Proceedings of International Conference on Computational Linguistics, \textbf{COLING 2022}, Republic of Korea. \href{https://arxiv.org/abs/2208.10310}{[\underline{Paper}]}\href{https://arxiv.org/abs/2208.10310}{ } | \href{https://github.com/ashishgupta2598/sacti}{[\underline{Code}]}\href{https://github.com/ashishgupta2598/sacti}{ }
    
    \item \textbf{Jivnesh Sandhan}, Amrith Krishna, Ashim, Gupta, Pawan Goyal, and Laxmidhar Behera. \emph{A Little Pretraining Goes a Long Way: A Case Study on Dependency Parsing Task for Low-resource Morphologically Rich Languages.} Proceedings of the European Chapter of the Association for Computational Linguistics, \textbf{EACL-SRW 2021}, Ukraine. \href{https://aclanthology.org/2021.eacl-srw.16/}{[\underline{Paper}]}\href{https://aclanthology.org/2021.eacl-srw.16/}{ } | \href{https://github.com/Jivnesh/LCM}{[\underline{Code}]}\href{https://github.com/Jivnesh/LCM}{ }

    \item \textbf{Jivnesh Sandhan}, Ayush Daksh, Om Adideva Paranjay, Laxmidhar Behera, Pawan Goyal. \emph{Prabhupadavani: A Code-mixed Speech Translation Data for 26 Languages.} Proceedings of International Conference on Computational Linguistics Workshop on Computational Linguistics for Cultural Heritage, Social Sciences, Humanities, \textbf{COLING-LaTeCH-CLfL 2022}, Gyeongju, Republic of Korea. \href{https://arxiv.org/abs/2201.11391}{[\underline{Paper}]}\href{https://arxiv.org/abs/2201.11391}{ } | \href{https://github.com/frozentoad9/CMST}{[\underline{Code}]}\href{https://github.com/frozentoad9/CMST}{ }
    
     \item \textbf{Jivnesh Sandhan}, Om Adideva, Digumarthi Komal, Laxmidhar Behera, Pawan Goyal. \emph{Evaluating Neural Word Embeddings for Sanskrit.} In Proceedings of the World Sanskrit Conference, \textbf{WSC 2022}, Canberra, Australia. \href{https://arxiv.org/abs/2104.00270}{[\underline{Paper}]}\href{https://arxiv.org/abs/2104.00270}{ } | \href{https://github.com/Jivnesh/EvalSan}{[\underline{Code}]}\href{https://github.com/Jivnesh/EvalSan}{ }
     
    \item \textbf{Jivnesh Sandhan}, Amrith Krishna, Pawan Goyal, and Laxmidhar Behera. \emph{Revisiting the role of feature engineering for compound type identification in Sanskrit.} In Proceedings of the 6th International Sanskrit Computational Linguistics Symposium, \textbf{ISCLS 2019}, Kharagpur, India. \href{https://aclanthology.org/W19-7503/}{[\underline{Paper}]}\href{https://aclanthology.org/W19-7503/}{ } | \href{https://github.com/Jivnesh/ISCLS-19}{[\underline{Code}]}\href{https://github.com/Jivnesh/ISCLS-19}{ }
    \end{enumerate}
}


\clearpage 
\setstretch{1.3} 

\acknowledgements{\addtocontents{toc}{\vspace{1em}} 
I would like to express my sincere gratitude to my advisor, Dr. Laxmidhar Behera, for inspiring me to work in Sanskrit Computational Linguistics (SCL) and for providing me with numerous opportunities to explore and collaborate. Dr. Behera's guidance, vision, and mentorship have been invaluable throughout my journey, and without his support, this thesis would not have been possible.
Additionally, I would like to extend my gratitude to Dr. Pawan Goyal, my mentor, for his knowledge, consideration, and easy-going nature. He has been a constant source of support and guidance, and his contributions have been instrumental in the completion of this thesis.

 My association with Dr. Amrith Krishna during the initial phase of PhD has helped me shape my perspective toward research. I extend my gratitude to Amba Kulkarni, Gerard Huet, and Oliver Hellwig for their contributions to the SCL community. I am grateful to Rathin Singha, Narein Rao, Ashish Gupta, Anshul Agarwal, Amruta Barbarikar and other collaborators who selflessly contributed to the works discussed in this thesis. I would also like to thank all collaborators and reviewers for their comments and feedback.

I am grateful to the Bhaktivedanta family (BVC) for introducing me to the teachings of H.D.G. A. C. Bhaktivedanta Swami Prabhupada. I owe a debt of gratitude to my BVC friends, and in particular, I want to thank Hariom Birla for his invaluable help, patience, and emotional support. In addition, I am deeply appreciative of my parents, sister-in-law, and brother for their unwavering love and encouragement.

}
\clearpage 


\pagestyle{fancy} 

\lhead{\emph{Contents}} 
\tableofcontents 

\lhead{\emph{List of Figures}} 
\listoffigures 

\lhead{\emph{List of Tables}} 
\listoftables 


\clearpage 




%




\clearpage 

\clearpage 

\setstretch{1.3} 
\pagestyle{empty} 
\dedicatory{Dedicated to My Beloved Teachers
} 
\addtocontents{toc}{\vspace{2em}} 


\mainmatter 

\pagestyle{fancy} 


\input{Chapters/Chapter1}
\input{Chapters/Chapter2} 
\input{Chapters/Chapter3}
\input{Chapters/Chapter4} 
\input{Chapters/Chapter5} 
\input{Chapters/Chapter6}

\input{Chapters/Chapter8} 

\addtocontents{toc}{\vspace{2em}} 

\appendix 


\input{Appendices/AppendixA}

\addtocontents{toc}{\vspace{2em}} 

\backmatter

\nocite{*}
\label{Bibliography}

\lhead{\emph{Bibliography}} 
\bibliographystyle{acl_natbib} 

\bibliography{Bibliography} 

\end{document}

%% file: Chapters/Chapter1.tex

\chapter{Introduction} 

\label{Chapter1}

\lhead{Chapter 1. \emph{Introduction}} 

Sanskrit, an ancient language of India, is known for its rich cultural heritage and its ability to preserve knowledge. With the advent of digitization, Sanskrit manuscripts have become more accessible \cite{goyal-etal-2012-distributed,adiga-etal-2021-automatic}, but their utility is still limited due to various linguistic phenomena and the user's lack of language expertise. To make these manuscripts more accessible, this research aims to develop neural-based Sanskrit Natural Language Processing (NLP) systems that can be accessed through a user-friendly web interface. However, the Sanskrit language poses several challenges for building deep learning solutions, including the \textit{sandhi} phenomenon, rich morphology, frequent compounding, flexible word order, and limited resources. This research identifies 4 essential tasks for processing Sanskrit texts: word segmentation, dependency parsing, compound type identification, and analysis of the aesthetic beauty of Sanskrit poetry.

\begin{figure}[!thb]
\centering
\includegraphics[width=0.6\textwidth]{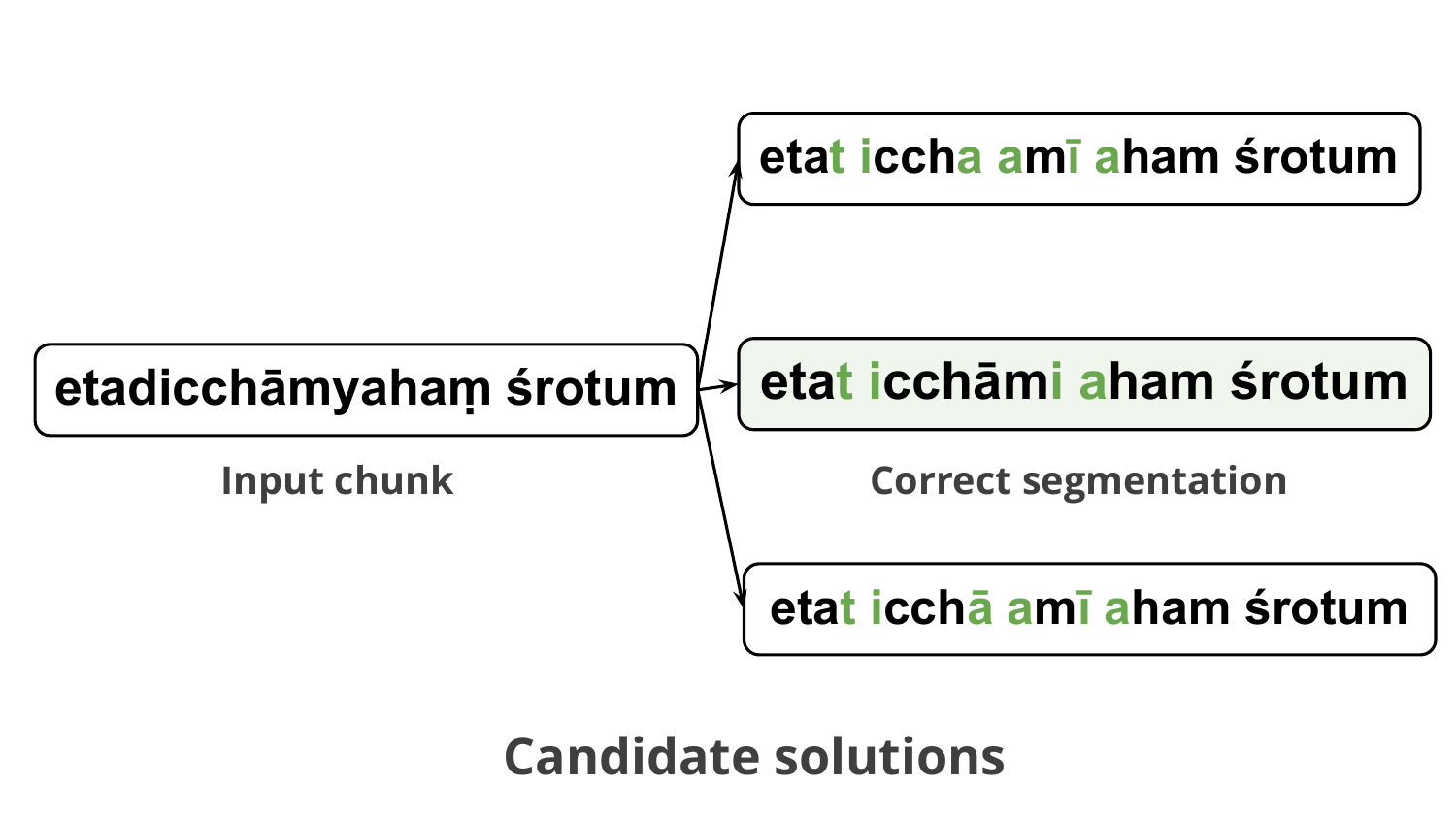}
\caption{An example that illustrates the difficulties posed by the \textit{sandhi} phenomenon for the SWS task.} 
\label{fig:sandhi_issue} 
\end{figure}

The conventional practice of Sanskrit word segmentation (SWS) is a crucial initial step in processing digitized manuscripts, as it enables accessibility and supports downstream tasks such as text classification \cite{sandhan-etal-2019-revisiting,krishna-etal-2016-compound}, morphological tagging \cite{gupta-etal-2020-evaluating,krishna-etal-2018-free}, dependency parsing \cite{sandhan-etal-2021-little,krishna-etal-2020-keep}, automatic speech recognition \cite{kumar2022linguistically} etc. Identifying Sanskrit word boundaries in word segmentation (SWS) is complicated due to the linguistic phenomenon of \textit{sandhi}, which involves phonetic transformations at word boundaries. This can obscure word boundaries and modify characters through deletion, insertion, and substitution operations. Figure~\ref{fig:sandhi_issue} provides examples of potential syntactical splits resulting from \textit{sandhi}, highlighting the difficulty of pinpointing the split location and the nature of the transformation at word boundaries.\footnote{We assume a unique valid solution for the input, though there may be multiple semantically valid candidates. However, this falls beyond the thesis scope.}

For the dependency parsing task, several strategies such as  data augmentation, sequential transfer learning, cross-lingual/mono-lingual pretraining, multi-task learning and self-training are tailored to enhance performance in low-resource scenarios.
 While these are well-known to the community, it is not trivial to select the best-performing combination of these strategies for a low-resource language that we are interested in, and not much attention has been given to measuring the efficacy of these strategies.   Assessing their utility for low-resource languages is essential before inventing novel ways to tackle data sparsity.

\begin{figure}[!thb]
\centering
\includegraphics[width=0.6\textwidth]{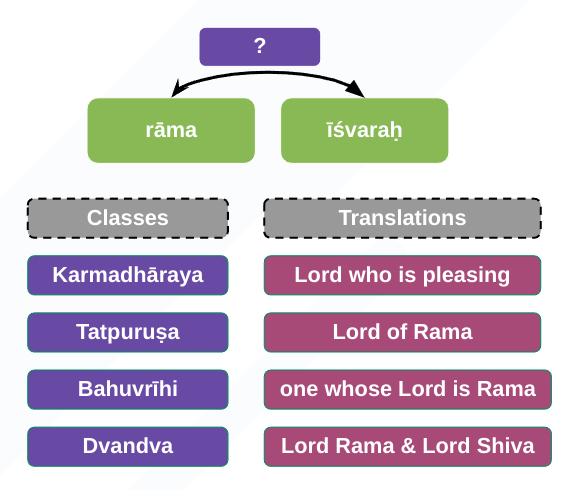}
\caption{Possible meanings of {\sl r\={a}ma-\={\i}\'{s}vara\d{h}} depending on the context-dependent semantic type} 
\label{fig:compo_meaning_intro} 
\end{figure}

The Sanskrit compound type identification (SaCTI) task is challenging and often depends upon the context or world knowledge about the entities involved \cite{krishna-etal-2016-compound}. For instance, as illustrated in Figure \ref{fig:compo_meaning_intro}, the semantic type of the compound {\sl r\={a}ma-\={\i}\'{s}vara\d{h}}  can be classified into one of the following semantic types depending on the context: {\sl Karmadh\={a}raya}\footnote{There are 4 broad semantic types of compounds: \textit{Avyay\={\i}bh\={a}va}, \textit{Bahuvr\={\i}hi}, \textit{Dvandva}, \textit{Tatpuru\d{s}a} and \textit{Dvandva}. {\sl Karmadh\={a}raya} is considered as sub-type of \textit{Tatpuru\d{s}a}. We encourage readers to refer \newcite{krishna-etal-2016-compound} for more details on these semantic types.}, {\sl Bahuvr\={\i}hi}  and {\sl Tatpuru\d{s}a}.
Although the compound has the same components as well as the final form, the implicit relationship between the components can be decoded only with the help of available contextual information \cite{kulkarni2013,krishna-etal-2016-compound}. Due to such instances, the downstream Natural Language Processing (NLP) applications for Sanskrit such as question answering \cite{terdalkar-bhattacharya-2019-framework} and machine translation \cite{aralikatte-etal-2021-itihasa}, etc. show sub-optimal performance when they stumble on compounds. 
For example, while translating {\sl r\={a}ma-\={\i}\'{s}vara\d{h}} into English, depending on the semantic type, there are four possible meanings (Figure~\ref{fig:compo_meaning_intro}): (1) Lord who is pleasing (in {\sl Karmadh\={a}raya})
(2) the one whose God is Rama (in {\sl Bahuvr\={\i}hi}) (3) Lord of Rama (in {\sl Tatpuru\d{s}a}) (4) Lord Rama and Lord Shiva ({\sl Dvandva}).
Therefore, the SaCTI task can be seen as a preliminary pre-requisite to building a robust NLP technology for Sanskrit.

Sanskrit literature has a rich and diverse tradition that has played a significant role in shaping the literary and cultural landscape of the Indian subcontinent for centuries \cite{pollock2006language,jamison2014rigveda}. The Sanskrit language, with its complex grammatical rules and nuanced vocabulary, has provided a fertile ground for poets to craft intricate and evocative verses that capture the essence of human experience \cite{pollock1996sanskrit}. From the ancient epics like the R\={a}m\={a}ya\d{n}a and the Mah\={a}bh\={a}rata, to the lyrical works of K\={a}lid\={a}sa and Bhart\d{r}hari etc. Sanskrit poetry has embodied the essence of Indian thought and culture, serving as a source of inspiration and contemplation for generations of readers and scholars. However, the computational linguistics researchers have not devoted much attention to uncovering the hidden beauty in Sanskrit poetry.

Summarily, this thesis aims to develop Sanskrit NLP technology by overcoming the above-mentioned challenges and making it accessible through a user-friendly web interface to make Sanskrit manuscripts more accessible.

\section{Objective}
The thesis aims to address how we can utilize natural language technologies to make Sanskrit texts more accessible. Given the above-mentioned challenges associated with analyzing Sanskrit texts, we now explain how to achieve this.

\noindent\textbf{How effectively can the linguistic-informed architectures and distributional information be incorporated into building data-driven systems for Sanskrit?}

Many different levels of linguistic information could be incorporated into a data-driven system for Sanskrit, including phonetics, morphology, syntax, and semantics. The challenge is to identify which levels of information are most relevant for the specific task at hand, and to design neural architectures that can effectively integrate this information. Sanskrit is a low-resource language, which means that there is limited annotated data available for training data-driven systems. To address this, transfer learning and unsupervised learning techniques can be used to leverage information from related languages or from unannotated text data. Overall, while there are challenges to incorporating linguistic-informed neural architectures and distributional information into data-driven systems for Sanskrit, with the right approaches and techniques, it is possible to build effective systems that can handle the complexities of this ancient language.

\noindent\textbf{How can we make our Sanskrit technology more accessible to end-users naive to deep learning?}
To make Sanskrit NLP technology more accessible to users naive to deep learning, we can take several steps such as:
\begin{itemize}
\item  Developing a user-friendly web application: Developing a web application that is user-friendly and easy to use can help users interact with the system in a much simpler way. The application should have an intuitive and simple interface that is easy to navigate and understand.
\item Offering proper documentation of the User-interface: Providing comprehensive documentation can help users understand how the system works and how to use it effectively. The documentation should include detailed instructions and examples that demonstrate the system's capabilities.
\end{itemize}
By taking these steps, we can make Sanskrit NLP technology more accessible to users unfamiliar with deep learning, enabling them to take advantage of these systems' benefits.

\noindent\textbf{How can we draw the attention of computational linguistics researchers / developers towards developing technology for Sanskrit?}
(1) Make datasets publicly available: Datasets are essential for developing and testing NLP systems. By making Sanskrit datasets publicly available, researchers can use these datasets to develop and evaluate their models. This can include both annotated datasets for various tasks.
(2) Publicly release source codes: Releasing the source code for Sanskrit NLP systems can help researchers reproduce and build upon previous work. It can also help to promote transparency and accountability in research. By releasing the source code, researchers can contribute to the development of the field and foster collaborations.
(3) Providing pre-trained models: Pre-trained models can be made available for download to help users get started quickly without needing to have deep knowledge of deep learning. This can help users save time and resources while working with the technology.
(4) Track Sanskrit NLP progress: Keeping track of Sanskrit NLP progress can help to make researchers aware of the state-of-the-art techniques and resources available for working with Sanskrit. This can include maintaining a leaderboard for benchmark datasets, publishing yearly progress reports, and hosting workshops and tutorials to teach new techniques and tools. By implementing these strategies, we can make Sanskrit NLP research more visible and accessible to researchers in computational linguistics, promoting collaborations and advancing the development of the field.

\section{Contribution}
This thesis explores how natural language technologies can be applied to make Sanskrit texts more accessible, considering morphological richness, compounding, relatively free word orderliness, and low
resource nature. We focus on tasks such as word segmentation, dependency parsing, compound type identification, and analysis of the aesthetics of Sanskrit poetry. We aim to address the language's low-resource nature by incorporating rich linguistic knowledge into our models and relying less on task-specific labelled data.

\noindent\textbf{Linguistically-informed neural architectures:} For the Sanskrit Word Segmentation (SWS) task, we propose the linguistically-informed tokenization module which accommodates language-specific \textit{sandhi} phenomenon and adds inductive bias. It also contains a novel soft-masked attention module that helps to add inductive bias for prioritizing potential candidates keeping mutual interactions between all candidates intact. Further, a novel path ranking algorithm is integrated to rectify the corrupted predictions. 
For the context-sensitive compound type identification (SaCTI) task, we propose a novel linguistically-informed context-sensitive multi-task learning architecture. We illustrate that morphological tagging and dependency parsing auxiliary tasks are helpful and serve as a proxy for explainability of system predictions for the SaCTI task.
 The dependency parsing literature has proposed several strategies to enhance performance in low-resource scenarios.  We propose a linguistically-informed ensembled architecture which augments several strategies such as data augmentation, sequential transfer learning, cross/mono-lingual pretraining, multi-task learning and self-training.
Finally, we propose to build an interpretable framework to analyze and classify Sanskrit poetry (k\={a}vya) \cite{baumann-etal-2018-analysis,kesarwani-etal-2017-metaphor} into levels of the best composition. We propose an automated, human-in-loop and interpretable framework consisting of several linguistically-informed modules: such as Rasa (the emotional essence of poetry), Ala\.{n}k\={a}ra (the use of rhetorical and figurative devices), Dhvani (the power of suggestion), Vakrokti (oblique), Aucitya (appropriateness) and r\={\i}ti (the appropriate use of style and language).

\noindent\textbf{State-of-the-art performance on available benchmark datasets for Sanskrit:} We build linguistically-informed architectures for word segmentation, morphological tagging, dependency parsing and compound type identification tasks. For all the above mentioned tasks, our systems report state-of-the-art performance. 
For the word segmentation task, TransLIST marks an average 7.2 points perfect match absolute gain over the current state-of-the-art system \cite{hellwig-nehrdich-2018-sanskrit}.
For the morphological tagging task, we train a neural-based architecture \cite[LemmaTag]{kondratyuk-etal-2018-lemmatag} on Sanskrit dataset \cite{hackathon}.  Currently, our system trained on the Sanskrit dataset stands first on the Hackathon dataset \cite{hackathon} leaderboard. 
For the dependency parsing task, we show a successful application of the ensembled system on a truly low-resource language Sanskrit. We find that the ensembled system outperforms the state-of-the-art system \cite{krishna-etal-2020-keep} for Sanskrit by 1.2 points absolute gain in terms of UAS and shows comparable performance in terms of LAS.  
In the context-sensitive compound type identification task, we report results with $7.71$ points (F1) absolute gains compared to the current state-of-the-art system by \newcite{krishna-etal-2016-compound}.

\noindent\textbf{Interpretability and multilingual extension:} 
In this thesis, we conduct probing analyses on attention modules in different NLP tasks, including word segmentation, compound type identification, and evaluation of aesthetic beauty of poetry. Specifically, we investigate the Soft masked attention (SMA) module in our system for its ability to effectively model inductive bias in the word segmentation task. Our findings indicate that the SMA module prioritizes candidate words containing the input character, with local information relevant for \textit{sandhi} split being captured by char-char interactions. Furthermore, the SMA module improves interactions between characters and words, and our analysis suggests its necessity in our word segmenter. We also probed the attention modules in a proposed system for the compound type identification task. The attention heatmaps revealed that all words mostly focus on the target compound word in the SaCTI task, while in the dependency parsing task, the focus shifts based on the context. Our analysis suggests that auxiliary tasks, such as morphological tagging and dependency prediction, add complementary signals to the system and serve as a proxy for explainability. To improve low-resource dependency parsing, we proposed an ensemble architecture that includes multi-task learning with the morphological prediction task. The morphological tag prediction head can also serve as a proxy for explainability for the correct / incorrect predictions as morphological information plays an important role in deciding the syntactic role that a nominal can be assigned in the sentence.
Lastly, we proposed a human-in-loop and interpretable framework for evaluating the aesthetic beauty of Sanskrit poetry. Our framework consists of 7 modules from k\={a}vya\'s\={a}stra and Chanda\'s\'s\={a}stra, which serve as explainability for the system's prediction. Overall, our findings highlight the importance of attention modules and auxiliary tasks in NLP tasks and emphasize the need for explainable AI systems.

We also showcase efficacy of our proposed approaches on multiple languages in low-resource settings. For the dependency parsing task, our exhaustive experimentation empirically establishes the effective generalization ability of the ensembled system on 7 languages and shows average absolute gains of 5.2/6.2 points Unlabelled/Labelled Attachment Score (UAS/LAS) over strong baseline \cite{dozat2017stanford}.  Notably, our ensembled system shows substantial improvements for the languages not covered in pretrained models.  To measure the efficacy of the proposed pretraining method, we further perform a series of experiments on 10 Morphologically Rich Languages (MRLs) in low-resource settings and show 2.0 points and 3.6 points average absolute gain  in terms of UAS and LAS, respectively. Our proposed pretraining also outperforms multilingual BERT \cite[mBERT]{devlin-etal-2019-bert} based multi-task learning model \cite[Udify]{kondratyuk-straka-2019-75} for the languages which are not covered in mBERT. For the context-sensitive compound type identification task, we show the efficacy of the proposed approach in English and Marathi.

\section{Organization of the Thesis}
This section offers a concise idea of how the thesis is structured.
\begin{itemize}
    \item Chapter 1 discusses how we can utilize natural language technologies to provide ease for making Sanskrit texts more accessible. In this chapter, the aims of the thesis are explained and the significant achievements of the thesis are outlined.

    \item Chapter 2 investigates the progress made in computational processing of Sanskrit by reviewing current literature. Additionally, it discusses techniques deployed in low-resource and language agnostic settings of NLP where similar tasks are addressed.

    \item Chapter 3 examines the challenges in Sanskrit Word Segmentation (SWS) due to the \textit{sandhi} phenomenon and the limitations of existing approaches using lexicon or deep learning. The proposed solution is the Transformer based Linguistically Informed Sanskrit Tokenizer (TransLIST), which uses a module to encode character input and latent-word information, soft-masked attention to prioritize candidate words, and a path ranking algorithm to rectify predictions.

    \item   Chapter 4 discuses the challenges of dependency parsing in low-resource settings for morphological rich languages (MRLs) due to morphological disambiguation and lack of powerful analyzers. It proposes a simple pretraining which linguistically motivated auxiliary tasks. The chapter also experiments with 5 low-resource strategies for an ensembled approach on 7 Universal Dependency (UD) low-resource languages and shows successful application on a truly low-resource language, Sanskrit.

    \item   Chapter 5 discusses the Sanskrit Compound Type Identification (SaCTI) task. It investigates whether recent advances in neural networks can outperform traditional hand-engineered feature-based methods in the context-agnostic SaCTI setting where an only compound word is used as input for the classification. Previous approaches only relied on lexical information, but this chapter proposes to incorporate contextual and syntactic information  in the context-sensitive setting using a multi-task learning architecture with morphological tagging and dependency parsing as auxiliary tasks.

    \item   Chapter 6 explores the intersection of Sanskrit poetry and computational linguistics by proposing a framework to classify and analyze Sanskrit poetry into levels of the best composition. This chapter proposes a human-in-the-loop approach that combines deterministic aspects delegated to machines and deep semantics left to human experts. It provides an analysis of a Sanskrit poem and various annotations for it, including compound, dependency, anvaya, meter, rasa, ala\.nk{\=a}ra , and r\={\i}ti. It aims to bridge the gap between k\={a}vya\'s\={a}stra and computational methods and pave the way for future research in this area.


    \item Chapter 7 provides a summary of the main contributions and suggests areas for future research.

\end{itemize}

%% file: Chapters/Chapter2.tex

\chapter{Literature Review}

\label{Chapter2}

\lhead{Chapter 2. \emph{Literature Review}} 
Researchers have worked on a range of applications related to the Sanskrit language. One area of focus has been the development of digital lexicons \cite{hellwig2010dcs}, thesauri, and wordnets \cite{kulkarni-etal-2019-introduction}, which help with vocabulary building and comprehension. Researchers also worked on developing rule-based systems for Sanskrit dependency parsing \cite{kulkarni-etal-2019-dependency,kulkarni-2013-deterministic}. Rule-based systems have also been developed for analyzing Sanskrit corpus, while simulation of P\={a}n\={\i}nian grammar is another area of interest \cite{goyal2016design}. Efforts have been put for the acquisition and maintenance of Sanskrit digital corpus, as well as for automated analysis of Sanskrit poetry and recognition of meter \cite{terdalkar-bhattacharya-2023-chandojnanam}. OCR recognition of romanized Sanskrit has also been explored \cite{krishna2018upcycle}, along with the development of annotation and editing tools \cite{maheshwari-etal-2022-benchmark}. User interface design is another area of interest, particularly in creating software tools for teaching Sanskrit. Researchers have also explored the possibility of Sanskrit speech recognition \cite{adiga-etal-2021-automatic}, which would have significant implications for language learning and accessibility. Google Translate has recently introduced an updated version of its automated translation system, specifically designed for languages with limited available resources, including Sanskrit.

\section{Word Segmentation}
\label{sws_literature_review}
Previous attempts at SWS mainly focused on rule-based Finite State Transducer systems \cite{gerard2003lexicon,mittal-2010-automatic}. One approach produced all possible solutions and recommended a solution based on a probabilistic score inferred from a dataset of 25,000 data splits \cite{mittal-2010-automatic}. Another approach attempted to solve the SWS task for sentences with one or two splits using a Bayesian approach and the same dataset \cite{natarajan}.
Recently, a "lexicon-driven" shallow parser was proposed, which aids in efficient selection of segmentation solutions aligned with the input sequence \cite{goyal2016}. This, along with the recent availability of segmentation datasets \cite{krishna-etal-2017-dataset,hellwig-nehrdich-2018-sanskrit,hackathon_data}, led to two categories of approaches: "lexicon-driven" and "purely engineering."
The "lexicon-driven" approaches mainly exploit the candidate space generated by the shallow parser, while "purely engineering" approaches avoid relying on the shallow parser and instead use sequence labeling or seq2seq models \cite{hellwig2015using,hellwig-nehrdich-2018-sanskrit,aralikatte-etal-2018-sanskrit,reddy-etal-2018-building}.
However, these existing approaches for SWS are either brittle in realistic scenarios or do not consider potentially useful external information. The TransLIST approach presented in this paper addresses these shortcomings by combining the strengths of both the "lexicon-driven" and "purely engineering" approaches, resulting in a new state-of-the-art solution that does not rely on hand-engineered features and is highly scalable and applicable in realistic scenarios.

\section{Dependency Parsing}
The P\={a}\d{n}inian framework, the oldest dependency grammar, is generative and deterministically irreversible. That leaves the scope of ambiguity when it is reversed for the Sanskrit analyzer.  \newcite{goyal2007analysis} proposed a rule-based system and pointed out the need for a hybrid scheme where the data-driven statistical method is integrated with the rule-based system to resolve ambiguity posed by the irreversibility of grammar rules. Later,~\newcite{DBLP:conf/sanskrit/Hellwig09} took forward this idea and improved performance using a hybrid approach, purely statistical parser augmented with simple syntactic rules from grammar. Other early attempts for developing dependency parser for the Sanskrit mainly focused on building rule-based systems ~\cite{kulkarni2010designing,kulkarni2013deterministic,kulkarni-etal-2019-dependency,Kulkarni2021SanskritPF}. \newcite{kulkarni2010designing} proposed a graph-based approach where each word is taken as a node and relations between words are identified using grammar rules. Based on heuristics, they rank the exhaustive candidate space. Later, they improved the computational aspect of their follow up work~\cite{kulkarni2013deterministic}. Earlier attempts for building dependency parser were targeted for simple prose order sentences so~\newcite{kulkarni-etal-2019-dependency} extended their previous work for poetry sentences. However, this system enumerates all possible solutions. Due to the increasing popularity of dependency parser across NLP,~\newcite{goyal2014converting} introduced a methodology to convert enriched constituency trees into unlabelled dependency tree. Recently,~\newcite{amrith2019thesis} proposed structured prediction framework for dependency parsing. This linguistically motivated model uses a graph-based parsing approach and currently reports a state-of-the-art performance for Sanskrit. Although the results of this system are very promising, it takes input from lexicon-driven shallow parsers. If parser fails to identify word, this system can not produce a parsed tree.
 Recently, \newcite{sandhan-etal-2021-little} proposed pretraining approach for low-resource dependency parsing for Sanskrit. Due to lack of powerful morphological analyzer, it is challenging to have this information during run time. Therefore, \newcite{sandhan-etal-2021-little} obviate the need of morphological information during run time with the help of proposed pretraining method. Data-driven approaches have shown tremendous achievement in the field of NLP~\cite{hellwig-nehrdich-2018-sanskrit,sandhan-etal-2019-revisiting}. Specifically, for dependency parsing, neural-based approaches have attracted intense attention of researchers due to state-of-the-art performance without explicit feature engineering~\cite{DBLP:conf/iclr/DozatM17,fernandez-gonzalez-gomez-rodriguez-2019-left,zhou-zhao-2019-head}. However, ~\newcite{krishna2020neural} reports that these \textit{purely data-driven} approaches do not match the performance of \textit{hybrid} counterparts due to data scarcity. 
 
 Thus, in this work, we investigate the following question: How far can we push a \textit{purely data-driven} approach using recently proposed strategies for low-resource settings? Can this simple ensemble approach outperform
the state-of-the-art of Sanskrit? We experiment with five strategies, namely, data augmentation, sequential transfer learning, pretraining, multi-task learning and self-training.

 \section{Compound Type Identification}
Semantic analysis of compounds is an essential preprocessing step for improving on overall downstream NLP applications such as information extraction, question answering, machine translation, and many more~\cite{start_related}. It has captivated much attention from the computational linguistics community, particularly on languages like English, Dutch, Italian, Afrikaans, and German~\cite{verhoeven2014automatic}. By rigorously studying Sanskrit compounding system and Sanskrit grammar, analysis of compounds in Hindi and Marathi has been done~\cite{amba_related}. Another interesting approach uses simple statistics on how to automate segmentation and type identification of compounds~\cite{processor}. ~\newcite{nastase2006learning} show that from two types of word meaning, namely, based on lexical resources and corpus-based, noun-modifier semantic relations can be learned. Another exciting work by~\newcite{seaghdha2013interpreting} has done noun-noun compound classification using statistical learning framework of kernel methods, where the measure of similarity between compound components is determined using kernel function. Based on \textit{A\d{s}\d{t}\={a}dhy\={a}y\={\i}} rules,~\newcite{kulkarni2013clues} has developed rule-based compound type identifier. This study helped to get more insights on what kind of information should be incorporated into lexical databases to automate this analysis.~\newcite{kulkarni2011statistical} proposed a constituency parser for Sanskrit compounds to generate paraphrase of the compound which helps to understand the meaning of compounds better.

Recently, neural models are widely used for different downstream NLP applications for Sanskrit. The error corrections in Sanskrit OCR documents is done based on a neural network based approach~\cite{ocr}. Another work used neural models for post-OCR text correction for digitising texts in Romanised Sanskrit~\cite{krishna2018upcycle}.~\newcite{hellwig2018sanskrit} proposed an approach for automating feature engineering required for the word segmentation task. Another neural-based approach for word segmentation based on seq2seq model architecture was proposed by ~\newcite{reddy2018building}, where they have shown significant improvement compared to the previous linguistically involved models. Feedforward networks are used for building Sanskrit character recognition system~\cite{dineshkumar2015sanskrit}.~\newcite{krishna2018free} proposed energy-based framework for jointly solving the word segmentation and morphological tagging tasks in Sanskrit. The pretrained word embeddings proposed by Mikolov~\shortcite{w2v} and Pennington~\shortcite{glove} had a great impact in the field of Natural Language Processing (NLP). However, these token based embeddings were unable to generate embeddings for out-of-vocabulary (OOV) words. To overcome this shortcoming, subword level information was integrated into recent approaches, where character-n-gram features~\cite{fasttext} have shown good performance over the compositional function of individual characters~\cite{wieting2015towards}. Another interesting approach~\cite{charcnn} is the use of character level input for word-level predictions. 

Prior to the deep learning era, various machine learning-based approaches have been proposed for Noun Compound Identification \cite{kim-baldwin-2005-automatic,o-seaghdha-copestake-2009-using,tratz-hovy-2010-taxonomy}. With the advent of deep-learning based approaches, \newcite{dima-hinrichs-2015-automatic} and \newcite{fares-etal-2018-transfer} proposed a neural-based architecture where concatenated representations of a compound were fed to a feed-forward network to predict a semantic relation between the compound's components. \newcite{shwartz-waterson-2018-olive} proposed an approach that combines labeling with paraphrasing. 
Recently, \newcite{ponkiya-etal-2021-framenet} proposed a novel approach using semantic label repository \cite[FrameNet]{ponkiya-etal-2018-towards} where continuous label space embeddings are used to predict unseen labels. To the best of our knowledge, a context has never been used for the classification task for NCI. In paraphrasing line of modeling, \newcite{ponkiya-etal-2020-looking} formulates paraphrasing as ``fill-in-the-blank'' problem to predict the ``missing'' predicate or preposition using pretrained language models.

Sanskrit Compound Type Identification task has garnered considerable attention of the researchers in the last decade.
In order to decode the meaning of a Sanskrit compound, it is essential to figure out its constituents \cite{huet2010,mittal-2010-automatic,hellwig-nehrdich-2018-sanskrit}, how the constituents are grouped \cite{Kulkarni2011StatisticalCP}, identify the semantic relation between them \cite{anil_thesis} and finally generate the paraphrase of the compound \cite{Kumar2009SanskritCP}.
 \newcite{Satuluri2013GenerationOS} and \newcite{kulkarni2013} proposed a rule-based approach where around 400 rules mentioned in P\={a}\d{n}ini's grammar \cite{panini} were analysed from the perspective of compound generation and type identification, respectively.
 Recently, \newcite{sandhan-etal-2019-revisiting} investigated whether a purely engineering data-driven approach competes with the performance of a linguistically motivated hybrid approach by \newcite{krishna-etal-2016-compound}. 
 Summarily, no attention has been given to incorporating contextual information, which is crucial and cheaply available. We address this research gap and mark the new state-of-the-art results with substantial improvements.

\section{Aesthetics of Sanskrit Poetry}
\noindent\textbf{The 6 main schools of k\={a}vya\'s\={a}stra (literary theory):}
k\={a}vya\'s\={a}stra is the traditional Indian science of poetics and literary criticism, which has played a significant role in shaping the development of literature and aesthetics in the Indian subcontinent for over two thousand years. The term "K\={a}vya" refers to poetry or literature, while "\'S\={a}stra" means science or knowledge, and k\={a}vya\'s\={a}stra is thus the systematic study of the nature, forms, and principles of poetry and literature.
The roots of k{\=a}vya\'s\={a}stra can be traced back to ancient India, where it developed alongside other branches of learning such as philosophy, grammar, and rhetoric. Over time, k\={a}vya\'{s}\={a}stra evolved into a complex and sophisticated system of literary theory, encompassing a wide range of concepts and techniques for analyzing and appreciating poetry, such as Rasa (the emotional essence of poetry), Ala\.{n}k\={a}ra (the use of rhetorical and figurative devices), Dhvani (the power of suggestion), Vakrokti (oblique), Aucitya (appropriateness) and r\={\i}ti (the appropriate use of style and language). Table \ref{table:kāvyaśāstra} gives a brief overview of these 6 schools in chronological order with founder \={a}c\={a}rya, their treatise and objectives.
 \begin{table*}[t]
\centering
\begin{adjustbox}{}
\small
\begin{tabular}{|l|l|l|L|L|}
\hline
\textbf{School}   & \textbf{Founder}        & \textbf{Treatise}                  & \textbf{Objective}                                                                     & \textbf{English meaning}                                                                                                                                                         \\\hline
Rasa     & Bharatamun\={\i}    & N\={a}\d{t}ya\'s\={a}stra              & \textit{na hi rasādṛte kaścidarthaḥ pravartate}                                         & No meaning proceeds, if a poetry does not carries Rasa.                                                                                                                      \\\hline
Ala\.{n}k\={a}ra & Bh\={a}maha        & K\={a}vy\={a}la\.{n}k\={a}ra              & \textit{rūpakādiralaṃkāraḥ tasyānyairvahudhoditaḥ na kāntamapi nirbhūṣaṃ vibhāti vanitāmukham } & Ala\.{n}k\={a}ras are vital for poetic beauty, just like ornaments for a charming woman's face. \\ \hline
R\={\i}ti    & V\={a}mana         & K\={a}vy\={a}la\.{n}k\={a}ra-s\={u}trav\d{r}tti & \textit{rītirātmā kāvyasya}                                                         & Poetic style (R\={\i}ti) is the soul of poetry.                                                                                                                                  \\\hline
Dhvani    & \={A}nandavardhana & Dhvany\={a}loka                & \textit{kāvyasyātmā dhvaniḥ }                                                       & Dhvani is the soul of poetry.                                                                                                                                                \\\hline
Vakrokti & Kuntaka        & Vakrokti-j\={\i}vitam         & \textit{vakroktiḥ kāvyajīvitam}                                                   & vakrokti is the vital element of the poetry.                                                                                                                                 \\\hline
Aucitya & K\d{s}emendra     & Aucitya-vic\={a}ra-carc\={a}  & \textit{aucityaṃ rasasiddhasya sthiraṃ kāvyasya jīvitam}                                    & Propriety is the stable vital element of the poetry full of rasa.    \\  \hline
\end{tabular}
\end{adjustbox}
\caption{Chronological overview of 6 schools of kāvyaśāstra with founder \={a}c\={a}rya, their treatise and objectives.} 
\label{table:kāvyaśāstra}
\end{table*}

\noindent\textbf{Chanda\'s\'s\={a}stra:} 
Chanda\'s\'s\={a}stra is the traditional Indian science of meter and versification in poetry, dating back to the Vedic period. It involves the systematic study of the principles, forms, and structures of meter and versification in poetry, including the use of various poetic devices and the study of various types of meters and poetic forms. While closely related to k\={a}vya\'s\={a}stra, Chanda\'s\'s\={a}stra is a separate branch of traditional Indian knowledge and is not considered one of the 6 main schools of k\={a}vya\'s\={a}stra. Key concepts and techniques include the classification of meters, rhyme and alliteration, and the principles of accentuation and stress. Chanda\'s\'s\={a}stra has played a significant role in the development of Indian poetics and literary theory, and continues to be a vital part of Indian cultural heritage.
By mastering the principles of Chanda\'s\'s\={a}stra, poets and writers are able to create verses that are both aesthetically pleasing and technically precise, resulting in some of the most beautiful and evocative poetry in the world.

\noindent\textbf{Efforts put in Computational Linguistics:}
In the field of computational poetry, researchers are tackling various problems using algorithms and computational methods, such as generating new poetry \cite{agarwal-kann-2020-acrostic,van-de-cruys-2020-automatic,li-etal-2020-rigid,hopkins-kiela-2017-automatically}, translating poetry \cite{yang-etal-2019-generating,ghazvininejad-etal-2018-neural} analyzing emotional tone, analyzing rhyme and meter patterns \cite{kao-jurafsky-2012-computational,greene-etal-2010-automatic,fang-etal-2009-adapting}, classifying poetry \cite{baumann-etal-2018-analysis,kesarwani-etal-2017-metaphor}, and recommending poetry \cite{10.1145/3341161.3342885,foley2019poetry}.
However, existing work does not focus much on analyzing the aesthetic aspect of poetry, which is crucial for improving the generation, translation, and recommendation applications. This gap motivates our work, which proposes an interpretable framework to classify Sanskrit poetry into different levels of  composition using computational linguistics.
We demonstrate the proposed framework by conducting a deep analysis of \'Sik\d{s}\={a}\d{s}\d{t}aka, a Sanskrit poem, from the perspective of 6 well-known k\={a}vya\'s\={a}stric schools. The key contributions of our paper include the proposed framework, the analysis of \'Sik\d{s}\={a}\d{s}\d{t}aka, the annotations, the web application, and the publicly released codebase for future research.

\section{Available Tools and Datasets}
Recently, the Sanskrit Computational Linguistics (SCL) field has seen significant growth in building web-based tools to help understand Sanskrit texts.
\newcite{goyal2016} introduced the Sanskrit Heritage Reader (SHR), a lexicon-driven shallow parser that aids in the selection of segmentation solutions.
Sams\={a}dhan{\=\i} is another web-based tool consisting of various rule-based modules.
Recently, \newcite{Sangrahaka,Chandojnanam} introduced a web-based annotation tool for knowledge-graph construction and a metrical analysis. 
In short, tools for NLP can be divided into two groups: rule-based and annotation tools. Rule-based tools have limitations such as not providing a final solution, limited vocabulary coverage, and lacking user-friendly annotation features. Annotation tools, on the other hand, don't have the recommendations of rule-based systems, relying solely on annotators. To address these limitations, a web-based annotation framework called SHR++ \cite{krishna-etal-2020-shr} was proposed. It combines the strengths of both types of tools by offering all possible solutions from rule-based system SHR for tasks like word segmentation and morphological tagging, allowing annotators to choose the best solution rather than starting from scratch. 

Our proposal, SanskritShala, goes a step further by integrating a neural-based NLP toolkit that combines state-of-the-art neural-based pre-trained models with rule-based suggestions through a web-based interface. Each module of SanskritShala is trained to predict the solutions from the exhaustive candidate solution space generated by rule-based systems. Hence, it makes predictions in real time using neural-based models that have already been trained. Thus, a complete solution is shown to the users / annotators, which was not possible in any of the previous attempts.

Further, annotators can easily correct the mispredictions of the system with the help of user-friendly web-based interface. This would significantly reduce the overall cognitive load of annotators. To the best of our knowledge, SanskritShala is the first NLP toolkit available for a range of tasks with a user friendly annotation interface integrated with the neural-based modules.  

%% file: Chapters/Chapter3.tex

\chapter{A Transformer-Based Linguistically Informed Sanskrit Tokenizer} 

\label{Chapter3} 

\lhead{Chapter 3. \emph{Sanskrit Word Segmentation}} 
Sanskrit Word Segmentation (SWS) is crucial for making digitized texts accessible and performing downstream tasks. However, it is challenging due to the \textit{sandhi} phenomenon that alters characters at word boundaries. Existing approaches for SWS are either lexicon-driven or purely engineering-based, but they have limitations in terms of handling out-of-vocabulary tokens or utilizing latent word information. To overcome these shortcomings, the proposed TransLIST approach combines the best of both approaches, using a transformer-based module that encodes character input with latent word information and accounts for \textit{sandhi} phenomenon. It also includes a soft-masked attention mechanism to prioritize candidate words and a path ranking algorithm to correct predictions. Experiments show that TransLIST outperforms the current state-of-the-art by an average of 7.2 points absolute gain in terms of the perfect match metric on benchmark datasets for SWS.\footnote{The codebase and datasets are publicly available at: \url{https://github.com/rsingha108/TransLIST}}

\section{Sanskrit Word Segmentation}

\begin{figure}[!thb]
\centering
\includegraphics[width=0.6\textwidth]{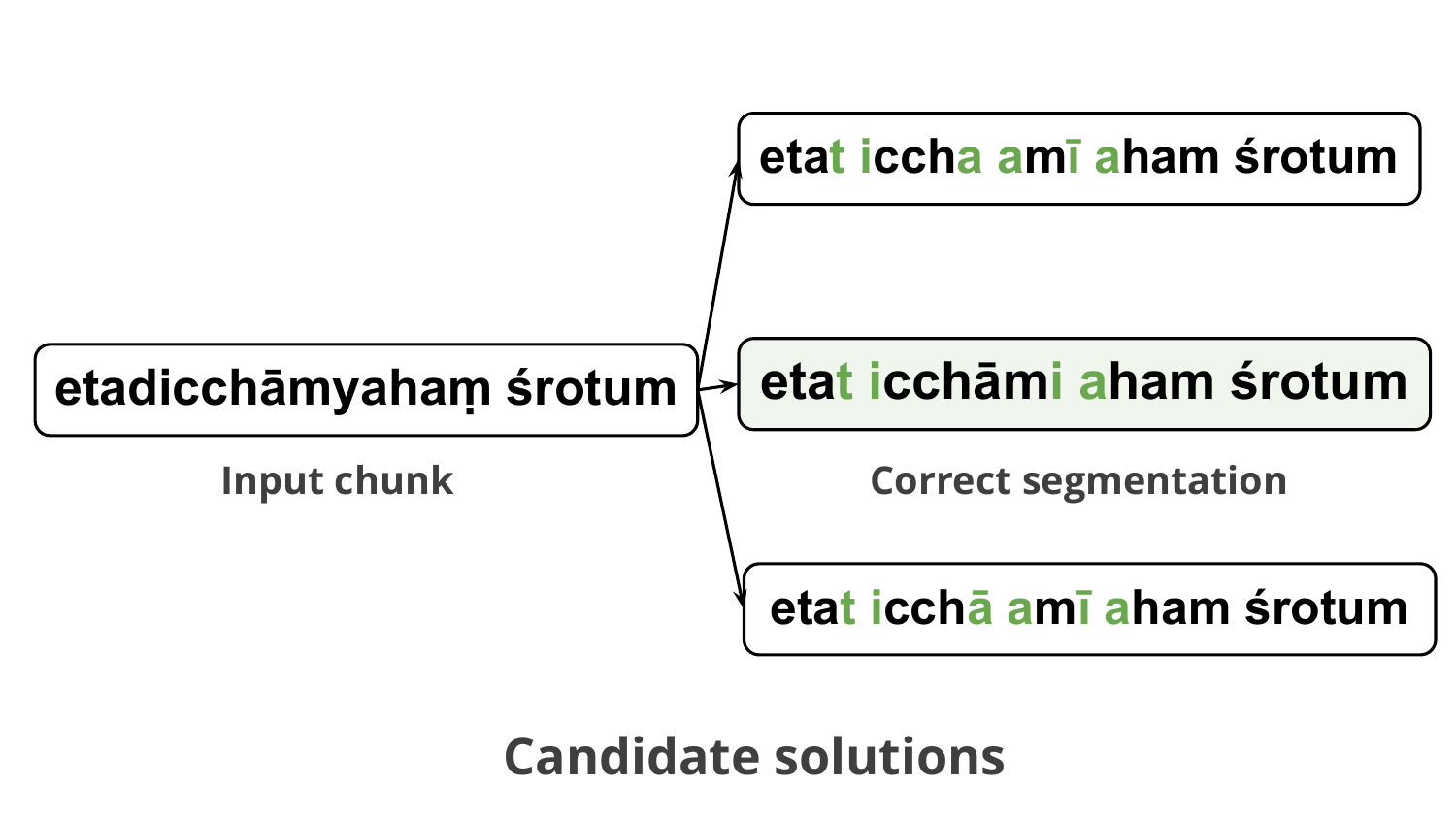}
\caption{An example that can demonstrate the difficulties caused by the phenomenon of \textit{sandhi} in the SWS task.} 
\label{fig:sandhi_issue_ch3} 
\end{figure}

The recent surge in SWS datasets \cite{krishna-etal-2017-dataset,hackathon_data} has led to various methodologies to handle SWS.
The current approaches that rely on a lexicon-based approach use a shallow parser called the Sanskrit Heritage Reader (SHR), which is popularly known for this purpose. These methods, such as the ones proposed in \cite{krishna-etal-2016-word,krishna-etal-2018-free,amrith21}, view the task as identifying the most accurate solution among the candidate solutions generated by SHR that are semantically and syntactically valid. By utilizing SHR to significantly reduce the exponential search space and incorporating linguistically-informed feature engineering, these lexicon-based systems such as those described in \cite{amrith21,krishna-etal-2018-free} achieve performance that is nearly close to the state-of-the-art for the SWS task.
However, these approaches rely on the completeness assumption of SHR, which is optimistic given that SHR does not use domain specific lexicons. These models are handicapped by the failure of this preliminary step.
Conversely, data-centric approaches that are purely engineering-based and lack explicit hand-crafted features and external linguistic resources, such as those proposed in \cite{hellwig-nehrdich-2018-sanskrit,reddy-etal-2018-building,aralikatte-etal-2018-sanskrit}, surprisingly achieve good performance. These purely engineering-based methods are known for their ease of scalability and deployment for training and inference. However, a disadvantage of these methods is that they do not consider the latent word information available through external resources.

There are also lattice-structured approaches \cite{zhang-yang-2018-chinese,gui-etal-2019-lexicon,li-etal-2020-flat} (originally proposed for Chinese Named Entity Recognition (NER), which incorporate lexical information in character-level sequence labelling architecture). However, these approaches cannot be directly applied for SWS; since acquiring word-level information is not trivial due to \textit{sandhi} phenomenon. To overcome these shortcomings,  we propose \textbf{Trans}former-based \textbf{L}inguistically \textbf{I}nformed \textbf{T}okenizer (TransLIST).

TransLIST is a combination of purely engineering-based and lexicon-driven approaches for the SWS task, which offers the following benefits: (1) It shares the scalability and ease of deployment of purely engineering-based methods during training and inference. (2) Like lexicon-driven approaches, it can take advantage of candidate solutions generated by the Sanskrit Heritage Reader (SHR) to enhance performance. (3) In contrast to lexicon-driven approaches, TransLIST is resilient and can operate even when the candidate solution space is partially available or unavailable.

Our key contributions are as follows:
(a) We propose the linguistically informed tokenization module (\S~\ref{LIST}) which accommodates language-specific \textit{sandhi} phenomenon and adds inductive bias for the SWS task.
(b) We propose a novel soft-masked attention (\S~\ref{sma_subsection}) that helps to add inductive bias for prioritizing potential candidates keeping mutual interactions between all candidates intact.
(c) We propose a novel path ranking algorithm (\S~\ref{constrained_inference}) to rectify the corrupted predictions. 
(d) We report an average 7.2 points perfect match absolute gain over the current state-of-the-art system \cite{hellwig-nehrdich-2018-sanskrit}.

We elucidate our findings by first describing TransLIST and its key components (\S~\ref{system_architecture}), followed by the evaluation of TransLIST against strong baselines on a test-bed of 2 benchmark datasets for the SWS task. Finally, we investigate and delve deeper into the capabilities of the proposed components and its corresponding modules (\S~\ref{fine_grained_analysis}).

\begin{figure*}[!tbh]
    \centering
    \subfigure[]{\includegraphics[width=0.45\textwidth]{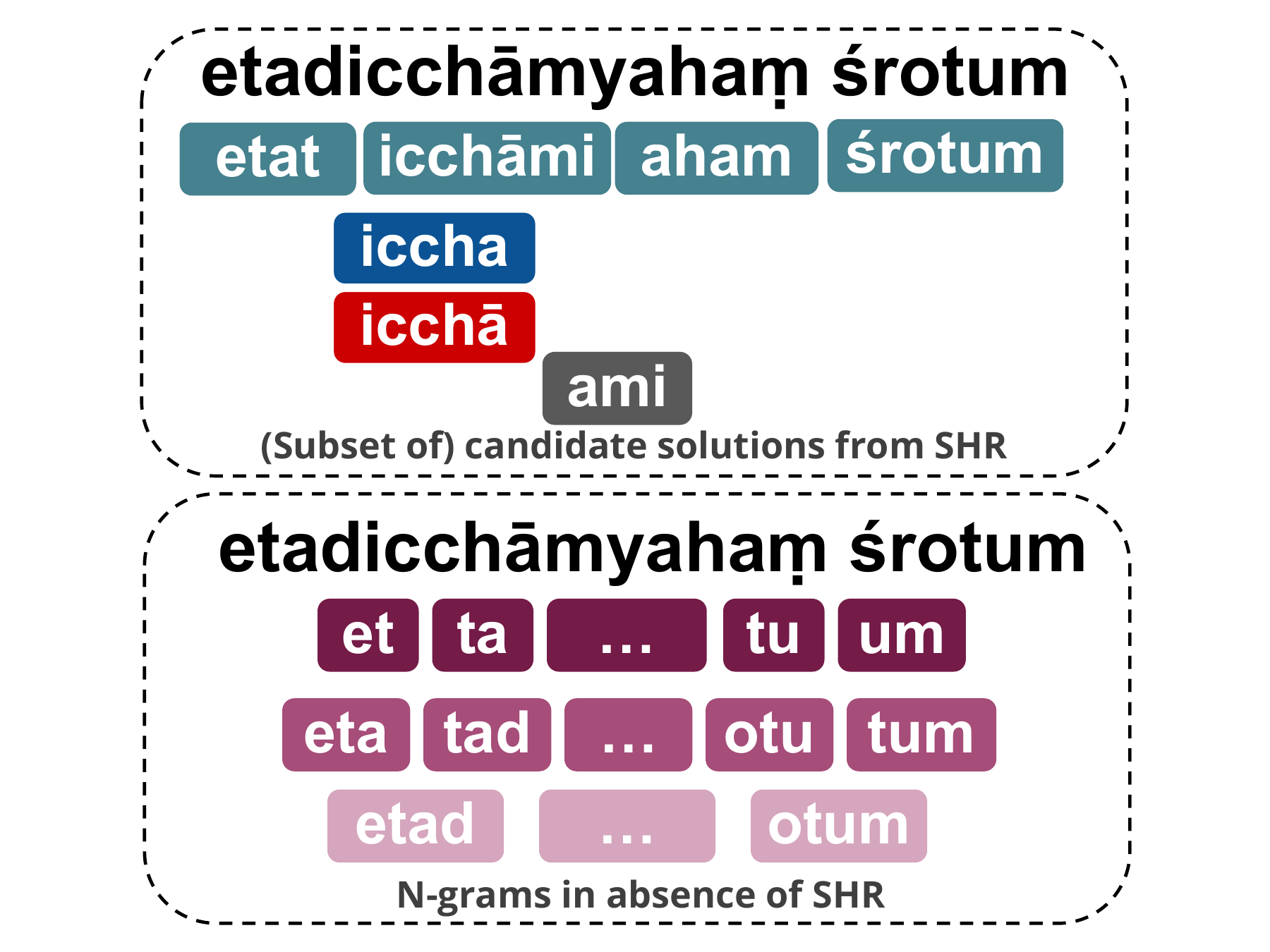}\label{fig:SHR2lattice}} 
    \subfigure[]{\includegraphics[width=0.45\textwidth]{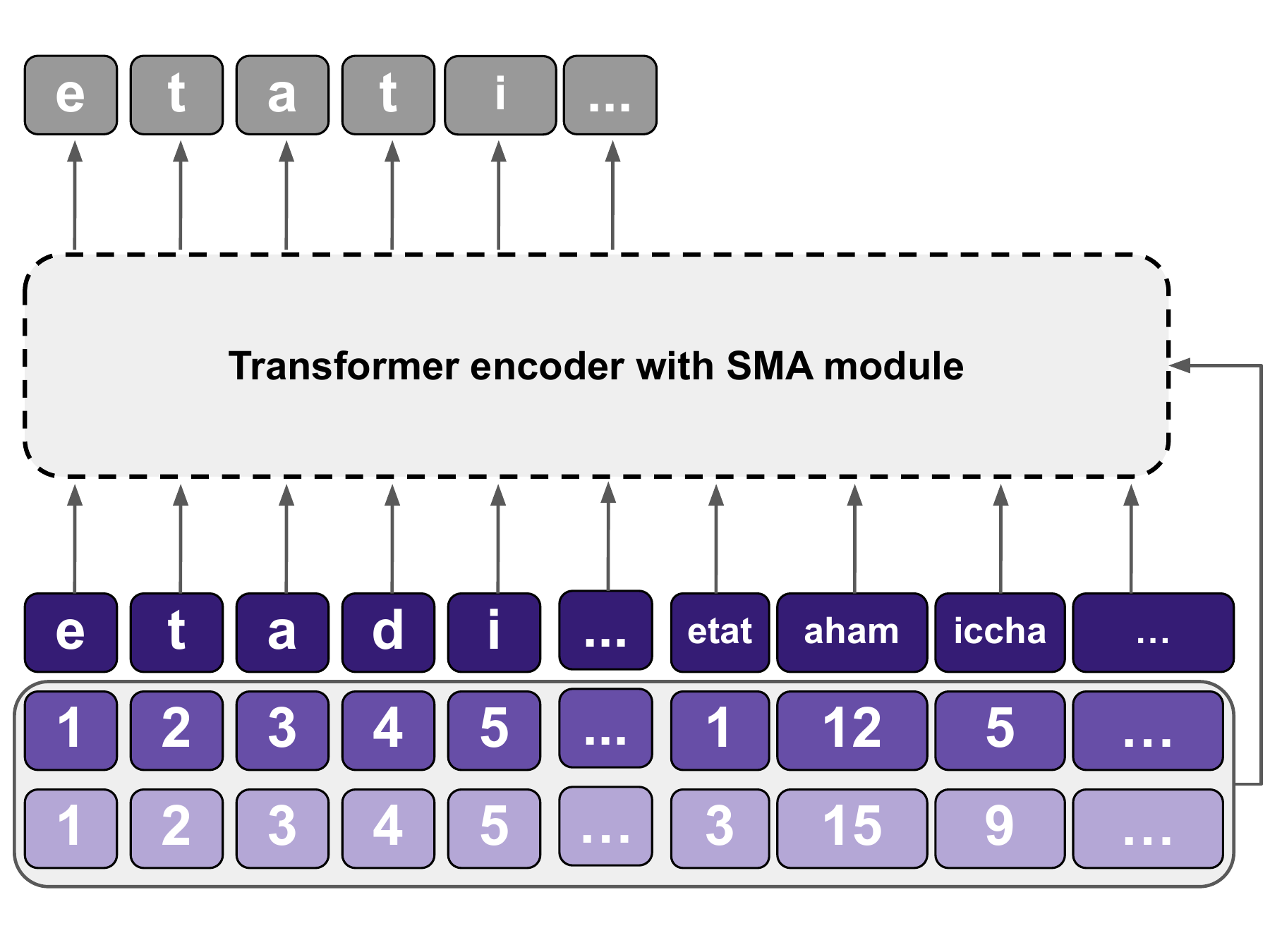}\label{fig:translist_architecture}} 
    \caption{Illustration of TransLIST with a toy example ``{\sl etadicchāmyahaṃ śrotum}''. Translation: ``I desire to hear this'' (a) LIST module: We use the candidate solutions (candidate solutions are shown, where \crule[g2]{0.25cm}{0.25cm} is the gold standard solution) from SHR if available; in the absence of SHR, we resort to using n-grams ($n \leq 4$). (b) TransLIST architecture: Span encoding is a method where each node is represented by the index positions of its head and tail characters in the input sequence. The colors \crule[dark]{0.25cm}{0.25cm}, \crule[med]{0.25cm}{0.25cm}, and \crule[light]{0.25cm}{0.25cm} denote tokens, heads, and tails, respectively. The Sanskrit Heritage Reader (SHR) is used to include words such as \textit{etat}, and \textit{ichha}, which have their boundaries modified due to the sandhi phenomenon in relation to the input sequence. Finally, the classification head is trained on top of the Transformer encoder to predict the gold standard output represented by \crule[special_red]{0.25cm}{0.25cm} for the corresponding input character nodes only.}
    \label{figure:complete_TransLIST_ch3}
\end{figure*}

\section{Methodology}
In this section, we will examine the key components of TransLIST which includes a linguistically informed tokenization module that encodes character input with latent-word information while accounting for SWS-specific \textit{sandhi} phenomena (\S~\ref{LIST}), a novel soft-masked attention to prioritise potential candidate words (\S~\ref{sma_subsection}) and a novel path ranking algorithm to correct mispredictions (\S~\ref{constrained_inference}).
\label{system_architecture}
\subsection{Linguistically Informed Sanskrit Tokenizer (LIST)}
\label{LIST}
\textit{Lexicon driven} approaches for SWS are brittle in realistic scenarios and \textit{purely engineering} based approaches do not consider the potentially useful latent word information. We propose a win-win/robust solution by formulating SWS as a character-level sequence labelling integrated with latent word information from the SHR as and when available.
TransLIST is illustrated with an example ``{\sl etadicchāmyahaṃ śrotum}'' in Figure~\ref{figure:complete_TransLIST_ch3}. SHR employs a Finite State Transducer (FST) in the form of a lexical juncture system to obtain a compact representation of candidate solution space aligned with the input sequence. As shown in Figure~\ref{fig:SHR2lattice}, we receive the candidate solution space from the SHR engine. Here, three syntactically possible splits are there. It does not suggest the final segmentation. The candidate space includes words such as \textit{etat} and \textit{ichha} whose boundaries are modified with respect to the input sequence due to \textit{sandhi} phenomenon. SHR gives us mapping (head and tail position) of all the candidate nodes with the input sequence. 
TransLIST is illustrated with an example ``{\sl etadicchāmyahaṃ śrotum}'' in Figure~\ref{figure:complete_TransLIST_ch3}. SHR employs a Finite State Transducer (FST) in the form of a lexical juncture system to obtain a compact representation of candidate solution space aligned with the input sequence. As shown in Figure~\ref{fig:SHR2lattice}, we receive the candidate solution space from the SHR engine. Here, three syntactically possible splits are there. It does not suggest the final segmentation. The candidate space includes words such as \textit{etat} and \textit{ichha} whose boundaries are modified with respect to the input sequence due to \textit{sandhi} phenomenon. SHR gives us mapping (head and tail position) of all the candidate nodes with the input sequence. 
In case such mapping is incorrect for some cases, we rectify it with the help of deterministic algorithm by matching candidate nodes with the input sentence and finding the closest match.
In the absence of SHR, we propose to use all possible n-grams ($n \leq 4$)\footnote{We do not observe significant improvements for $n > 4$.} which helps to add inductive bias about neighboring candidates in the window size of 4.\footnote{Our probing analysis (Figure~\ref{fig:sma_probing}) suggests that char-char attention mostly focuses on immediate neighbors. Refer to \S~\ref{fine_grained_analysis} for detailed ablations on LIST variants.}
We feed the candidate words/n-grams to the Transformer encoder and the classification head learns to predict gold standard output for the corresponding input character nodes only.
The output vocabulary consists of unigram characters (e.g., e, t), bigrams and tri-grams. The output vocabulary contains `\_' to represent spacing between words.
Consequently, TransLIST is capable of using both character-level modelling as well as latent word information as and when available. On the other hand, \textit{purely engineering} approaches rely only on character-level modelling  and \textit{Lexicon driven} approaches rely only on word-level information from SHR to handle \textit{sandhi}.
\subsection{Soft Masked Attention (SMA)}
\label{sma_subsection}
Recent studies have demonstrated the effectiveness of Transformers \cite{vasvani_attention} in capturing long-range dependencies within a sequence. Transformers have a self-attention mechanism that enables them to interact effectively with both the available latent word information and the character information.
There are two preliminary prerequisites for effective modelling of inductive bias for tokenization: (1) Allow interactions between the candidate words/characters within and amongst chunks. (2) Prioritize candidate words containing the input character for which a prediction is being made (e.g., in Figure~\ref{fig:translist_architecture}, \textit{etat} is prioritized amongst the candidate words when predicting for the character \textit{d}).\footnote{We find that failing to meet any one of the prerequisites leads to drop in performance (\S~\ref{fine_grained_analysis}).} 
The vanilla self-attention \cite{vasvani_attention} can address both the requirements; however, it has to self-learn the inductive bias associated with prioritisation. It may not be an effective solution in low-resourced settings. On the other hand, if we use hard-masked attention to address the second prerequisite, we lose mutual interactions between the candidates. Hence, we propose a novel soft-masked attention which helps to address both the requirements effectively. To the best of our knowledge, there is no existing soft-masked attention similar to ours. We formally discuss this below.

\textbf{Self-attention} maps a query and a set of key-value pairs to an output as discussed in \newcite{vasvani_attention}. For an input $x=(x_1, ..., x_n)$ where $x_i\in R^{d_x}$, self-attention gives an output $z=(z_1, ..., z_n)$ where $z_i\in R^{d_z}$. We presume the standard formulation of vanilla self-attention \cite{vasvani_attention} where $d_x$ is the dimension of input word representation and $d_z$ is the projection dimension. Here, $W^Q,W^K,W^V \in R^{d_x \times d_z}$ are parameter matrices. For simplicity, we ignore multi-head attention in equations \ref{eq1_SMA}, \ref{eq2_SMA} and \ref{eq3_SMA}.
\begin{align}
       z_i&= \sum_{j=1}^{n}\alpha_{ij}(x_jW^V)\label{eq1_SMA}\\
       \alpha_{ij}&= \frac{\exp{(e_{ij}})}{\sum_{k=1}^{n}\exp{(e_{ik})}}\label{eq2_SMA}\\
       e_{ij}&=\frac{(x_iW^Q)(x_jW^K)^T}{\sqrt{d_z}}\label{eq3_SMA}
\end{align}

In \textbf{soft-masked attention}, we provide a prior about interactions between candidate words and the  input characters using a span encoding ($s_{ij} \in R^{d_z}$) \cite{li-etal-2020-flat}. Intuitively, it helps inject inductive bias associated with prioritisation whilst maintaining mutual interactions between the candidates.

Formally, we modify Equation \ref{eq2_SMA} to define soft masked attention as:
\begin{align}
       \alpha^{SM}_{ij}&=\frac{M_{ij}\exp{(e_{ij}})}{\sum_{k=1}^{n}M_{ik}\exp{(e_{ik})}}
\end{align}
where $M \in R^{n \times n}$, $M_{ij}\in [0,1]$. 
$M_{ij}$ is defined as:
\begin{align}
       M_{ij}&=\frac{(x_iW^Q)(s_{ij}W^R)^T}{\sqrt{d_z}}
\end{align}
 $W^R \in R^{d_z \times d_z}$ is a learnable parameter which projects $s_{ij}$ into a location-based key vector space. Summarily, the proposed SMA module helps to prioritize potential candidate words with the help of separation, inclusion and intersection information between nodes. Finally, we calculate the output $z$ with the help of the proposed SMA as follows:
\begin{align}
       z_i&= \sum_{j=1}^{n}\alpha^{SM}_{ij}(x_jW^V)
\end{align}
Next, we discuss the span position encoding.

\textbf{Span position encoding} is one of the backbones of the proposed soft-masked module. It is utilized to capture the interactions between the candidate words and the sequence of input characters. Each span/node (which is a character/word and its corresponding position in the input sentence) is represented by the head and tail which denote the position index of the initial and final characters of the token in the input sequence, as shown in Figure~\ref{fig:translist_architecture}. The span of character is characterized by the same head and tail position index. For example, $head[i]$ and $tail[i]$ represent the head and tail index of span $x_i$, respectively. The separation, inclusion and intersection information between nodes $x_i$ and $x_j$ can be captured by  the four distance equations \ref{eq7}-\ref{eq10}.

\begin{align}    
       d_{ij}^{(hh)}&=head[i]-head[j]\label{eq7}\\
       d_{ij}^{(ht)}&=head[i]-tail[j]\label{eq8}\\
       d_{ij}^{(th)}&=tail[i]-head[j]\label{eq9}\\
       d_{ij}^{(tt)}&=tail[i]-tail[j]\label{eq10}
\end{align}
 The final span encoding is a non-linear transformation of these 4 distances:
\begin{align}
        s_{ij}&=\text{ReLU}(w_s(p_{d_{ij}^{(hh)}}\oplus p_{d_{ij}^{(ht)}}\oplus p_{d_{ij}^{(th)}}\oplus p_{d_{ij}^{(tt)}}))
\end{align}
where $w_s \in R$ is a learnable parameter, $\oplus$ is a concatenation operation and $p_d \in R^{\frac{d_z}{4}}$ is a sinusoidal position encoding similar to \newcite{vasvani_attention}.

\subsection{Path Ranking for Corrupted Predictions (PRCP)}
\label{constrained_inference}
Our investigation of errors in our system (discussed in \S~\ref{fine_grained_analysis}) revealed instances where the system predicted words that were not part of the candidate solution space. To rectify these errors, we can use the candidate solutions generated by SHR and appropriately substitute suitable candidates. We refer to the prediction corresponding to a chunk that does not fall in the candidate solution space as a ``corrupted prediction." We define a ``path" as the sequence of characters in a candidate solution for a given input. To solve this problem, we consider a ``path ranking problem" where we enumerate all possible directed paths corresponding to the input (with a corrupted prediction) and formulate a path scoring function. When designing the path scoring function, we consider the following criteria: (1) select a path consisting of semantically coherent candidate words by using an integrated judgment from two sources: a high log-likelihood (LL) score as per TransLIST to choose a semantically coherent path, and the perplexity score ($\rho$) for the path from the character-level language model to reinforce the scoring function (S), and (2) to avoid paths consisting of over-generated segmentation provided by SHR, we use a penalty proportional to the number of words ($|W|$) present in the path to prefer paths with a fewer number of words. The correct solution is highlighted with \crule[g2]{0.25cm}{0.25cm} colors in Figure~\ref{fig:SHR2lattice}.
This gives us the following path scoring function (S):
\begin{equation*}
      S = \frac{LL_{TransLIST}}{\rho_{CharLM} \times |W|}
\end{equation*}
where
\begin{align*}
 LL_{TransLIST} &= \text{log-likelihood by TransLIST}\\
 \rho_{CharLM}    &=  \text{Perplexity score by CharLM} \\ 
 |W| &= \text{Number of words present in path} \\
\end{align*}

\section{Experiments}
\label{experiments}
\textbf{Data and Metrics:}
\label{data_metrics}
%
Currently, the Digital Corpus of Sanskrit \cite[DCS]{hellwig2010dcs} contains over 600,000 lines of text that have been morphologically tagged. This corpus includes compositions in both prose and poetry, spanning a period of 3000 years and representing various writing styles. For our Sanskrit word segmentation (SWS) task, we use two benchmark datasets: \cite[SIGHUM]{krishna-etal-2017-dataset}\footnote{\url{https://zenodo.org/record/803508\#.YRdZ43UzaXJ}} and \cite[Hackathon]{hackathon_data}, both of which are subsets of DCS and come with candidate solution space generated by SHR for SWS. We chose to use SIGHUM over a larger dataset \cite{hellwig-nehrdich-2018-sanskrit} to save time and effort required for obtaining candidate solution space. We obtained the ground truth segmentation solutions from DCS. We did not use DCS10k due to missing gold standard segmentation for almost 50\% of the data points. SIGHUM contains 97,000, 3,000, and 4,200 sentences for the train, dev, and test sets, respectively, while Hackathon has 90,000, 10,332, and 9,963 sentences for the train, dev, and test sets, respectively. We evaluated our system using word-level metrics including macro-averaged Precision, Recall, F1-score, and the percentage of sentences with perfect matching.

 \textbf{Hyper-parameter settings:} For the implementation of TransLIST, we build on top of codebase by \newcite{li-etal-2020-flat}. We use the following hyper-parameters for the best configuration of TransLIST: number of epochs as
50 and a dropout rate of 0.3 with a learning rate of 0.001. We release our codebase and datasets publicly under the Apache license 2.0. All the artifacts used in this chapter are publicly available for the research purpose.
For all the systems, we do not use any pretraining. All the input representations are randomly initialized.
 We use GeForce RTX 2080, 11 GB GPU memory computing infrastructure for our experiments.

\noindent\textbf{Baselines:}
\label{baselines}
 We consider two \textit{lexicon-driven} approaches where \newcite[\textbf{SupPCRW}]{krishna-etal-2016-word} formulate SWS as an iterative query expansion problem and   \newcite[\textbf{Cliq-EBM}]{krishna-etal-2018-free} deploy a structured prediction framework.
Next, we evaluate four \textit{purely-engineering} based approaches, namely, Encoder-Decoder framework  \cite[\textbf{Seq2Seq}]{reddy-etal-2018-building},  character-level sequence labelling system with combination of recurrent and convolution element  \cite[\textbf{rcNN-SS}]{hellwig-nehrdich-2018-sanskrit}, vanilla \textbf{Transformer} \cite{vasvani_attention} and character-level Transformer with relative position encoding  \cite[\textbf{TENER}]{yan2019tener}. 
Finally, we consider lattice-structured approaches originally proposed for Chinese NER which incorporate lexical information in character-level sequence labelling architecture. 
These approaches consist of lattice-structured LSTM \cite[\textbf{Lattice-LSTM}]{zhang-yang-2018-chinese}, graph neural network (GNN) based architecture \cite[\textbf{Lattice-GNN}]{gui-etal-2019-lexicon} and Transformer based architecture \cite[\textbf{FLAT-Lattice}]{li-etal-2020-flat}.
\textbf{TransLIST:}  As per \S~\ref{LIST}, we report two variants: (a) TransLIST\textsubscript{ngrams} which makes use of only n-grams, and (b) TransLIST which makes use of SHR candidate space.
 \begin{table*}[!hbt]
\centering
\begin{adjustbox}{width=0.8\textwidth}
\small
\begin{tabular}{|c||c|c|c|c||c|c|c|c|}
\hline
\multirow{1}{*}{  } & \multicolumn{4}{c||}{\textbf{SIGHUM}} &\multicolumn{4}{c|}{\textbf{Hackathon}}\\\hline
\textbf{Model}                           & \textbf{P}     & \textbf{R}     & \textbf{F}    & \textbf{PM} & \textbf{P}     & \textbf{R}     & \textbf{F}    & \textbf{PM}   \\ \hline
Seq2seq                      & 73.44 & 73.04 & 73.24 & 29.20&72.31&72.15&72.23&20.21   \\ \hline
SupPCRW &76.30  & 79.47 & 77.85  & 38.64&-&-&-&-   \\ \hline
TENER                           & 90.03  & 89.20 & 89.61 & 61.24 & 89.38&87.33&88.35&49.92 \\ \hline
Lattice-LSTM                    & 94.36  & 93.83 & 94.09 & 76.99&91.47&89.19&90.31&65.76  \\ \hline
Lattice-GNN                     & 95.76 & 95.24 & 95.50 & 81.58&92.89&94.31&93.59&70.31  \\ \hline
Transformer                    & 96.52 & 96.21  & 96.36 & 83.88&95.79&95.23&95.51&77.70  \\ \hline
FLAT-Lattice                    & 96.75 & 96.70  & 96.72 & 85.65&\underline{96.44}&\underline{95.43}&\underline{95.93}&\underline{77.94}  \\ \hline
Cliq-EBM                      &96.18 & \underline{97.67} & \underline{96.92} & 78.83&-&-&-&-  \\ \hline 
rcNN-SS               & \underline{96.86} & 96.83  & 96.84 & \underline{87.08}&96.40&95.15&95.77&77.62   \\ \hline\hline
TransLIST\textsubscript{ngrams}  & 96.97 &96.77  &96.87  &86.52&96.68&95.74&96.21&79.28  \\ \hline
TransLIST  & \textbf{98.80} & \textbf{98.93} & \textbf{98.86} & \textbf{93.97}&\textbf{97.78}&\textbf{97.44}&\textbf{97.61}&\textbf{85.47}  \\ \hline
\end{tabular}
\end{adjustbox}
\caption{Performance evaluation between baselines in terms of P, R, F and PM metrics. The significance test between the best baselines, rcNN-ss, FLAT-lattice and TransLIST  in terms of recall/perfect-match metrics: $p < 0.05$ (as per t-test, for both the datasets).   We do not report the performance of SupPCRW and Cliq-EBM on Hackathon dataset due to unavailability of codebase.  On SIGHUM, we report numbers from their papers. The best baseline's results for the corresponding datasets are underlined. The overall best results per column are highlighted in bold.
}
\label{table:main_result}
\end{table*}
\noindent\textbf{Results:}
\label{results}
Table~\ref{table:main_result} reports the results for the best performing configurations of all the baselines on the test set of benchmark datasets for the SWS task.\footnote{We do not compare with recently proposed variant of Clique-EBM \cite{amrith21} and seq2seq baseline \cite{aralikatte-etal-2018-sanskrit} due to unavailability of codebase. Also, they do not report performance on these two datasets.}  Except \textit{purely engineering} based systems (Seq2seq, TENER, Transformer and rcNN-SS), all systems leverage linguistically refined candidate solution space generated by SHR.
Among the lattice-structured systems, FLAT-Lattice demonstrates competing performance against rcNN-SS. 
We find that rcNN-SS and FLAT-Lattice perform the best among all the baselines on SIGHUM and Hackathon datasets, respectively.

Both the TransLIST variants outperforms all the baselines in terms of all the evaluation metrics with TransLIST providing an average 1.8 points (F) and 7.2 points (PM) absolute gain with respect to the best baseline systems, rcNN-SS (on SIGHUM) and FLAT-Lattice (on Hackathon). Even when the SHR candidate space is not available, the proposed system can use TransLIST\textsubscript{ngrams}, which provides an average 0.11 points (F) and 0.39 points (PM) absolute gain over the best baselines. TransLIST\textsubscript{ngrams} gives comparable performance to rcNN-SS on SIGHUM dataset, while on the Hackathon dataset, it performs significantly better than FLAT-Lattice ($p<0.05$ as per t-test). The wide performance gap between TransLIST and TransLIST\textsubscript{ngrams} demonstrates the effectiveness of using SHR candidate space, when available. 
Summarily, we establish a new state-of-the-art results with the help of meticulously stitched LIST, SMA and PRCP modules. The knowledge of the candidate space by SHR gives an extra advantage to TransLIST. Otherwise, natural choice is the proposed purely engineering variant TransLIST\textsubscript{ngrams} when that is not available.  

\section{Analysis}
\label{fine_grained_analysis}
In this section, we investigate various questions to dive deeper into the proposed components and investigate the capabilities of various modules. We use SIGHUM dataset for the analysis.

\begin{figure}[h]
\centering
  \subfigure[]{\includegraphics[width=0.35\textwidth]{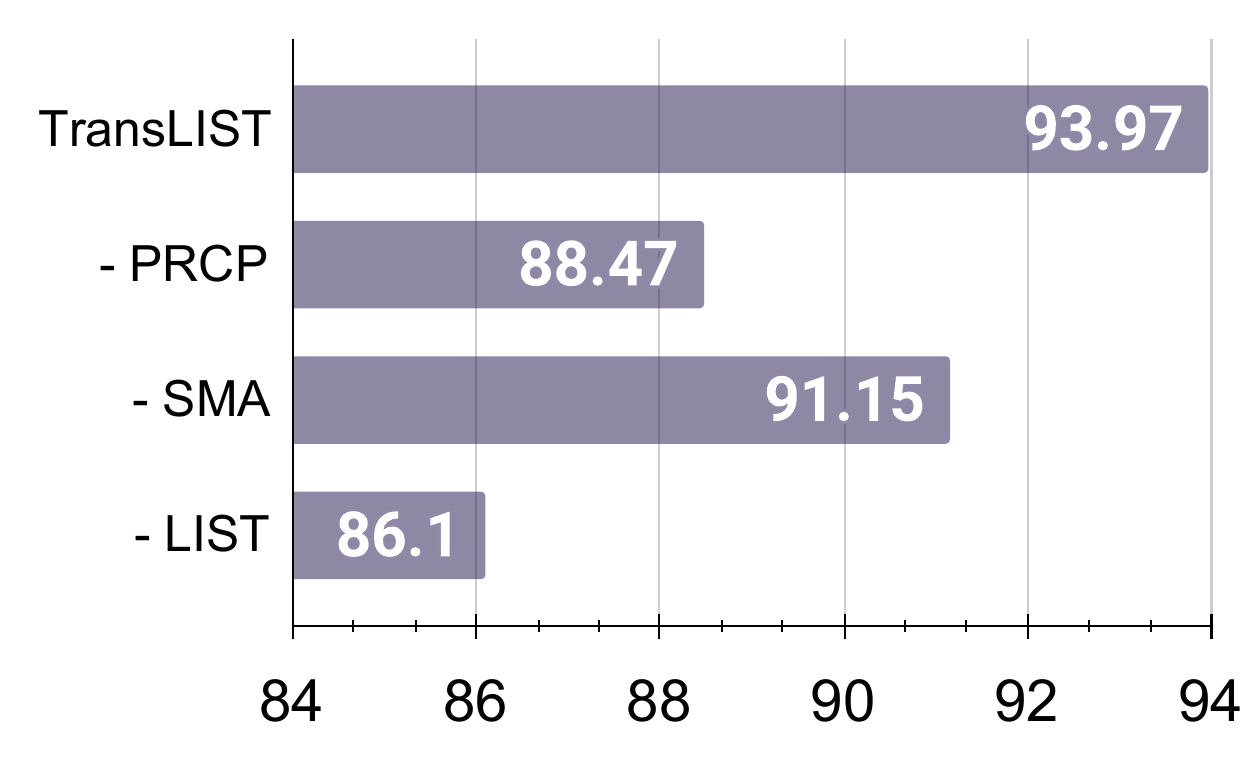}\label{fig:abl}} 
    \subfigure[]{\includegraphics[width=0.35\textwidth]{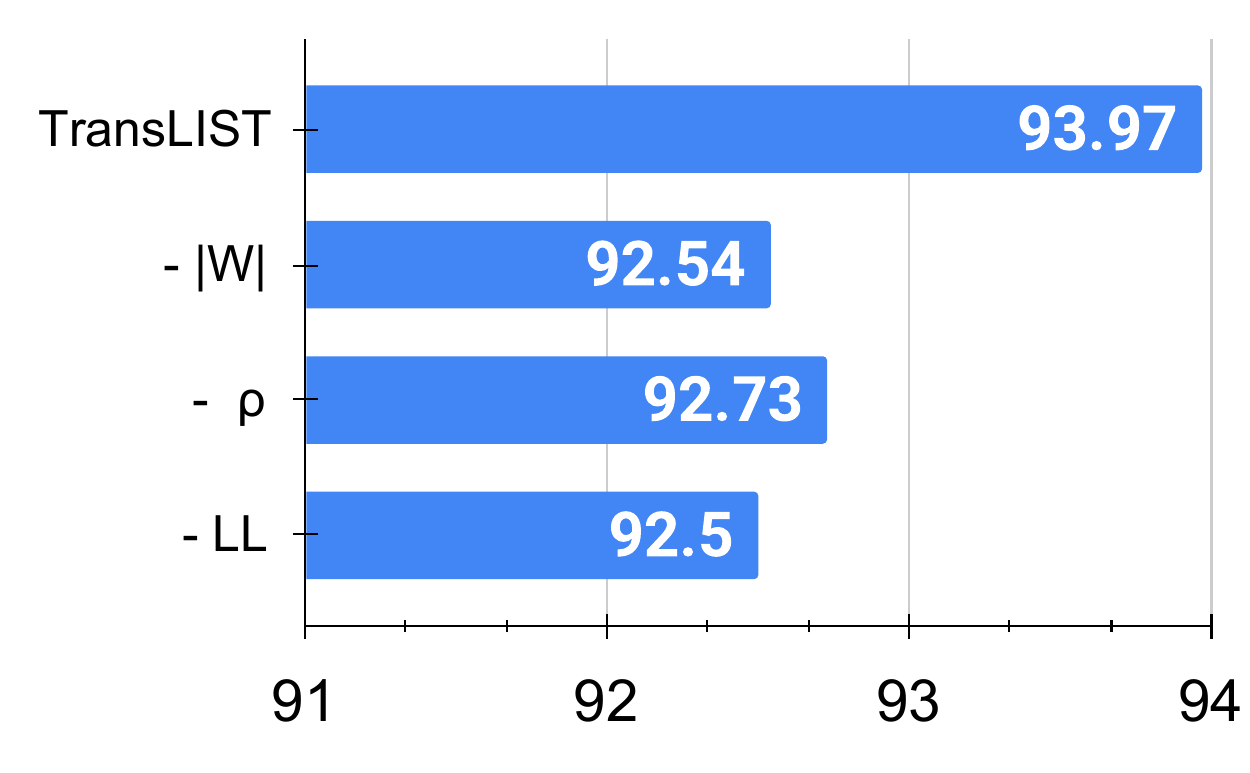}\label{fig:ci}} 
\caption{Ablations on (a) TransLIST (b) PRCP module in terms of PM (SIGHUM-test). Each ablation in (a) removes a single module from TransLIST. For example, ``-SMA'' removes SMA from TransLIST. For (b), ablations are shown by removing a particular term from path scoring function ($S$).} 
\label{fig:ablation} 
\end{figure}

\noindent\textbf{(1) Ablation analysis:} Here, we study the contribution of different modules towards the final performance of TransLIST. Figure~\ref{fig:abl} illustrates ablations in terms of PM when a specific module is removed from TransLIST. For instance, `-LIST' corresponds to character-level transformer encoder with SMA and PRCP. Removal of any of the modules degrades the performance. Figure~\ref{fig:abl} shows that LIST module is the most crucial for providing inductive bias of tokenization. Also, removal of `PRCP' module has a large impact on the performance. We observe that the PRCP module gets activated for 276 data points out of 4,200 data points in the test set.   We then deep dive into the PRCP path scoring function in Figure~\ref{fig:ci},  which consists of 3 terms, namely, penalty ($|W|$), perplexity score by CharLM ($|\rho|$) and log-likelihood ($LL$) by TransLIST, respectively. We remove a single term at a time from the path scoring function, and observe each of the terms used in the scoring function plays a major role in the final performance.

\noindent\textbf{(2) Comparative analysis of potential LIST module variants to add inductive bias for tokenization:} 
We evaluate possible LIST variants which can help inject inductive bias for tokenization via auxiliary (word) nodes illustrated in Figure~\ref{fig:translist_architecture}: (a) \textit{sandhi} rules: We use \textit{sandhi} rules as a proxy to indicate potential modifications at specific position in the input sequence. For example, if input chunk contains the character \textit{`o'} then it can be substituted with two possibilities \textit{\={o} $\rightarrow$ a-\={u}/a\d{h}}. We provide this proxy information through auxiliary nodes. (b) Sanskrit vocab: We obtain a list of vocabulary words from DCS corpus \cite{hellwig2010dcs} and add the words which can be mapped to the input character sequence using  a string matching algorithm. (c) n-grams: This is TransLIST\textsubscript{ngrams}  (d) SHR: We follow the exact settings as described in \S~\ref{LIST} except that we do not use the PRCP component. In Table~\ref{table:LIST_module}, we compare these with the \textit{purely engineering} variant of TransLIST (Base system: only character-level Transformer) where no inductive bias for tokenization is injected. Clearly, due to availability of enriched candidate space, SHR variant outperforms all its peers. However, competing performance of n-gram variant is appealing because it completely obliviates the dependency on SHR and remains unaffected in the absence of SHR's candidate space. 
\begin{table}[ht]
\centering
\begin{adjustbox}{width=0.6\textwidth}
\small
\begin{tabular}{|c|c|c|c|c|}
\hline
\textbf{System}                           & \textbf{P}     & \textbf{R}     & \textbf{F}  & \textbf{PM}    \\ \hline
Base system   & 92.75  & 92.62  &92.69 & 72.33 \\\hline
+\textit{sandhi} rules & 93.53 & 93.70 & 93.62 & 75.71  \\\hline
+Sanskrit Vocab & 96.75 & 96.70 & 96.72 & 85.65 \\\hline
+n-grams  &96.97 &96.77  &96.87  &86.52 \\\hline
+SHR & \textbf{97.79} & \textbf{97.45} & \textbf{97.62}& \textbf{88.47} \\ \hline
\end{tabular}
\end{adjustbox}
\caption{The comparison (on SIGHUM-test set) in between LIST variants. `+' indicates system where the corresponding variant is augmented with the base system. We do not activate PRCP for any of these systems.} 
\label{table:LIST_module}
\end{table}
\begin{figure}[!htb]
\centering
\includegraphics[width=0.9\textwidth]{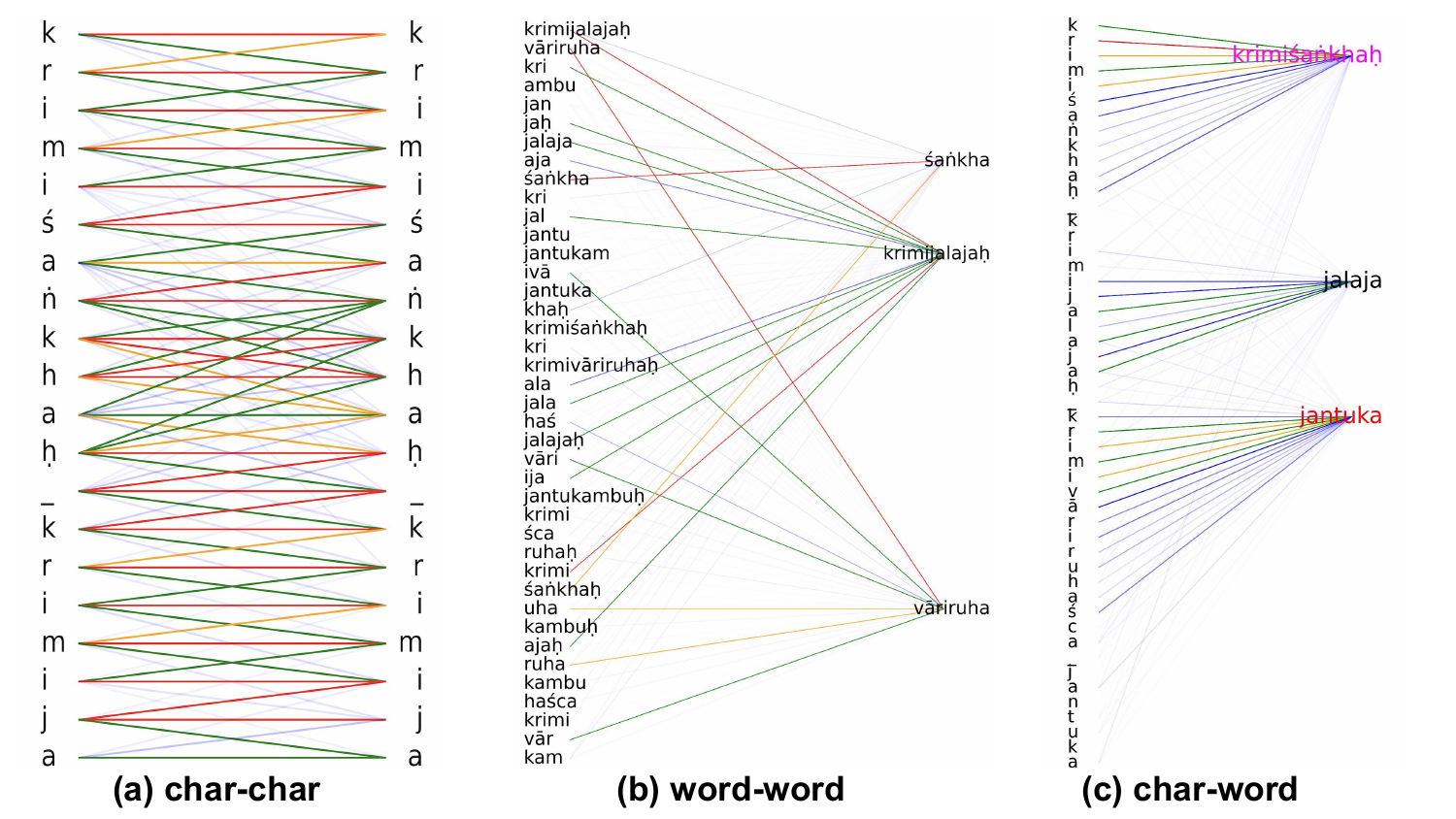}
\caption{SMA probing: Illustration of char-char, char-word and word-word interactions. The strength of the SMA decreases in the following order: red, orange, green and blue. Char-char attention mostly focuses on characters present in the vicinity of window size 1. Word-word interactions are able to capture whether a word is subword of another word or not. Finally, we find that quality of attention goes down for char-word as we move as per the following order: in vocabulary gold words (pink), in vocabulary non-golds (black) and out-of-vocabulary words (red). Some of the attentions are invisible due to very low attention score.} 
\label{fig:sma_probing} 
\end{figure}

\noindent\textbf{(3) Probing analysis on SMA:} Here we analyze whether SMA upholds the prerequisite for effective modelling of inductive bias, i.e., prioritize candidate words which contain the input character for which the prediction is being made. Figure~\ref{fig:sma_probing} illustrates three types of interactions, namely, char-char, char-word and word-word. We use color coding scheme to indicate the strength of attention weight. The attention weight decreases in the following order: Red, Orange, Green and Blue. Char-char attention mostly focuses on characters present in the vicinity of window size 1. This local information is relevant to make decisions regarding possible {\sl sandhi} split. 
Word-word interactions are able to capture whether a word is subword of another word or not. Finally, for char-word attention, we find that quality of attention goes down as we move as per the following order: in vocabulary gold words (pink), in vocabulary non-golds (black) and out-of-vocabulary (unseen in training but recognized by SHR) gold words (red). While the drop in attention from in-vocabulary gold tokens to out-of-vocabulary gold tokens is expected, the drop in attention from gold tokens to non-gold tokens is desired. 
Thus, this probing analysis suggests that SMA module helps to improve intra/inter interactions between character/words and this substantiates the need of SMA module in TransLIST.

\noindent\textbf{(4) How does TransLIST perform in a non-trivial situation where multiple \textit{sandhi} rules are applicable?}
In Table~\ref{table:sandhi_special}, we report the comparison  with rcNN-SS for a critical scenario of a \textit{sandhi} phenomenon. Table~\ref{table:sandhi_special} represents the possible \textit{sandhi} rules that generate the surface character \textit{\={a}}.  
Following \newcite{design_goyal}, the sandhi rewrite rules are formalized as $u|v \rightarrow f/x_{--}$    \cite{kaplan-kay-1994-regular} where $x, v, f \in \Sigma$ , and $u \in \Sigma^{+}$. $\Sigma$ is the collection of phonemes, $\Sigma^{*}$: a set of all possible strings over $\Sigma$, and $\Sigma^{+}$ = $\Sigma^{*} - \epsilon$.
For example, the potential outputs for the input ā can be \={a}, \={a}-\={a}, \={a}-a, a-a and a\d{h}. The correct rule can be decided based on the context.
The presence of numerous rules creates difficulty in determining which rule is applicable for a specific situation. Therefore, it is important to evaluate TransLIST's ability to generalize semantically by comparing it to the current state-of-the-art system. Our observations indicate that TransLIST consistently performs better than rcNN-SS across all metrics.
\footnote{Follwing  \citet{hellwig-nehrdich-2018-sanskrit}, we report character-level F-score metric. $P = \frac{|S_{g} \cap S_{p}|}{|S_{p}|}$ ; $R = \frac{|S_{g} \cap S_{p}|}{|S_{g}|}$,  $F1 =\frac{2PR}{P+R}$,  ($S_{g}$) : Set of locations where the rule occurs in gold output, ($S_{p}$) : Set of locations where the rule is predicted.} 
Table~\ref{table:sandhi_special} lists the \textit{sandhi} rules in descending order of their frequency. It is worth noting that TransLIST outperforms the current state-of-the-art system significantly, especially for infrequent \textit{sandhi} rules. This finding reinforces the notion that TransLIST has superior performance compared to the current state-of-the-art system.
\begin{table}[h]
\centering
\begin{adjustbox}{width=0.8\textwidth}
\small
\begin{tabular}{|c|c|c|c||c|c|c||c|c|c|}
\hline
\multirow{1}{*}{  } & \multicolumn{3}{c||}{\textbf{rcNN-SS}} &\multicolumn{3}{c|}{\textbf{TransLIST\textsubscript{ngrams}}}&\multicolumn{3}{c|}{\textbf{TransLIST}}\\\hline
\textbf{Rules}                           & \textbf{P}     & \textbf{R}     & \textbf{F}    & \textbf{P}     & \textbf{R}     & \textbf{F} & \textbf{P}     & \textbf{R}     & \textbf{F} \\ \hline
\={a}                        & 99.3 & 99.3 & 99.4 & 99.2 & 99.3 & 99.2 & 99.7 & 99.6 & 99.6 \\
a-a                                           & 95.4 & 96.6 & 96.0   & 95.3 & 96.8 & 96.0   & 96.6 & 97.8 & 97.2 \\
\={a}-a                     & 88.4 & 83.1 & 86.5 & 87.3 & 82.4 & 84.8 & 90.5 & 83.8 & 87.0   \\
\={a}\d{h} & 76.7 & 70.1 & 73.7 & 75.3 & 72.6 & 73.9 & 77.2 & 80.1 & 78.0   \\
\={a}-\={a}& 50.1 & 42.1 & 45.7 & 48.1 & 40.1 & 43.7 & 80.0   & 70.9 & 70.4\\ \hline
\end{tabular} 
\end{adjustbox}
\caption{The comparison (on SIGHUM-test set) in terms of P, R and F metrics between rcNN-SS and the TransLIST for ambiguous \textit{sandhi} rules leading to the same surface character \textit{\={a}}. The proposed model consistently outperforms rcNN-SS in all the metrics.} 
\label{table:sandhi_special}
\end{table}

\begin{figure}[h]
    \centering
    \includegraphics[width=0.5\textwidth]{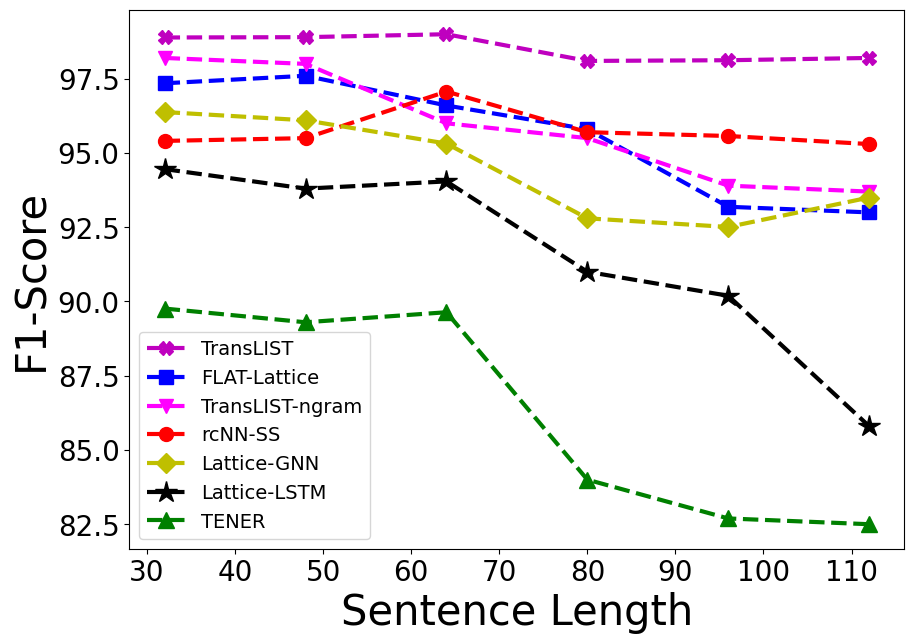}
    \caption{F1-score against sentence length (no. of characters) over the SIGHUM dataset}
    \label{fig:error_analysis}
\end{figure}

\noindent\textbf{(5) How robust is the system when sentence length is varied?}
 In Figure~\ref{fig:error_analysis}, we analyze the performance of the baselines with different sentence lengths. 
We plot the F1-score against sentence length. Clearly, while all the systems show superior performance for shorter sentences, TransLIST is much more robust for longer sentences compared to other baselines. The lattice-structured baselines give competing F1-scores over short sentences but relatively sub-par performance over long sentences.

\noindent\textbf{(6) Illustration of PRCP with an example:} 
Table~\ref{table:case_study} illustrates  an example that probes the effectiveness of PRCP in TransLIST. We compare TransLIST with rcNN-SS and observe that TransLIST also predicts words out of candidate solution space when PRCP module is not activated. However, the degree of such mistakes in TransLIST is comparatively less due to effective modelling of inductive bias for tokenization using LIST and SMA modules. In Table~\ref{table:case_study}, rcNN-SS predicts three words which are not part of candidate space, namely, \textit{vāmbike, yak\d{s}avapu\d{h}} and \textit{caka}. These are mistakes that can be rectified with the help of available candidate space.
Interestingly, TransLIST commits only a single mistake in this category by predicting out of solution space word \textit{aambike}. PRCP aids in mitigating such mistake by appropriately substituting suitable candidates.
 \begin{table*}[t]
\centering
\begin{adjustbox}{width=0.9\textwidth}
\small
\begin{tabular}{|c|c|c|}
\hline
&\textbf{Sentence}& \textbf{F-score}\\\hline
\textbf{Input sentence}& kimetadīśe bahuśobhamāne vā\d{m}bike yak\d{s}avapuścakāsti& -\\&\textit{Translation:} What is this body resembling a \textit{Yaksha} that glows, &\\
&oh Ambika! You who lord over! You who shine!& \\\hline 
\textbf{Correct segmentation}&kim etat īśe bahu śobhamāne vā ambike yak\d{s}a vapu\d{h} cakāsti &- \\ \hline
\textbf{SHR candidate space}& kim, etat, īśe, bahu, śobhamāne, śobham, āne, śobha, māne, &- \\
&mā, vā, ambike, yak\d{s}a, vapu\d{h}, cakāsti, ca, kā, asti&\\
&\textit{Word-word meaning:} what, this, the one who lord, very much,&\\
& the one who shine, bright, mouth, I respect, never, or, Parvati, &\\
&a kind of celestial being, body, glows, and, who (female), is there (be).&\\  \hline \hline
\textbf{rcNN-SS} &kim etat īśe bahu śobhamāne \textbf{vāmbike yak\d{s}avapu\d{h} caka asti}  & 52.60 \\\hline 
\textbf{TransLIST-PRCP} &kim etat īśe bahu śobhamāne vā \textbf{aambike} yak\d{s}a vapu\d{h} cakāsti  & 90.00\\\hline 
\textbf{TransLIST} &kim etat īśe bahu śobhamāne vā ambike yak\d{s}a vapu\d{h} cakāsti  & 100.00 \\\hline
\end{tabular}
\end{adjustbox}
\caption{An example to illustrate the effectiveness of PRCP module of TransLIST. Bold represents incorrect segmentation for the input sequence.} 
\label{table:case_study}
\end{table*}

 \textbf{(7) Average run times:} Table~\ref{table:run_time} shows the average training time in hours and inference time per sentence in milliseconds for all competing baselines. We find that pure engineering-based techniques (TENER, rcNN-SS) outperform lattice-structured architectures (Lattice-LSTM, Lattice-GNN, FLAT-Lattice) in terms of run time. When the inference times of TransLIST and TransLIST\textsubscript{ngrams} are compared, TransLIST takes longer owing to the PRCP module. It would be interesting to explore approaches to optimise the inference time of the PRCP module.
 
\begin{table}[h]
\centering
\resizebox{0.5\textwidth}{!}{
\begin{tabular}{|c|c|c|}
\hline
\textbf{System}                            & \textbf{Train (Hours)}  & \textbf{Test (ms)}  \\ \hline

TENER                           & 4 H & 7 ms\\ \hline
Lattice-LSTM                    & 16 H  & 110 ms\\ \hline
Lattice-GNN                     & 64 H & 95 ms \\ \hline
FLAT-Lattice                    & 5 H & 14 ms\\ \hline
rcNN-SS               & 4 H & 5 ms\\ \hline
Cliq-EBM                      &10.5 H & 750 ms \\ 
 \hline 
 TransLIST\textsubscript{ngrams}  & 8 H & 14ms \\ \hline
TransLIST  & 8 H & 105 ms\\ \hline
\end{tabular}}
\caption{Average training time (in hours) and inference time per sentence (in milliseconds)  for all the competing baselines.}
\label{table:run_time}
\end{table}

\section{Summary}
This chapter focuses on Sanskrit word segmentation and aims to address the shortcomings of existing approaches by combining the strengths of both engineering and lexicon-driven approaches. TransLIST is introduced as a solution that uses the LIST module to induce an inductive bias for tokenization and the SMA module to prioritize relevant candidate words. The PRCP module is also proposed to rectify corrupted predictions using linguistic resources. Experimental results show that TransLIST outperforms the best baselines by an average of 7.2 points in terms of perfect matching percentage. Additionally, a detailed analysis of TransLIST's performance is provided. 

SHR's lexicon database is occasionally updated, which may improve its performance in certain cases. However, TransLIST remains unaffected by these updates, as it does not rely on completeness assumption on the candidate solution space and is robust to partially available SHR’s candidate space. As a result, no retraining of TransLIST is needed for SHR updates. Remarkably, these updates might assist our PRCP inference module in enhancing post-correction by leveraging the improved version of SHR module.


%% file: Chapters/Chapter4.tex

\chapter{Low-resource Dependency parsing}

\label{Chapter4}

\lhead{Chapter 4. \emph{Dependency Parsing}}
This chapter focuses on dependency parsing for morphological rich languages (MRLs) in a low-resource setting. Although morphological information is essential for the dependency parsing task, the morphological disambiguation and lack of powerful analyzers pose challenges to get this information for MRLs. To address these challenges, we propose simple auxiliary tasks for pretraining.
We perform experiments on 10 MRLs in low-resource settings to measure the efficacy of our proposed pretraining method and observe an average absolute gain of 2 points (UAS) and 3.6 points (LAS).

Several strategies are tailored to enhance performance in low-resource scenarios.
 While these are well-known to the community, it is not trivial to select the best-performing combination of these strategies for a low-resource language that we are interested in, and not much attention has been given to measuring the efficacy of these strategies.
 We experiment with 5 low-resource strategies for our ensembled approach on 7 Universal Dependency (UD) low-resource languages.
Our exhaustive experimentation on these languages supports the effective improvements for languages not covered in pretrained models.
We show a successful application of the ensembled system on Sanskrit.

\section{Pretraining Tailored for Low-Resource Dependency Parsing}
\label{intro}

Dependency parsing has greatly benefited from neural network-based approaches. While these approaches simplify the parsing architecture and eliminate the need for hand-crafted feature engineering \cite{chen-manning-2014-fast,dyer-etal-2015-transition,kiperwasser-goldberg-2016-simple,DBLP:conf/iclr/DozatM17,kulmizev-etal-2019-deep}, their performance has been less exciting for several morphologically rich languages (MRLs) and low-resource languages \cite{more-etal-2019-joint,seeker-cetinoglu-2015-graph}. In fact, the need for large labeled treebanks for such systems has adversely affected the development of parsing solutions for low-resource languages~\cite{vania-etal-2019-systematic}. 
\newcite{zeman-etal-2018-conll} observe that data-driven parsing on 9 low resource treebanks resulted not only in low scores but those outputs ``are hardly useful for downstream applications".

Several approaches have been suggested for improving the parsing performance of low-resource languages. This includes data augmentation strategies,  cross-lingual transfer \cite{vania-etal-2019-systematic} and using unlabelled data with semi-supervised learning \cite{clark-etal-2018-semi} and self-training \cite{rotman2019deep}. Further, incorporating morphological knowledge substantially improves the parsing performance for MRLs, including low-resource languages ~\cite{vania-etal-2018-character,dehouck-denis-2018-framework}. This aligns well with the linguistic intuition of the role of morphological markers, especially that of case markers, in deciding the syntactic roles for the words involved \cite{wunderlich2001interaction,sigurdhsson2003case,kittila2011introduction}. However, obtaining the morphological tags for input sentences during run time is a challenge in itself for MRLs \cite{more-etal-2019-joint} and use of predicted tags from taggers, if available, often hampers the performance of these parsers.  In this chapter, we primarily focus on one such morphologically-rich low-resource language, Sanskrit. 

We propose a simple pretraining approach, where we incorporate encoders from simple auxiliary tasks by means of a gating mechanism \cite{sato-etal-2017-adversarial}. This approach outperforms multi-task training and transfer learning methods under the same low-resource data conditions ($\sim$500 sentences). The proposed approach when applied to \newcite{dozat2017stanford}, a neural parser, not only obviates the need for providing morphological tags as input at runtime, but also outperforms its original configuration that uses gold morphological tags as input. Further, our method performs close to DCST \cite{rotman2019deep}, a self-training based extension of \newcite{dozat2017stanford}, which uses gold morphological tags as input for training. 

To measure the efficacy of the proposed method, we further perform a series of experiments on 10 MRLs in low-resource settings and show 2 points and 3.6 points average absolute gain (\S~\ref{results}) in terms of UAS and LAS, respectively. Our proposed method also outperforms multilingual BERT \cite[mBERT]{devlin-etal-2019-bert} based multi-task learning model \cite[Udify]{kondratyuk-straka-2019-75} for the languages which are not covered in mBERT (\S~\ref{Comparison with mBERT Pretraining}).

\subsection{Methodology}
\label{proposed_model}
\begin{figure*}[!tbh]
\centering
\subfigure[\label{fig:tagging_schem_SRW}]{\includegraphics[width=0.45\linewidth]{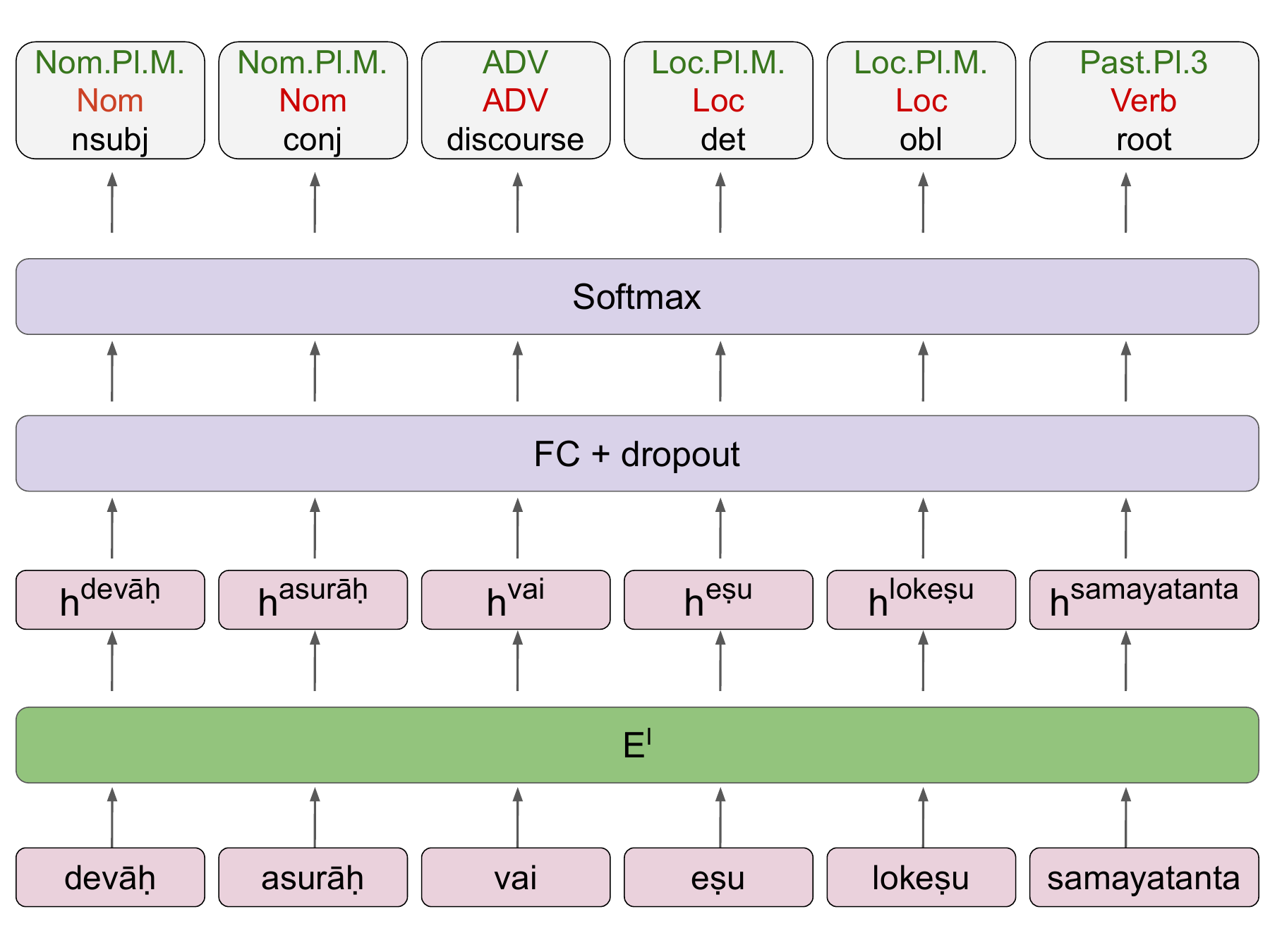}}
\subfigure[\label{fig:gating_SRW}]{\includegraphics[width=0.45\linewidth]{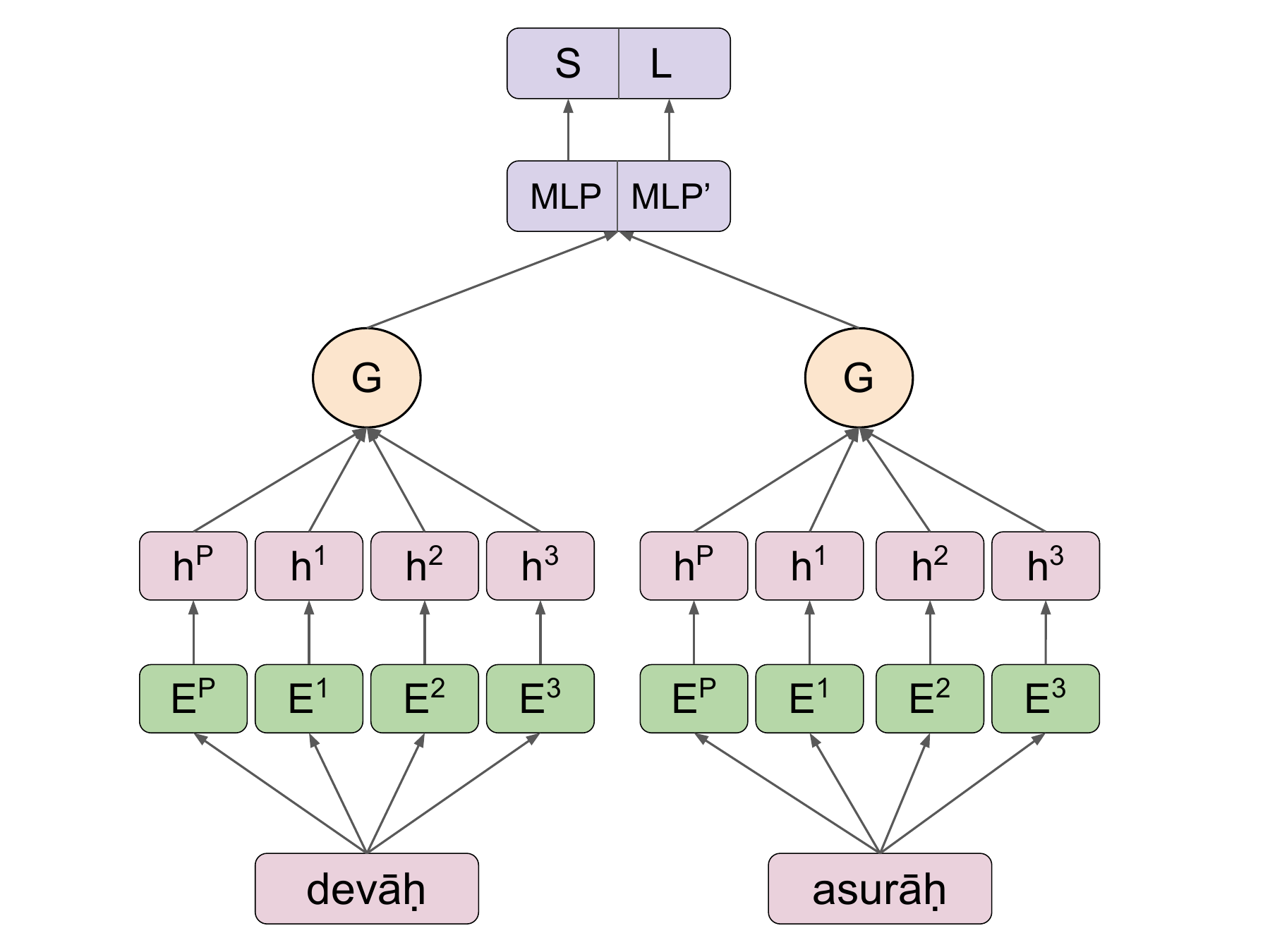}}
\caption{Illustration of proposed architecture for a Sanskrit sequence. English translation: ``Demigods and demons had tried with equal effort for these planets". (a) Pretraining step: For an input word sequence, tagger predicts labels as per three proposed auxiliary tasks, namely, Morphological Tag (green), Case Tag (red) and Label Tag (black). (b) Parser with gating: $E^{(P)}$ is encoder of a neural parser like~\newcite{DBLP:conf/iclr/DozatM17} and $E^{(1)-(3)}$ are the encoders pre-trained with proposed auxiliary tasks. 
Gating mechanism combines representations of all the encoders which,
for each word pair, is passed to two MLPs to predict the probability of arc score (S) and label (L).
}
\label{fig:model_SRW}
\end{figure*}
Our proposed pretraining approach
essentially attempts to combine word representations from encoders trained on multiple sequence level supervised tasks, as auxiliary tasks, with that of the default encoder of the neural dependency parser. While our approach is generic and can be used with any neural parser,  we use   BiAFFINE parser~\cite{DBLP:conf/iclr/DozatM17}, hence forth referred to as BiAFF, in our experiments.This is a graph-based neural parser that makes use of biaffine attention and a biaffine classifier.
Figure~\ref{fig:model_SRW} illustrates the proposed approach using an example sequence from Sanskrit. Our pipeline-based approach consists of two steps: (1) Pretraining step (2) Integration step. Figure~\ref{fig:tagging_schem_SRW} describes the pretraining step with three auxiliary tasks to pretrain the corresponding encoders $E^{(1)-(3)}$. Finally, in the integration step, these pretrained encoders along with the encoder for the BiAFF model $E^{(P)}$ are then combined using a gating mechanism (\ref{fig:gating_SRW}) as employed in ~\newcite{sato-etal-2017-adversarial}. \footnote{Our proposed approach is inspired from \newcite{rotman2019deep}.}

All the auxiliary tasks are trained independently as separate models, but using the same architecture and hyperparameter settings which differ only in terms of the output label they use. The models for the pretraining components are trained using BiLSTM encoders, similar to the encoders in \newcite{DBLP:conf/iclr/DozatM17} and then decoded using two fully connected layers, followed by a softmax layer~\cite{huang2015bidirectional}. These sequential tasks involve prediction of the morphological tag \textbf{(MT)}, dependency label (relation) that each word holds with its head (\textbf{LT}) and further we also consider task where the case information of each nominal forms the output label \textbf{(CT)}. Other grammatical categories did not show significant improvements over the case (\S~\ref{Experiments on additional auxiliary}). This aligns well with the linguistic paradigm that the case information plays an important role in deciding the syntactic role that a nominal can be assigned in the sentence. For words with no case-information, we predict their coarse POS tags. Here, the morphological information is automatically leveraged using the
pre-trained encoders, and thus during runtime the morphological tags need not be provided as inputs. It also helps in reducing the gap between UAS and LAS (\S~\ref{results}).

\subsection{Experiments}
\label{experiments}
\noindent\textbf{Data and Metric:} We use 500, 1,000 and 1,000 sentences from the Sanskrit Treebank Corpus \cite[STBC]{kulkarni2010designing} as the training, dev and test data respectively for all the models. For the proposed auxiliary tasks, all the sequence taggers are trained with additional previously unused 1,000 sentences from STBC along with the training sentences used for the dependency parsing task. For the Label Tag (LT) prediction auxiliary task, we do not use gold dependency information; rather we use predicted tags from BiAFF parser. For the remaining auxiliary tasks, we use gold standard morphological information.

  For all the models, input representation consists of FastText~\cite{grave-etal-2018-learning}\footnote{\url{https://fasttext.cc/docs/en/crawl-vectors.html}} embedding of 300-dimension and convolutional neural network (CNN) based 100-dimensional character embedding~\cite{NIPS2015_5782}. For character level CNN architecture, we use following setting: 100 number of filters with kernel size equal to 3.
We use standard Unlabelled and Labelled Attachment Scores
(UAS, LAS) to measure the parsing performance and use t-test for statistical significance  ~\cite{dror-etal-2018-hitchhikers}.


 For STBC treebank, the original data does not have morphological tag entry, so the Sanskrit Heritage reader~\cite{huet2013design,goyal2016design} is used to obtain all the possible morphological analysis and only those sentences are chosen which do not have any word showing homonymy or syncretism~\cite{amrith21}.
For other MRLs, we restrict to the same training setup as Sanskrit and use 500 annotated sentences as labeled data for training. Additionally, we use 1000 sentences with morphological information as unlabelled data for pretraining sequence taggers.\footnote{The predicted relations on unlabelled data by the model trained with 500 samples are used for Label Tagging task.} We use all the sentences present in original development and test split data for development and test data. For languages where multiple treebanks are available, we chose only one available treebank to avoid domain shift. Note that STBC adopts a tagging scheme based on the grammatical tradition of Sanskrit, specifically based on K\={a}raka \cite{kulkarni-sharma-2019-paninian,kulkarni2010designing}, while the other MRLs including Sanskrit-Vedic use UD.

 \noindent\textbf{Hyper-parameters:} 
 We utilize the BiAFFINE parser (BiAFF) implemented by~\newcite{ma-etal-2018-stack}. We employ the following hyper-parameter setting for pretraining sequence taggers and base parser BiAFF: the batch size of 16,  number of epochs as 100, and a dropout rate of 0.33 with a learning rate equal to 0.002. The hidden representation generated from n-Stacked-LSTM layers of size 1,024 is passed through two fully connected layers of size 128 and 64. Note that LCM and MTL models use 2-Stacked LSTMs. We keep all the remaining parameters the same as that of~\newcite{ma-etal-2018-stack}.
 
 For all TranSeq variants, one BiLSTM layer is added on top of three augmented pretrained layers from an off-the-shelf morphological tagger~\cite{ashim2020evaluate} to learn task-specific features. In TranSeq-FEA, the dimension of the non-linearity layer of the adaptor module is 256, and in TranSeq-UF, after every 20 epochs, one layer is unfrozen from top to down fashion. In TranSeq-DL, the learning rate is decreased from top to down by a factor of 1.2.
 We have used default parameters to train Hierarchical Tagger~\footnote{\url{https://github.com/ashim95/sanskrit-morphological-taggers}} and baseline models.

\noindent\textbf{Models:} All our experiments are performed as augmentations on two off the shelf neural parsers,  BiAFF~\cite{DBLP:conf/iclr/DozatM17}
and Deep Contextualized Self-training (DCST), which integrates self-training with BiAFF~\cite{rotman2019deep}. Hence their default configurations become the baseline models \textbf{(Base)}.
We also use a system that simultaneously trains the BiAFF (and DCST) model for dependency parsing along with the sequence level case prediction task in a multi task setting \textbf{(MTL)}. For MTL model, we also experiment with morphological tagging, as an auxiliary task. However, we do not find significant improvement in performance compared to case tagging.  Hence, we consider case tagging as an auxiliary task to avoid sparsity issue due to the monolithic tag scheme for morphological tagging.
As a transfer learning variant (\textbf{TranSeq}), we extract first three layers from a hierarchical multi-task morphological tagger~\cite{ashim2020evaluate}, trained on 50k examples from DCS~\cite{hellwig2010dcs}. Here each layer corresponds to different grammatical categories, namely, number, gender and case. Note that number of randomly initialised encoder layers in BiAFF (and DCST) are now reduced from 3 to 1. We fine-tune these layers with default learning rate and experiment with four different fine-tuning schedules. Finally, our proposed configuration (in \S \ref{proposed_model}) is referred to as the \textbf{LCM} model.\footnote{LCM denotes Label, Case and Morph tagging schemes.} We also train a version of each of the base models which expects morphological tags as input and is trained with gold morphological tags. During runtime, we report two different settings, one which uses predicted tags as input (\textbf{Predicted MI}) and other that uses gold tag as input (\textbf{Oracle MI}). We obtain the morphological tags from a Neural CRF tagger \cite{yang2018ncrf} trained on our training data. Oracle MI will act as an upper-bound on the reported results.

\noindent\textbf{BiAFFINE Parser (BiAFF)}
\label{supple_BiaFF}
BiAFF~\cite{DBLP:conf/iclr/DozatM17} is a graph-based dependency parsing approach similar to \newcite{kiperwasser-goldberg-2016-simple}. It uses biaffine attention instead of using a traditional MLP-based attention mechanism. For input vector $\vec{h}$, the affine classifier is expressed as $W\vec{h}+b$, while the biaffine classifier is expressed as $W'(W\vec{h}+b)+b'$. The choice of biaffine classifier facilitates the key benefit of representing the prior probability of word $j$ to be head and the likelihood of word $i$ getting word $j$ as the head. In this system, during training, each modifier in the predicted tree has the highest-scoring word as the head. This predicted tree need not be valid. However, at test time, to generate a valid tree MST algorithm~\cite{edmonds1967optimum} is used on the arc scores.

\noindent\textbf{Deep Contextualized Self-training (DCST)}
\label{supple_DCST}
~\newcite{rotman2019deep} proposed a self-training method called Deep Contextualized Self-training (DCST).\footnote{\url{https://github.com/rotmanguy/DCST}} 
In this system, the base parser BiAFF~\cite{DBLP:conf/iclr/DozatM17} is trained on the labelled dataset. 
Then this trained base parser is applied to the unlabelled data to generate automatically labelled dependency trees. In the next step, these automatically-generated trees are transformed into one or more sequence tagging schemes. Finally, the ensembled parser is trained on manually labelled data by integrating base parser with learned representation models.  The gating mechanism proposed by~\newcite{sato-etal-2017-adversarial} is used to integrate different tagging schemes into the ensembled parser. This approach is in line with the representation models based on language modeling related tasks~\cite{peters-etal-2018-deep,devlin-etal-2019-bert}. In summary, DCST demonstrates a novel approach to transfer information learned on labelled data to unlabelled data using sequence tagging schemes such that it can be integrated into final ensembled parser via word embedding layers.

\noindent\textbf{Results}
\label{results}
Table~\ref{table:san_results_SRW} presents results for dependency parsing on Sanskrit.
We observe that BiAFF + LCM outperforms all corresponding BiAFF models including Oracle MI. This is indeed a serendipitous outcome as one would expect Oracle MI to be an upper bound owing to its use of gold morphological tags at runtime. The DCST variant of our pretraining approach is also the best among its peers, although the performance of Oracle MI model in this case is indeed the upper bound.
 \begin{table}[H]
  \begin{small}
\centering
\begin{tabular}{ccccc}
\cmidrule(r){1-5}
 &\multicolumn{2}{c}{BiAFF}
&
\multicolumn{2}{c}{DCST} \\\cmidrule(r){2-3}\cmidrule(l){4-5}
Model &UAS&LAS    & UAS &LAS      \\\cmidrule(r){2-3}\cmidrule(l){4-5}
Base    & 70.67          & 56.85          & 73.23          & 58.64          \\
Predicted MI&69.02&    53.11&    71.15&    51.75\\
\cmidrule(r){2-3}\cmidrule(l){4-5}
MTL     & 70.85          & 57.93          & 73.04          & 59.12          \\
TranSeq  & 71.46 & 60.58 & 74.58 & 62.70\\
LCM    & \textbf{75.91} & \textbf{64.87} & \textbf{75.75} &\textbf{64.28}\\
\cmidrule(r){2-3}\cmidrule(l){4-5}
Oracle MI&\textit{74.08}&    \textit{62.48}&    \textit{76.66}&    \textit{66.35}\\

\hline
\end{tabular}
    \caption{Results on Sanskrit dependency parsing. Oracle MI is an upper bound and is not comparable.}
     \label{table:san_results_SRW}
     \end{small}
\end{table}
On the other hand, using predicted morphological tags instead of gold tags at run time degrades results drastically, especially for LAS, possibly due to the cascading effect of incorrect morphological information~\cite{nguyen-verspoor-2018-improved}. 
This shows that morphological information is essential in filling the UAS-LAS gap and substantiates the need for pretraining to incorporate such knowledge even when it is not available at run time. Interestingly, both MTL, and TranSeq, show improvements as compared to the base models, though do not match with that of our pretraining approach. In our experiments, the pretraining approach, even with \textit{a little training data}, clearly outperforms the other approaches.

\subsection{Analysis}
\noindent\textbf{Ablation:}
We perform further analysis on Sanskrit to study the effect of training set size as well as the impact of various tagging schemes as auxiliary tasks.
First, we evaluate the impact on performance as a function of the training size (Table~\ref{table:ablation_train_size}). 
Noticeably, for training size 100, we observe a 12 (UAS) and 17 (LAS) points increase for BiAFF+LCM over BiAFF, demonstrating the effectiveness of our approach in a very low-resource setting. This improvement is consistent for larger training sizes, though the gain reduces.
 \begin{table}[t]
  \begin{small}
\centering
\begin{tabular}{ccccc}
\hline
Training     & BiAFF       & DCST        & BiAFF+LCM            \\
Size     &UAS/LAS&UAS/LAS&UAS/LAS\\
\hline
100  & 58.0/42.3 & 64.0/44.0 & \textbf{70.4/59.9} \\
500  & 70.7/56.9 & 73.2/58.6 & \textbf{75.9/64.9} \\
750  & 74.0/61.8 & 75.2/62.3 & \textbf{77.3/66.8}  \\
1000 & 74.4/62.9 & 76.0/64.1 & \textbf{77.9/67.3} \\
1250 & 75.6/64.7 & 76.7/65.2 & \textbf{78.5/68.3}\\
\hline
\end{tabular}
    \caption{Performance as a function of training set size.}
     \label{table:ablation_train_size}
     \end{small}
\end{table}
\begin{figure}[h]
\centering
\includegraphics[width=0.5\textwidth]{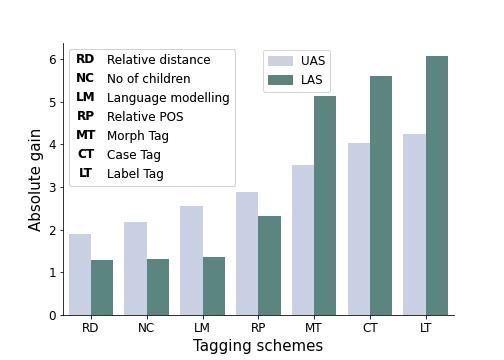}
\caption{Comparison of proposed tagging schemes (MT, CT, LT) with those in DCST (RD, NC, LM, RP).} 
\label{fig:tagging_impact} 
\end{figure}
In Figure~\ref{fig:tagging_impact}, we compare our tagging schemes with those used in self-training of DCST, namely, Relative Distance from root (RD), Number of Children for each word (NC), Language Modeling (LM) objective where task is to predict next word in sentence, and Relative POS (RP) of modifier from root word.  Here, we integrate each pretrained model (corresponding to each tagging scheme) individually on top of the BiAFF baseline using the gating mechanism and report the absolute gain over the BiAFF in terms of UAS and LAS metric.
Interestingly, our proposed tagging schemes, with an improvement of 3-4 points (UAS) and 5-6 points (LAS), outperform those of DCST and help bridge the gap between UAS-LAS.
    \begin{table}[h]
\centering
\begin{tabular}{cccc}
\hline
Auxiliary Task &F-score&Gain\\
\hline
Relative Distance (RD)             & 58.71   & 1.9/1.3 \\
No of children (NC)               & 52.82   & 2.2/1.3 \\
Relative POS  (RP)              & 46.52   & 2.9/2.3 \\
Lang Model (LM)                 & 41.54   & 2.6/1.4 \\
\hline
Coarse POS (CP)                     & 13.02   & 1.6/0.8 \\
Head Word (HW)                   & 40.12   & 1.5/0.4 \\
POS Head Word (PHW)         & 38.98   & 2.0/1.2 \\
Number Tagging (NT)                & 13.33   & 1.9/0.9 \\
Person Tagging (PT)                & 12.27   & 1.6/0.7 \\
Gender Tagging (GT)                & 0.28    & 1.3/0.2 \\
\hline
Morph Tagging (MT)               & 62.84   & 3.5/5.1 \\
Case Tagging (CT)                 & 73.51   & 4.0/5.6 \\
Label Tagging (LT)                & 71.51   & 4.2/6.0\\
\hline
\end{tabular}
    \caption{Comparison of different auxiliary tasks. F-score: Task performance, Gain: Absolute gain (when integrated with BiAFF) in terms of UAS/LAS score compared to BiAFF scores.}
     \label{table:Additional_tag_scheme}
\end{table}

\noindent\textbf{Additional auxiliary tasks}
\label{Experiments on additional auxiliary}
  With our proposed pretraining approach, we experiment with using the prediction of different grammatical categories as auxiliary tasks, namely, Number Tagging (NT), Person Tagging (PT), and Gender Tagging (GT). As the results in Table~\ref{table:Additional_tag_scheme} demonstrate, the improvements observed in these cases are much smaller than those for our proposed auxiliary tasks. Similar results are observed when considering other auxiliary tasks (see Table~\ref{table:Additional_tag_scheme}).
We find that combining these auxiliary tasks with our proposed ones did not provide any notable improvements. 
  One possible reason for under performance of these tagging schemes compared to the proposed ones could be that either when the training set is small, sequence taggers are not able to learn discriminative features only from surface form of words  (F-score is less than 40 in all such cases in Table~\ref{table:Additional_tag_scheme}) or the learned features are not helpful for the dependency parsing task.

\noindent\textbf{Experiments on other MRLs}
We choose 10 additional MRLs from Universal Dependencies (UD) dataset~\cite{mcdonald2013universal,nivre2016universal}, namely,  
Arabic (ar), Czech (cs), German (de), Basque (eu), Greek (el),    Finnish (fi),    Hungarian (hu), Polish (pl), Russian (ru) and Swedish (sv).\footnote{We choose MRLs that have the explicit morphological information with following grammatical categories: case, number, gender, and tense.}
Then we train them in low-resource setting (500 examples) to investigate the applicability of our approach for these MRLs.
\begin{table*}[bht]
\begin{small}
    \centering
\resizebox{1\textwidth}{!}{%
 \begin{tabular}{ccccccccccc|ccc}
\toprule
 &\multicolumn{2}{c}{eu} &\multicolumn{2}{c}{el} &\multicolumn{2}{c}{sv} &\multicolumn{2}{c}{pl}&\multicolumn{2}{c}{ru}
  &\multicolumn{2}{|c}{avg}
 \\\cmidrule(r){2-3}\cmidrule(l){4-5}\cmidrule(l){6-7}\cmidrule(l){8-9}\cmidrule(l){10-11} \cmidrule(l){12-13}
Model &UAS&LAS    & UAS &LAS   & UAS &LAS & UAS &LAS & UAS &LAS &UAS&LAS    
\\\cmidrule(r){2-3}\cmidrule(l){4-5}\cmidrule(l){6-7}\cmidrule(l){8-9}\cmidrule(l){10-11} \cmidrule(l){12-13}
BiAFF      & 63.18          & 54.52          & 79.64          & 75.01          & 71.73          & 64.83          & 78.33          & 70.83          & 73.98          & 67.42          & 73.37          & 66.52          \\
DCST       & 69.60          & 60.65          & 83.48          & 78.61          & 77.03          & 69.62          & 81.40          & 73.09          & 78.61          & 72.07          & 78.02          & 70.81          \\
\cmidrule(r){2-3}\cmidrule(l){4-5}\cmidrule(l){6-7}\cmidrule(l){8-9}\cmidrule(l){10-11}\cmidrule(l){12-13}
DCST+MTL       & 70.38          & 61.52          & 83.74          & 79.31          & 76.70          & 69.88          & 81.25          & 73.34          & 78.46          & 72.08          & 78.11          & 71.23          \\
DCST+TranSeq   & 70.70          & 62.96          & 84.69          & 80.37          & 77.30          & 70.85          & 82.84          & 75.02          & 78.95          & 73.18          & 78.90          & 72.48          \\
BiAFF+LCM        & \textbf{72.40} & \textbf{65.50} & \textbf{86.56} & \textbf{83.18} & 77.95          & 72.20          & \textbf{84.08} & \textbf{77.65} & 79.97          & 74.47          & 80.20          & 74.60          \\
DCST+LCM       & 72.01          & 65.33          & 85.94          & 82.22          & \textbf{78.72} & \textbf{73.04} & 83.83          & 77.63          & \textbf{80.62} & \textbf{75.26} & \textbf{80.22} & \textbf{74.70} \\
\cmidrule(r){2-3}\cmidrule(l){4-5}\cmidrule(l){6-7}\cmidrule(l){8-9}\cmidrule(l){10-11}\cmidrule(l){12-13}
BiAFF+Oracle MI  & \textit{72.16} & \textit{66.08} & \textit{83.05} & \textit{79.81} & \textit{76.50} & \textit{71.17} & \textit{83.27} & \textit{77.83} & \textit{77.83} & \textit{73.13} & \textit{78.56} & \textit{73.60} \\
DCST+Oracle MI & \textit{77.47} & \textit{71.55} & \textit{85.99} & \textit{82.72} & \textit{80.33} & \textit{75.00} & \textit{86.03} & \textit{80.46} & \textit{82.21} & \textit{77.54} & \textit{82.41} & \textit{77.45}\\
\hline
\\
 &\multicolumn{2}{c}{ar} &\multicolumn{2}{c}{hu} &\multicolumn{2}{c}{fi} &\multicolumn{2}{c}{de}&\multicolumn{2}{c}{cs}
  &\multicolumn{2}{|c}{avg}
 \\\cmidrule(r){2-3}\cmidrule(l){4-5}\cmidrule(l){6-7}\cmidrule(l){8-9}\cmidrule(l){10-11} \cmidrule(l){12-13}
Model &UAS&LAS    & UAS &LAS   & UAS &LAS & UAS &LAS & UAS &LAS &UAS&LAS    
\\\cmidrule(r){2-3}\cmidrule(l){4-5}\cmidrule(l){6-7}\cmidrule(l){8-9}\cmidrule(l){10-11} \cmidrule(l){12-13}
BiAFF      & 76.24          & 68.07          & 70.00          & 62.81          & 60.93          & 50.68          & 67.77          & 59.94          & 65.75          & 57.43          & 70.30          & 62.62          \\
DCST       & 79.05          & 71.18          & 74.62          & 67.00          & 66.04          & 54.76          & 73.22          & 65.18          & 74.15          & 65.52          & 75.61          & 67.70          \\
DCST+Predicted MI & 77.17 & 66.63 & 61.55 & 36.18 & 56.48 & 39.67 & 65.31 & 47.12 & 72.03 & 58.37 & 68.72 & 52.61 \\
\cmidrule(r){2-3}\cmidrule(l){4-5}\cmidrule(l){6-7}\cmidrule(l){8-9}\cmidrule(l){10-11}\cmidrule(l){12-13}
DCST+MTL       & 79.35          & 71.37          & 74.49          & 66.70          & 66.30          & 55.29          & 73.98          & 66.05          & 74.66          & 65.95          & 75.84          & 67.99          \\
DCST+TranSeq-FT   & 79.66          & 72.17          & 75.22          & 68.25          & 67.04          & 56.57          & 74.66          & 67.27          & \textbf{75.15} & 67.02          & 76.40          & 69.11          \\
BiAFF+LCM        & \textbf{79.68} & \textbf{72.55} & \textbf{76.15} & \textbf{69.53} & 69.05          & 59.41          & 75.85          & 68.80          & 74.94          & \textbf{67.58} & 76.91          & 70.13          \\
DCST+LCM       & 79.60          & 72.38          & 75.71          & 68.93          & \textbf{69.15} & \textbf{60.06} & \textbf{76.12} & \textbf{69.20} & 74.81          & 67.54          & \textbf{76.99} & \textbf{70.22} \\
\cmidrule(r){2-3}\cmidrule(l){4-5}\cmidrule(l){6-7}\cmidrule(l){8-9}\cmidrule(l){10-11}\cmidrule(l){12-13}
BiAFF+Oracle MI  & \textit{77.52} & \textit{71.46} & \textit{75.89} & \textit{70.63} & \textit{70.80} & \textit{64.64} & \textit{72.63} & \textit{66.53} & \textit{72.39} & \textit{66.22} & \textit{74.99} & \textit{69.20} \\
DCST+Oracle MI & \textit{80.43} & \textit{74.79} & \textit{78.43} & \textit{73.19} & \textit{75.30} & \textit{68.90} & \textit{77.70} & \textit{71.66} & \textit{78.54} & \textit{72.38} & \textit{79.09} & \textit{73.40}\\
\toprule
\end{tabular}}
    \caption{Evaluation on 10 MRLs. Results of BiAFF+LCM and DCST+LCM are statistically significant compared to strong baseline DCST as per t-test ($p < 0.01$). 
    Last two columns denote the average performance. Models using Oracle MI are not comparable.}
    \label{table:multilingual_results}
    \end{small}
\end{table*}

For all MRLs, the trend is similar to what is observed for Sanskrit. While all four models improve over both the baselines, BiAFF+LCM and DCST+LCM consistently turn out to be the best configurations. Note that these models are not directly comparable to Oracle MI models since Oracle MI models use gold morphological tags instead of the predicted ones. 
The performance of BiAFF+LCM and DCST+LCM is also comparable. Across all 11 MRLs, BiAFF+LCM shows the average absolute gain of 2 points (UAS) and 3.6 points (LAS) compared to the strong baseline DCST.

\noindent\textbf{Comparison with mBERT Pretraining}
\label{Comparison with mBERT Pretraining}
We compare the proposed method with multilingual BERT \cite[mBERT]{devlin-etal-2019-bert} based multi-task learning  model \cite[Udify]{kondratyuk-straka-2019-75}. This single model trained on 124 UD treebanks covers 75 different languages and produces state of the art results for many of them.
Multilingual BERT leverages large scale pretraining on wikipedia for 104 languages. 
   \begin{table}[H]
  \begin{small}
\centering
\begin{tabular}{cccccc}
\hline
Lang &BiAFF&BiAFF+LCM&Udify\\
\hline
Basque&	63.2/54.5&	72.4/65.5&	\textbf{76.6/69.0}\\
German&	67.7/60.0&	75.8/68.8&	\textbf{83.7/77.5}\\
Hungarian&	70.0/62.8&	76.2/69.5&	\textbf{84.4/76.7}\\
Greek&	69.6/75.0&	86.6/83.2&	\textbf{90.6/87.0}\\
Polish&	78.3.70.8&	84.1/77.7&	\textbf{90.7/85.0}\\
\hline
Sanskrit&	70.7/56.8&\textbf{75.9/64.9}&	69.4/53.2\\
Sanskrit-Vedic&	56.0/42.3& \textbf{61.6/48.0}&	47.4/28.3\\
Wolof&	75.3/67.8&\textbf{78.4/71.3}&		70.9/60.6\\
Gothic&	61.7/53.3&\textbf{69.6/61.4}&		63.4/52.2\\
Coptic&	84.3/80.2&\textbf{86.2/82.7}&		32.7/14.3\\
\hline
\end{tabular}
    \caption{The proposed method outperforms Udify for the languages (down) not covered in mBERT and under performs for the languages (top) which are covered in mBERT.}
     \label{table:mBERT_udify_compare}
     \end{small}
\end{table}
In our experiments, we find that Udify outperforms the proposed method for languages covered during mBERT's  pretraining. Notably, not only the proposed method but also a simple BiAFF parser with randomly initialized embedding outperforms Udify (Table~\ref{table:mBERT_udify_compare}) for languages which not available in mBERT.
Out of 7,000 languages, only a handful of languages can take advantage of mBERT pretraining~\cite{joshi-etal-2020-state} which substantiates the need of our proposed pretraining scheme.

\noindent\textbf{Experiments on TranSeq Variants}
\label{supple_TranSeq}

In TranSeq variations, instead of pretraining with three auxiliary tasks, we use a hierarchical multi-task morphological tagger~\cite{ashim2020evaluate} trained on 50k training data from DCS~\cite{hellwig2010dcs}.
In TranSeq setting, we extract the first three layers from this tagger and augment them in baseline models and experiment with five model sub-variants.
 \begin{table}[!hbt]
  \begin{small}
\centering
\begin{tabular}{ccccc}
\cmidrule(r){1-5}
 &\multicolumn{2}{c}{BiAFF}
&
\multicolumn{2}{c}{DCST} \\\cmidrule(r){2-3}\cmidrule(l){4-5}
Model &UAS&LAS    & UAS &LAS      \\\cmidrule(r){2-3}\cmidrule(l){4-5}

Base    & 70.67          & 56.85          & 73.23          & 58.64          \\
Base$\star$ & 69.35    &52.79 &    72.31&    54.82 \\
\cmidrule(r){2-3}\cmidrule(l){4-5}
TranSeq-FE  & 66.54          & 55.46          & 71.65          & 60.10          \\
TranSeq-FEA & 69.50          & 58.48          & 73.48          & 61.52          \\
TranSeq-UF  & 70.60          & 59.74          & 73.55          & 62.39          \\
TranSeq-DL  & 71.40          & 60.58          & 74.52          & 62.73 \\
TranSeq-FT  & \textbf{71.46} & \textbf{60.58} & \textbf{74.58} & \textbf{62.70}\\
\cmidrule(r){2-3}\cmidrule(l){4-5}
Oracle MI&\textit{74.08}&    \textit{62.48}&    \textit{76.66}&    \textit{66.35}\\

\hline
\end{tabular}
    \caption{Ablation analysis for TranSeq variations. Oracle MI is not comparable. It can be considered as upper bound for TranSeq.}
     \label{table:san_results_TranSeq}
     \end{small}
\end{table}
To avoid catastrophic forgetting~\cite{MCCLOSKEY1989109,french1999catastrophic}, we gradually increase the capacity of adaptations for each variant. \textbf{TranSeq-FE:} Freeze the pre-trained layers and use them as Feature Extractors (FE).  \textbf{TranSeq-FEA:} In the feature extractor setting, we additionally integrate adaptor modules~\cite{pmlr-v97-houlsby19a,pmlr-v97-stickland19a} in between two consecutive pre-trained layers.
     \textbf{TranSeq-UF:} Gradually Unfreeze (UF) these three pre-trained layers in the top to down fashion~\cite{felbo-etal-2017-using,howard-ruder-2018-universal}. \textbf{TranSeq-DL:} In this setting, we use discriminative learning (DL) rate~\cite{howard-ruder-2018-universal} for pre-trained layers, i.e., decreasing the learning rate as we move from top-to-bottom layers. \textbf{TranSeq-FT:} We fine-tune (FT), pre-trained layers with default learning rate used by~\newcite{DBLP:conf/iclr/DozatM17}.
     
    In the TranSeq setting, as we move down across its sub-variants in Table~\ref{table:san_results_TranSeq},  performance improves gradually, and TranSeq-FT configuration shows the best performance with 1-2 points improvement over Base. The Base$\star$ has one additional LSTM layer compared to Base such that the number of parameters are same as that of TranSeq-FT variation. The performance of Base$\star$ decreases compared to Base but TranSeq-FT outperforms Base. This shows that transfer learning definitely helps to boost the performance.

\section{Low-Resource Dependency Parsing for Multiple Low-Resource Languages}
Recently, the supervised learning paradigm has dramatically increased the state-of-the-art performance for the dependency parsing task for resource-rich languages~\cite{chen-manning-2014-fast,dyer-etal-2015-transition,kiperwasser-goldberg-2016-simple,DBLP:conf/iclr/DozatM17,kulmizev-etal-2019-deep}.
However, only a handful of resource-rich languages are able to take advantage, and many low-resource languages are far from these benefits \cite{joshi-etal-2020-state,more19,zeman-etal-2018-conll}. 

  In literature, several strategies have been proposed to enhance performance in low-resource scenarios, such as data augmentation, cross/mono-lingual pretraining \cite{conneau-etal-2020-unsupervised,peters-etal-2018-deep,kondratyuk-straka-2019-75}, sequential transfer learning \cite{ruder-etal-2019-transfer}, multi-task learning \cite{nguyen-verspoor-2018-improved}, cross-lingual transfer~\cite{cross_survey} and self-training \cite{rotman2019deep,clark-etal-2018-semi}.
    However, not much attention has been given to measuring the efficacy of the existing low-resource strategies well-known to the community for low-resource dependency parsing \cite{vania-etal-2019-systematic}. 
  This is essential to assess their utility for low-resource languages before inventing novel ways to tackle data sparsity.

In this chapter, we systematically explore 5 pragmatic strategies for low-resource settings on 7 languages. We experiment with low-resource strategies such as data augmentation, sequential transfer learning, cross/mono-lingual pretraining, multi-task learning and self-training. We investigate: (1) How is the trend in performance of each strategy across various languages? Whether the choice of best performing variant of each strategy is language dependent? (2) We integrate the best performing variant of each strategy and call the resulting system as the ensembled system. Do all the strategies contribute towards performance gain in the ensembled system? How well does this ensemble approach generalize across multiple low-resource languages?  (3) 
How far can we push a purely data-driven ensemble system using the best-performing low-resource strategies? Can this simple ensemble approach outperform the state-of-the-art of a low-resource language? We argue that while it may sound like a simple application of techniques well known to the community; it is non-trivial to select the best performing combination for a target low-resource language.

Our exhaustive experimentation empirically establishes the effective generalization ability of the ensembled system on 7 languages and shows average absolute gains of 5.2/6.2 points Unlabelled/Labelled Attachment Score (UAS/LAS) over strong baseline \cite{dozat2017stanford}. 
Notably, our ensembled system shows substantial improvements for the languages not covered in pretrained models.
Finally, we show a successful application of the ensembled system on a truly low-resource language Sanskrit.
We find that the ensembled system outperforms state-of-the-art system \cite{krishna-etal-2020-keep} for Sanskrit by 1.2 points absolute gain in terms of UAS and shows comparable performance in terms of LAS (\S~\ref{experiments}).

\subsection{Investigation of Strategies Tailored for Low-resource Settings}
\label{5_strategies}
 We explore 5 strategies specially tailored for low-resource settings, on 7 languages and integrate the best performing strategy of each category in our ensembled system (Table~\ref{table:multilingual_results_EACL}). We utilize \newcite{DBLP:conf/iclr/DozatM17} as a base system for all the experiments, henceforth referred to as BiAFF.

\noindent\textbf{Language selection criteria:} 
 We choose low-resource languages with less than 2,500 training samples from 4 different typological families such that each language belongs to a unique sub-family.  In order to accommodate a low-resource tailored pretraining \cite{sandhan-etal-2021-little}, we choose languages that have explicit morphological information. Additionally, we divide the set of languages into the languages covered/not-covered in the multilingual language model’s pretraining: (1) Covered: Arabic (ar), Greek (el), Hungarian (hu) (2) Not covered:  Wolof (wo), Gothic (got), Coptic (cop) and Sanskrit (san). 
\begin{figure}[!h]
\centering
{\includegraphics[width=0.6\linewidth]{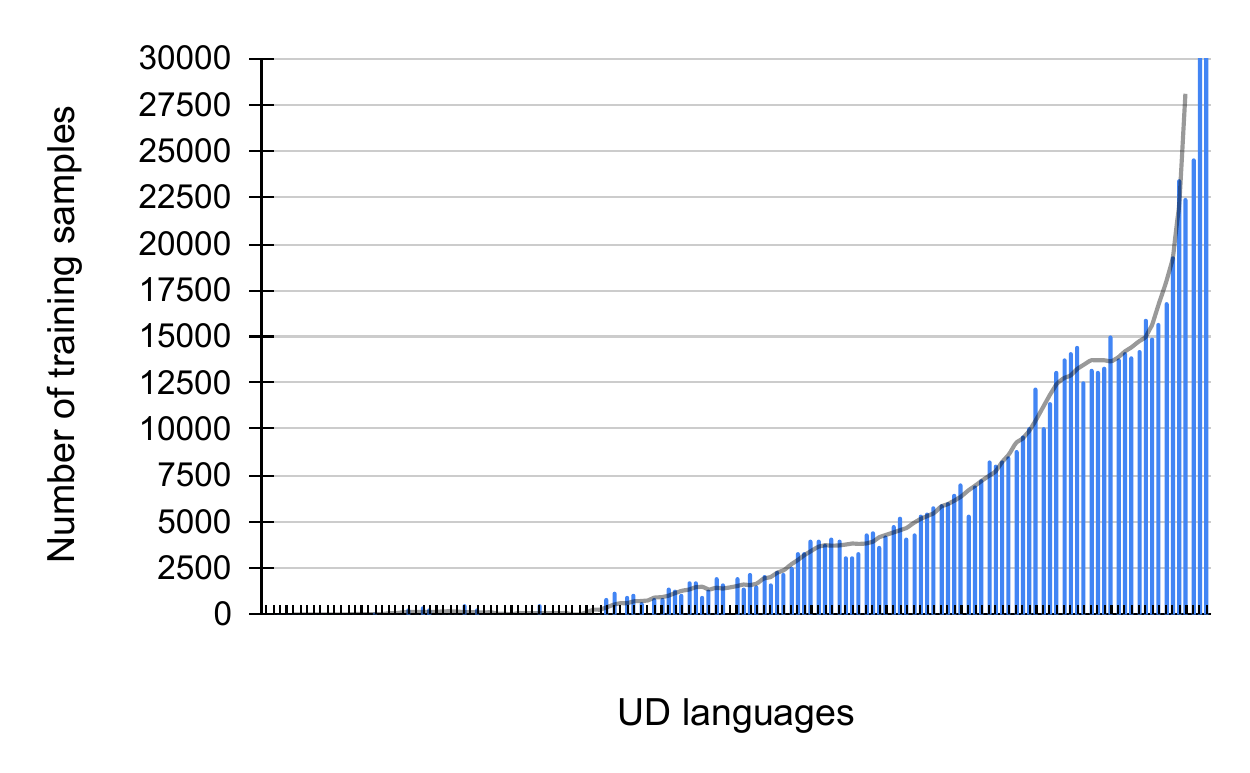}}
\caption{Plot for number of training samples vs. UD languages available in UD-2.6.}
\label{fig:UD_vs_train}
\end{figure}
\noindent Figure~\ref{fig:UD_vs_train} illustrates the number of training samples available for all UD languages in UD-2.6.
  \newcite{hedderich-etal-2021-survey} ask ``How low is low-resource?'' and suggest that the threshold of low-resource is task and language-dependent. The low-resource settings can be seen as a continuum of resource availability due to the absence of a hard threshold. More efforts should focus on evaluating low-resource strategies across multiple languages for a fair comparison between these strategies. Therefore, we select a threshold for the languages with less than 2,500 training samples (Figure~\ref{fig:UD_vs_train}). 
   We restrict ourselves to the setting where the target low-resource language does not have a high-resource related language that could possibly facilitate the positive cross-lingual or zero-shot transfer 
   \cite{vulic-etal-2019-really,sogaard-etal-2018-limitations}. Thus, we do not consider low-resource languages with only a test set available. Also, we do not consider cross-lingual transfer~\cite{duong2015low,ahmad-etal-2019-cross,vania-etal-2019-systematic,cross_survey} strategy in our study.
 
\noindent\textbf{Dataset and metric:} For each of these  7 low-resource languages from Universal Dependencies (UD-2.6) \cite{10.1162/coli_a_00402},
  following \newcite{rotman2019deep}, we use 500 data points for training (allocating equal power to each language for fair comparison) and the original dev/test split as dev/test set. Additionally, 1000 morphologically tagged data points (without dependency annotations) are used for self-training and pretraining.
   We use sentence level macro averaged UAS/LAS metric for evaluation.

 \begin{table*}[bht]
\begin{small}
    \centering
\resizebox{1\textwidth}{!}{
 \begin{tabular}{ccccccccccccccccc}
\toprule
& &\multicolumn{2}{c}{el} &\multicolumn{2}{c}{ar} &\multicolumn{2}{c}{hu} &\multicolumn{2}{c}{got}&\multicolumn{2}{c}{cop}
  &\multicolumn{2}{c}{wo}&\multicolumn{2}{c}{san}
 \\\cmidrule(r){2-2}\cmidrule(r){3-4}\cmidrule(l){5-6}\cmidrule(l){7-8}\cmidrule(l){9-10}\cmidrule(l){11-12} \cmidrule(l){13-14} \cmidrule(l){15-16}
Strategy&Model &UAS&LAS    & UAS &LAS   & UAS &LAS & UAS &LAS & UAS &LAS &UAS&LAS  &UAS&LAS  
\\\cmidrule(r){2-2}\cmidrule(r){3-4}\cmidrule(l){5-6}\cmidrule(l){7-8}\cmidrule(l){9-10}\cmidrule(l){11-12} \cmidrule(l){13-14} \cmidrule(l){15-16}
&BiAFF      & 86.61          & 82.23          & \textbf{79.89} & \textbf{73.13} & \textbf{80.51} & \textbf{75.11} & 75.66          & \textbf{69.42} & 87.33          & \textbf{84.48} & \textbf{82.63} & \textbf{77.83} & 75.47          & 65.77          \\
\cmidrule(r){2-2}\cmidrule(r){3-4}\cmidrule(l){5-6}\cmidrule(l){7-8}\cmidrule(l){9-10}\cmidrule(l){11-12} \cmidrule(l){13-14} \cmidrule(l){15-16}

&Cropping &      85.98          & 81.76          & 79.42          & 73.03          & 77.46          & 71.62          & 75.72          & 68.25          & 86.05          & 82.95          & 80.38          & 75.39          & 73.82          & 63.46          \\
Data aug.&Rotation &      86.39          & 82.19          & 79.22          & 72.79          & 77.33          & 71.16          & 76.19          & 68.89          & 86.27          & 83.27          & 80.66          & 76.05          & 75.58          & 65.14          \\
&Nonce   &    \textbf{87.43} & \textbf{82.60} & 79.52          & 73.00          & 79.56          & 69.50          & \textbf{76.42} & 67.74          & \textbf{87.64} & 83.52          & 81.97          & 76.37          & \textbf{77.25} & \textbf{66.30} \\
\cmidrule(r){2-2}\cmidrule(r){3-4}\cmidrule(l){5-6}\cmidrule(l){7-8}\cmidrule(l){9-10}\cmidrule(l){11-12} \cmidrule(l){13-14} \cmidrule(l){15-16}
&BiAFF+mBERT &     91.41          & 87.89          & 83.50          & 76.30          & 85.20          & 77.50          & 64.20          & 53.20          & 33.70          & 15.60          & 71.50          & 61.40          & 71.40          & 55.12          \\
Pretraining&BiAFF+XLM-R &  \textbf{93.61} & \textbf{90.85} & \textbf{86.04} & \textbf{79.55} & \textbf{89.05} & \textbf{83.80} & -              & -              & -              & -              & -              & -              & 78.43          & 66.72          \\
&BiAFF+LCM  &     89.00          & 85.83          & 82.49          & 76.67          & 83.22          & 78.49          & \textbf{79.88} & \textbf{74.65} & \textbf{88.79} & \textbf{86.02} & \textbf{85.85} & \textbf{81.66} & \textbf{81.63} & \textbf{73.86} \\
\cmidrule(r){2-2}\cmidrule(r){3-4}\cmidrule(l){5-6}\cmidrule(l){7-8}\cmidrule(l){9-10}\cmidrule(l){11-12} \cmidrule(l){13-14} \cmidrule(l){15-16}
&SelfTrain  &    86.78          & 82.25          & 80.86          & 74.45          & 80.62          & 75.09          & 76.96          & 70.15          & 87.95          & 85.33          & 83.83          & 78.80          & 77.53          & 66.59          \\
Self-training&CVT  &     80.53          & 77.37          & 76.21          & 71.87          & 75.21          & 70.01          & 69.43          & 63.59          & 79.32          & 74.21          & 73.21          & 69.50          & 69.21          & 56.21          \\
&DCST    &   \textbf{88.26} & \textbf{84.09} & \textbf{82.21} & \textbf{75.78} & \textbf{82.85} & \textbf{77.65} & \textbf{79.52} & \textbf{72.91} & \textbf{88.85} & \textbf{85.53} & \textbf{85.51} & \textbf{80.71} & \textbf{78.55} & \textbf{69.10} \\
\cmidrule(r){2-2}\cmidrule(r){3-4}\cmidrule(l){5-6}\cmidrule(l){7-8}\cmidrule(l){9-10}\cmidrule(l){11-12} \cmidrule(l){13-14} \cmidrule(l){15-16}
&SeqTraL-FE  &   88.43          & 85.43          & 81.86          & 76.60          & 82.97          & 78.46          & 80.15          & 75.24          & 88.08          & 85.57          & 85.61          & 81.77          & 81.20          & 73.70          \\
&SeqTraL-UF     &88.50          & 85.36          & 82.52          & 76.79          & \textbf{83.83} & \textbf{79.24} & \textbf{80.79} & \textbf{75.65} & \textbf{88.87} & 86.30          & 85.78          & 81.54          & 81.51          & 73.65          \\
SeqTraL&SeqTraL-DL        & \textbf{89.06} & \textbf{85.88} & 82.57          & 76.66          & 83.36          & 78.57          & 80.29          & 74.89          & 88.78          & 86.14          & \textbf{86.25} & \textbf{81.85} & 81.17          & 73.10          \\
&SeqTraL-FT       &88.80          & 85.47          & \textbf{82.66} & \textbf{76.83} & 83.79          & 78.95          & 80.13          & 75.11          & 88.86          & \textbf{86.31} & 86.03          & 81.64          & \textbf{81.84} & \textbf{73.94} \\
\cmidrule(r){2-2}\cmidrule(r){3-4}\cmidrule(l){5-6}\cmidrule(l){7-8}\cmidrule(l){9-10}\cmidrule(l){11-12} \cmidrule(l){13-14} \cmidrule(l){15-16}
&MTL-Case    & \textbf{86.73} & \textbf{82.47} & \textbf{80.49} & \textbf{74.08} & \textbf{80.73} & \textbf{75.52} & -              & -              & 86.45          & 83.82          & 82.86          & 77.46          & 76.15          & 65.36          \\
Multi-tasking&MTL-Label     & 86.13          & 81.55          & 79.86          & 72.72          & 80.07          & 73.92          & 75.52          & 69.30          & \textbf{87.44} & \textbf{84.62} & 83.08          & 77.94          & 76.02          & 65.20          \\
&MTL-Morph      & 86.30          & 82.23          & 80.02          & 73.55          & 80.49          & 74.70          & \textbf{77.33} & \textbf{71.05} & 87.00          & 84.22          & \textbf{83.25} & \textbf{78.75} & \textbf{76.71} & \textbf{66.69} \\
\cmidrule(r){2-2}\cmidrule(r){3-4}\cmidrule(l){5-6}\cmidrule(l){7-8}\cmidrule(l){9-10}\cmidrule(l){11-12} \cmidrule(l){13-14} \cmidrule(l){15-16}
&BiAFF    &86.61          & 82.23          & 79.89          & 73.13          & 80.51          & 75.11          & 75.66          & 69.42          & 87.33          & 84.48          & 82.63          & 77.83          & 75.47          & 65.77          \\
&+Pretraining     & \textbf{93.61} & \textbf{90.85} & \textbf{86.04} & \textbf{79.55} & \textbf{89.05} & \textbf{83.80} & 79.88          & 74.65          & 88.79          & 86.02          & 85.85          & 81.66          & 81.63          & 73.86          \\
&+MTL      &89.99          & 86.49          & 82.47          & 76.24          & 84.35          & 79.74          & 80.33          & 75.15          & 88.42          & 85.94          & 85.91          & 81.56          & 81.30          & 73.49          \\
Prop. system&+SeqTraL      &90.31          & 86.70          & 82.70          & 76.57          & 84.58          & 80.15          & \textbf{80.79} & \textbf{75.65} & \textbf{88.87} & \textbf{86.30} & \textbf{86.05} & \textbf{81.85} & \textbf{81.84} & \textbf{73.94} \\
&+Self-training      & 89.83          & 86.09          & 82.08          & 75.92          & 84.12          & 79.66          & 80.08          & 75.24          & 88.78          & 86.07          & 85.73          & 81.77          & 79.89          & 72.28          \\
&+Data. aug. &89.11          & 85.87          & 82.08          & 75.92          & 84.12          & 79.66          & 79.56          & 73.53          & 88.31          & 84.67          & 85.73         & 81.77          & 79.52          & 71.89  \\
\cmidrule(r){2-2}\cmidrule(r){3-4}\cmidrule(l){5-6}\cmidrule(l){7-8}\cmidrule(l){9-10}\cmidrule(l){11-12} \cmidrule(l){13-14} \cmidrule(l){15-16}
Evaluation&BiAFF      & 87.10           & 83.06          & 80.92         & 75.02          & 80.31          &74.16 & 77.73          & 70.72          & 88.50          & 85.32          & 80.92          &75.02         & 79.33         & 67.92          \\
on test set&Prop. system & \textbf{93.66}           & \textbf{90.68}          & \textbf{86.43}         & \textbf{79.88}          & \textbf{88.50}          &\textbf{82.67} & \textbf{82.52}          & \textbf{77.07}          & \textbf{89.31} & \textbf{86.38}          & \textbf{87.50}          &\textbf{82.95}         & \textbf{83.59}         & \textbf{74.83}   \\ 
\hline
\end{tabular}}
    \caption{Evaluation of low-resource strategies on 7 languages. Experiments are first performed on dev set to find best performing combination of strategies for each language. The best results from strategies from each family are bold and statistically significant compared to its peer baselines belonging to the same family as per t-test ($p < 0.01$). The second last block shows ablations when the best variant from each family is added to the ensembled system. For example, +Data. aug. refers to the system with the best variant from all 5 strategies. The best performing system as per dev set is finally compared with BiAFF on the test set. XLM-R is not compatible with 3 languages and case information of Gothic (\textit{got}) language is missing; hence we do not report their results.}
    \label{table:multilingual_results_EACL}
    \end{small}
\end{table*}
\noindent\textbf{Sequential Transfer Learning (SeqTraL):}
Following \newcite{sandhan-etal-2021-little}, we pretrain three encoders (similar to BiAFF) on three sequence labelling auxiliary tasks and integrate them with the BiAFF encoder using a gating mechanism. We adapt these pretrained encoders with various optimization schemes, proposed for reducing a catastrophic forgetting~\cite{french1999catastrophic,MCCLOSKEY1989109}.
\textbf{SeqTraL-FE:} We treat newly integrated layers as Feature Extractors (FE) by freezing them.  
 \textbf{SeqTraL-UF:} Gradually Unfreeze (UF) these new layers in the top to down order~\cite{howard-ruder-2018-universal,felbo-etal-2017-using}. \textbf{SeqTraL-DL:} The discriminative learning rate (DL)  is used for newly added layers~\cite{howard-ruder-2018-universal}, the learning rate is decreased from top-to-bottom layers. \textbf{SeqTraL-FT:} We fine tune (FT) all the newly added layers with the default learning rate.

\noindent\textbf{Cross/mono-lingual Pretraining:}
We experiment with two multilingual pretrained models, namely, multilingual BERT \cite[\textbf{mBERT}]{devlin-etal-2019-bert} based system \cite{kondratyuk-straka-2019-75} and XLM-Roberta \cite[\textbf{XLM-R}]{conneau-etal-2020-unsupervised} based system \cite{nguyen2021trankit}. 
We also consider supervised pretraining specially tailored for low-resource dependency parsing \cite[\textbf{LCM}]{sandhan-etal-2021-little} which essentially combines three sequence labelling auxiliary tasks.  
We pretrain it on 1,000 morphologically tagged data points without dependency annotations. 

\noindent\textbf{Self-training:} 
Another line of modelling focuses on self-training~\cite{goldwasser-etal-2011-confidence,clark-etal-2018-semi,rybak2018semi} to overcome the bottleneck of task-specific labelled data. Earlier attempts failed to prove effectiveness of self-training for dependency parsing~\cite{rush-etal-2012-improved}. However, \newcite[\textbf{CVT}]{clark-etal-2018-semi} and \newcite[\textbf{DCST}]{rotman2019deep}, show successful application, thus, we consider these two systems. Also, we generate dependency data by applying a pretrained BiAFF system on 1000 unlabelled data points. We augment this predicted data with gold data and retrain BiAFF in \textbf{Self-Train} setting.

\noindent\textbf{Multi-task Learning:}
We simultaneously train BiAFF and a sequence labelling based auxiliary task in a multi-task setting (\textbf{MTL}).  We experiment with the following auxiliary tasks: prediction of the morphological label (\textbf{MTL-Morph}),
dependency relation between a word and its head (\textbf{MTL-Label}) and the case label (\textbf{MTL-Case}).

\noindent\textbf{Data Augmentation:}
\label{data_augmentation}
\newcite{sahin-steedman-2018-data} introduce \textbf{Cropping:} delete some parts of a sentence to create multiple short meaningful sentences, and \textbf{Rotation:} permute the siblings of headword restricted to a set of relations. 
Both operations modify a set of words or configurational information; however, they do not change the dependencies. 
\textbf{Nonce:} \newcite{gulordava-etal-2018-colorless} propose to create nonce sentences by substituting a few words which share the same syntactic labels. For each variant, we use additional 1,000 augmented data points. 

\noindent\noindent\textbf{Results on multilingual experiments:}
 Table~\ref{table:multilingual_results_EACL} first reports results of all 5 strategies on dev set of 7 languages.
 Next, the second last block of Table~\ref{table:multilingual_results_EACL} (Prop. system) shows ablations on dev set where the best variant from each family is gradually added into the ensembled system. For example, \textit{+Data.aug.} row refers to the system with the best variant from all 5 strategies. Finally, the best performing system as per dev set is compared with BiAFF on the test set.
 We observe that (1) the best performing variant from augmentation, SeqTraL and MTL families is language dependent. (2) DCST variant of self-training wins over its peer for all the languages. (3) XLM-R outperforms for the languages which are covered in its pretraining (except Sanskrit\footnote{Maybe due to limited coverage of corpus for Sanskrit.}) and LCM outperforms for the rest of the languages which are truly low-resource. (4) Notably, we find effective generalization ability of the proposed approach on languages covered in cross-lingual pretraining (only pretraining helps)
 and for the rest of the languages (pretraining, MTL and SeqTraL helps). 

\begin{figure*}[!tbh]
\centering
\subfigure[\label{fig:tagging_schem}]{\includegraphics[width=0.45\linewidth]{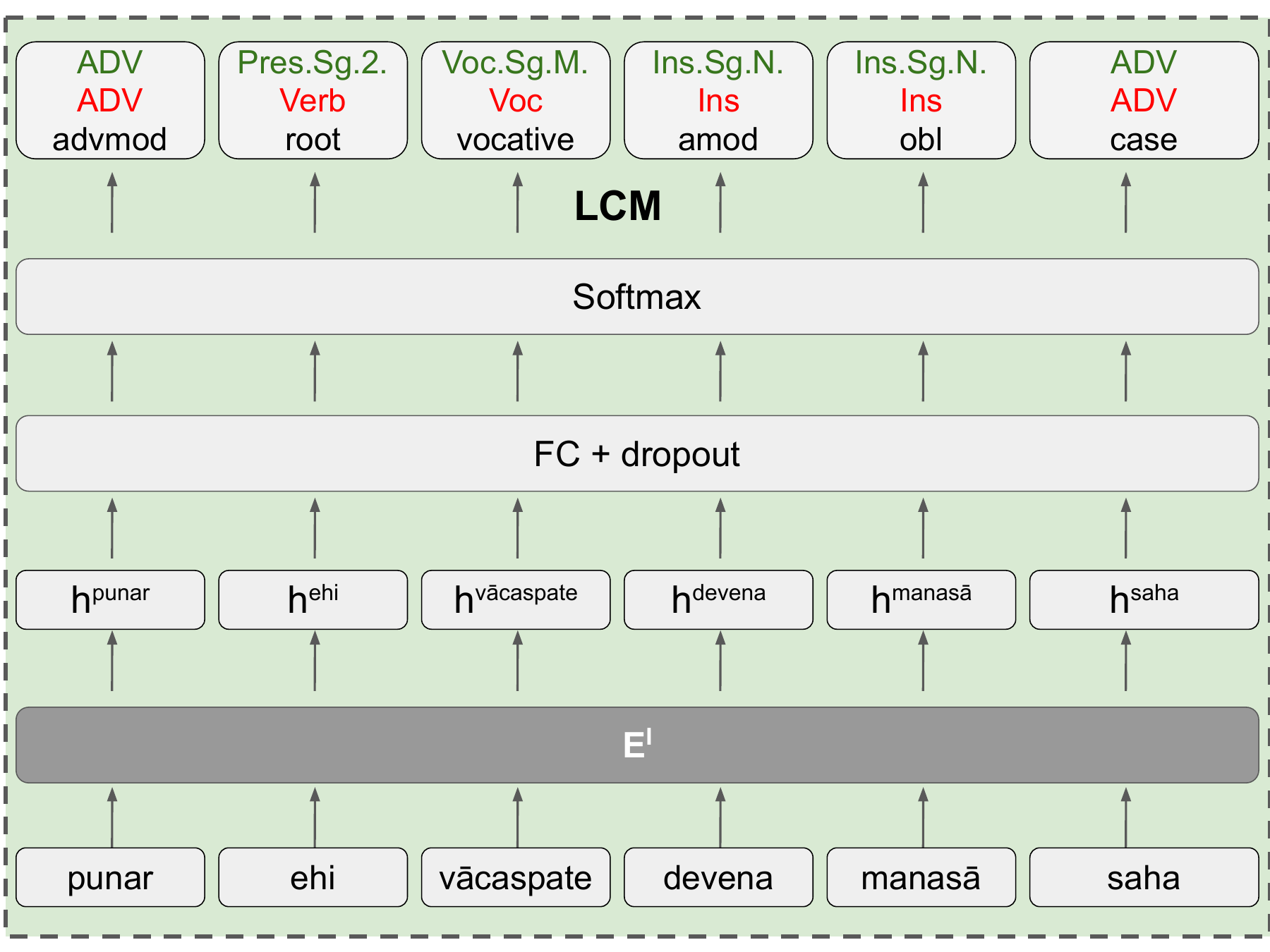}}
\subfigure[\label{fig:gating}]{\includegraphics[width=0.45\linewidth]{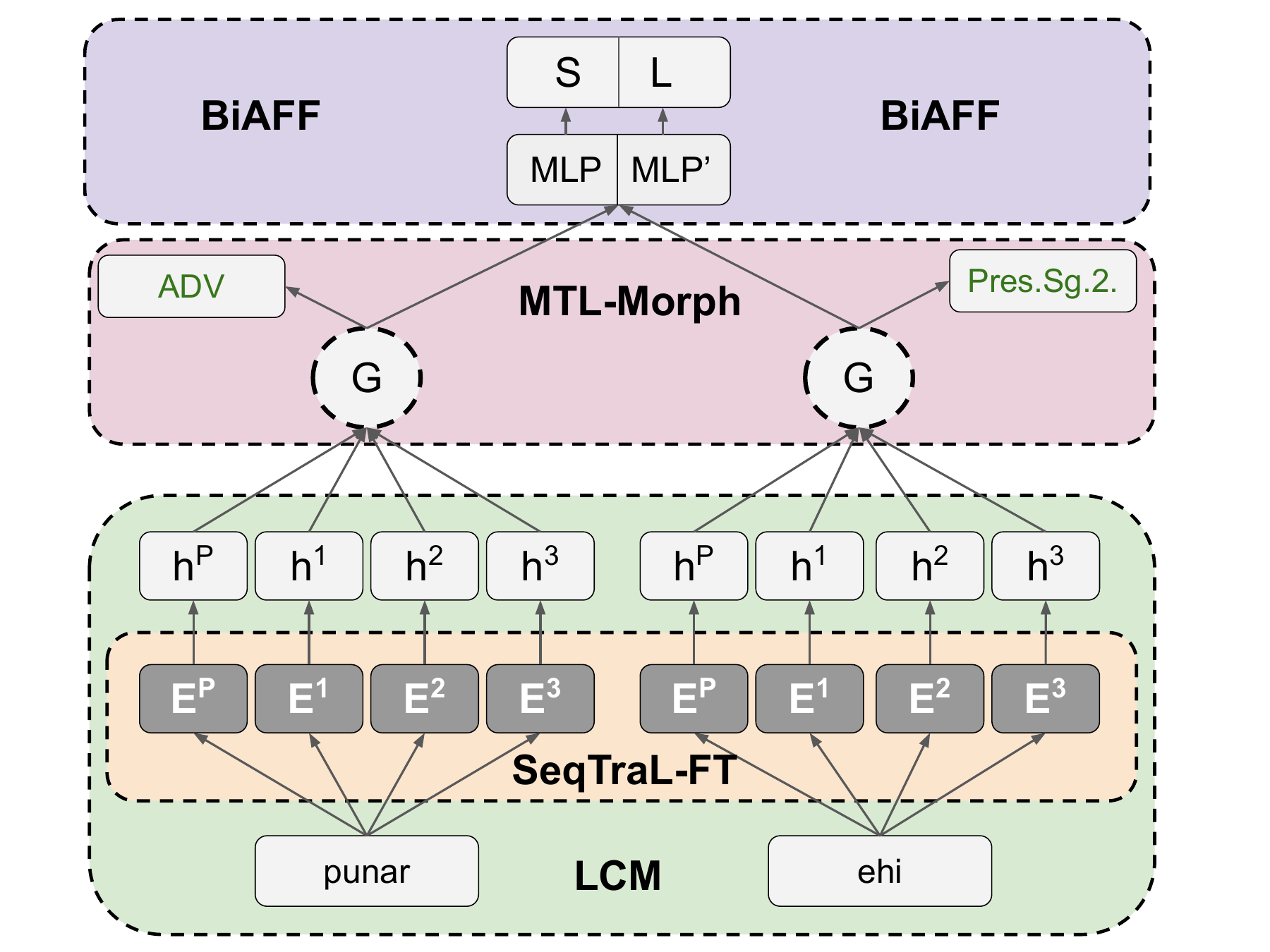}}
\caption{The ensembled system for Sanskrit. Translation: ``Oh V\={a}caspate! Come again with divine mind".
}
\label{fig:model}
\end{figure*}
\subsection{Application on Sanskrit}
\label{experiments}
\noindent\textbf{Data:} 
We use two standard benchmark datasets available for Sanskrit. We use 1,700, 1,000 and 1,300 sentences (prose domain) from the Sanskrit Treebank Corpus \cite[STBC]{kulkarni2010designing} as train, dev and test set, respectively.
 We also evaluate on the Vedic Sanskrit Treebank \cite[VST]{hellwig-etal-2020-treebank} consisting of 1,500 , 1,024 and 1,473 sentences (poetry-prose mixed) as train, dev and test data, respectively. For both data, the final results on the test set are reported using systems trained with combined gold train and dev set.
   
\noindent\textbf{Baselines:}
We use \newcite[\textbf{YAP}]{more-etal-2019-joint} and \newcite[\textbf{L2S}]{chang2016} from transition-based dependency parsing family.
  \newcite[\textbf{BiAFF}]{DBLP:conf/iclr/DozatM17} is a graph-based approach with BiAFFINE attention mechanism.  \newcite[\textbf{MG-EBM}]{krishna-etal-2020-keep} extends \newcite[\textbf{Tree-EBM-F}]{amrith21} using multi-graph formulation. 
  Systems marked with (*) are hybrid systems which leverage linguistic rules from P\={a}\d{n}ini.\\
\noindent\textbf{The ensembled system:}
Figure~\ref{fig:model} shows the ensembled system for Sanskrit as per Table~\ref{table:multilingual_results_EACL}. It consists of two steps, namely, pretraining (\textbf{LCM} \crule[lightg]{0.25cm}{0.25cm}) and integration. As shown in Figure~\ref{fig:tagging_schem}, \textbf{LCM} pretrains three encoders $E^{(1)-(3)}$ using three independent auxiliary tasks, namely,  morphological label prediction, case label prediction and relation label prediction.  Thereafter, as shown in Figure \ref{fig:gating}, these pretrained encoders are integrated with the BiAFF  encoder $E^{(P)}$  using a gating mechanism  as employed in ~\newcite{sato-etal-2017-adversarial}. We use \textbf{SeqTraL-FT} \crule[lorg]{0.25cm}{0.25cm} optimization scheme to update the weights of these four encoders. Next, \textbf{MTL-Morph} \crule[lpink]{0.25cm}{0.25cm} component adds morphological tagging as an auxiliary task to inject complementary signal in the model. Finally, the combined representation of a pair of words in passed to \textbf{BiAFF} \crule[lvio]{0.25cm}{0.25cm} to calculate probability of arc score (S) and label (L).
 \begin{table}[h]
\centering
\resizebox{0.45\textwidth}{!}{
\begin{tabular}{|c|c|c|c|c|}
\hline
 &\multicolumn{2}{c|}{\textbf{STBC}}
&
\multicolumn{2}{c|}{\textbf{VST}} \\\hline
\textbf{System} &\textbf{UAS}&\textbf{LAS}    & \textbf{UAS} &\textbf{LAS}      \\\hline
YAP                & 75.31 & 66.02&70.37&56.09\\
L2S                & 81.97 & 74.14&72.44&62.76 \\
Tree-EBM-F                & 82.65 & 79.28&-&- \\
BiAFF              & 85.88 & 79.55&77.23&67.68 \\
Ours              & \textbf{88.67} & \textbf{83.47}&\textbf{79.71}&\textbf{69.89} \\
\hline
Tree-EBM-F*               &  \textit{85.32} & \textit{83.93} &-&- \\
MG-EBM* & \textit{87.46} & \textit{84.70} &-&- \\ \hline
\end{tabular}} 
    \caption{Results on test set for Sanskrit. \textit{Hybrid} systems, marked with (*) use extra-linguistic knowledge and are not directly comparable with our system. Our results are statistically significant compared to BiAFF as per t-test ($p < 0.01$). Results are averaged over 3 runs.} 
     \label{table:san_results}
\end{table} \\
\noindent\textbf{Results:}
On STBC, the ensembled system outperforms the state of the art \textit{purely data-driven} system (BiAFF) by 2.8/3.9 points (UAS/LAS) absolute gain.
Interestingly, it also supersedes the performance of the \textit{hybrid} state of the art system \cite[\textbf{MG-EBM}]{krishna-etal-2020-keep} by 1.2 points (UAS) absolute gain and shows comparable performance for LAS metric. We observe that performance of transition-based systems (\textbf{YAP/L2S}) is significantly low compared to graph-based counterparts (\textbf{BiAFF/Ours}). We also obtain a similar performance trend for VST data.
The VST data is a mixture of dependency labelled trees from both poetry and prose domain. As a result, the overall performance for VST is low compared to STBC due to loss of configurational information.\footnote{We do not evaluate Tree-EBM-F* and MG-EBM* on VST data due to the unavailability of the codebase.}


\noindent\textbf{LCM Pretraining}
\label{lcm_pretraining}
 \newcite[LCM]{sandhan-etal-2021-little} focused on a low-resource dependency parsing scenario, where obtaining morphological information is troublesome. To mitigate this challenge, the authors propose a supervised pretraining, which automatically leverages morphological information using the pretrained encoders.
In a nutshell, LCM integrates word representations from multiple encoders trained on three independent auxiliary tasks into the encoder of the neural dependency parser. LCM follows a pipeline-based approach consisting of two steps: pretraining and integration. Pretraining uses a sequence labelling paradigm and trains encoders for three independent auxiliary tasks. Later, these pretrained encoders are combined with the encoder of the neural parser via a gating mechanism similar to ~\newcite{sato-etal-2017-adversarial}. The LCM consists of three sequence labelling-based auxiliary tasks, namely, predicting the dependency label between a modifier-modified pair (\textbf{LT}), the monolithic morphological label \textbf{(MT)}, and the case attribute of each nominal \textbf{(CT)}.
Each auxiliary task leverages the same hyper-parameter settings and neural architecture and deviates only in terms of the output tags.
 The neural architecture employed for these auxiliary tasks consists of BiLSTM encoders similar to \newcite{DBLP:conf/iclr/DozatM17} and a decoder with two fully connected layers followed by a softmax layer. We encourage readers to refer \newcite[LCM]{sandhan-etal-2021-little} for more details.

 \subsection{Error Analysis}
\label{Error_analysis}
\begin{figure*}[!tbh]
\centering
\subfigure[\label{fig:UAS_error}]{\includegraphics[width=0.3\linewidth]{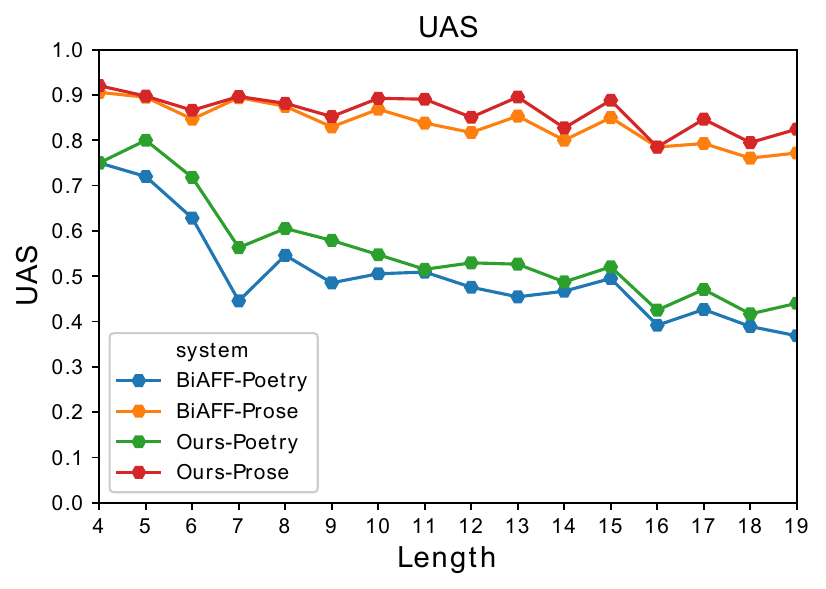}}
\subfigure[\label{fig:label-vocab}]{\includegraphics[width=0.3\linewidth]{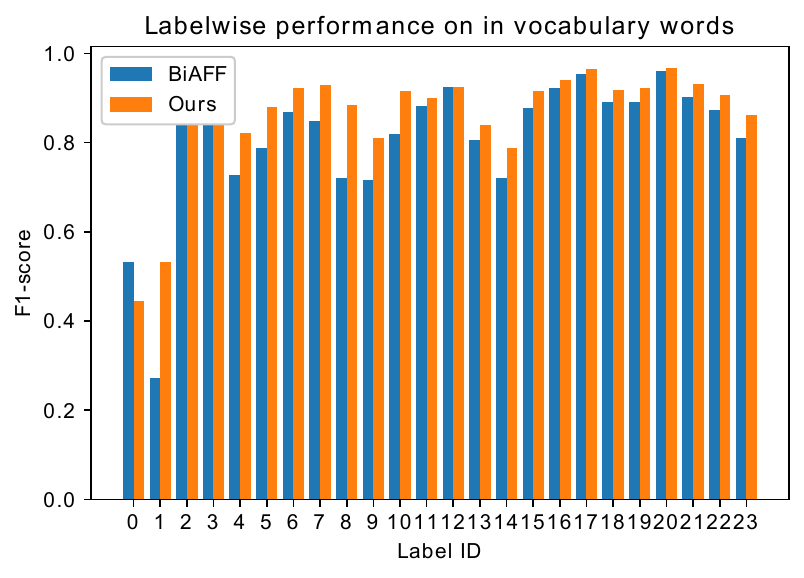}}
\subfigure[\label{fig:label-OOV}]{\includegraphics[width=0.3\linewidth]{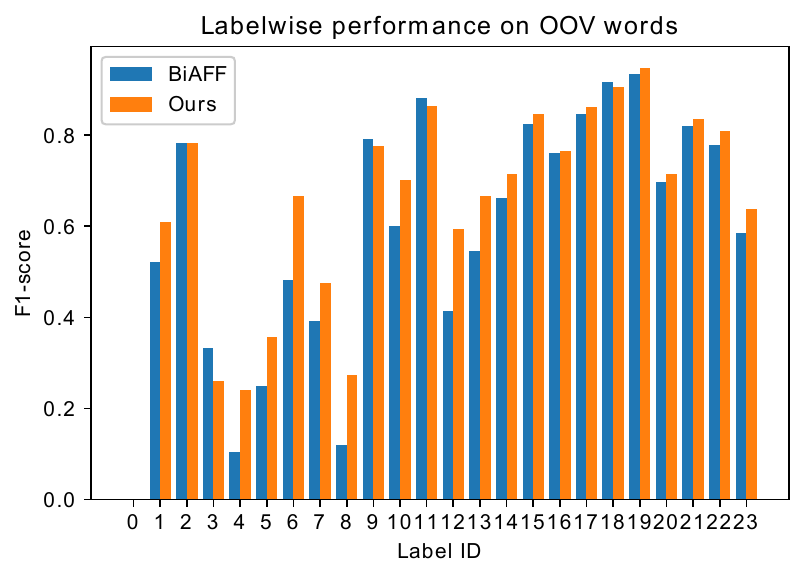}}\\
\subfigure[\label{fig:dependency_length}]{\includegraphics[width=0.3\linewidth]{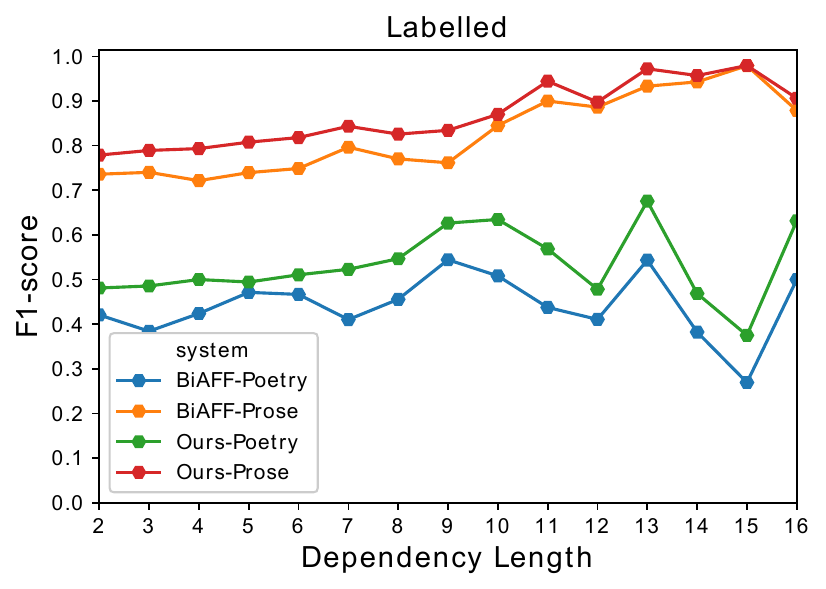}}
\subfigure[\label{fig:distance2root}]{\includegraphics[width=0.3\linewidth]{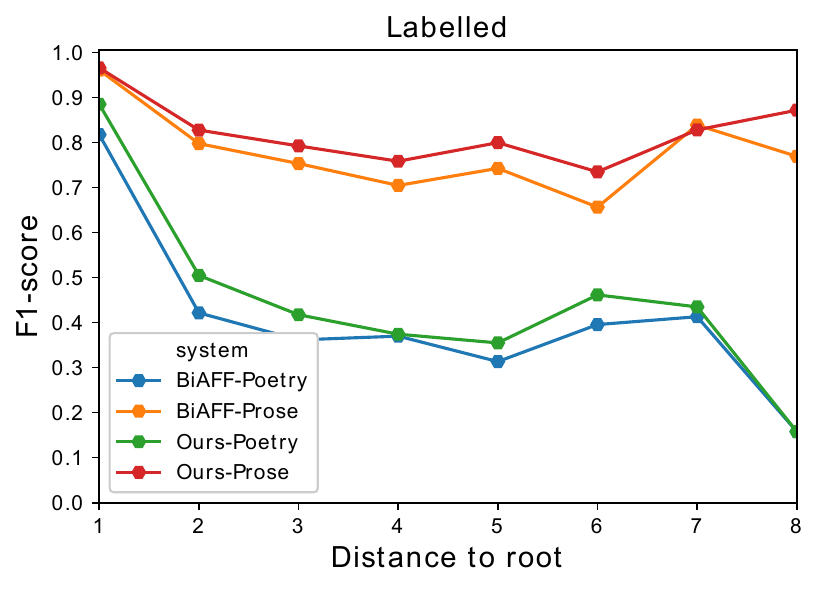}}
\subfigure[\label{fig:nonprojectivity}]{\includegraphics[width=0.3\linewidth]{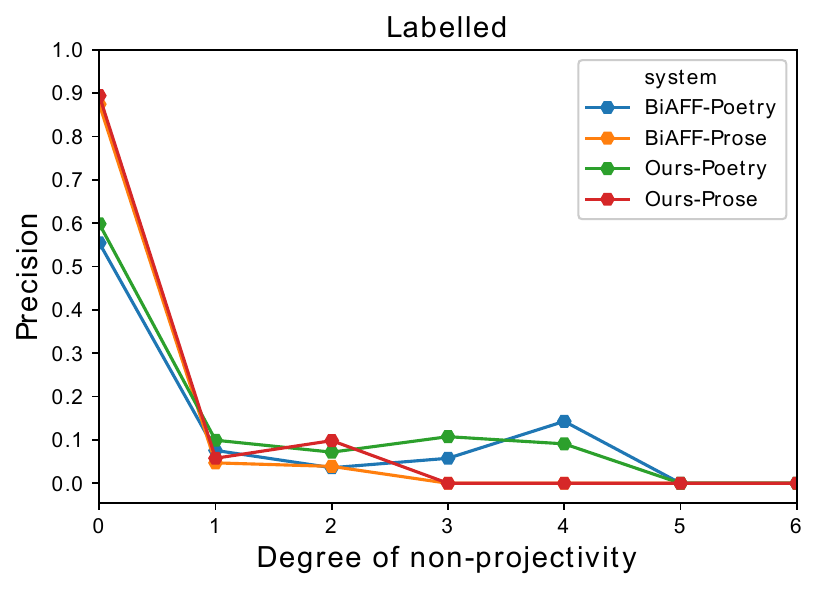}}
\caption{(a) Performance on prose and poetry domain as sentence length increases. The dependency label-wise performance on (b) in vocabulary (c) out of vocabulary words. From left to right the label frequency increases. Performance in terms of (d) dependency length (e) distance to root and (f) degree of non-projectivity.}
\label{fig:length_analysis}
\end{figure*}

\noindent We compare the ensembled system with the strong baseline BiAFF.\footnote{We could not compare with Tree-EBM-F* and MG-EBM* due to unavailability of codebase and system outputs.}
We investigate the robustness of systems based on the following language specific phenomenona such as (1) relatively free word order nature: Here,  we verify the robustness to the configurational information by evaluating on poetry and prose domains. 
We compare the strong baseline BiAFF and the ensembled system on test set of STBC dataset.
 The test set of STBC contains 300 datapoints from \textit{\`{S}i\`{s}up\={a}lavadha} \cite{ryalichallenges}. These 300 data points are also available in poetry order. Thus, we use this poetry version to report the results for the poetry.
Table~\ref{table:OOV_analysis} illustrates that the ensembled system shows consistently superior performance over BiAFF; however, both systems are brittle to poetry domain; on an average the performance degrades by 36.8/34.0 points (UAS/LAS) compared to prose counterpart. The prose/poetry data has 44.0/0.6\% word pairs with dependency length as 1, 13.5/30.7\% word pairs with dependency length more than 5 and 7.1/19.2\% non-projective arc. Clearly, these structural divergences explain the low performance on poetry domain due to systems trained on prose domain.   (2) morphologically rich nature: is investigated by focusing on out of vocabulary phenomenon.
 Being a morphologically rich language, the OOV phenomenon is crucial for modelling robust systems. The test set of STBC consists of 34\% words that are OOV.
 The prominent contribution to average drop (reported in Table~\ref{table:OOV_analysis}) of 7.9/12.6 points (UAS/LAS) comes from failure to predict rarely occurring dependency relation, corresponding to OOV words correctly. In order to verify the performance in OOV vocabulary scenario, we evaluate the dependency label-wise performance for out of vocabulary (OOV) words (Figure~\ref{fig:label-OOV}) and in vocabulary words (Figure~\ref{fig:label-vocab}).  The contrast between Figure~\ref{fig:label-vocab} and~\ref{fig:label-OOV} demonstrates that both systems perform poorly when a word is OOV and dependency label is rarely occurring.\footnote{Opposite observation for the frequency marked with 1 is due to the fact that the total number of candidates belonging to this label class is very less so systems are not able to generalize well for this class.}
  \begin{table}[h]
\centering
\begin{tabular}{|c|c|c|c|c|}
\hline
 &\multicolumn{2}{c|}{\textbf{BiAFF}}
&
\multicolumn{2}{c|}{\textbf{Ours}} \\\hline
\textbf{Domain} &\textbf{UAS}&\textbf{LAS}    & \textbf{UAS} &\textbf{LAS}      \\\hline
Prose  &85.88 & 79.55&\textbf{88.67} & \textbf{83.47}\\
Poetry&47.75&44.32&\textbf{52.86}&\textbf{50.45}\\\hline
In Vocab              & 86.25 & 81.76&\textbf{89.16} &
\textbf{85.64} \\
OOV              &79.01&69.97 &\textbf{80.37}&\textbf{72.26} \\ \hline
\end{tabular} 
    \caption{Fine-grained analysis in between the strong baseline: BiAFF and the ensembled system on test set of STBC dataset. The test set of STBC contains 300 datapoints from \textit{\`{S}i\`{s}up\={a}lavadha} \cite{ryalichallenges}. These 300 data points are also available in poetry order. Thus, we use these poetry version to report the results for the poetry.
     \label{table:OOV_analysis}}
\end{table}
 
 Figure~\ref{fig:UAS_error} illustrates the  empirical performance in terms of sentence length to validate the robustness of the ensembled system against varied sentence lengths. We notice downwards trend in the performance as the sentence length increases. Particularly, both systems drastically go down in poetry domain. Following \newcite{kulmizev-etal-2019-deep,mcdonald-nivre-2011-analyzing}, we analyze the performance in terms of  dependency length\footnote{The dependency length is defined as the distance between head-dependent pair when arranged linearly in a sentence.} (Figure~\ref{fig:dependency_length}), distance to root\footnote{The distance of a node to its root in a dependency tree.} (Figure~\ref{fig:distance2root}) and  degree of non-projectivity\footnote{The degree of non-projectivity for a head($x$)/dependent($y$) pair  is defined as the number of words occurrences in between $x$ and $y$ which are not part of decedents of $x$ and modify a word outside $x$ and $y$ window.} (Figure~\ref{fig:nonprojectivity}). If we contrast between the performance of both the systems in poetry and prose domain then  they show the declined performance in terms of sentence length, distance to root and degree of non-projectivity. In Figure~\ref{fig:dependency_length}, we observe slightly improved performance for both the systems in prose domain as the dependency length increases. This can be attributed to ability of graph-based parsers to capture long range dependencies well. Figure~\ref{fig:length_analysis} (d-f) only reports the performance on labelled score; however, we observe a similar trend for unlabelled score. 
Summarily, we conclude that both systems perform similarly in the all aspects of error analysis where the ensembled system shows consistent improvements. 

\section{Summary}
We focused on dependency parsing for low-resource MRLs, where getting morphological information itself is a challenge. To address low-resource nature and lack of morphological information, we proposed a simple pretraining method based on sequence labeling that does not require complex architectures or massive labelled or unlabelled data. We show that little supervised pretraining goes a long way compared to transfer learning, multi-task learning, and mBERT pretraining approaches (for the languages not covered in mBERT).
One primary benefit of our approach is that it does not rely on morphological information at run time; instead this information is leveraged using the pretrained encoders. Our experiments across 10 MRLs showed that proposed pretraining provides a significant boost with an average 2 points (UAS) and 3.6 points (LAS) absolute gain compared to DCST.

Finally, we focused on low-resource dependency parsing for multiple languages. We found that our ensembled system can benefit the languages not covered in pretrained models. While multi-lingual pretraining (mBERT and XLM-R) is helpful for the languages covered in pretrained models, LCM pretraining (which simply uses an additional 1,000 morphologically tagged data points) is helpful for the remaining languages. Thus, these findings would help community to pick strategies suitable for their language of interest and come up with robust parsing solutions.
Specifically for Sanskrit, our ensembled system superseded the performance of the state-of-the-art \textit{hybrid} system MG-EBM* by 1.2 points (UAS) absolute gain and showed comparable performance in terms of LAS.

\noindent\textbf{Applicability for the downstream tasks:} There exist diverse downstream tasks, including machine translation, word order linearisation (conversion of poetry to prose), and context-sensitive compound type identification, in which dependency parsing holds promising potential for utility. For instance, in our next contributing chapter concerning compound type identification, we have demonstrated the efficacy of dependency parsing in enhancing contextual information, resulting in improved performance in the downstream task. Theoretically, both machine translation and word order linearisation also exhibit a robust correlation with the dependency parsing task

%% file: Chapters/Chapter5.tex

\chapter{Compound Type Identification Task in Sanskrit}

\label{Chapter5} 

\lhead{Chapter 5. \emph{Compound Type Identification}}
 The phenomenon of compounding is ubiquitous in Sanskrit. It serves for achieving brevity in expressing  thoughts, while simultaneously enriching the lexical and structural formation of the language. In this chapter, we focus on the {\sl Sanskrit Compound Type Identification (SaCTI)} task, where we consider the problem of identifying semantic relations between the components of a compound word. 

\section{Task Description}
\label{task}
Compounding is a process whereby individual components or words are combined into a single word. Compounds are formed when there is compatibility or congruity between the components. However, in cases where there is no compatibility between the components and a compound is still formed, they are referred to as \textit{asamartha} compounds. These compounds are capable of conveying the intended sense despite the lack of compatibility between the components. The term ``compound'' is applied even when there is no mutual compatibility (\textit{asamartha})  between the words, due to the principle of \textit{gamakatva} (based on its frequent usage).
The process of decoding an implicit semantic relation between the components of a compound in Sanskrit is called {\sl Sanskrit Compound Type Identification (SaCTI)}. Alternatively, it is also termed as {\sl Noun Compound Interpretation (NCI)} \cite{ponkiya-etal-2021-framenet,ponkiya-etal-2020-looking}. In the literature, the NCI problem has been formulated in two related but different ways. Let's take {\sl mango juice} as an example. In the first formulation, the relation between the two components is labeled from  a set of semantic relations ({\sl MADEOF}) \cite{dima-hinrichs-2015-automatic,fares-etal-2018-transfer,ponkiya-etal-2021-framenet}. The second formulation uses paraphrasing to illustrate the semantic relations ({\sl a juice made from mango}) \cite{lapata-keller-2004-web,ponkiya-etal-2018-treat,ponkiya-etal-2020-looking}. In this chapter, we use the first formulation that frames the task as a multi-class classification problem. We propose an automated approach for semantic class identification of compounds in Sanskrit. It is essential to extract semantic information hidden in compounds for improving overall downstream Natural Language Processing (NLP) applications such as information extraction, question answering, machine translation, and many more.

In this chapter, we systematically investigate the following research question: Can recent advances in neural network outperform traditional hand engineered feature based methods on the semantic level multi-class compound classification task for Sanskrit?
Contrary to the previous methods, our method does not require feature engineering. For well-organized analysis, we categorize neural systems based on Multi-Layer Perceptron (MLP), Convolution Neural Network (CNN) and Long Short Term Memory (LSTM) architecture and feed input to the system from one of the possible levels, namely, word level, sub-word level, and character level. Our best system with LSTM architecture and FastText embedding with end-to-end training has shown promising results in terms of F-score (0.73) compared to the state of the art method based on feature engineering (0.74) and outperformed in terms of accuracy (77.68\%). 
 
 Earlier approaches solely rely on the lexical information obtained from the components and ignore the most crucial contextual and syntactic information useful for SaCTI. However, the SaCTI task is challenging primarily due to the implicitly encoded context-sensitive semantic relation between the compound components. Thus, we propose a novel multi-task learning architecture which incorporates the contextual information and enriches the complementary syntactic information using morphological tagging and dependency parsing as two auxiliary tasks. Experiments on the benchmark datasets for SaCTI show $6.1$ points (Accuracy) and $7.7$ points (F1-score) absolute gain compared to the state-of-the-art system. Further, our multi-lingual experiments demonstrate the efficacy of the proposed architecture in English and Marathi languages.

\section{Challenges in Compound Type Identification Task}
Compounds in Sanskrit can be categorized into 30 possible classes based on how granular categorizations one would like to have~\cite{lowe2015syntax}. There are slightly altered set of categorizations considered by~\newcite{gillon2009tagging},~\newcite{olsen2000composition},~\newcite{bisetto2005classification} and ~\newcite{tubb2007scholastic}.
Semantically \textit{A\d{s}\d{t}\={a}dhy\={a}y\={\i}} categorizes the Sanskrit compounds into four major semantic classes, namely, \textit{Avyay\={\i}bh\={a}va, Tatpuru\d{s}a, Bahuvr\={\i}hi} and \textit{Dvandva}~\cite{processor}. Similar to prior computational  approaches in Sanskrit compounding \cite{krishna2016compound,processor}, we follow this four class coarse level categorization of the semantic classes in compounds. 
Compounding in Sanskrit is extremely productive, or rather recursive, resulting in compound words with multiple components~\cite{lowe2015syntax}. Further, it is observed that compounding of a pair of components may result in compounds of different semantic classes. 
\textit{Avyay\={\i}bh\={a}va} and \textit{Tatpuru\d{s}a} may likely be confusing due to particular sub-category of \textit{Tatpuru\d{s}a} if the first component is an \textit{avyaya}. For example,  \textit{upa j\={\i}vata\d{h}} has the first component as \textit{avyaya} which is strong characteristic of \textit{Avyay\={\i}bh\={a}va}. However, this compound belongs to  \textit{Tatpuru\d{s}a} class. Likewise, a negation \textit{avyaya} in the first component can create confusion between \textit{Tatpuru\d{s}a} and \textit{Bahuvr\={\i}hi} classes. The instances mentioned above reveal the difficulties associated with distinguishing the semantic classes of a compound.

 The phenomenon of compounding is ubiquitous in Sanskrit. It serves for achieving brevity in expressing  thoughts, while simultaneously enriching the lexical and structural formation of the language. In this chapter, we focus on the {\sl Sanskrit Compound Type Identification (SaCTI)} task, where we consider the problem of identifying semantic relations between the components of a compound word. Earlier approaches solely rely on the lexical information obtained from the components and ignore the most crucial contextual and syntactic information useful for SaCTI. However, the SaCTI task is challenging primarily due to the implicitly encoded context-sensitive semantic relation between the compound components.

Thus, we propose a novel multi-task learning architecture which incorporates the contextual information and enriches the complementary syntactic information using morphological tagging and dependency parsing as two auxiliary tasks. Experiments on the benchmark datasets for SaCTI show $6.1$ points (Accuracy) and $7.7$ points (F1-score) absolute gain compared to the state-of-the-art system. Further, our multi-lingual experiments demonstrate the efficacy of the proposed architecture in English and Marathi languages.\footnote{The code and datasets are publicly available at: \url{https://github.com/ashishgupta2598/SaCTI}}

\section{Context-Sensitive Compound Type Identification}
The compound type identification task is challenging and often depends upon the context or world knowledge about the entities involved \cite{krishna-etal-2016-compound}. For instance, the semantic type of the compound {\sl r\={a}ma-\={\i}\'{s}vara\d{h}}  can be classified into one of the following semantic types depending on the context: {\sl Karmadh\={a}raya}\footnote{There are 4 broad semantic types of compounds: \textit{Avyay\={\i}bh\={a}va}, \textit{Bahuvr\={\i}hi}, \textit{Dvandva}, and \textit{Tatpuru\d{s}a}. {\sl Karmadh\={a}raya} is considered as sub-type of \textit{Tatpuru\d{s}a}. We encourage readers to refer \newcite{krishna-etal-2016-compound} for more details on these semantic types.}, {\sl Bahuvr\={\i}hi}  and {\sl Tatpuru\d{s}a}.
Although the compound has the same components as well as the final form, the implicit relationship between the components can be decoded only with the help of available contextual information \cite{kulkarni2013,krishna-etal-2016-compound}. Due to such instances, the downstream Natural Language Processing (NLP) applications for Sanskrit such as question answering \cite{terdalkar-bhattacharya-2019-framework} and machine translation \cite{aralikatte-etal-2021-itihasa}, etc. show sub-optimal performance when they stumble on compounds. 
For example, while translating {\sl r\={a}ma-\={\i}\'{s}vara\d{h}} into English, depending on the semantic type, there are four possible meanings: (1) Lord who is pleasing (in {\sl Karmadh\={a}raya})
(2) the one whose God is Rama (in {\sl Bahuvr\={\i}hi}) (3) Lord of Rama (in {\sl Tatpuru\d{s}a}) (4) Lord Rama and Lord Shiva ({\sl Dvandva}).
Therefore, the SaCTI task can be seen as a preliminary pre-requisite to building a robust NLP technology for Sanskrit. Further, this dependency on contextual information rules out the possibility of storing and doing a lookup to identify a compound's semantic types. 

 With the advent of recent contextual models \cite{peters-etal-2018-deep,devlin-etal-2019-bert,conneau-etal-2020-unsupervised}, there has been upsurge in performance of various downstream NLP applications \cite{kondratyuk-straka-2019-75,roberta,NEURIPS2019_dc6a7e65}.
 Nevertheless, there have been no efforts to build context-dependent models in SaCTI.  This may be attributed to the fact that while most of the natural language technology is built for resource-rich languages such as English \cite{joshi-etal-2020-state}, compounding is not a predominant phenomenon in them \cite{krishna-etal-2016-compound}. 
 There is also lack of task-specific context-sensitive labeled data.

Earlier approaches \cite{kulkarni2013,krishna-etal-2016-compound,sandhan-etal-2019-revisiting} for SaCTI solely rely on lexical information obtained from the components and are blind to potentially useful contextual and syntactic information. 
The context is the most feasible, cheaply available information. 
As per Pa\d{n}ini's grammar \cite{panini,kulkarni2013}, the morphological features have direct correlation with the semantic types. Sometimes, the dependency information also helps is disambiguation and can provide a medium to enrich contextual information. Thus, we propose a novel multi-task learning approach which (1) incorporates the contextual information, and (2) enriches the complementary syntactic information using morphological tagging and dependency parsing auxiliary tasks without any additional manual labeling. 
  Summarily, our key contributions are:
  \begin{itemize}
      \item We propose a novel context-sensitive multi-task learning architecture for SaCTI (\S~\ref{proposed-system}).
      \item We illustrate that morphological tagging and dependency parsing auxiliary tasks  are helpful and serve as a proxy for explainability of system predictions (\S~\ref{analysis}) for the SaCTI task.
      \item We report results with $7.71$ points (F1) absolute gains compared to the current state-of-the-art system by \newcite{krishna-etal-2016-compound} (\S~\ref{main-results}). 
       \item We show the efficacy of the proposed approach in English and Marathi (\S~\ref{analysis}).
       \item We release our codebase and pre-processed datasets and web-based tool (\S~\ref{analysis}) for using our pretrained models.  
  \end{itemize}

\subsection{Methodology}
\label{proposed-system}
\begin{figure}[!htb]
\centering
\includegraphics[width=0.6\textwidth]{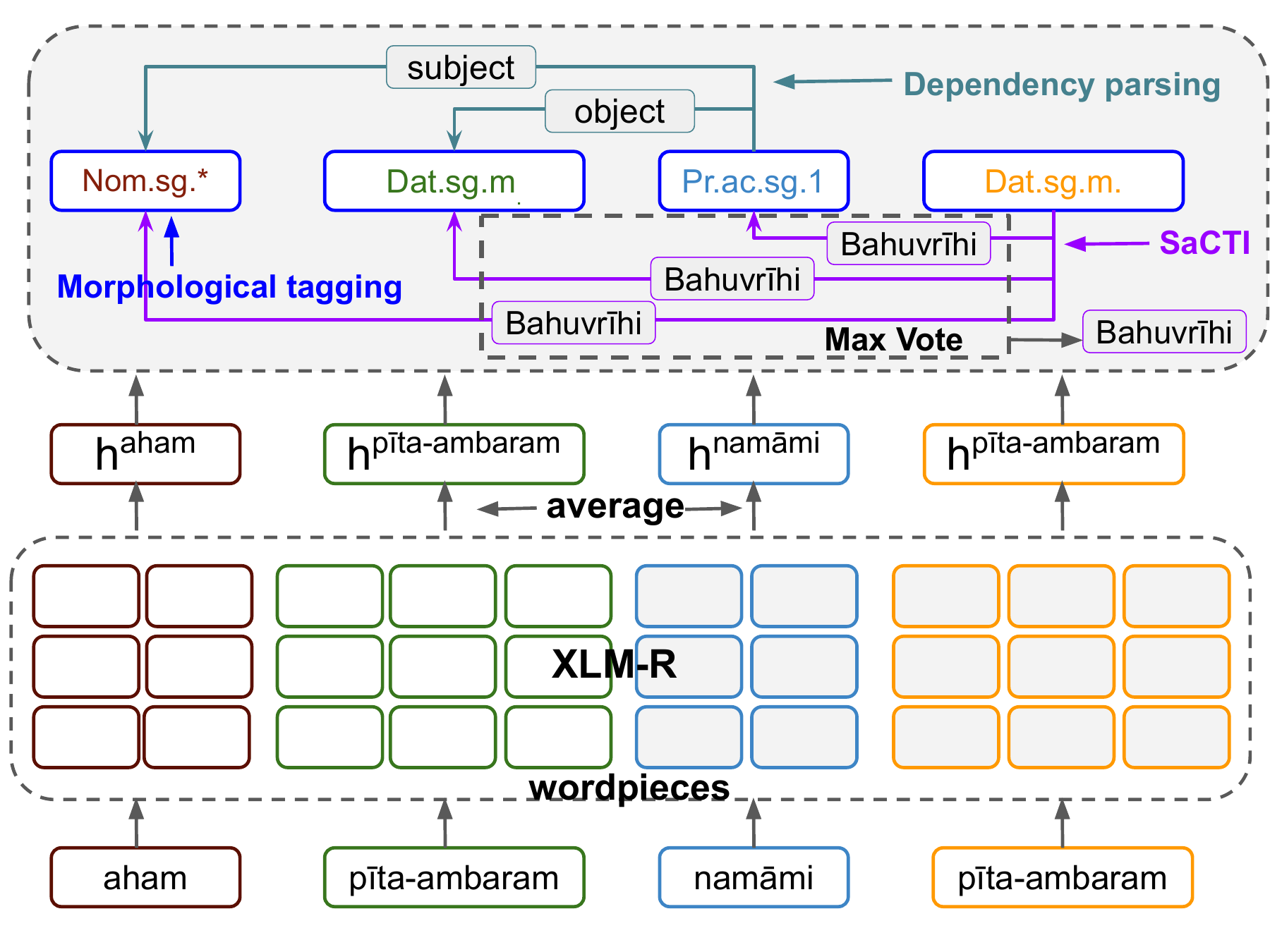}
\caption{Illustration of the proposed multi-task learning architecture with an example ``{\sl aham p\={\i}ta-ambaram nam\={a}mi}'' (Translation: ``I pray to P\={\i}t\={a}mbara (Lord Vishnu).'') where `{\sl p\={\i}ta-ambaram}' is a compound word belonging to the {\sl Bahuvr\={\i}hi} semantic type as per the given context.  We feed this context and the compound word at the end as an input to the system. Each token is split into wordpieces using a multi-lingual tokenizer \cite{kudo-richardson-2018-sentencepiece}. This sequence of wordpieces is passed to multi-lingual pretrained XLM-R encoder \cite{conneau-etal-2020-unsupervised}. The hidden representation of each token is the average of its wordpieces' representations obtained from the encoder. We apply our multi-task learning architecture which consists of three tasks, namely, Sanskrit compound type identification (SaCTI), morphological tagging and dependency parsing over the hidden representations. We formulate SaCTI as a pair-wise (a context word and the compound) classification task where the objective is to predict the semantic type of the target compound word ({\sl Bahuvr\={\i}hi}). At test time, we apply the maximum voting policy to select a single prediction from the set of semantic relations predicted by such $n$ pairs.} 
\label{fig:main_model} 
\end{figure}
Figure~\ref{fig:main_model} illustrates the proposed multi-task learning architecture with an example context, ``{\sl aham p\={\i}ta-ambaram nam\={a}mi}'' (Translation: ``I pray to P\={\i}t\={a}mbara (Lord Vishnu).'') where `{\sl p\={\i}ta-ambaram}' is a compound belonging to the {\sl Bahuvr\={\i}hi} semantic type as per the given context.  As shown in Figure~\ref{fig:main_model}, we feed this context along with the compound word concatenated at the end, as an input to the system, and obtain hidden representations from the multi-lingual encoder as described below.  
On top of the hidden representations as obtained via the encoder module, we apply our multi-task learning architecture consisting of three tasks: SaCTI, morphological tagging, and dependency parsing. We formulate the SaCTI task as a pair-wise (a context word paired with the compound word) classification task where the objective is to predict the semantic type of the target compound word ({\sl Bahuvr\={\i}hi} for {\sl p\={\i}ta-ambaram} compound word in this example) for all the pairs as shown in Figure~\ref{fig:main_model}. At test time, we apply the maximum voting policy 
to select a single prediction from the set of semantic types predicted by these pairs. We formally discuss the details of each component below.

\noindent\textbf{Multilingual Encoder:} Sanskrit is a morphologically rich and low-resource language. In order to build powerful contextual representations for Sanskrit words, morphological richness poses the out-of-vocabulary problem and low-resource nature poses an unlabelled data scarcity problem. Thus, we opt for a multi-lingual encoder \cite[XLM-R]{conneau-etal-2020-unsupervised} to mitigate these issues. 

Given a compound $c_p$ and its context $C= [c_1, c_2, ..., c_n]$ such that $p^{th}$ position $(1 \leq p \leq n)$ in the context is the compound word, we append the compound word to the context such that $C= [c_1, c_2, ..., c_n, c_{n+1}]$ where $c_{n+1} = c_p$. Each token $(c_i)$ is further split into wordpieces $(c_i = [c_i^1,c_i^2,..., c_i^{m_i}])$ using a multi-lingual subword tokenizer \cite[Sentencepiece]{kudo-richardson-2018-sentencepiece,kudo-2018-subword}. Next, we pass the overall sequence $(C = [c_1^1,c_1^2,..., c_n^{m_n},c_{n+1}^{1}..., c_{n+1}^{m_{n+1}}])$ of wordpieces corresponding to the context $C$ into the pretrained transformer. Finally, we obtain the contextual representation of all tokens as $h = (h_1, h_2, h_3, ..., h_n, h_{n+1})$ where 
\begin{equation}
    h_i = \frac{1}{m_i}\sum_{k=1}^{m_i}Transformer(c_i^k)
\end{equation}
\noindent\textbf{SaCTI:} Our context-sensitive classifier uses Bi-Affine attention, henceforth referred to as BiAFF. 
Given the hidden representations of $i^{th}$, $j^{th}$ context words as $h_i$, $h_j$ from the multi-lingual encoder, the scoring function $s_{i,j}$ 
indicates the system's belief that the
latter $(j^{th})$ (Eqn.~\ref{eq1}) should be related to the former $(i^{th})$  in identifying the semantic type of the latter, where $q_i^{T}z_i$ indicates bias to capture the prior of contextual information in the $i^{th}$ word.
\begin{equation}
s_{i,j} = z_i^{T}Uz_j + q_i^{T}z_i \\ \label{eq1}
\end{equation}
where $z_i = MLP(h_i)$, $U$ and $q_i$ are learnable parameters, $MLP$ denotes a multi-layered perceptron. 
Similarly, a score for $k^{th}$ possible semantic type relation between every pair of $i^{th}$ context word ($\forall i \in [1,n]$) and the compound is computed by:
\begin{equation}
r_{i,k} = z_i^{'T}U_k^{'}z_{n+1}^{'} + q_k^{'T}[z_i^{'};z_{n+1}^{'}] +b_k^{'}\\
\end{equation}
where $z_i^{'} = MLP^{'}(h_i)$, $U^{'}$, $b_k^{'}$ and $q_k^{'}$ are learnable parameters. Finally, model maximizes the following objective function during training.
\begin{equation}
  \sum_{i=1}^{n} p(y_{n+1}|c_i,\theta) + p(y_{n+1}^l|c_i,y_{n+1},\theta) \\ \label{eq3}
\end{equation}
where $y_{n+1}$ is the target compound appended in context $C$, $y_{n+1}^{l}$ is the semantic type of $(y_{n+1}/c_i,\theta)$, $\theta$ denotes system's parameters, $p(y_{n+1}|c_i,\theta) \propto exp(s_{i,n+1})$, $p(y_{n+1}^l|c_i,y_{n+1},\theta) \propto exp(r_{i,l})$. At test time, we apply maximum voting policy to select a single prediction from the set of semantic relations predicted for $n$ pairs $(c_i,c_{n+1}), \forall i \in [1,n]$. 

\noindent\textbf{Morphological tagging:}  The primary motivation behind using morphological tagging as an auxiliary task aligns well with grammatical rules from Pa\d{n}ini's grammar \cite{panini,kulkarni2013}. For instance, {\sl Avyay\={\i}bh\={a}va} compounds are in neuter gender. {\sl Tatpuru\d{s}a} is a function of `case' attribute of morphological features. The number attribute of a compound depends on the semantic type of the compound. Also, there are constraints based on inflection/derivational suffix.  Summarily, these morphological features have direct correlation with the semantic classes. 
In our proposed system, the morphological tagging task leverages hidden representations from the multi-lingual encoder and decodes the pseudo-labels\footnote{The benchmark datasets do not have a gold standard morphological information. We use predicted morphological information as pseudo-labels (\S~\ref{datasets}).} using a fully connected layer followed by a softmax layer.\footnote{Note that all the parameters present in the multi-lingual encoder are trainable during the task-specific training.} 
In this process, morphological information useful for the SaCTI task is enriched in the hidden representations. This can be seen as an implicit way to encode the grammatical rules in the system.

\noindent\textbf{Dependency parsing:} 
\newcite{syntax} argues that compounding is mostly a syntactic phenomenon. {\sl Bahuvr\={\i}hi} compounds are ``exocentric'' in nature, which attribute a property to an entity external to the compound with the adjective relationship.  Sometimes, syntactic information can provide a complementary signal useful for compound type disambiguation. For instance, consider the following example: {\sl aham n{\=\i}la-utpala\d{h} sara\d{h} pa\'sy\=ami} (Translation: I watch the pond having a blue-lotus.) Here, {\sl n{\=\i}la-utpalam} qualifies to be {\sl Bahuvr\={\i}hi} due to presence of its referent {\sl sara\d{h}} with an adjective relationship. However, in the absence of {\sl sara\d{h}} in the context, ambiguity pops up in between {\sl Bahuvr\={\i}hi} and {\sl Tatpuru\d{s}a}.\footnote{The ambiguity is whether I am seeing the blue lotus or the pond having a blue lotus.}
This motivates us to investigate the usefulness of syntactic information for the SaCTI task. The benchmark datasets do not have a gold standard dependency information. We use predicted dependency trees as pseudo-labels (\S~\ref{datasets}).  Our dependency parsing component leverages Bi-Affine parser \cite{DBLP:conf/iclr/DozatM17} over hidden representations from multi-lingual encoder.

\subsection{Experiments}
\noindent\textbf{Dataset and metrics}
\label{datasets}
Table~\ref{table:data} reports the unique number of compounds, data statistics and the number of semantic types for the respective datasets used in this chapter.  We restrict to binary compounds (compounds with two components) in all the datasets. These datasets consist of components, context and semantic type of a compound. The SaCTI datasets for Sanskrit are available with two levels of annotations: coarse (4 broad types) and fine-grained (15 sub-types).
\begin{table}[h]
\centering
\begin{adjustbox}{width=0.6\textwidth}
\small
\begin{tabular}{|c|c|c|c|c|c|}
\hline
\textbf{Datasets}  &\textbf{\#Unique}  & \textbf{\#Train}     & \textbf{\#Dev}     & \textbf{\#Test}  & \textbf{\#Types}    \\ \hline
SaCTI-base   &8,594& 9,356  & 2,339 &2,994& 4 (15) \\\hline
SaCTI-large &48,132&59,133 & 6,571 & 7,301 & 4 (15)  \\\hline
English &4,676& 4,163 & 1,041 & 1,301 & 7 \\\hline
Marathi  &368&659 &99  &114  &4 \\\hline
\end{tabular}
\end{adjustbox}
\caption{Data statistics for all the datasets} 
\label{table:data}
\end{table}

\noindent\textbf{Sanskrit:} We evaluate on two available context-sensitive benchmark datasets: {\sl SaCTI-base} and {\sl SaCTI-large}.
We follow the same experimental settings as \newcite{krishna-etal-2016-compound} in {\sl SaCTI-base} to keep our results comparable with their state-of-the-art results. {\sl SaCTI-base} is a subset of {\sl SaCTI-large} dataset. In due course of time, more annotated data resulted in {\sl SaCTI-large} dataset.

\noindent\textbf{English:} We use instance-based (context-dependent) noun-noun compound dataset released by \newcite{fares-2016-dataset}.
The compounds used in this dataset are extracted from the Wall Street Journal (WSJ) portion in the Penn Treebank (PTB). 

\noindent\textbf{Psuedo-labels for auxiliary tasks:} The benchmark datasets do not have a gold standard dependency and morphological information. We use predicted labels as pseudo-labels. For Sanskrit, we obtain pseudo-labels from the Trankit model \cite{nguyen2021trankit} trained on STBC dataset \cite{krishna-etal-2020-keep} and  morphological pseudo-labels from the LemmaTag model \cite{kondratyuk-etal-2018-lemmatag} trained on Hackathon dataset \cite{hackathon_data}.  For English, we use pseudo-labels from English XLM-R model\footnote{\url{https://spacy.io/models/en}} for morphological and dependency parsing task. For Marathi, we do not find any such data or pretrained model to obtain pseudo-labels. Therefore, we do not activate morphological tagging and dependency parsing components in the proposed system while training on Marathi data.

 \noindent\textbf{Hyper-parameter settings:} For the implementation of the proposed system, we modify on top of codebase by \newcite{nguyen2021trankit}. We use the following hyper-parameter settings for the best configuration of the proposed system: number of epochs as 70, batch size 50 and a embedding dropout rate of 0.3 with a learning rate of 0.001. In our multi-task loss function, we penalize dependency component's loss function by 0.01 to prioritize the performance on SaCTI. This penalty is identified based on hyper-parameter tuning. The rest of the hyper-parameters are kept the same as \newcite{nguyen2021trankit}. For multi-lingual baselines, we used Huggingface’s transformers repository \cite{wolf-etal-2020-transformers}.
 We release our codebase and datasets publicly under the licence CC-BY 4.0. All the artifacts used in this chapter are publicly available for the research purpose. 
 
 \noindent\textbf{Computing infrastructure used:} We use a single GPU with Tesla P100-PCIE, 16 GB GPU memory, 3584 GPU Cores computing infrastructure for our experiments. Our proposed system takes approximately 1 hour for training SaCTI-base coarse w/o context setting dataset. 

\noindent\textbf{Evaluation metrics:} Following \newcite{krishna-etal-2016-compound,sandhan-etal-2019-revisiting}, we report macro averaged \textbf{P}recision, \textbf{R}ecall and \textbf{F1}-score for all our experiments. 
We also report micro-averaged \textbf{A}ccuracy. We use Scikit-learn software \cite{scikit-learn} to compute these metrics.

\noindent\textbf{Baselines:}  
We consider two context agnostic systems where \newcite[\textbf{ISCLS19}]{sandhan-etal-2019-revisiting} formulate SaCTI as a purely neural-based multi-class classification approach using static word embeddings of components of a compound and   \newcite[\textbf{COLING16}]{krishna-etal-2016-compound} deploy a hybrid system which leverages linguistically involved hand-crafted feature engineering with distributional information from Adaptor Grammar \cite{Johnson2006AdaptorGA}. The COLING16 system is the current state-of-the-art system for the SaCTI task. Next, we opt for multi-lingual contextual language models due to the lack of sufficiently large unlabelled data available for Sanskrit.
We consider three multi-lingual pretrained language models, namely, \newcite[\textbf{IndicALBERT}]{kakwani2020indicnlpsuite} which is ALBERT model trained on 12 Indic languages excluding Sanskrit, BERT \cite[\textbf{mBERT}]{devlin-etal-2019-bert} trained on 104 languages having largest Wikipedia's excluding Sanskrit and \newcite[\textbf{XLM-R}]{conneau-etal-2020-unsupervised} trained on 100 languages including Sanskrit. Finally, we consider a mono-lingual ALBERT model~\cite[\textbf{SanALBERT}]{jivnesh_eval} trained on Sanskrit corpus \cite{hellwig2010dcs} from scratch. In all the contextual baselines, we pass the classification token [CLS] representation of a sentence pair (compound word and its context separated by [SEP] token) to the classification head.
\textbf{Ours:}  This is our proposed system from \S~\ref{proposed-system}.

\noindent\textbf{Results}
\label{main-results}
Table~\ref{table:main_results} shows the performance for the best performing configurations of all the baselines on the test set of benchmark datasets for SaCTI.
We report results on two levels of annotations (coarse and fine-grained) and two settings (w/o context and w/ context).\footnote{For the systems that require context, we feed the compound word only as the context.}
Except for ISCLS19 and COLING16 systems, all systems utilize available context along with components of a compound.\footnote{These baselines cannot utilize the context; therefore, we report the same numbers in w/context as w/o context.}
XLM-R reports the best performance among all the baselines while using the context information. 
 \begin{table*}[bht]
\begin{small}
    \centering
\resizebox{1\textwidth}{!}{%
 \begin{tabular}{ccccccccccccccccccc}
\toprule
& &\multicolumn{8}{c}{\textbf{Sanskrit (coarse)}} &\multicolumn{8}{c}{\textbf{Sanskrit (fine-grained)}} 
 \\\cmidrule(r){3-10}\cmidrule(l){11-18}
 & &\multicolumn{4}{c}{\textbf{w/o context }} &\multicolumn{4}{c}{\textbf{w/ context}} &\multicolumn{4}{c}{\textbf{w/o context }} &\multicolumn{4}{c}{\textbf{w/ context}} 
 \\
 \cmidrule(r){2-2}\cmidrule(r){3-6}\cmidrule(l){7-10}\cmidrule(l){11-14} \cmidrule(l){15-18}
\textbf{Datasets}&\textbf{System} &\textbf{A}&\textbf{P}    & \textbf{R} &\textbf{F1}   &\textbf{A}&\textbf{P}    & \textbf{R} &\textbf{F1}&\textbf{A}&\textbf{P}    & \textbf{R} &\textbf{F1}&\textbf{A}&\textbf{P}    & \textbf{R} &\textbf{F1} 
\\\cmidrule(r){2-2}\cmidrule(r){3-6}\cmidrule(l){7-10}\cmidrule(l){11-14} \cmidrule(l){15-18}
&ISCLS19     & 77.68          & 76.00          & 71.00          & 73.00          & 77.68          & 76.00          & 71.00          & 73.00          & 70.64          & 67.53          & 63.18          & 64.58          & 70.64          & 67.53          & 63.18          & 64.58          \\
&COLING16    & 77.39          & \textbf{78.00} & 72.00          & \textbf{74.00} & 77.39          & 78.00          & 72.00          & 74.00          & -              & -              & -              & -              & -              & -              & -              & -              \\
&SanALBERT   & 72.01          & 72.10          & 70.00          & 71.40          & 77.40          & 77.31          & 73.00          & 74.20          & 69.39          & 62.58          & 58.13          & 58.22          & 75.40          & 70.14          & 61.22          & 62.04          \\
SaCTI-base&IndicALBERT & 71.87          & 51.60          & 53.90          & 52.00          & 78.63          & 77.47          & 75.83          & 76.47          & 68.06          & 53.05          & 53.19          & 52.30          & 71.93          & 57.29          & 56.49          & 55.25          \\
&mBERT       & 76.12          & 77.43          & 71.68          & 73.10          & 81.42          & 83.12          & 73.54          & 78.09          & 75.00          & 67.65          & 72.69          & 68.56          & 78.36          & 77.41          & 69.59          & 69.37          \\
&XLM-R        & 78.19          & 73.54          & 73.10          & 73.31          & 81.00          & 82.01          & 77.00          & 79.10          & 73.75          & 66.38          & 63.96          & 65.16          & 77.94          & 71.72          & 70.18          & 70.94          \\
&Ours        & \textbf{80.21} & 72.31          & \textbf{74.50} & 73.38          & \textbf{83.45} & \textbf{79.65} & \textbf{83.87} & \textbf{81.71} & \textbf{78.25} & \textbf{72.94} & \textbf{73.81} & \textbf{72.68} & \textbf{82.47} & \textbf{76.87} & \textbf{79.08} & \textbf{77.20}  \\
\cmidrule(r){2-2}\cmidrule(r){3-6}\cmidrule(l){7-10}\cmidrule(l){11-14} \cmidrule(l){15-18}
&ISCLS19     & 90.69          & 75.66          & 72.09          & 73.76          & 90.69          & 75.66          & 72.09          & 73.76          & 76.66          & 71.62          & 65.40          & 68.09          & 76.66          & 71.62          & 65.40          & 68.09          \\
&SanALBERT   & 88.17          & 66.65          & 61.31          & 62.85          & 87.84          & 64.15          & 65.44          & 63.63          & 79.38          & 69.90          & 75.34          & 67.57          & 80.73          & 71.62          & 74.60          & 72.69          \\
SaCTI-large&IndicALBERT & 87.77          & 68.82          & 49.58          & 56.05          & 92.95          & 84.98          & 74.90          & 79.32          & 79.03          & 68.00          & 64.14          & 63.67          & 83.13          & 74.23          & 79.77          & 76.11          \\
&mBERT       & 92.29          & 78.51          & 77.45          & 77.41          & 93.52          & 83.13          & 80.82          & 81.83          & 81.56          & 70.82          & 76.91          & 72.74          & 80.98          & 70.00          & 79.13          & 72.80          \\
&XLM-R        & 92.61          & 79.91          & 79.00          & 79.42          & 93.85          & 86.64          & 79.67          & 82.78          & 81.84          & 74.46          & 77.93          & 75.68          & 83.12          & 73.97          & 81.07          & 76.20          \\
&Ours        & \textbf{93.54} & \textbf{81.30} & \textbf{81.65} & \textbf{81.47} & \textbf{94.78} & \textbf{83.89} & \textbf{87.61} & \textbf{85.64} & \textbf{82.85} & \textbf{74.94} & \textbf{78.12} & \textbf{76.49} & \textbf{84.73} & \textbf{78.53} & \textbf{80.30} & \textbf{77.13}  \\
\hline
\end{tabular}}
    \caption{Evaluation on Sanskrit datasets in two levels of annotations (coarse and fine-grained) and two settings (w/o context and w/ context).  The best results are bold.  Results are averaged over 4 runs.
    The significance test between the best baseline XLM-R and the proposed system in terms of Recall/Accuracy metrics: $p < 0.01$ (as per t-test). We could not perform significance test with COLING16 (SaCTI-base-coarse-w/o) and report its results on all the datasets due to unavailability of its predictions and codebase. 
    }
    \label{table:main_results}
    \end{small}
\end{table*}

Our proposed system outperforms all the competing systems in terms of all the evaluation metrics and reports $6.1$ points (A) and $7.7$ points (F1) absolute gain with respect to the current state-of-the-art system COLING16 (on SaCTI-base coarse w/context dataset). COLING16 outperforms the proposed system  on SaCTI-base coarse w/o context. 
Notably, our proposed system outperforms the strong baseline XLM-R with large margins in fine-grained setting on SaCTI-base dataset (low-resourced setting). This confirms the usefulness of the proposed system in low-resourced settings with fine-grain labels.
The large performance gap between the proposed system with context and COLING16/ISCLS19 baselines illustrates the efficacy of using contextual information and syntactic information. 
Summarily, we mark new state-of-the-art results with the help of the novel architecture, where the contextual component is integrated with syntax-based auxiliary tasks such as morphological tagging and dependency parsing.
We find a similar trend in performance for the SaCTI-large dataset.

\subsection{Analysis}
\label{analysis}
In this section, we dive deep into the proposed system architecture for a detailed analysis as well as generalizability. We use SaCTI-base coarse dataset in the w/ context setting for the analysis.

\noindent\textbf{(1) Ablation analysis:} Here, we investigate the contribution of various components towards the overall improvements of the proposed system. Table~\ref{table:ablation} reports ablations in terms of all the evaluation metrics when a particular component is inactivated from the proposed system. For example, ``-DP'' denotes the system where the dependency parsing component is removed from the proposed system. We see that elimination of any of the components deteriorates the performance. Table~\ref{table:ablation} illustrates that `context' component is the most critical towards improvements. Also, the deletion of the `BiAFF' component has the second largest impact on the final performance.\footnote{In the absence of the proposed `BiAFF' component, we use [CLS] token for the sentence-level prediction, where this system is similar to XLM-R + DP + morph.} 

\begin{table}[h]
\centering
\begin{adjustbox}{width=0.45\textwidth}
\small
\begin{tabular}{|c|c|c|c|c|}
\hline
\textbf{System}                           & \textbf{A}     & \textbf{P}     & \textbf{R}  & \textbf{F1}    \\ \hline
Ours        & 83.45 & 79.65 & 83.87 & 81.71 \\\hline
-context    & 80.21 & 72.31 & 74.50 & 73.38 \\\hline
-BiAFF      & 81.00 & 82.01 & 77.00 & 79.10 \\\hline
-morph  & 82.87 & 80.00 & 81.21 & 80.60 \\\hline
-DP    & 81.89 & 79.35 & 81.62 & 80.26 \\ \hline
-morph -DP & 81.50 & 82.34 & 77.67 & 79.93 \\ \hline
\end{tabular}
\end{adjustbox}
\caption{Ablations of the proposed system
in terms of all the metrics. Each ablation deletes a single component from the proposed system. For example, “-DP” deletes the dependency parsing task from the proposed system.} 
\label{table:ablation}
\end{table}

\begin{figure}[!h]
\centering
\subfigure[]{\includegraphics[width=2in]{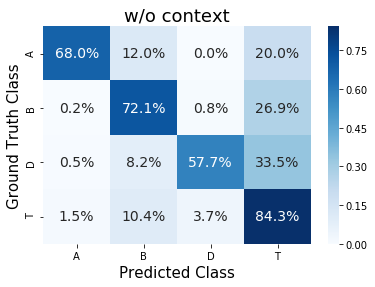}}
\subfigure[]{\includegraphics[width=2in]{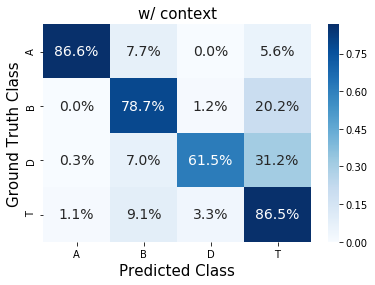}}\\
\caption{The confusion matrix for the proposed system trained (a) w/o context (b) w/ context. Semantic types: \textit{\textbf{A}vyay\={\i}bh\={a}va}, \textit{\textbf{B}ahuvr\={\i}hi}, \textit{\textbf{D}vandva}, and \textit{\textbf{T}atpuru\d{s}a}} 
\label{fig:error} 
\end{figure}

\noindent\textbf{(2) How effective is the proposed system in reducing confusion between conflicting classes?}
Figure~\ref{fig:error} illustrates the confusion matrices in w/ context and w/o context scenarios. We observe a similar trend in both the scenarios.  (1) Both systems mis-classify the predictions into the most populated type (\textit{Tatpuru\d{s}a}). This can be attributed to the imbalanced nature of the dataset.\footnote{In this chapter, we do not consider any strategy to tackle imbalanced classification. We plan to address this in future.} (2) The confusion between \textit{Avyay\={\i}bh\={a}va} and \textit{Tatpuru\d{s}a} is due to these compounds having their first component as an indeclinable word. Notably, the system with context is able to reduce confusion by 15\%. (3) One of the reasons for conflict between \textit{Tatpuru\d{s}a} and \textit{Bahuvr\={\i}hi} is due to the specific subcategory of both the classes where the first component is a negation. With the help of enriched information, a system with context can reduce this miss-classification by 7\%. Summarily, the system with contextual information always performs superior to one with no context.
This substantiates the importance of contextual information in the proposed system. 

\noindent\textbf{(3) How well can we generalize the proposed system for other languages?}
The primary motivation is to illustrate the efficacy of our language agnostic approach. The semantic type of compounds of the language of interest need not be similar to that of Sanskrit for the model to work. It is purely language agnostic model. To study the generalization ability of the proposed system, we consider 2 additional languages, namely,  
English (en) and Marathi (ma).  We choose English due to its availability of context-sensitive annotated data and Marathi due to its closeness to Sanskrit.
 \begin{table*}[bht]
\begin{small}
    \centering
\resizebox{1\textwidth}{!}{%
 \begin{tabular}{cccccccccccccccccc}
\toprule
 &\multicolumn{8}{c}{\textbf{English}} &\multicolumn{8}{c}{\textbf{Marathi}} 
 \\\cmidrule(r){2-9}\cmidrule(l){10-17}
  &\multicolumn{4}{c}{\textbf{w/o context}} &\multicolumn{4}{c}{\textbf{w/ context}} &\multicolumn{4}{c}{\textbf{w/o context}} &\multicolumn{4}{c}{\textbf{w/ context}}  \\
 \cmidrule(r){2-5}\cmidrule(l){6-9}\cmidrule(l){10-13} \cmidrule(l){14-17}
\textbf{System} & \textbf{A}     & \textbf{P}     & \textbf{R}     & \textbf{F1}    & \textbf{A}     & \textbf{P}     & \textbf{R}     & \textbf{F1}    & \textbf{A} & \textbf{P} & \textbf{R} & \textbf{F1} & \textbf{A} & \textbf{P} & \textbf{R} & \textbf{F1} \\
 \cmidrule(r){2-5}\cmidrule(l){6-9}\cmidrule(l){10-13} \cmidrule(l){14-17}
ISCLS19     & 67.43          & 71.81          & 64.38          & 66.26          & 67.43          & 71.81          & 64.38          & 66.26          & 68.62          & 70.78          & 52.89          & 56.85          & 68.62          & 70.78          & 52.89          & 56.85          \\
IndicALBERT & 68.23          & 57.25          & 59.70          & 57.96          & 70.22          & 68.01          & 69.31          & 68.65          & 67.65          & 33.33          & 45.15          & 38.08          & 60.00          & 45.17          & 40.62          & 40.95          \\
mBERT       & 70.98          & 70.00          & 69.87          & 69.97          & 72.48          & 76.08          & 73.47          & 74.30          & 77.45          & 58.70          & 61.28          & 59.48          & 71.05          & 51.45          & 53.34          & 52.18          \\
BERT        & 74.19          & 72.79          & 71.12          & 71.62          & 74.71          & 75.83          & 74.48          & 75.12          & 78.43          & 71.21          & 68.04          & 69.20          & 75.43          & 67.07          & 66.07          & 66.43          \\
XLM-R        & 72.21          & 71.53          & 68.88          & 69.20          & 74.33          & 77.12          & 76.86          & 76.57          & 76.47          & 67.36          & 62.90          & 63.33          & 74.56          & 65.50          & 62.07          & 62.75          \\
Ours        & \textbf{74.69} & \textbf{74.79} & \textbf{75.19} & \textbf{75.19} & \textbf{77.81} & \textbf{79.17} & \textbf{79.19} & \textbf{79.12} & \textbf{78.12} & \textbf{71.98} & \textbf{70.00} & \textbf{70.57} & \textbf{80.43} & \textbf{66.54} & \textbf{77.00} & \textbf{69.12}     \\
\hline
\end{tabular}}
    \caption{Evaluation on English and Marathi languages. The best results are bold. The significance test between the best baseline XLM-R and our system in terms of Recall/Accuracy metrics: $p < 0.01$ (as per t-test).  ISCLS19 do not have power to utilize the context information; therefore, we report the same numbers in w/context as w/o context.}
    \label{table:multilingual_results}
    \end{small}
\end{table*} 
Table~\ref{table:multilingual_results} reports the results for these two languages. For English, all the baselines (with context) improve over their counterpart (without context). However, we do not find similar trend in Marathi possibly due to (1) lack of sufficient task-specific data, and (2) lack of both the auxiliary task~\footnote{We could not activate auxiliary tasks due to lack of datasets for Marathi.}.
For both languages, our system consistently outperforms all the competing systems.  Across both the languages, it shows the average absolute gain of $4.7$ points (A) and $4.4$ points (F1) compared to the strong baseline XLM-R.  Summarily, these empirical results prove the proposed approach's efficacy in languages other than Sanskrit.

\noindent\textbf{(4) Multi-lingual training and zero-shot cross-lingual transfer experiments for Marathi:} Here, we investigate the transferability of the SaCTI task for low-resourced languages. We experiment with the Marathi language (w/ context). Since the label space of Marathi is the same as that of the SaCTI coarse dataset, this makes it possible to experiment with cross-lingual zero-shot transfer and multi-lingual training. There is an isomorphic semantic type system with 4 types for Marathi as is the case for most of the Indian languages, due to a close connection with / inheritance from Sanskrit.  In Table~\ref{table:multi-lingual_experiments}, we consider the mono-lingual (training on Marathi) results as a baseline.  In multi-lingual training, we train the proposed system with mix of Marathi and SaCTI-base coarse dataset and evaluate on test set of Marathi.\footnote{Here, we use pretrained models of Sanskrit for both the auxiliary tasks to obtain psuedo-labels for Marathi.} In the zero-shot transfer, we leverage the model trained on the SaCTI-base coarse dataset to get predictions on the test set of Marathi. In multi-lingual training experiment, we observe substantial improvements ($3.4$ points F1) over mono-lingual training. However, zero-shot cross-lingual transfer does not show encouraging results, where system predicts \textit{Tatpuru\d{s}a} type for majority test samples ($80\%$ predictions).
\begin{table}[h]
\centering
\begin{adjustbox}{width=0.6\textwidth}
\small
\begin{tabular}{|c|c|c|c|c|}
\hline
\textbf{Tasks}                           & \textbf{A}     & \textbf{P}     & \textbf{R}  & \textbf{F1}    \\ \hline
ours (Marathi)         & 80.43          & 66.54          & 77.00          & 69.12          \\\hline
zero-shot transfer     & 48.09          & 33.85          & 48.36          & 31.08          \\\hline
multi-lingual training & \textbf{83.10} & \textbf{71.00} & \textbf{80.48} & \textbf{73.70}\\\hline
\end{tabular}
\end{adjustbox}
\caption{Performance of the multilingual training and cross-lingual zero-shot transfer on Marathi.} 
\label{table:multi-lingual_experiments}
\end{table}
\begin{figure*}[!h]
\centering
\subfigure[]{\includegraphics[width=2.5in]{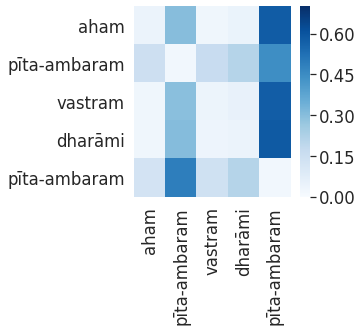}}
\subfigure[]{\includegraphics[width=2.5in]{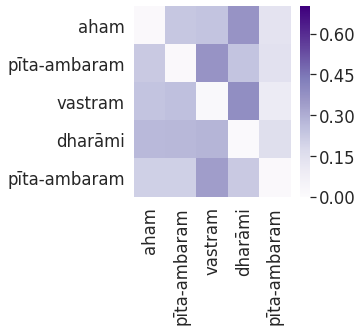}}\\
\subfigure[]{\includegraphics[width=2.5in]{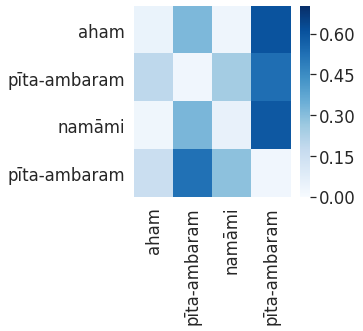}}
\subfigure[]{\includegraphics[width=2.5in]{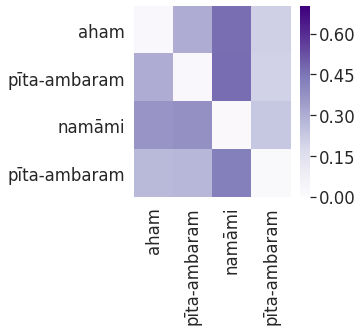}}\\
\caption{Attention heatmaps for the SaCTI (Blue) and dependency parsing (Purple) tasks.  The SaCTI heatmap shows how different context words contribute towards final prediction (Blue) and dependency parsing heatmap (Purple) serve as proxy for interpretation.  We illustrates how the same compound ({\sl p\={\i}ta-ambaram}) in two different contexts [(a-b) {\sl aham \textbf{p\={\i}ta-ambaram} vastram dhar\={a}mi} (I wear a yellow cloth) and (c-d) {\sl aham \textbf{p\={\i}ta-ambaram} nam\={a}mi} (I pray to the Lord Vi\d{s}nu)] leads to different semantic type predictions (\textit{Tatpuru\d{s}a}: yellow cloth and \textit{Bahuvr\={\i}hi}: Lord Vi\d{s}nu). In the dependency heatmap of the first case, \textit{p\={\i}ta-ambaram} focuses on \textit{vastram} (cloth) and in latter case it focuses on \textit{nam\={a}mi} (the action of praying).}
\label{fig:probing} 
\end{figure*}

\noindent\textbf{(5) Probing analysis:}
Here, we probe the attention modules of the proposed system to investigate (1) How do different context words contribute towards final prediction? (2) Do these attentions serve as a proxy for explainability of correct/incorrect predictions? Figure~\ref{fig:probing} illustrates attention heatmaps for the SaCTI (Blue) and dependency parsing (Purple) tasks.  The SaCTI heatmap shows how different context words contribute towards final prediction (Blue) and dependency parsing heatmap (Purple) serves as proxy for interpretation. We notice in the SaCTI attentions that all words mostly focus on the target compound word.  Figure~\ref{fig:probing} illustrate how the same compound ({\sl p\={\i}ta-ambaram}) in two different contexts [(a-b) {\sl aham \textbf{p\={\i}ta-ambaram} vastram dhar\={a}mi} (I wear a yellow cloth) and (c-d) {\sl aham \textbf{p\={\i}ta-ambaram} nam\={a}mi} (I pray to the Lord Vi\d{s}nu)] leads to different semantic type predictions (\textit{Tatpuru\d{s}a}: yellow cloth and \textit{Bahuvr\={\i}hi}: Lord Vi\d{s}nu). In the dependency heatmap of the first case, \textit{p\={\i}ta-ambaram} focuses on \textit{vastram} (cloth) and in latter case, it focuses on \textit{nam\={a}mi} (the action of praying).  As per the grammatical rules, the morphological tagging task correctly predicts the gender information as neuter and masculine in these cases, respectively.
Thus, this probing analysis suggests that auxiliary tasks not only help add complementary signals to the system but also serve as a proxy for explainability. 

\noindent\textbf{(6) Additional auxiliary tasks:}
\label{additional_tasks}
  With our proposed multi-task learning approach, we experiment with a few more additional sequence labeling auxiliary tasks (on SaCTI-base w/ context dev set), namely, the prediction of the case grammatical category (C), lemma prediction (L) and prediction of a relation (R) between modifier and its headword. The results in Table~\ref{table:additional_aux} show that except for the relation prediction task, all the remaining auxiliary tasks report improvements over the base system (with no auxiliary task). However, none of the combinations of these auxiliary tasks could outperform the proposed combination of morphological parsing and dependency parsing tasks. Therefore, we do not consider these additional auxiliary tasks in our final system.
\begin{table}[!h]
\centering
\begin{adjustbox}{width=0.6\textwidth}
\small
\begin{tabular}{|c|c|c|c|c|}
\hline
\textbf{Tasks}                           & \textbf{A}     & \textbf{P}     & \textbf{R}  & \textbf{F1}    \\ \hline
BiAFF            & 87.99 & 85.48 & 87.90  & 86.64 \\\hline
+case (C)            & 87.64 & 85.65 & 88.90  & 87.19 \\\hline
+morph (M)             & 88.01 & 88.90  & 85.75 & 87.26 \\\hline
+relation (R)     & 86.61 & 85.53 & 84.99 & 85.22 \\\hline
+lemma (L)            & 87.43 & 88.27 & 86.21 & 87.18 \\\hline
+Dep. parse (DP)               & 89.14 & 86.49 & 90.10  & 88.01 \\\hline \hline
M+C       & 87.55 & 88.30  & 85.28 & 86.72 \\\hline
M+C+L & 87.08 & 87.74 & 84.98 & 86.30  \\\hline
M+C+R & 86.10  & 86.04 & 83.28 & 84.55 \\\hline
M+DP         & \textbf{88.11} & \textbf{86.12} & \textbf{89.23} & \textbf{88.43} \\ \hline
\end{tabular}
\end{adjustbox}
\caption{The comparison (on SaCTI-base w/ context dev set) in between auxiliary tasks. `+' denotes a system where the corresponding task is integrated with BiAFF.} 
\label{table:additional_aux}
\end{table}

\noindent\textbf{(7) Sensitivity analysis for context window length:} In the present study, we consider the default setting where the complete sentence serves as the context for the compound word under consideration. Within this section, our focus is directed toward an examination of the effects arising from training the system using contexts of varying window sizes. Specifically, we embark on a sensitivity analysis pertaining to the context length. In order to conduct this analysis, we establish a contextual window of length $n$ surrounding the target compound word, encompassing $n$ neighboring words both preceding and succeeding it. To systematically investigate the impact of different context lengths, we conduct multiple experiments, ranging from context lengths of $0$ to $5$. Throughout each experimental iteration, the context length remains consistent and fixed for both the training and testing phases.
\begin{figure}[!hbt]
\centering
\includegraphics[width=0.35\textwidth]{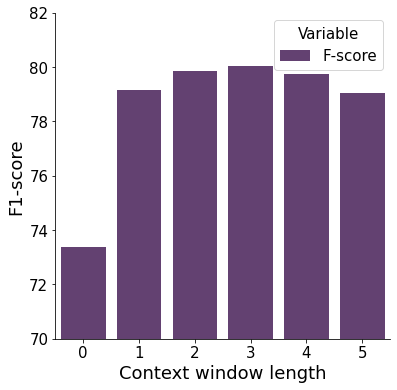}
\caption{Sensitivity analysis for different context window size varying from $0$ to $5$.} 
\label{fig:window_impact} 
\end{figure}
Figure~\ref{fig:window_impact} illustrates that as context window size increases, the performance slightly increases  and then again decreases.

\noindent\textbf{(8) Web-based tool:}
 We deploy our pretrained models as a web-based tool which facilitates the following advantages: (1) A naive user with no prior deep-learning expertise can use it for pedagogical purposes. (2) It can serve as a semi-supervised annotation tool keeping a human in the loop for the error corrections.  (3) Our tool helps the user interpret the model prediction using model confidence on each semantic type and the probing analysis.  (4) It can be used for any general purpose classification task.

\begin{figure*}[!hbt]
\centering
\includegraphics[width=0.8\textwidth]{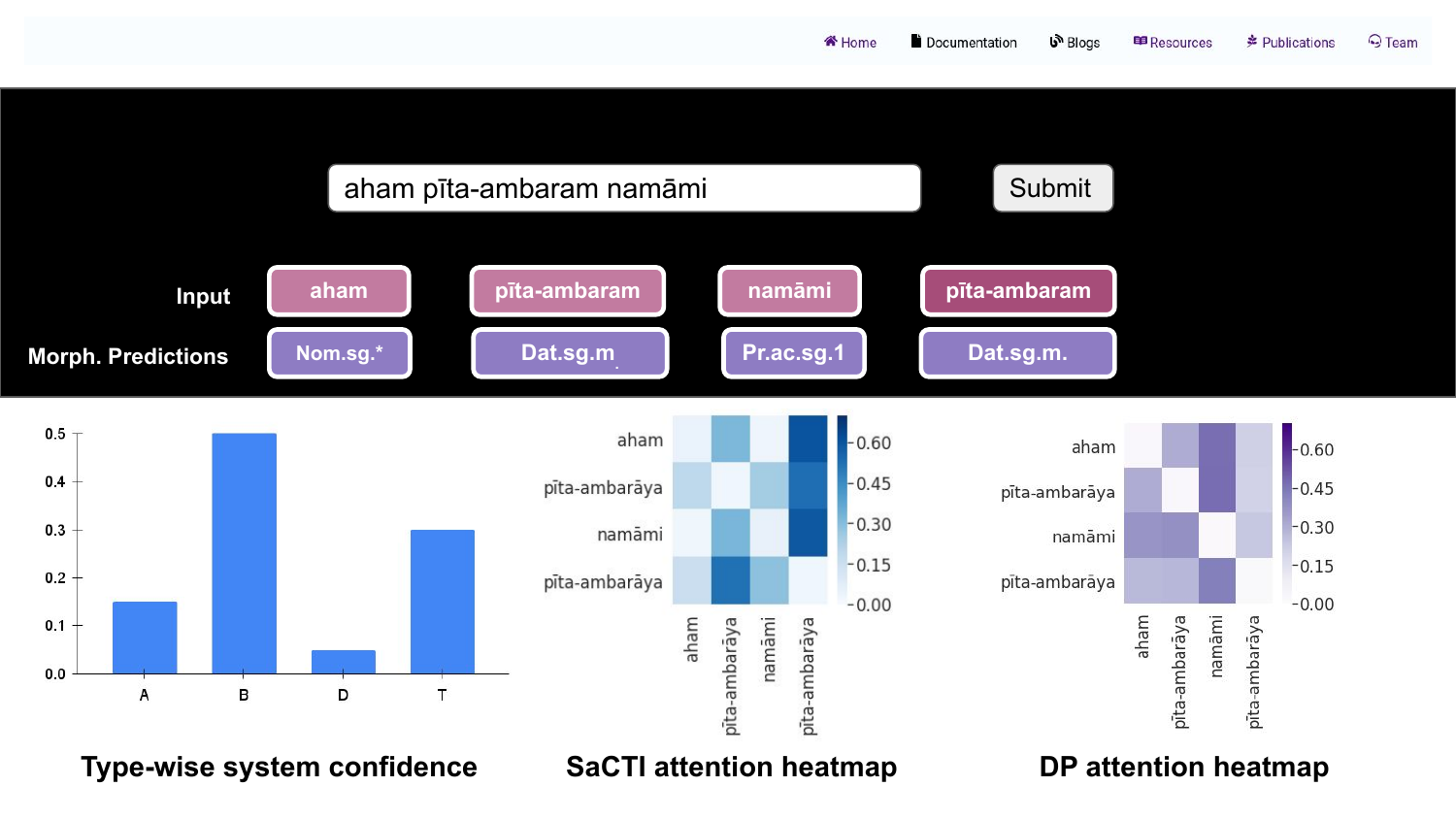}
\caption{Illustration of web-based tool integrated with our best performing pretrained system. Input: ``{\sl aham p\={\i}ta-ambaram nam\={a}mi}'' (Translation: ``I pray to P\={\i}t\={a}mbara (Lord Vishnu).'') where `{\sl p\={\i}ta-ambaram}' is a compound word. Our interface shows predicted morphological tags (color-coded with violet boxes), type-wise system confidence (bar plot), attention heatmaps.} 
\label{fig:interface_interpret} 
\end{figure*}

\section{Summary}
For resource-rich languages, deep learning based models have helped in improving the state of the art for most of the NLP tasks, and have now replaced the need for feature engineering with the choice of a good model architecture. In this chapter, we systematically investigated the following research question: Can the recent advances in neural network outperform traditional hand engineered feature based methods on the semantic level multi-class compound classification task for Sanskrit? We experimented with some of the basic architectures, namely, MLP, CNN, and LSTM, with input representation at the word, sub-word, and character level. The experiments suggest that the end-to-end trained LSTM architecture with FastText embedding gives an F-score of 0.73 compared to the state of the art baseline (0.74) which utilized a lot of domain specific features including lexical lists, grammar rules, etc. This is clearly an important result.

 The Sanskrit compound type identification, where the task is to decode the semantic information hidden in the compound, which can be {\sl context-sensitive}. This poses a limitation to the existing \textit{context agnostic} approaches, thus we propose a novel multi-task learning architecture which incorporates the contextual information and also enriches it with complementary syntactic information using morphological tagging and dependency parsing auxiliary tasks. Our probing analysis showcased that these auxiliary tasks also serve as a proxy for model prediction explainability. To the best of our knowledge, this is the first time that the importance of these auxiliary tasks has been showcased for SaCTI.
Our experiments on benchmark datasets showed that the proposed system provides stunning improvements with $6.1$ points (A) and $7.7$ points (F1) absolute gain compared with the current state-of-the-art system. Our fine-grained analysis showcased some light on the inner engineering of the proposed system. 
Our multi-lingual experiments on English and Marathi languages proved the efficacy of the proposed system in other languages.

 We limit our study to the {\sl purely engineering} data-driven settings.
 We plan to extend the current work by augmenting logical rules \cite{li-srikumar-2019-augmenting,nandwani2019primal} derived from P\={a}\d{n}inian grammar in the proposed approach. 

%% file: Chapters/Chapter6.tex

\chapter{Aesthetics of Sanskrit Poetry from the Perspective of Computational Linguistics: A Case Study Analysis on \'Sik\d{s}\={a}\d{s}\d{t}aka}

\label{Chapter6} 

\lhead{Chapter 6. \emph{Aesthetics of Sanskrit Poetry from the Computational Perspective}}

Sanskrit poetry has played a significant role in shaping the literary and cultural landscape of the Indian subcontinent for centuries. However, not much attention has been devoted to uncovering the hidden beauty of Sanskrit poetry in computational linguistics. This article explores the intersection of Sanskrit poetry and computational linguistics by proposing a roadmap of an interpretable framework to analyze and classify the qualities and characteristics of fine Sanskrit poetry. We discuss the rich tradition of Sanskrit poetry and the significance of computational linguistics in automatically identifying the characteristics of fine poetry. We also identify various computational challenges involved in this process, including subjectivity, rich language use, cultural context and lack of large labeled datasets. The proposed framework involves a human-in-the-loop approach that combines deterministic aspects delegated to machines and deep semantics left to human experts.

We provide a deep analysis of \'Sik\d{s}\={a}\d{s}\d{t}aka, a Sanskrit poem, from the perspective of 6 prominent k\={a}vya\'s\={a}stric schools, to illustrate the proposed framework. Additionally, we provide compound, dependency, anvaya (prose order linearised form), meter, rasa (key elements), ala\.nk{\=a}ra (figure of speech), and r\={\i}ti (writing style) annotations for \'Sik\d{s}\={a}\d{s}\d{t}aka and a web application to illustrate the poem's analysis and annotations. Our key contributions include the proposed framework, the analysis of \'Sik\d{s}\={a}\d{s}\d{t}aka, the annotations and the web application\footnote{Link for interactive analysis of \'Sik\d{s}\={a}\d{s}\d{t}aka: \url{https://sanskritshala.github.io/shikshastakam/}} for future research. We aim to bridge the gap between k\={a}vya\'s\={a}stra and computational methods and pave the way for future research in this area.

\section{Introduction}
Sanskrit literature has a rich and diverse tradition that has played a significant role in shaping the literary and cultural landscape of the Indian subcontinent for centuries \cite{pollock2006language,jamison2014rigveda}. The Sanskrit language, with its complex grammatical rules and nuanced vocabulary, has provided a fertile ground for poets to craft intricate and evocative verses that capture the key elements of human experience \cite{pollock1996sanskrit}. From the ancient epics like the R\={a}m\={a}ya\d{n}a and the Mah\={a}bh\={a}rata, to the lyrical works of K\={a}lidasa and Bhart\d{r}hari etc. Sanskrit poetry has embodied the key elements of Indian thought and culture, serving as a source of inspiration and contemplation for generations of readers and scholars. However, the researchers in the Computational Linguistics community have not devoted much attention to uncovering the hidden beauty in Sanskrit poetry.

The computational linguistic research community should be interested in identifying for various reasons, including the need to evaluate machine-generated poetry \cite{agarwal-kann-2020-acrostic,van-de-cruys-2020-automatic,li-etal-2020-rigid,hopkins-kiela-2017-automatically}, translating poetry \cite{yang-etal-2019-generating,ghazvininejad-etal-2018-neural,krishna-etal-2019-poetry}, to gain a deeper understanding of poetic language \cite{haider-2021-metrical,hamalainen-alnajjar-2019-lets,waltz-1975-understanding}, and to develop recommender systems \cite{10.1145/3341161.3342885,foley2019poetry} for readers interested in poetry. By analyzing large corpora of poetry \cite{haider-2021-metrical,haider-etal-2020-po,gopidi-alam-2019-computational,fairley-1969-stylistic} and identifying the most outstanding examples, researchers can improve evaluation metrics for machine-generated poetry \cite{yi-etal-2018-automatic,liu-etal-2019-rhetorically,ghazvininejad-etal-2017-hafez,goncalo-oliveira-2017-survey}, develop new models and techniques for natural language processing \cite{hu-sun-2020-generating}, and promote and preserve cultural heritage. Furthermore, personalized recommendations for readers can be developed based on their preferences and the most highly regarded poetry in different genres.

Analyzing and identifying the characteristics of fine poetry automatically presents several computational challenges, including: (1) Subjectivity: The perception of what makes a poem ``good" is subjective and can vary widely among individuals and cultures \cite{sheng-uthus-2020-investigating}. Developing a computational model that can accurately capture and evaluate the aesthetic qualities of a poem is therefore challenging.
(2) Rich language use: Poetry often employs complex language use, including metaphors, similes, allusions, and wordplay, which can be difficult for computational models to understand and generate \cite{chakrabarty-etal-2021-dont}.
(3) Cultural context: Poetry is often deeply rooted in cultural and historical contexts \cite{gopidi-alam-2019-computational}, which can be challenging for automated systems to comprehend and incorporate.
(4) Lack of large labeled datasets: The development of automated systems for identifying the characteristics of fine poetry relies on large, labeled datasets, which can be difficult to create due to the subjective nature of poetry and the diversity of cultural and linguistic contexts \cite{al-ghamdi-etal-2021-dependency,horvath-etal-2022-elte}.
Overall, the computational challenges of identifying the characteristics of fine poetry automatically require the development of sophisticated algorithms that can account for the subjective nature of poetry, understand complex language use, incorporate cultural context and work with limited labeled data.

In this article, we aim to address the question of whether we can build an automated and interpretable framework to analyze and classify Sanskrit poetry (k\={a}vya) \cite{baumann-etal-2018-analysis,kesarwani-etal-2017-metaphor} into levels of the characteristics of fine composition. This framework has the potential to answer interesting questions such as given two k\={a}vyas, one is more beautiful and why. It could also serve as a valuable tool for pedagogical purposes. The K\={a}vya\'s\={a}stra provides various perspectives for the analysis of poetry, and we select a \'Sik\d{s}\={a}\d{s}\d{t}aka composition to analyze from these different perspectives. We also aim to evaluate whether computational methods can aid in similar analysis and discuss the limitations of existing state-of-the-art methods.
Our proposed framework involves a human-in-the-loop approach \cite{zhipeng-etal-2019-jiuge,ghazvininejad-etal-2017-hafez}, where deterministic aspects are delegated to machines, and deep semantics are left to human experts. We hope that with further development, this framework can eventually become fully automated, enhancing the appreciation of Sanskrit poetry's inner beauty for neophytes and Sanskrit enthusiasts alike. We believe that this work can serve as a stepping stone, bridging the gap between the tradition of k\={a}vya\'s\={a}stra and computational methods, and paving the way for future research in this area \cite{kao-jurafsky-2015-computational}.

Our key contributions are as follows:
\begin{enumerate}
    \item We propose an automated, human-in-loop and interpretable framework to classify Sanskrit poetry into levels of the characteristics of fine composition.
    \item We provide a deep analysis of \'Sik\d{s}\={a}\d{s}\d{t}aka from the perspective of 6 prominent k\={a}vya\'s\={a}stra schools to illustrate the proposed framework.
    \item We provide compound, dependency, anvaya, meter, rasa, ala\.{n}k{\=a}ra, r\={\i}ti annotations for \'Sik\d{s}\={a}\d{s}\d{t}aka.
    \item We also provide web application to illustrate our \'Sik\d{s}\={a}\d{s}\d{t}aka analysis and annotations.
    \item We publicly release our codebase of framework and web-application for future research.
\end{enumerate}

\section{Background}
 \begin{table*}[t]
\centering
\begin{adjustbox}{}
\small
\begin{tabular}{|p{0.08\linewidth}|p{0.17\linewidth}|p{0.13\linewidth}|p{0.22\linewidth}|p{0.2\linewidth}|}
\hline
\textbf{School}   & \textbf{Founder}        & \textbf{Treatise}                  & \textbf{Objective}                                                                     & \textbf{English meaning}                                                                                                                                                         \\\hline
Rasa     & Bharatamun\={\i}    & N\={a}\d{t}ya\'s\={a}stra              & \textit{na hi rasād\d{r}te kaścidartha\d{h} pravartate}                                         & No meaning proceeds, if a poetry does not carries Rasa.                                                                                                                      \\\hline
Ala\.{n}k\={a}ra & Bh\={a}maha        & K\={a}vy\={a}la\.{n}k\={a}ra              & \textit{rūpakādirala\d{m}kāra\d{h} tasyānyairvahudhodita\d{h} na kāntamapi nirbhū\d{s}a\d{m} vibhāti vanitāmukham } & Ala\.{n}k\={a}ras are vital for poetic beauty, just like ornaments for a charming woman's face. \\ \hline
R\={\i}ti    & V\={a}mana         & K\={a}vy\={a}la\.{n}k\={a}ra-s\={u}trav\d{r}tti & \textit{rītirātmā kāvyasya}                                                         & Poetic style (R\={\i}ti) is the soul of poetry.                                                                                                                                  \\\hline
Dhvani    & \={A}nandavardhana & Dhvany\={a}loka                & \textit{kāvyasyātmā dhvani\d{h} }                                                       & Dhvani is the soul of poetry.                                                                                                                                                \\\hline
Vakrokti & Kuntaka        & Vakrokti-j\={\i}vitam         & \textit{vakrokti\d{h} kāvyajīvitam}                                                   & vakrokti is the vital element of the poetry.                                                                                                                                 \\\hline
Aucitya & K\d{s}emendra     & Aucitya-vic\={a}ra-carc\={a}  & \textit{aucitya\d{m} rasasiddhasya sthira\d{m} kāvyasya jīvitam}                                    & Propriety is the stable vital element of the poetry full of rasa.    \\  \hline
\end{tabular}
\end{adjustbox}
\caption{Chronological overview of 6 schools of kāvyaśāstra with founder \={a}c\={a}rya, their treatise and objectives.} 
\label{table:kāvyaśāstra}
\end{table*}
\noindent\textbf{The 6 main schools of k\={a}vya\'s\={a}stra (literary theory):}
k\={a}vya\'s\={a}stra is the traditional Indian science of poetics and literary criticism, which has played a significant role in shaping the development of literature and aesthetics in the Indian subcontinent for over two thousand years. The term "K\={a}vya" refers to poetry or literature, while "\'S\={a}stra" means science or knowledge, and k\={a}vya\'s\={a}stra is thus the systematic study of the nature, forms, and principles of poetry and literature.
The roots of k{\=a}vya\'s\={a}stra can be traced back to ancient India, where it developed alongside other branches of learning such as philosophy, grammar, and rhetoric. Over time, k\={a}vya\'{s}\={a}stra evolved into a complex and sophisticated system of literary theory, encompassing a wide range of concepts and techniques for analyzing and appreciating poetry, such as Rasa (the emotional key elements of poetry), Ala\.{n}k\={a}ra (the use of rhetorical and figurative devices), Dhvani (the power of suggestion), Vakrokti (oblique), Aucitya (appropriateness) and r\={\i}ti (the appropriate use of style and language). Table \ref{table:kāvyaśāstra} gives a brief overview of these 6 schools in chronological order with founder \={a}c\={a}rya, their treatise and objectives.

\noindent\textbf{Chanda\'s\'s\={a}stra:} 
Chanda\'s\'s\={a}stra is the traditional Indian science of meter and versification in poetry, dating back to the Vedic period. It involves the systematic study of the principles, forms, and structures of meter and versification in poetry, including the use of various poetic devices and the study of various types of meters and poetic forms. While closely related to k\={a}vya\'s\={a}stra, Chanda\'s\'s\={a}stra is a separate branch of traditional Indian knowledge and is not considered one of the 6 main schools of k\={a}vya\'s\={a}stra. Key concepts and techniques include the classification of meters, rhyme and alliteration, and the principles of accentuation and stress. Chanda\'s\'s\={a}stra has played a significant role in the development of Indian poetics and literary theory, and continues to be a vital part of Indian cultural heritage.
By mastering the principles of Chanda\'s\'s\={a}stra, poets and writers are able to create verses that are both aesthetically pleasing and technically precise, resulting in some of the most beautiful and evocative poetry in the world.

\noindent\textbf{Efforts put in Computational Linguistics:}
In the field of computational poetry, researchers are tackling various problems using algorithms and computational methods, such as generating new poetry \cite{agarwal-kann-2020-acrostic,van-de-cruys-2020-automatic,li-etal-2020-rigid,hopkins-kiela-2017-automatically,krishna-etal-2020-graph}, translating poetry \cite{yang-etal-2019-generating,ghazvininejad-etal-2018-neural} analyzing emotional tone, analyzing rhyme and meter patterns \cite{kao-jurafsky-2012-computational,greene-etal-2010-automatic,fang-etal-2009-adapting}, classifying poetry \cite{baumann-etal-2018-analysis,kesarwani-etal-2017-metaphor}, and recommending poetry \cite{10.1145/3341161.3342885,foley2019poetry}.
However, existing work does not focus much on analyzing the aesthetic aspect of poetry, which is crucial for improving the generation, translation, and recommendation applications. This gap motivates our work, which proposes an interpretable framework to classify Sanskrit poetry into different levels of  composition using computational linguistics.
We demonstrate the proposed framework by conducting a deep analysis of \'Sik\d{s}\={a}\d{s}\d{t}aka, a Sanskrit poem, from the perspective of 6 well-known k\={a}vya\'s\={a}stric schools. The key contributions of our article include the proposed framework, the analysis of \'Sik\d{s}\={a}\d{s}\d{t}aka, the annotations, the web application, and the publicly released codebase for future research.

\section{Basics of schools, Computational Aspect and Analysis}
This section aims to explicate the rationale behind the selection of \'Sik\d{s}\={a}\d{s}\d{t}aka for our analysis, followed by the presentation of 7 modules of our framework, each corresponding to one of the 6 schools of k\={a}vya\'s\={a}stra and one module from Chanda\'s\'s\={a}stra. In each subsection, we provide a comprehensive overview of the respective school's fundamentals, the feasibility of developing a computational system based on the same, the intriguing insights gleaned from the analysis of \'Sik\d{s}\={a}\d{s}\d{t}aka from the said school's perspective.

\noindent\textbf{Why \'Sik\d{s}\={a}\d{s}\d{t}aka:}
To facilitate a more thorough and nuanced analysis of k\={a}vya, we set a few criteria to guide our selection of the most appropriate composition for our study. These criteria are as follows: (1) the composition should be sufficiently small to enable comprehensive scrutiny from diverse k\={a}vya\'s\={a}stra perspectives; (2) the poet who has authored the composition ought not to possess a significant reputation in the k\={a}vya\'s\={a}stra domain to obviate the potential for biases that may arise from establishing the composition as Uttama k\={a}vya; (3) the composition should have made a substantive contribution to the traditions from a non-literary perspective to explore the role of k\={a}vya\'s\={a}stra aspects in its accomplishment.

The \'Sik\d{s}\={a}\d{s}\d{t}aka composition is well-suited for our analysis as it satisfies all three criteria that we have established. First, it meets the requirement of being compact enough to allow for in-depth examination from various k\={a}vya\'s\={a}stra perspectives as it consists of only eight stanzas or verses.\footnote{We encourage readers to go through this composition: \url{https://sanskritshala.github.io/shikshastakam/shloka.html} where word-to-word meanings and translations are given.} Additionally, not all A\d{s}\d{t}akas can be considered uttama k\={a}vya (the fine poetry), which further supports its appropriateness as a subject of study.
Second, the author of \'Sik\d{s}\={a}\d{s}\d{t}aka, Chaitanya Mah\={a}prabhu, is not primarily known as a poet or literary figure. Thus, he meets the second criterion of not being a well-established poet in the k\={a}vya\'s\={a}stra domain, which helps to minimize the potential biases that could arise from claiming the composition as uttama k\={a}vya. 
Finally, \'Sik\d{s}\={a}\d{s}\d{t}aka has made significant contributions to the traditions of Gau\d{d}\={\i}ya Vai\d{s}\d{n}avism from a non-literary perspective, which satisfies the third criterion.
This composition has significantly influenced the lifestyle of the religious community adhering to the principles of Gaudiya Vaishnavism, serving as the fundamental backbone of their philosophy for the past 500 years. The enduring prevalence and practice of these teachings within the community underscore the profound impact of this composition on their way of life.
In summary, the \'Sik\d{s}\={a}\d{s}\d{t}aka composition satisfies all three criteria of our selection process, making it an excellent choice for our analysis.

\subsection{Metrical analysis}
\noindent\textbf{Basics:} The bulk of Sanskrit literature consists of poetic works that conform to the conventions of Sanskrit prosody, or Chanda\'s\'s\={a}stra, which involves the study of Sanskrit meters, or chandas. The primary objective of utilizing chandas is to infuse rhythm into the text to facilitate memorization, while also aiding in the preservation of accuracy to some degree \cite{deo_2007,Melnad}. Pi\.{n}gala is considered to be the father of Chanda\'s\'s\={a}stra. Prominent scholars and their work in this field include Pi\.{n}gala (Chanda\'s\'s\={u}tra), Bharata Mun\={\i} ( N\={a}\d{t}ya\'s\={a}stra), Pi\.{n}gala Bha\d{t}\d{t}a (Chandomanjar\={\i}), V\d{r}ttan\={a}tha (V\d{r}ttaratn\={a}kara), Jagann\={a}tha Pa\d{n}\d{d}ita (Rasaga\.{n}g\={a}dhara), etc. Although approximately 1,398 meters are described in V\d{r}ttaratn\={a}kara \cite{Rajagopalan2018AUT}, not all are widely used.

\noindent\textbf{Computational aspect:} In chanda\'s\={a}stra, the classification of each syllable (ak\d{s}ara) in Sanskrit as either laghu (short) or guru (long) enables the representation of every line of Sanskrit text as a binary sequence comprised of laghu and guru markers. The identification of specific patterns within these sequences leads to the recognition of various types of chandas, thus enabling the deterministic identification of a meter. In recent years, commendable efforts have been made by the computational community to develop user-friendly toolkits for meter identification \cite{neill-2023-skrutable,Rajagopalan2018AUT,terdalkar-bhattacharya-2023-chandojnanam}.

\noindent\textbf{\'Sik\d{s}\={a}\d{s}\d{t}aka analysis:} The a\d{s}\d{t}aka, which comprises a series of 8 verses, is typically composed in a single meter. In the case of the \'Sik\d{s}\={a}\d{s}\d{t}aka, there are 5 meters employed, namely, \'Sārdūlavikrī\d{d}ita, 
Vasantatilaka, Anu\d{s}\d{t}up, Viyogin\={\i} and Upaj\={a}ti (Indrava\.m\'sa and Va\d{m}\'sastha).
Further details on how to identify meters can be found in the relevant literature \cite{neill-2023-skrutable,Rajagopalan2018AUT,terdalkar-bhattacharya-2023-chandojnanam}. The identification of meters has aided in the identification of typographical errors (tanuja $\rightarrow$ tanūja,  dhūli$\rightarrow$ dhūlī, gadagada$\rightarrow$ gadgada) in some of the existing manuscripts of the \'Sik\d{s}\={a}\d{s}\d{t}aka. One may wonder about the use of 5 different meters in the \'Sik\d{s}\={a}\d{s}\d{t}aka. Some scholars, or \={a}c\={a}ryas, have argued that the verses were not composed in a single sitting. Rather, they were collected in Rupa Goswami's (disciple of the author of \'Sik\d{s}\={a}\d{s}\d{t}aka) Pady\={a}val\={\i}, which is an anthology of compositions by expert poets in the Gau\d{d}\={\i}ya Vai\d{s}\d{n}ava tradition. K\d{r}\d{s}\d{n}adas Kaviraj Goswami later compiled and arranged them in a meaningful order in his composition, the Chaitanya Carit\={a}m\d{r}ta. We posit that the use of multiple meters serves as an evidence supporting this claim.

\noindent\textbf{Future direction:} In future directions, it would be interesting to explore the correlation between the intrinsic mood of the meter and the rasa of the poetry. It would also be worthwhile to investigate the relationship between poets and the meters they have employed. This could involve examining patterns in their favorite choices and attempting to identify the poet based on the composition itself.  Additionally, there is potential for exploring whether the occurrence of 
Rasado\d{s}a (obstruction in the enjoyment of mellow) can be predicted through automatic identification of the composition's rasa. These investigations could contribute to a deeper understanding of the relationship between meters, poetry, and emotion in Sanskrit literature.

\subsection{Ala\.{n}k\={a}ra school}
\noindent\textbf{Basics:}  This school focuses on the use of figurative language or literary devices to enhance the beauty and aesthetic appeal of poetry. It includes the study of metaphors, similes, personification, hyperbole, and other literary devices. The most important exponent of this school is Bh\={a}maha, who wrote the K\={a}vy\={a}la\.{n}k\={a}ra, one of the earliest treatises on poetic embellishments. Other notable figures associated with this school include Da\d{n}\d{d}in, Udbha\d{t}a, and Rudra\d{t}a. There are two broad categories of Ala\.{n}k\={a}ras: (1) \'{S}abd\={a}la\.{n}k\={a}ra: refers to a figure of speech that relies on the pleasing sound or choice of words, which loses its effect when the words are substituted with others of similar meaning.  (2) Arth\={a}la\.{n}k\={a}ra: is a figure of speech that is based on the meaning of words and is not affected by sound (in contrast to \'{S}abd\={a}la\.{n}k\={a}ra).

\noindent\textbf{Computational aspect:} The process of identifying \'{S}abd\={a}la\.{n}k\={a}ra is considered deterministic and involves verifying the occurrence of specific patterns of syllables or words. In contrast, the identification of Arth\={a}la\.{n}k\={a}ra presents a significant semantic challenge, even for experienced annotators. The feasibility and difficulty of developing supervised systems for this purpose can be better understood through empirical investigations.
There is no system currently available for automated ala\.{n}k\={a}ra analysis for Sanskrit. To develop a supervised data-driven system, several important questions must be addressed. For instance, it is crucial to determine the amount of data that needs to be annotated, as well as the appropriate methods for marking ala\.{n}k\={a}ra annotations, such as whether they should be marked at the level of a complete \'{s}loka and a phrase.
To address these concerns, it is necessary to develop a standardized scheme for ala\.{n}k\={a}ra annotation. This would enable researchers to collect and analyze annotated data consistently and systematically, which would facilitate the development of accurate and reliable automated systems for ala\.{n}k\={a}ra analysis.

\noindent\textbf{\'Sik\d{s}\={a}\d{s}\d{t}aka analysis:} In our analysis of the \'Sik\d{s}\={a}\d{s}\d{t}aka, we have employed the ala\.{n}k\={a}ra categorization outlined by Mamma\d{t}a in his influential work ``K\={a}vyaprak\={a}\'{s}a". This text provides a comprehensive overview of 67 ala\.{n}k\={a}ras or literary ornaments that can be utilized to embellish and enhance the expression of poetry. According to the Ala\.{n}k\={a}ra school, ala\.{n}k\={a}ra is considered the soul of poetry, and without its incorporation, a poetic composition lacks vitality. In other words, k\={a}vya without ala\.{n}k\={a}ra is likened to a lifeless entity, deemed by some poets of this school to be comparable to a widow.\footnote{\textit{ala\.{n}kārarahitā vidhavaiva sarasvatī|} 2.334, Agnipur\={a}\d{n}a} 
Furthermore, it is worth noting that certain poets from this tradition place greater emphasis on arth\={a}la\.{n}k\={a}ra, or figurative language, over \'{s}abd\={a}la\.{n}k\={a}ra, which deals primarily with sound patterns and word repetition. The \'Sik\d{s}\={a}\d{s}\d{t}aka, a devotional composition in Sanskrit, exhibits the mood of separation (Vipralambha-\'{s}\d{r}\.{n}gāra\d{h}) as its primary rasa or emotional key elements. The author portrays himself as a devotee, and his beloved is Lord Govinda. The appropriate use of ala\.{n}k\={a}ras (figures of speech) in this composition serves to express the rasa without hindrance. The author employs \'{s}abd\={a}la\.{n}k\={a}ras (Anupr\={a}sa\footnote{\textit{var\d{n}as\=amyamanupr\=asa\d{h}}| 9.79, K\={a}vyaprak\={a}\'{s}a }) and arth\={a}la\.{n}k\={a}ras (R\={u}paka\footnote{\textit{tadr\=upakambhedo ya\d{h} upam\=anopameyayo\d{h}|} 10.93, K\={a}vyaprak\={a}\'{s}a}, Upam\={a}\footnote{\textit{s\=adharmyamupam\=a bhede|}1.87, K\={a}vyaprak\={a}\'{s}a}, Vyatireka\footnote{\textit{upam\=an\=ad yadanyasya vyatireka\d{h} sa\d{h} eva sa\d{h}}, 10.105, K\={a}vyaprak\={a}\'{s}a}, Tulyayogitā\footnote{\textit{niyat\=an\=a\.m sak\d{r}ddharma\d{h} s\=a puna\d{h} tulyayogit\=a|} 10.104, K\={a}vyaprak\={a}\'{s}a} and Viśe\d{s}okti\footnote{\textit{viśe\d{s}oktirakha\d{n}\d{d}e\d{s}u k\=ara\d{n}e\d{s}u phalāvaca\d{h}|}
10.108, K\={a}vyaprak\={a}\'{s}a} ) in \'Sik\d{s}\={a}\d{s}\d{t}aka. The use of R\={u}paka (metaphor)  and Upam\={a} is frequent, with the former appearing 6 times and the later 4 times. The metaphors are employed to equate two things directly, while similes use ``like" or ``as" to make comparisons.
 
In the first verse, the author uses a series of 5 metaphors to describe the victory of \'{S}rik\d{r}\d{s}\d{n}a Sa{\.n}k\={\i}rtana. The use of these metaphors creates a garland-like effect, which serves as an offering to acknowledge the victory of \'{S}r\={\i}k\d{r}\d{s}\d{n}a Sa{\.n}k\={\i}rtana. Additionally, the use of Anty\={a}nupr\=asa \'Sabd\=ala\.nk\=ara (a figure of speech where every word ends with a similar suffix) creates a rhythmic effect. 
In the second verse, the author expresses his sorrow of not having an attachment to such a glorious Sa{\.n}k\={\i}rtana, where the Lord invests all his potencies without keeping any rules and regulations. Although there is every reason to get attached to such Sa{\.n}k\={\i}rtana, the author cannot get attached, which is expressed through Uktanimitta vi\.se\d{s}yokti ala\.nk\=ara. This ala\.{n}k\={a}ra is used to denote that all the causes are present for action to happen, yet the action does not occur.
In the third verse, the author talks about the prerequisites of developing attachment for the Sa\.nk\={\i}rtana: humility and tolerance. The author employs Vyatireka ala\.nk\=ara to embellish this verse by comparing humility with a grass and tolerance with a tree. In this figure of speech, the subject of illustration (upam\=ana) has higher qualities than the subject of comparison (upameya). Thus, the author says that one should be lower than a grass and more tolerant than a tree.
In the fourth verse, multiple things (wealth, followers, beautiful women and flowery language of Vedas) are coupled with a single property ``na k\=amaye'' (do not desire). This device is called tulyayogitā ala\.{n}k\={a}ra.
In the fifth verse, the author beautifully couples metaphor and simile to express his eternal servitude to  Govinda. The author compares the way water drops fall from a lotus into the mud, similarly, he has fallen from Govinda's lotus feet to the ocean of nescience (material existence) as a dust-like servant. The author requests Govinda to place him back at his lotus feet.
In the sixth verse, again tulyayogitā ala\.nk\=ara is employed to connect different symptoms mentioned to a single property.
In the seventh verse, the author expresses his intense separation from Govinda by deploying three similes (upam\=a ala\.nk\=ara). The author uses these similes to express his love in the mood of separation. He compares a moment to a great millennium, his eyes to the rainy season, and the entire world to void. This garland of similes (m\={a}lopam\=a ala\.nk\=ara) serves as an offering to express the author's love for Govinda.
The final verse employs the anuktanimitta-viśe\d{s}okti device, wherein a condition is expressed but the consequent effect is absent, and no explanation is offered for the lack of effect. This constitutes the anukatanimitta-viśe\d{s}okti ala\.nk\=ara. The author enumerates several potential causes for being despised, yet the anticipated outcome of animosity does not manifest. Despite hatred conditions, the author maintains devotion to the lord. The rationale for the absence of hatred is not mentioned here.
In conclusion, the \'Sik\d{s}\={a}\d{s}\d{t}aka employs a range of ala\.{n}k\={a}ras to express its emotional key elements (rasa) without hindrance. The appropriate use of metaphors, similes, and other figures of speech serves to create a garland-like effect that acknowledges Govinda's glory and expresses the author's devotion and love.

\noindent\textbf{Future direction:} Moving forward, several directions for research on ala\.nk\=aras can be explored. One crucial area is the formulation of the problem of identifying ala\.nk\=aras. Since ala\.nk\=aras can be assigned to complete or partial verses, or sequences of syllables or pairs of words, determining the appropriate approach for identifying them is non-trivial. Different formulations, such as sentence classification, sequence labeling, or structured prediction, can be employed.
Another interesting direction for research is to investigate which ala\.{n}k\={a}ras are more beautiful and how we can compare two compositions based solely on their use of ala\.nk\=aras. Defining a basis for evaluating beauty in poetry is a challenging problem in itself. However, understanding which ala\.{n}k\={a}ras contribute more to the aesthetic appeal of a poem and how to measure this appeal can be valuable for poets and scholars alike.
Finally, exploring the correlation of ala\.{n}k\={a}ra school with other schools of k\={a}vya\'s\={a}stra, such as rasa and dhvani, is a promising area for future research. Understanding how these different aspects of poetry interact can provide insight into the complex mechanisms of poetic expression and perception. Overall, these future directions offer exciting opportunities to deepen our understanding of ala\.nk\=aras and their role in poetry.

\subsection{Rasa school}
\noindent\textbf{Basics:}  The rasa school of Indian aesthetics places significant emphasis on the emotional impact of poetry on the reader or audience. Bharata mun\={\i}, the author of the  N\={a}\d{t}ya\'s\={a}stra, is considered the most influential thinker associated with this school. His treatise on Indian performing arts includes a detailed discussion of rasa theory, which posits that rasa is the soul of poetry. According to Bharata, rasa is the ultimate emotional pleasure that can be derived from a work of art. 
Rasa is the heightened emotional response to the text. Bharata also classified the various emotions depicted in performing arts into eight different rasas, or flavors, which are \'S\d{r}\.ng\=ara (romance), H\=asya (comedy), Karu\d{n}a (piteous), Raudra (anger), V\={\i}ra (heroism), Bhay\=anaka (fear), B\={\i}bhatsa (disgust), and Adbhuta (wonder). The nineth emotion, \'S\=anta (peace) is also enlisted by the other scholars.\footnote{\textit{na yatra du\d{h}kha\d{m} na sukha\d{m} na dve\d{s}o nāpi matsara\d{h}| sama\d{h} sarve\d{s}u bhūte\d{s}u sa śānta\d{h} prathito rasa\d{h} ||}, N\={a}\d{t}ya\'s\={a}stra . } The \'S\=anta rasa is not compatible for dramas but other kinds of poetry can be composed in this rasa. 
The concept of rasa theory remains an integral part of Indian culture and has greatly influenced the performing arts in India and beyond.

In Bharata's  N\={a}\d{t}ya\'s\={a}stra, a formula\footnote{\textit{vibhāvānubhāvavyabhicāri\-sa\d{m}yogādrasani\d{s}patti\d{h}|} ,N\={a}\d{t}ya\'s\={a}stra} for the arousal of rasa is presented, wherein the combination of vibh\=ava, anubh\=ava, and vyabhic\=ar\={i} bh\=ava is necessary to evoke the experience of rasa. The term `vibh\=ava' refers to the stimulants of emotions, while `anubh\=ava' represents the physical responses that accompany emotional reactions, and `vyabhic\=ar\={\i} bh\=ava' denotes the transitory emotions. The basic emotions or bh\=avas of the reader or spectator are activated by the vibh\=avas, while the anubh\=avas and vyabhic\=ar\={\i} bh\=avas serve to indicate the emotional response experienced. This formula suggests that the experience of rasa is not solely dependent on a single aspect of the literary text or performance, but rather requires the combination of multiple elements to evoke a complete emotional response.

\noindent\textbf{Computational aspect:} Computational identification of rasa in poetry  is an interesting and challenging research area. 
One of the main challenges in identifying rasa computationally is that the interpretation of rasa is subjective. Different readers or critics may have different opinions on the dominant emotion evoked by a particular work of poetry.
Another challenge is the complexity of the vibh\=ava, anubh\=ava, and vyabhic\=ar\={\i} bh\=ava that contribute to the formation of rasa. These elements are often implicit or subtle in the text, and their identification requires a deep understanding of the cultural and literary context in which the text was produced.
To overcome these challenges, possible computational approaches that can be taken include using machine learning algorithms to identify patterns in the text that are associated with particular emotions, and incorporating contextual information such as the author, genre, and historical period. In addition, the use of computational linguistics techniques, such as sentiment analysis and topic modeling, may also be helpful in identifying the dominant emotions expressed in a text.
However, it is important to note that while computational systems may be able to identify the dominant rasa of a work of poetry, they may not be able to fully capture the richness and complexity of the vibh\=ava, anubh\=ava, and vyabhic\=ar\={\i} bh\=ava that contribute to the formation of rasa. The identification of these elements requires a deep understanding of the cultural and literary context in which the text was produced, and is likely to remain the purview of human literary critics and scholars.
In conclusion, while computational approaches may be able to provide some insight into the identification of rasa in poetry, the complexity and subjectivity of rasa theory poses significant challenges. Further research in this area is needed to develop computational models that can better capture the cultural and literary context of poetic works and the nuances of the vibh\=ava, anubh\=ava, and vyabhic\=ar\={\i} bh\=ava that contribute to the formation of rasa.

\noindent\textbf{\'Sik\d{s}\={a}\d{s}\d{t}aka analysis:} A fundamental aspect of evoking rasa involves combining three elements, namely vibh\=ava, anubh\=ava, and vyabhic\=ar\={\i} bh\=ava. The vibh\=avas activate the basic emotions or bh\=avas of the reader or spectator, while the anubh\=avas and vyabhic\=ar\={\i} bh\=avas indicate the emotional response experienced. In the composition under discussion, the author assumes the mood of a devotee and reciprocates with his beloved Lord K\d{r}\d{s}\d{n}a. The author employs several words to indicate that the vibh\=ava is K\d{r}\d{s}\d{n}a, including śrī-k\d{r}\d{s}\d{n}a-sa\.{n}kīrtanam (congregational chanting of K\d{r}\d{s}\d{n}a's holy name), nāmnām (K\d{r}\d{s}\d{n}a's holy name), Hari\d{h} (another name of K\d{r}\d{s}\d{n}a), nanda-tanuja (K\d{r}\d{s}\d{n}a, the son of Nanda Maharaj), tava nāma-graha\d{n}e (chanting K\d{r}\d{s}\d{n}a's holy name), Govinda (another name of K\d{r}\d{s}\d{n}a), and mat-prā\d{n}a-nātha\d{h} (the master of my life).
The anubh\=ava of the author is evident in the seventh verse, where he expresses intense feelings of separation. He considers even a moment to be a great millennium, and tears flow from his eyes like torrents of rain. He perceives the entire world as void. The author employs several words to indicate the vipralambha s\d{r}\.ng\=ara, such as nānurāga\d{h} (no attachment to K\d{r}\d{s}\d{n}a), a-darśanāna (K\d{r}\d{s}\d{n}a not being visible), govinda-virahe\d{n}a (separation from Govinda), and mat-prā\d{n}a-nātha\d{h} (the master of my life). In conclusion, it is evident from our analysis of vibh\=ava and anubh\=ava in this composition that the rasa evoked is vipralambha s\d{r}\.ng\=ara, which is characterized by intense feelings of separation and longing. The author effectively employs vibh\=avas, such as the names and descriptions of K\d{r}\d{s}\d{n}a, to activate the basic emotions of the reader or spectator, while his anubh\=avas, such as his intense feelings of separation and tears, serve to indicate the emotional response experienced. The composition is a powerful example of how the combination of vibh\=ava and anubh\=ava can be used to evoke rasa in Indian classical literature.

\noindent\textbf{Future direction:} The future direction of computational linguistics in identifying rasa in poetry requires a comprehensive approach that covers all the crucial dimensions of the problem. 
Data annotation is one crucial aspect of the future direction of computational linguistics in identifying rasa in poetry. Data annotation involves the manual tagging of texts with information such as the dominant rasa, vibh\=ava, anubh\=ava, and vyabhic\=ar\={\i} bh\=ava. 
Possible approaches that can be taken include using supervised machine learning algorithms to identify patterns in the text that are associated with particular emotions. This involves training a model on a labeled dataset of texts annotated with the dominant rasa and associated vibh\=ava, anubh\=ava, and vyabhic\=ar\={\i} bh\=ava. 
Another approach is to use unsupervised machine learning algorithms such as clustering and topic modeling to identify the dominant emotions expressed in the text. This approach requires minimal human annotation.
In addition, incorporating contextual information such as the author, genre, and historical period can also be useful in identifying the dominant rasa in a text. 

\subsection{R\={\i}ti school}
\noindent\textbf{Basics:} In this study, the focus is on the concept of R\={\i}ti, also referred to as M\=arga or Sa\.ngha\d{t}ana, in poetry. This aspect of poetry has been heavily emphasized by the scholar V\=amana, who considers R\={\i}ti as the soul of poetry. According to V\=amana, there are three R\={\i}tis, namely vaidarbh\={\i}, Gau\d{d}\={\i}, and P\=a\~nc\=al\={\i}, which were named after the geographical areas from where poets constructed in these styles. The deterministic elements of R\={\i}ti include the letters, length and quantity of compounds, and the complexity or simplicity of the composition.
The Vaidarbh\={\i} style is considered the most delightful style of poetry, consisting of all ten gu\d{n}as\footnote{\textit{samagragu\d{n}opet\=a vaidarbh\={\i}}| 2.11, K\={a}vy\={a}la\.{n}k\={a}ra-s\={u}trav\d{r}tti} explained by V\=amana. The construction of Vaidarbh\={\i} contains a shorter compounds.\footnote{We need to perform empirical analysis of existing annotated poetry to get threshold of length of short compounds.} The Gau\d{d}\={\i} style, dominated by Oja and K\=anti gu\d{n}as,\footnote{\textit{oja\d{h}k\=antimat\={\i} gau\d{d}\={\i}y\=a}| 2.12, K\={a}vy\={a}la\.{n}k\={a}ra-s\={u}trav\d{r}tti} lacks Sukum\=arat\=a and M\=adhurya gu\d{n}as. The composition possesses long compounds and too many joint consonants. 
P\=a\~nc\=al\={\i} style has a simple composition having no compounds.

V\=amana also describes ten qualities of sound and ten qualities of meaning, including \'Sle\d{s}a, Pras\=ada, Samat\=a, M\=adhurya, Sukumarat\=a, Arthavyakti, Ud\=arata, Oja, K\=anti, and Sam\=
adhi.\footnote{\textit{oja\d{h} prasādaśle\d{s}asamatāsamādhimādhurya\-saukumāryodāratārthavyaktikāntayo bandhagu\d{n}ā\d{h}||} 3.4, K\={a}vy\={a}la\.{n}k\={a}ra-s\={u}trav\d{r}tti} These 10 qualities of poetry are compressed into 3 qualities suggested by letters, compounds, and diction.
M\=adhurya (pleasantness), which leads to the melting of the mind, is suggestive of the spar\'sa(mute) consonants with the exception of those in the `\d{t}a' class (\d{t}, \d{t}h, \d{d}, \d{d}h), combined with the last consonants of the class (\.{n}, ñ, n, m), the consonants `r' and  `\d{n}' when followed by a short vowel, and when there are no or medium length compounds. Oja, the cause of lustrous expansion of mind, is suggestive of complex diction consisting of combinations of first consonants of the class with the third, second consonants with the third one, any consonant with `r' in any order, any consonant with itself, the `\d{t}a' class without `\d{n}', `ś' and `\d{s}', and consisting of long compounds. Pras\=ada is present in all the rasas and all kinds of poetries and R\={\i}tis equally. In Pras\=ada gu\d{n}a, words can be comprehended right after the hearing of it.
In conclusion, R\={\i}ti plays an essential role in the composition of poetry, as emphasized by V\=amana. Vaidarbh\={\i}, Gau\d{d}\={\i}, and P\=a\~nc\=al\={\i} are the three R\={\i}tis, each having its own unique features and characteristics. Additionally, the ten qualities of sound and ten qualities of meaning and their compression into three gu\d{n}as suggested by letters, compounds, and diction are important elements of R\={\i}ti. The proper deployment of these elements can significantly enhance the impact of poetry on the reader or listener.

\noindent\textbf{Computational aspect:} The identification of the R\={\i}ti of a composition can be considered a deterministic process. K\={a}vya\'s\={a}stra provides various clues that enable the classification of a R\={\i}ti category. For instance, if a composition comprises of soft syllables, no joint consonants, and a dearth of compounds or short compounds, it is considered as Vaidarbh\={\i} R\={\i}ti. Conversely, if a composition contains more joint consonants, long compounds, the usage of fricatives, and specific combinations of joint syllables, it is considered as Gau\d{d}\={\i} r\={\i}ti. The remaining compositions, which do not fall into these two categories, are classified as P\=a\~nc\=al\={\i} r\={\i}ti. To the best of our knowledge,  currently no computational system exists that can determine the R\={\i}ti of a composition.

\noindent\textbf{\'Sik\d{s}\={a}\d{s}\d{t}aka analysis:} The \'Sik\d{s}\={a}\d{s}\d{t}aka is an example of a composition that embodies the M\=adhurya gu\d{n}a and Vaidarbh\={\i} r\={\i}ti. This is apparent through the simple composition, usage of soft consonants, limited occurrence of long compounds, and the restricted usage of joint consonants. Moreover, the rasa of the k\={a}vya is vipralambha \'s\d{r}\.ng\=ara, which is aptly brought out through the use of Vaidarbh\={\i} r\={\i}ti and M\=adhurya gu\d{n}a. The utilization of any other R\={\i}ti would have led to a rasa do\d{s}a, highlighting the significance of choosing the appropriate R\={\i}ti to convey the intended rasa.

\noindent\textbf{Future direction:} A natural direction for future research would be to consolidate the provided clues and develop a rule-based computational system for the identification of R\={\i}ti in various compositions by different poets. This would enable automated analysis and assessment of R\={\i}ti in k\={a}vya. An empirical investigation could be carried out to explore the degree to which R\={\i}ti is effective in identifying the rasa of a k\={a}vya. Additionally, further research could explore the correlation of R\={\i}ti school with other k\={a}vya\'s\={a}stra schools. V\=amana, the founder of the R\={\i}ti school, maintains that different r\={\i}ti can provide unique experiences for readers. Therefore, it would be of interest to conduct a cognitive analysis of brain signals of readers after exposing them to compositions of different r\={\i}tis. This would provide valuable insights into the effect of r\={\i}ti on the reader's cognitive processes and emotional responses to k\={a}vya.

\subsection{Dhvani school}
\noindent\textbf{Basics:} The school of Dhvani, a prominent school of Indian poetics, is associated with \=Anandavardhana, who authored the seminal work Dhvany\=aloka, which expounds on the theory of suggestion in poetry. Other significant figures in this school include Abhinavagupta, Mamma\d{t}a. According to the Dhvani school, poetry's key elements lies in suggestion or Dhvani, which alludes to the verbal, hidden, and profound aspects of poetry. This school emphasizes the power of suggestion and implication in poetry and argues that a poem's true meaning is not merely in its literal content, but in the emotional response it elicits in the reader through suggestion and implication.
The meaning is distinguished in three categories : (1) V\=acya or literal meaning, (2) Lak\d{s}a\d{n}a or indirect meaning, which is preferred when the literal meaning is inadequate and contradictory, and (3) Vya\.ngya or poetic/metaphysical meaning, which is a deeper meaning that does not conflict with the literal meaning and evokes wonder by reaching out to the soul of poetry. When the Vya\.ngy\=artha is felt more delightful than the other meanings, the Vya\.ngy\=artha is called as Dhvani.\footnote{\textit{yatrārtha\d{h} śabdo vā tamarthamupasarjanīk\d{r}tasvārthau |
vya\.{n}kta\d{h} kāvyaviśe\d{s}a\d{h} sa dhvaniriti sūribhi\d{h} kathita\d{h}||} 1.13, Dhvany\=aloka } Based on the degree of Dhvani, a poem is classified into three categories that demonstrate high, medium, or low conformity with the standards of fine Sanskrit poetry, as articulated by experts in K\={a}vya\'s\={a}stra: (1) Uttama k\={a}vya or high composition, (2) Madhyama k\={a}vya or medium composition, and (3) Adhama k\={a}vya or low= composition.
In Uttama k\={a}vya, Dhvani is present, evoking wonder in the reader. In Madhyama k\={a}vya, Vya\.ngy\=artha is present, but it does not evoke wonder compared to V\=acy\=artha, while in Adhama k\={a}vya, Dhvani is absent. 
Furthermore, Dhvani is divided into three broad categories: (1) Vastu-Dhvani, which implies some rare fact or idea, (2) Ala\.nk\=ara-Dhvani, which suggests a figure of speech, and (3) Rasa-Dhvani, which evokes a feeling or mood in the reader.

\noindent\textbf{Computational aspect:} Identifying Dhvani in k\={a}vya computationally is a complex and challenging task due to the subjective nature of literary interpretation and the intricacy of the language. Dhvani, which encompasses various layers of meaning, including literal, indirect, and metaphoric, presents a challenge to computational identification due to  the absence of annotated data. Poetic devices such as simile, metaphor, and alliteration add to the complexity of creating a unified set of features for the identification of Dhvani. Furthermore, identifying Dhvani requires additional information, such as the context, mood of the composer, his biography, and his other compositions. The cultural and historical context further complicates the identification of Dhvani, which often requires scholars' commentaries to decode the layers of meanings. Even humans may not be able to identify Dhvani without these inputs, indicating the level of expertise required for such tasks. At present, to the best of our knowledge, no system is available to provide Dhvani analysis for Sanskrit poetry. 

\noindent\textbf{\'Sik\d{s}\={a}\d{s}\d{t}aka analysis:} The \'Sik\d{s}\={a}\d{s}\d{t}aka is a set of 8 verses containing sublime instructions delivered without any specific audience. In contrast to the Bhagavad-g\={\i}t\=a, which is directed towards a particular audience, the \'Sik\d{s}\={a}\d{s}\d{t}aka is meant for inner contemplation and self-instruction. 
We can draw three possible dhvanis from these 8 verses. \textit{The first dhvani deals with the Sambandha, Abhidheya, and Prayojana for a devotee.} Sambandha refers to the relation of the subject with the Abhidheya of devotion, Abhidheya refers to the means of devotion which is being stated, and Prayojana refers to the fruit of devotion. The first five verses of the \'Sik\d{s}\={a}\d{s}\d{t}aka discuss the Sambandha aspect, while the entire set of verses serves as the Abhidheya. The last three verses focus on prayojana, which involves absorption in the remembrance of K\d{r}\d{s}\d{n}a and the intense emotions that this experience brings.

\textit{The second dhvani presents the higher stages of devotional service.} The first five verses describe S\=adhana Bhakti, or the devotional service that is practiced through discipline and rules. The last 3 verses discuss Prema, the ultimate perfection of devotion.
\textit{The third dhvani is the biography of the author.} In the initial verse of the \'Sik\d{s}\={a}\d{s}\d{t}aka, the author's proficiency in Sanskrit grammar and poetry is conveyed by the phrase ``vidyā-vadhū-jīvanam." The Chaitanya Caritam\d{r}ta (CC) further provides a detailed description of the author's logical expertise, wherein he is seen to skillfully refute and reestablish his own arguments, as well as his prowess in teaching devotion to K\d{r}\d{s}\d{n}a through grammar. Additionally, CC narrates an incident where the author defeated the great poetry scholar Ke\'sava K\=a\'smir\={\i}. The phrase ``para\.m vijayate śrī-k\d{r}\d{s}\d{n}a-sa\.{n}kīrtanam" illustrates the author's victorious preaching of the message of sa\.{n}kīrtanam, or congregational chanting of the holy names of K\d{r}\d{s}\d{n}a. CC even goes on to mention how animals like tigers, deers and elephants were inspired to participate in sa\.{n}kīrtanam, resulting in a blissful embrace between a tiger and a deer.
The second verse of the \'Sik\d{s}\={a}\d{s}\d{t}aka reveals the author's egalitarian nature, as he considered everyone to be eligible to participate in sa\.{n}kīrtanam without any discrimination. The third verse introduces the author's b\={\i}ja-mantra, which urges one to develop a taste for sa\.{n}kīrtanam and show mercy towards all living entities. The author's adherence to the principle of humility is evident in his desire for devotees to follow the same. The fourth verse recounts the author's renunciation of wealth, followers, and even his beautiful wife at the age of 24 to embrace the life of a sany\=as\={\i}, or a renunciant. The last three verses of the \'Sik\d{s}\={a}\d{s}\d{t}aka reflect the author's mood of separation in the last 24 years of his life. The CC provides a vivid description of the same. In summary, the author's teachings in the \'Sik\d{s}\={a}\d{s}\d{t}aka are a reflection of his own life experiences and practices.

\noindent\textbf{Is \'Sik\d{s}\={a}\d{s}\d{t}aka an Uttama k\={a}vya?} The criterion for a composition to be qualified as an Uttama k\={a}vya is the presence of Dhvani, which evokes wonder in the reader.\footnote{\textit{idamuttamamatiśayini vya\.{n}gace vācyād dhvanirbudhai\d{h} kathita\d{h}|}1.4, K\={a}vyaprak\={a}\'{s}a} Dhvani is the soul of a k\={a}vya and is of three types: Vastu-Dhvani, Ala\.nk\=ara-Dhvani, and Rasa-Dhvani. Vastu-Dhvani signifies rare facts or ideas; Ala\.nk\=ara-Dhvani indicates figures of speech, and Rasa-Dhvani evokes feelings or moods in the reader.
In the context of Vastu-Dhvani, \'Sik\d{s}\={a}\d{s}\d{t}aka explores the sambandha, abhidheya, and prayojana for a devotee, gradations of devotional service, and the biography of the author. The composition provides an insightful glimpse into the key elements of Bhakti and its various stages, making it a source of rare knowledge for the reader.
In terms of Ala\.nk\=ara-dhvani, \'Sik\d{s}\={a}\d{s}\d{t}aka employs a range of figures of speech such as metaphors, similes, and other literary devices to express its emotional key elements (rasa) without hindrance. The appropriate use of these devices creates a garland-like effect that acknowledges Govinda's glory and expresses the author's devotion and love. The composition's poetic beauty is further enhanced by its rhythm and musicality, making it a delight for readers.
Moreover, \'Sik\d{s}\={a}\d{s}\d{t}aka is a powerful example of how the combination of vibh\=ava and anubh\=ava can be used to evoke vipralambha s\d{r}\.ng\=ara rasa, making it an exemplar of Rasa-Dhvani. The composition conveys the feeling of separation from the divine, which is central to the Vai\d{s}\d{n}ava tradition of Bhakti, and generates a deep emotional response in the reader.
In conclusion, the evidence presented above establishes that \'Sik\d{s}\={a}\d{s}\d{t}aka is indeed an Uttama k\={a}vya. Its exploration of rare knowledge, its ala\.{n}k\={a}ra beauty, and its evocation of deep emotions through Rasa-Dhvani which renders greater amusement than the other meanings make it a masterpiece of Sanskrit literature.

\noindent\textbf{Future direction:} In future research, it is crucial to address the challenges of identifying suggestive meaning in poetry by formulating a well-defined problem statement. One potential direction is to explore the use of machine learning techniques to identify Dhvani in k\={a}vya. This could involve developing annotation schemes that capture different levels of meaning, including literal, indirect, and metaphoric, which can be used to train machine learning models.
Evaluation of performance is another important direction for future research. Metrics could be developed to measure the accuracy of the system's predictions of Dhvani. These metrics could be based on human evaluations or derived automatically by comparing the system's output to expert annotations.
One addional aspect that could be explored in future research is the use of multi-lingual and cross-lingual approaches for identifying Dhvani. Many works of Sanskrit poetry have been translated into other languages, and it would be interesting to investigate whether the same Dhvani can be identified across languages, and whether insights from one language can be used to improve the identification of Dhvani in another language.
Lastly, researchers could explore ways of integrating multiple sources of information to improve the accuracy of Dhvani identification. For example, incorporating biographical information about the composer, their other works, and the cultural and historical context of the composition could help to disambiguate multiple possible meanings and identify the intended Dhvani more accurately.

\subsection{Vakrokti school}
\noindent\textbf{Basics:} Kuntaka's Vakrokti school is a prominent school of Indian poetics that emphasizes the use of oblique expressions in poetry. Kuntaka's seminal work, Vakroktij\={i}vita, lays out the principles of this school and classifies the levels of expression of Vakrokti into six categories (Var\d{n}aviny\=asa, Padap\=urv\=ardha, Pratyay\=a\'srita, V\=akya, Prakara\d{n}a, Prabandha). The first category is phonetic figurativeness, which involves the skillful use of syllables to embellish the sound of the poem. This is closely related to r\={\i}tis and anupr\=as ala\.{n}k\={a}ra, which are types of \'sabd\=ala\.nk\=ara. The second category is Lexical figurativeness, which involves the use of oblique expressions at the level of the root word without suffix. The third category is the obliqueness in the suffixes. The fourth category is involves the figurativeness, which involves the use of oblique expressions at the level of the sentence. In this category, one has to rely on a complete sentence to understand the oblique meaning. Different arth\=ala\.nk\=aras are categoriesed under this category. The fifth category is sectional figurativeness, which involves the twist in the chapter-wise arrangement of the poetry. The sixth category is compositional figurativeness, which involves understanding the oblique meaning by relying on the complete composition. 
Kuntaka considers Vakrokti to be the soul of poetry, and argues that it is the only embellishment possible to the word and its meaning. The use of Vakrokti is essential to create effective poetry, and mastery of the six levels of expression is fundamental to the creation of effective poetry in the Vakrokti school.

\noindent\textbf{Computational aspect:} Upon analysis of Kuntaka's Vakrokti school, it becomes apparent that his theory of oblique expression incorporates various elements of the previously established schools of R\={\i}ti, Rasa, Dhvani, and Ala\.{n}k\={a}ra. Despite this, Kuntaka's contribution to the field lies in his systematic classification of 6 levels of oblique expression. However, the computational challenges posed by the deep semantics of this theory are similar to those discussed in the other schools. Therefore, it is difficult to develop a module capable of accurately identifying Vakrokti in a composition. To date, no computational system exists for computing Vakrokti, as it remains a complex and challenging task.

\noindent\textbf{\'Sik\d{s}\={a}\d{s}\d{t}aka analysis:} The present analysis examines the use of oblique expressions in \'Sik\d{s}\={a}\d{s}\d{t}aka. The study identifies all 6 categories of oblique expressions deployed in the poem. The first category, phonetic figurativeness, concerns the skillful use of syllables to enhance the poetic sound, and we find that \'Sik\d{s}\={a}\d{s}\d{t}aka employs various types of anupr\=asa ala\.nk\=aras.
The second category, word-root level figurativeness, relates to the use of oblique expressions at the level of individual words, and the poem utilizes R\=upak ala\.{n}k\={a}ra for this purpose. The third category, sentential figurativeness, focuses on oblique expressions at the level of sentences, and employs several arthala\.{n}k\={a}ras. The fourth category, Contextual Figurativeness, relies on the poem's context and utilizes a series of R\=upaka ala\.{n}k\={a}ras to create a garland-like effect. The fifth category, compositional figurativeness, involves understanding the oblique meaning by examining the poem's different section. \'Sik\d{s}\={a}\d{s}\d{t}aka uses this category to discuss stages of Bhakti and the concept of sambandha, abhidheya, and prayojana in an oblique manner. The final category, refers to the use of complete composition to convey oblique. We find that the author talks about his biography in the entire composition in an oblique manner. Overall, the study reveals that \'Sik\d{s}\={a}\d{s}\d{t}aka employs all six categories of oblique expressions, highlighting the author's mastery of poetic devices and his ability to convey complex spiritual ideas in a highly nuanced and layered manner.

\noindent\textbf{Future direction:} As the field of NLP advances, there is growing interest in exploring the use of oblique expressions in computational models. In particular, the study of oblique expressions can contribute to the development of natural language generation, sentiment analysis, and machine translation systems.
One potential area of research is the identification and classification of oblique expressions in large corpora of texts. Machine learning algorithms can be trained to recognize patterns of oblique expression usage and distinguish between different types of oblique expressions, such as the six categories identified in the present analysis of \'Sik\d{s}\={a}\d{s}\d{t}aka. This would require the creation of annotated datasets that can be used for training and testing purposes.
Another area of research is the development of computational models that can generate oblique expressions in natural language. Such models could be trained to generate oblique expressions based on the context of the text and the intended meaning. This would require a deep understanding of the different types of oblique expressions and their functions, as well as the ability to generate language that is both grammatically correct and semantically appropriate.
Despite the potential benefits of using oblique expressions in computational models, there are also several challenges that need to be addressed. One of the main challenges is the ambiguity of oblique expressions, which can make it difficult for computational models to accurately interpret their meaning. Another challenge is the variability of oblique expression usage across different languages and cultural contexts, which requires a more nuanced and context-specific approach.
Overall, the study of oblique expressions from a computational perspective holds great promise for advancing our understanding of language and communication, but also presents several challenges that must be addressed through interdisciplinary research and collaboration between linguists, computer scientists, and other experts.

\subsection{Aucitya school}
\noindent\textbf{Basics:} K\d{s}emendra, a renowned poet of the 11th century, introduced the concept of Aucitya in his seminal work Aucitya-vic\=ara-carc\=a. Aucitya, meaning appropriateness, suggests that the elements of poetry, such as Rasa, Ala\.{n}k\={a}ra, and R\={\i}ti, should be used in an appropriate manner.\footnote{\textit{ucitasthānavinyāsādala\.{n}k\d{r}tirala\.{n}k\d{r}ti\d{h} | \-aucityādacyutā nitya\d{m} bhavantyeva gu\d{n}ā gu\d{n}ā\d{h}||} 1.6, Aucitya-vic\=ara-carc\=a} Aucitya is considered the soul of a poem, and appropriateness is regarded as the secret of beautiful composition. This theory has been widely accepted by scholars without any argument and is often referred to as the `Theory of Coordination' since it regulates all aspects of  N\={a}\d{t}ya\'s\={a}stra.
Aucitya is a tool that aids writers in wielding their poem according to their will and effectively delivering their ideas. K\d{s}emendra categorized Aucitya into several classes, namely Pada (word form), V\=akya (sentence), Prabandhan\=artha (meaning of the whole composition), gu\d{n}a (qualities), ala\.{n}k\={a}ra (poetic figures), Rasa (sentiments), etc.
In conclusion, K\d{s}emendra's concept of Aucitya has had a significant impact on the field of poetry. The idea of appropriateness and the use of poetry elements in a balanced manner are essential components of successful composition. The categorization of Aucitya into multiple classes provides a comprehensive framework for analyzing and understanding the nature of poetry. Therefore, Aucitya is an invaluable tool for writers, enabling them to deliver their ideas efficiently and effectively.

\noindent\textbf{Computational aspect:} According to the available literature, no computational system currently exists that provides an analysis of Aucitya. In order to develop a module for Aucitya analysis, we propose considering the mutual compatibility of r\={\i}ti, rasa, and ala\.nk\=ara in the initial implementation. For instance, Vaidarbh\={\i} is considered a compatible r\={\i}ti when deploying the s\d{r}\.ng\=ara rasa, while Gau\d{d}\={\i} is appropriate for the V\={\i}ra rasa. Additionally, certain ala\.nk\=aras are suitable for specific rasas. At a later stage, we may annotate corpora and build data-driven metrics to automatically measure the Aucitya of a composition. However, to accomplish this, it is essential to establish standardized data annotation policies in advance. One major challenge in evaluating poetic appropriateness across various languages, cultures, and time periods is the lack of standardized rules or models. Furthermore, the subjective nature of poetic appropriateness makes it challenging to reach a consensus among scholars and experts on the definition of Aucitya.

\noindent\textbf{\'Sik\d{s}\={a}\d{s}\d{t}aka analysis:} Our analysis suggests that the \'Sik\d{s}\={a}\d{s}\d{t}aka is an exemplary composition that demonstrates Aucitya. The composition employs the Vipralambha s\d{r}\.ng\=ara rasa, and the Vaidarbh\={\i} r\={\i}ti with the M\=adhurya gu\d{n}a, which are perfectly compatible. Additionally, the use of ala\.{n}k\={a}ras is deployed in an appropriate manner, without overwhelming the expression of the rasa. The composition's rasa remains consistent and does not change abruptly. Instead, it gradually intensifies throughout the composition, reaching its pinnacle in the last verse. This uniformity in the rasa's expression contributes to the effectiveness of the composition in evoking the desired emotional response.

\noindent\textbf{Future direction:} Moving forward, the development of a computational system for Aucitya analysis presents an exciting area of research. In terms of future directions, several questions can be explored. Firstly, the effectiveness of the proposed mutual compatibility approach between R\={\i}ti, Rasa, and Ala\.nk\=ara needs to be evaluated by analyzing the performance of the initial module. Furthermore, additional features could be incorporated to improve the accuracy of Aucitya analysis, such as word choice, meter, and rhyme patterns.
Moreover, there is a need to establish standardized guidelines for the annotation of corpora in different languages and cultural contexts. This would facilitate the creation of a large, diverse dataset that can be used to train the Aucitya analysis system. Additionally, research could focus on the development of data-driven metrics that automatically evaluate poetic appropriateness.
Another potential direction for research is to investigate the interplay between language, culture, and poetic appropriateness. This involves analyzing the extent to which poetic conventions vary across different languages and cultures and how these variations affect the evaluation of Aucitya. The impact of historical and temporal changes on poetic conventions can also be explored to understand the evolution of poetic conventions over time.
Finally, collaboration between experts in literature and computer science can help establish a consensus on what constitutes Aucitya. This could lead to the development of a standardized set of rules and models for evaluating poetic appropriateness, which could be applied across different languages, cultures, and time periods.

\section{The proposed framework}
\begin{figure*}[!tbh]
    \centering
    \includegraphics[width=0.9\textwidth]{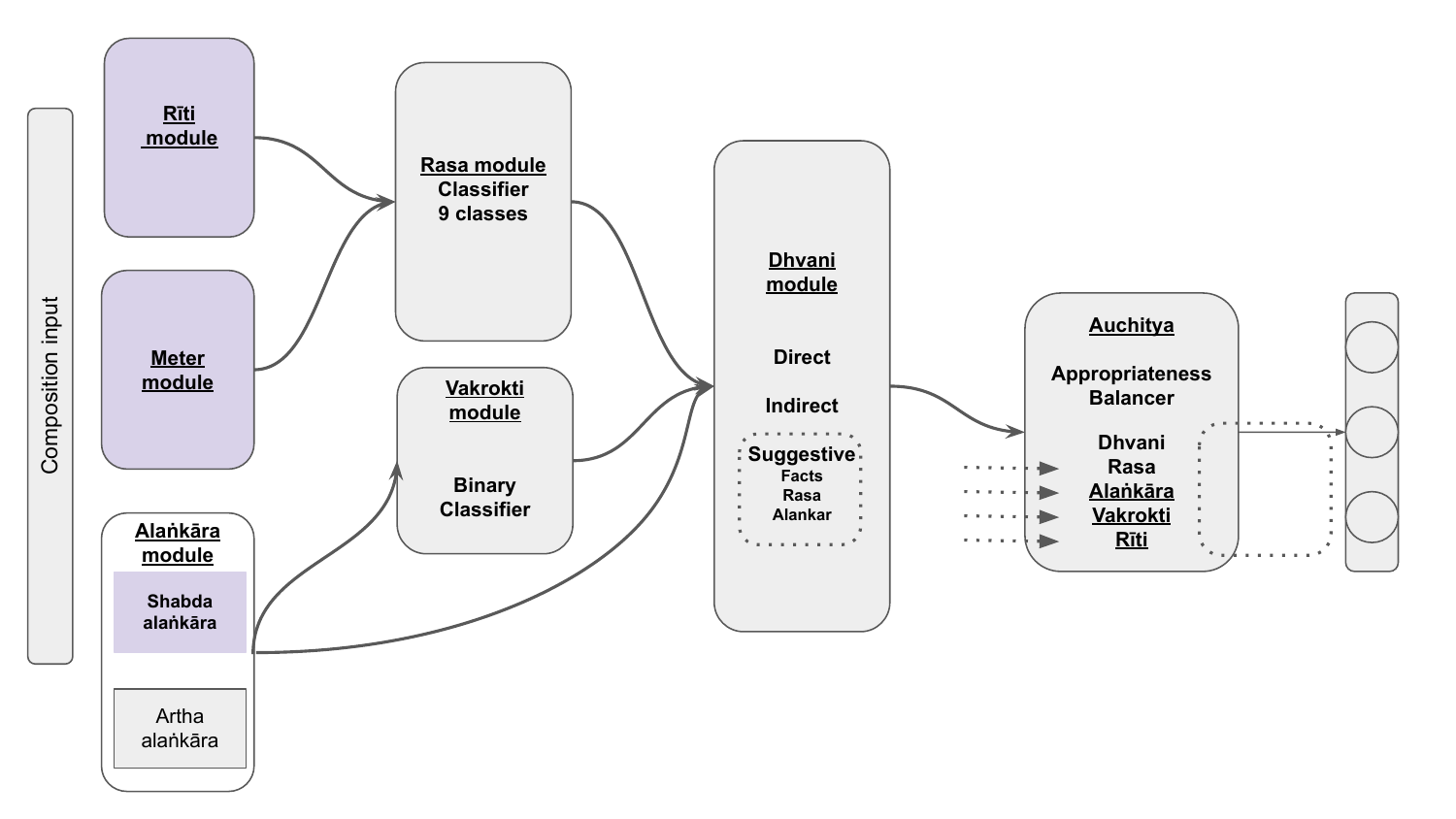}
    \caption{The proposed framework for analysis and classification of poetry using a hierarchical ensembling architecture. It consists of 7 modules, with violet modules being deterministic and the rest being supervised modules, currently supported by human experts. All modules are trained independently. The outputs and features of previous modules are helpful for the next phase modules, and the Aucitya module learns the weight of each module using all their outputs except the meter module. The composition is finally classified into three classes that demonstrate high, medium, or low conformity with the standards of fine Sanskrit poetry, as articulated by experts in K\={a}vya\'s\={a}stra.}
    \label{figure:kāvya_framework}
\end{figure*}
This section presents the description of a proposed framework for the analysis and classification of poetry. The framework is designed to categorize compositions into 3 classes that demonstrate high, medium, or low conformity with the standards of fine Sanskrit poetry, as articulated by experts in K\={a}vya\'s\={a}stra. Figure \ref{figure:kāvya_framework} depicts the proposed framework, which employs a hierarchical ensembling architecture consisting of 7 modules. The violet modules are deterministic, while the remaining modules are learning-based and currently supported by human experts. Each module can be trained independently, and the outputs and features of the previous modules are utilized to enhance the performance of the next phase modules. The Aucitya module is responsible for learning the weight of each module by employing all their outputs, except the meter module.

The first layer of the framework consists of 3 modules namely, R\={\i}ti, Meter and Ala\.{n}k\={a}ra module.
The R\={\i}ti module is the first module in the proposed framework, and it is responsible for identifying the R\={\i}ti of a composition. This is deterministic, and the classification is enabled by various clues provided by the K\={a}vya\'s\={a}stra. The next module, the meter identification module, is also deterministic, and commendable efforts have been made by the SCL community in recent years to develop user-friendly toolkits for the identification of meter \cite{neill-2023-skrutable,Rajagopalan2018AUT,terdalkar-bhattacharya-2023-chandojnanam}.
In the Ala\.{n}k\={a}ra module, the identification of \'{S}abd\=ala\.nk\=ara is considered deterministic, and it involves verifying the occurrence of specific patterns of syllables or words. In contrast, the identification of Arth\=ala\.nk\=ara presents a significant semantic challenge, even for experienced annotators. The module relies on a supervised paradigm of binary classification to identify whether any Arth\=ala\.nk\=ara is present or not.
The Rasa and Vakrokti (oblique) modules are the next two modules in the framework. For the Rasa module, a classification problem with 9 classes of Rasas is framed. These Rasas are \'S\d{r}\.ng\=ara (love and beauty), H\=asya (comedy), Karu\d{n}a (piteous), Raudra (anger), V\={\i}ra (heroism), Bhay\=anaka (fear), B\={\i}bhatsa (disgust), Adbhuta (wonder), and \'S\=anta (peace). It is argued that r\={\i}ti and meter may be helpful and can serve as auxiliary information to identify Rasa; therefore, they are used as features in the Rasa module. The vakrokti module is framed as a binary classification problem to identify the presence of oblique. Here, the Ala\.{n}k\={a}ra module has a strong correlation in contributing to oblique identification.
The next layer of the module is the Dhvani module, which is also a classification problem with 3 classes: direct meaning, indirect meaning, and suggestive meaning. Suggestive meaning has a strong correlation with Rasa, Ala\.{n}k\={a}ra, and Vakrokti. Therefore, these features are plugged into the Dhvani module. Finally, the Aucitya module learns the weight of each module using a supervised paradigm. These weighted features of all modules are used for the final classification of the composition into three classes that demonstrate high, medium, or low conformity with the standards of fine Sanskrit poetry, as articulated by experts in K\={a}vya\'s\={a}stra.
In the absence of annotated data, human experts are relied upon for the modules that are supervised modules. In the future, it is hoped that all the modules can be automated, and researchers can consider training these modules in an end-to-end fashion.
Overall, the proposed framework presents a systematic approach to analyze and classify poetry into different categories based on various features. The framework has the potential to assist poets, scholars, and critics in the evaluation of compositions and provide insights into the nuances of poetry.

\section{Resources and Web Interface}
\noindent\textbf{Annotation:}
This work employs compound tagging, morphological tagging, K\=araka dependency tagging, and Anvaya to annotate the \'Sik\d{s}\={a}\d{s}\d{t}aka. The primary objective of this annotation exercise is to facilitate a literal understanding of the composition. By providing grammatical information, this annotation serves as a baseline to evaluate the Dhvani, or poetic suggestion, of the composition. The verses were annotated by 4 annotators, each an expert in one of the aforementioned tasks. In cases where there was a discrepancy in the annotation process, the annotators discussed and resolved the issue. All annotators hold a minimum academic qualification of a Master in Arts in Sanskrit. The annotations are publicly available via an interactive web-based platform as well as standard text annotation format.
\begin{figure*}[!tbh]
    \centering
    \includegraphics[width=0.9\textwidth]{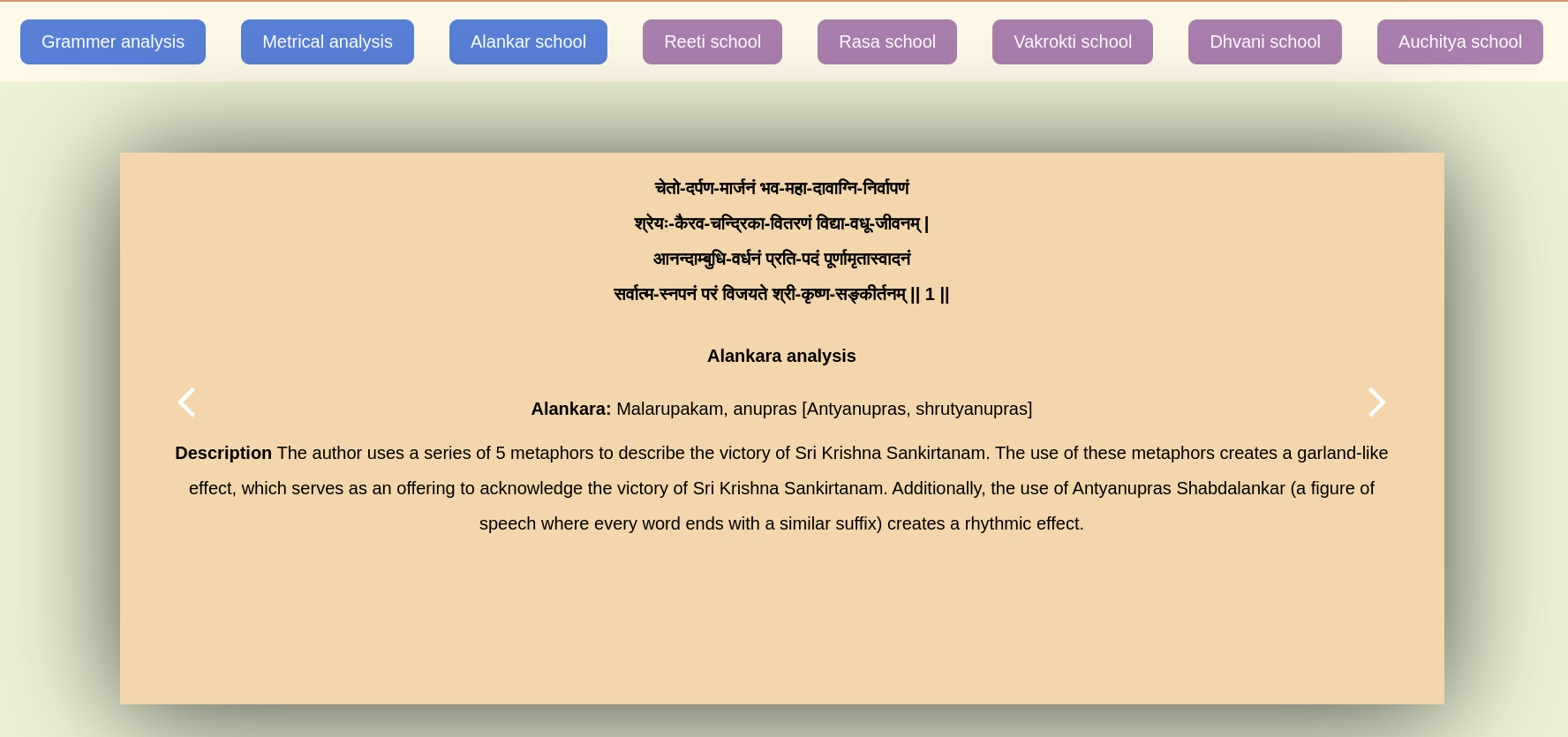}
    \caption{Web-based interactive modules to illustrate annotations and poetry analysis. Here, we showcase the verse-level analysis for Ala\.nk\=ara module.}
    \label{figure:website}
\end{figure*}

\noindent\textbf{\'Sik\d{s}\={a}\d{s}\d{t}aka analysis:}
In the first verse, the author elucidates the initial remarkable effects and, at the culmination, unveils that these extraordinary outcomes arise solely from the practice of Śrīk\d{r}\d{s}\d{n}a Sa\.{n}kīrtana. The verse's structure exudes a jubilant ambiance to portray the triumphal mood of Sa\.{n}kīrtana. Each compound word concludes with an action noun, instilling a sense of dynamism and activity.
The second verse employs a highly intricate grammar, incorporating infrequently utilized verb forms. The grammatical structure assumes a formal character, aiming to evoke sentiments of reverence, awe, or a perceived distance. Through this structure, the author intends to convey the notion of your exalted stature and my own abject insignificance. In instances of formal relationships, such as court proceedings or scholarly discourse, the employment of highly formal language is customary. However, in intimate bonds shared between lovers, simplicity of expression prevails.
The third verse emphasizes brevity as a key feature. In the fourth verse, a profound correlation emerges with King Kulaśekhara's Mukundamālā-stotra, underscoring the potent influence of the Ālvār sampradāya (school) upon the Bhakti school. The word ``kavitā" in this context offers diverse interpretations. Various commentators view it either as the elegant and poetic language found in the ritualistic section of the Vedas (Karmakānda) or as an adjective for ``Sundarī" signifying the depiction of beautiful women in poetry or an ideal feminine figure.
The fifth verse signifies a notable increase in intimacy. For the first time, the beloved is addressed by their personal name, ``Nandatanuja" (the son of Nanda). Previously, the author referred to the revered Lord by a formal and generic appellation. This creates a striking contrast, as the author acknowledges the exalted greatness of the beloved and simultaneously presents themselves as the most fallen servant (ki\.{n}kara).
The sixth verse embodies an intense sentiment of separation, as the author ardently yearns for such an experience to manifest. In this state of longing, the presence of the lover becomes more substantial than their physical existence. Although the beloved is not physically present with the author, their presence is still deeply felt. This poignant contrast evokes a sense of beauty.
In the seventh verse, captivating comparisons (millennium and moment, eyes and monsoon, zero and everything) are employed to depict the separation from the beloved. By juxtaposing these opposing elements, the verse evokes powerful imagery in the reader's mind, giving rise to cognitive dissonance. The final line commences with ``Govinda" and sets the stage for placing blame on the beloved K\d{r}\d{s}\d{n}a.
Within the final verse, an alternative perspective emerges, wherein the author directs an accusation towards his esteemed master. The author asserts that his master's powerful embrace has left him crushed and discarded, resulting in his metaphorical prostration at the master's feet. While acknowledging the master's freedom to exercise personal autonomy, the author laments his own lack of alternative choices, as his affections remain solely reserved for his beloved master. This portrayal reveals the absence of reverential sentiment, emphasizing an intimate and deeply personal connection instead.  

\noindent\textbf{Web-based application}
Figure \ref{figure:website} shows our web application that offers an interactive module to illustrate annotations and poetry analysis. The motivation to build a web-based application to illustrate poetry analysis \cite{delmonte-2015-visualizing,delmonte-prati-2014-sparsar} is that it can help to make poetry analysis more accessible, engaging, and effective, providing students with a rich learning experience that encourages deeper understanding and appreciation of this art form. This template can serve as a proxy to illustrate how an automated system should produce the poetry analysis. 

There are 8 dimensions to our analysis which can be classified into two broad categories, namely, verse-level (grammar, metrical and Ala\.nk\=ara analysis) and global-level (R\={\i}ti, Vakrokti, Rasa, Dhvani and Aucitya analysis). As the names suggest, verse-level analysis shows the analysis for each verse and global analysis considers complete composition to provide the respective analysis. The verse-level modules are shown in blue color and the rest in purple.  Our platform can provide the analysis in two possible ways. First, a user can choose to analyze any verse through all the modules. Otherwise, the user can also opt for choosing any module and analyzing all the verses from the perspective of the chosen module. Our web template leverages the flex module to provide interactive features. We publicly release of the web application along with the codebase to facilitate future research in this direction.

\section{Summary}
In conclusion, this chapter presents an innovative approach to bridge the gap between Sanskrit poetry and computational linguistics by proposing a framework that classifies Sanskrit poetry into levels of the characteristics of fine composition. The proposed framework takes a human-in-the-loop approach, combining the deterministic aspects delegated to machines and the deep semantics left to human experts. This approach addresses the 5 computational challenges involved in the process, including subjectivity, complex language use, cultural context, lack of large labeled datasets, and multi-modality.
The chapter provides a deep analysis of \'Sik\d{s}\={a}\d{s}\d{t}aka, a Sanskrit poem, from the perspective of 6 prominent K\={a}vya\'s\={a}stra schools to illustrate the proposed framework. The analysis establishes that \'Sik\d{s}\={a}\d{s}\d{t}aka is indeed the uttama k\={a}vya composition. Moreover, the article provides compound, dependency, anvaya, meter, rasa, ala\.nk\=ara, and r\={\i}ti annotations for \'Sik\d{s}\={a}\d{s}\d{t}aka and a web application to illustrate the poem's analysis and annotations.
The proposed framework opens new possibilities for automatic analysis and classification of Sanskrit poetry while preserving the rich tradition of Sanskrit poetry.

Future work can explore extending the framework to other classical languages and literary traditions, addressing the challenges in cross-lingual and cross-cultural poetry analysis. Overall, this research demonstrates the potential of interdisciplinary research to enrich our understanding of ancient literature and advance computational linguistics.

%% file: Chapters/Chapter8.tex

\chapter{Conclusion and Future Work}

\label{Chapter7} 

\lhead{Chapter 7. \emph{Conclusion and Future Work}}
This thesis discusses how natural language technologies can make Sanskrit texts more accessible. 
The first contributing chapter (Chapter 3) examines the challenges in Sanskrit Word Segmentation (SWS) and proposes a solution called TransLIST, which uses a module to encode character input and latent-word information, soft-masked attention, and a path ranking algorithm.
Chapter 4 discusses the challenges of dependency parsing in low-resource settings for morphological rich languages (MRLs) and proposes a simple pretraining approach with linguistically motivated auxiliary tasks, and experiments with five low-resource strategies for an ensembled approach.
Chapter 5 investigates whether recent advances in neural networks can outperform traditional hand-engineered feature-based methods in the context-agnostic SaCTI setting and proposes a multi-task learning architecture with morphological tagging and dependency parsing as auxiliary tasks.
Chapter 6 proposes a framework to classify and analyze Sanskrit poetry into levels of the best composition, using a human-in-the-loop approach that combines deterministic aspects delegated to machines and deep semantics left to human experts.
The final chapter (Chapter 7) introduces SanskritShala, a neural-based Sanskrit NLP toolkit that facilitates various computational linguistic analyses for tasks such as word segmentation, morphological tagging, dependency parsing, and compound type identification.
In summary, we strongly feel that this work would be stepping stone towards making digitized Sanskrit texts more accessible.
\section{Summary of the Contributions}
\begin{itemize}
    \item We focused on Sanskrit word segmentation task. To address the shortcomings of existing \textit{purely engineering} and \textit{lexicon driven} approaches, we demonstrate the efficacy of TransLIST as a win-win solution over drawbacks of the individual lines of approaches. TransLIST induces inductive bias for tokenization in a character input sequence using the LIST module, and prioritizes the relevant candidate words with the help of soft-masked attention (SMA module). Further, we propose a novel path ranking algorithm to rectify corrupted predictions using linguistic resources on availability (PRCP module).

    \item We focused on dependency parsing for low-resource MRLs, where getting morphological information itself is a challenge. To address low-resource nature and lack of morphological information, we proposed a simple pretraining method based on sequence labeling that does not require complex architectures or massive labelled or unlabelled data. We show that little supervised pretraining goes a long way compared to transfer learning, multi-task learning, and mBERT pretraining approaches (for the languages not covered in mBERT).

\item We focused on low-resource dependency parsing for multiple languages. We found that our ensembled system can benefit the languages not covered in pretrained models. While multi-lingual pretraining (mBERT and XLM-R) is helpful for the languages covered in pretrained models, LCM pretraining (which simply uses an additional 1,000 morphologically tagged data points) is helpful for the remaining languages. Thus, these findings would help community to pick strategies suitable for their language of interest and come up with robust parsing solutions. Specifically for Sanskrit, our ensembled system superseded the performance of the state-of-the-art \textit{hybrid} system MG-EBM* by 1.2 points (UAS) absolute gain and showed comparable performance in terms of LAS.

\item  Earlier approaches on compound type identification tasks were \textit{context agnostic}; however contextual information is very crucial for this task. Thus, we propose a novel multi-task learning architecture which incorporates contextual information and also enriches it with complementary syntactic information using morphological tagging and dependency parsing auxiliary tasks. Our probing analysis showcased that these auxiliary tasks also serve as a proxy for model prediction explainability. To the best of our knowledge, this is the first time that the importance of these auxiliary tasks has been showcased for SaCTI.


\item We presented an innovative approach to bridge the gap between Sanskrit poetry and computational linguistics by proposing a framework that classifies Sanskrit poetry into levels of the best composition. The proposed framework takes a human-in-the-loop approach, combining the deterministic aspects delegated to machines and the deep semantics left to human experts.
The paper provides a deep analysis of \'Sik\d{s}\={a}\d{s}\d{t}aka, a Sanskrit poem, from the perspective of 6 prominent K\={a}vya\'s\={a}stra schools to illustrate the proposed framework. The analysis establishes that \'Sik\d{s}\={a}\d{s}\d{t}aka is indeed the uttama k\={a}vya composition. 
The proposed framework opens new possibilities for automatic analysis and classification of Sanskrit poetry while preserving the rich tradition of Sanskrit poetry.

\item We offered the first neural-based Sanskrit NLP toolkit, SanskritShala which facilitates diverse linguistic analysis for tasks such as word segmentation, morphological tagging, dependency parsing and compound type identification. It is set up as a web-based application to make the toolkit easier to use for teaching and annotating. All the codebase, datasets and web-based applications are publicly available. We also release word embedding models trained on publicly available Sanskrit corpora and various annotated datasets for 4 intrinsic evaluation tasks to access the intrinsic properties of word embeddings. We strongly believe that our toolkit will benefit people who are willing to work with Sanskrit and will eventually accelerate the Sanskrit NLP research.
\end{itemize}

\section{Future Work}
The main objective of the thesis was to create advanced neural models that can handle various linguistic tasks in Sanskrit, including lexical, syntactic, and semantic tasks. During the research process, we discovered several new avenues for further exploration that can build on our work. In this section, we briefly mention some of these directions.

\begin{itemize}
    \item In the Sanskrit word segmentation task, The preliminary requirement to extend TransLIST for the languages which also exhibit {\sl sandhi} phenomenon is {\sl lexicon-driven} shallow parser similar to Sanskrit Heritage Reader (SHR). Otherwise, the natural choice is the proposed purely engineering variant TransLIST\textsubscript{ngram}. It would be interesting to check if TransLIST and TransLIST\textsubscript{ngram} can be used together.

    \item  For the dependency parsing task, we could not evaluate on complete UD due to limited available compute resources (single GPU), hence we selected 7 representative languages for our experiments.

    \item  In  context-sensitive compound type identification problem, we limit our study to the {\sl purely engineering} data-driven settings.  However, it will be interesting to see whether rules extracted from P\={a}\d{n}inian grammar can help to improve the performance of less populated classes.

    \item We plan to extend SanskritShala  by integrating more downstream tasks such as Post-OCR correction, named entity recognition, verse recommendation, word order linearisation, and machine translation. Improving the performance of existing tasks would be important. For example, the current dependency parser is very fragile (performance drops by 50\%) in the poetry domain.
     We plan to extend the current work by augmenting logical rules \cite{li-srikumar-2019-augmenting,nandwani2019primal} derived from P\={a}\d{n}inian grammar in the proposed approach. 
\end{itemize}

\noindent\textbf{Future Directions in Aesthetics of Sanskrit Poetry:}
This section aims to provide future directions for 7 modules of our framework, each corresponding to one of the 6 schools of k\={a}vya\'s\={a}stra and one module from Chanda\'s\'s\={a}stra. In each subsection, the feasibility of developing a computational system based on the same, and potential avenues for future research in this regard.

\noindent\textbf{Meter school:} In future directions, it would be interesting to explore the correlation between the meter's intrinsic mood and the poetry's rasa. It would also be worthwhile to investigate the relationship between poets and the meters they have employed. This could involve examining patterns in their favorite choices and attempting to identify the poet based on the composition itself.  Additionally, there is potential for exploring whether the occurrence of 
Rasado\d{s}a (obstruction in the enjoyment of mellow) can be predicted through automatic identification of the composition's rasa. These investigations could contribute to a deeper understanding of the relationship between meters, poetry, and emotion in Sanskrit literature.

\noindent\textbf{Ala\.{n}k\={a}ra school:} Moving forward, several directions for research on ala\.nk\=aras can be explored. One crucial area is the formulation of the problem of identifying ala\.nk\=aras. Since ala\.nk\=aras can be assigned to complete or partial verses, or sequences of syllables or pairs of words, determining the appropriate approach for identifying them is non-trivial. Different formulations, such as sentence classification, sequence labeling, or structured prediction, can be employed.
Another interesting direction for research is to investigate which ala\.{n}k\={a}ras are more beautiful and how we can compare two compositions based solely on their use of ala\.nk\=aras. Defining a basis for evaluating beauty in poetry is a challenging problem in itself. However, understanding which ala\.{n}k\={a}ras contribute more to the aesthetic appeal of a poem and how to measure this appeal can be valuable for poets and scholars alike.
Finally, exploring the correlation of ala\.{n}k\={a}ra school with other schools of k\={a}vya\'s\={a}stra, such as rasa and dhvani, is a promising area for future research. Understanding how these different aspects of poetry interact can provide insight into the complex mechanisms of poetic expression and perception. Overall, these future directions offer exciting opportunities to deepen our understanding of ala\.nk\=aras and their role in poetry.

\noindent\textbf{Rasa school:} The future direction of computational linguistics in identifying rasa in poetry requires a comprehensive approach that covers all the crucial dimensions of the problem. 
Data annotation is one crucial aspect of the future direction of computational linguistics in identifying rasa in poetry. Data annotation involves the manual tagging of texts with information such as the dominant rasa, vibh\=ava, anubh\=ava, and vyabhic\=ar\={\i} bh\=ava. 
Possible approaches that can be taken include using supervised machine learning algorithms to identify patterns in the text that are associated with particular emotions. This involves training a model on a labeled dataset of texts annotated with the dominant rasa and associated vibh\=ava, anubh\=ava, and vyabhic\=ar\={\i} bh\=ava. 
Another approach is to use unsupervised machine learning algorithms such as clustering and topic modeling on words to identify the dominant emotions expressed in the text. This approach requires minimal human annotation.
In addition, incorporating contextual information such as the author, genre, and historical period can also be useful in identifying the dominant rasa in a text. 

\noindent\textbf{R\={\i}ti school} A natural direction for future research would be to consolidate the provided clues and develop a rule-based computational system for the identification of R\={\i}ti in various compositions by different poets. This would enable automated analysis and assessment of R\={\i}ti in k\={a}vya. An empirical investigation could be carried out to explore the degree to which R\={\i}ti is effective in identifying the rasa of a k\={a}vya. Additionally, further research could explore the correlation of R\={\i}ti school with other k\={a}vya\'s\={a}stra schools. V\=amana, the founder of the R\={\i}ti school, maintains that different r\={\i}ti can provide unique experiences for readers. Therefore, it would be of interest to conduct a cognitive analysis of brain signals of readers after exposing them to compositions of different r\={\i}tis. This would provide valuable insights into the effect of r\={\i}ti on the reader's cognitive processes and emotional responses to k\={a}vya.

\noindent\textbf{Dhvani school:} In future research, it is crucial to address the challenges of identifying suggestive meaning in poetry by formulating a well-defined problem statement. One potential direction is to explore the use of machine learning techniques to identify Dhvani in k\={a}vya. This could involve developing annotation schemes that capture different levels of meaning, including literal, indirect, and metaphoric, which can be used to train machine learning models.
Evaluation of performance is another important direction for future research. Metrics could be developed to measure the accuracy of the system's predictions of Dhvani. These metrics could be based on human evaluations or derived automatically by comparing the system's output to expert annotations.
One addional aspect that could be explored in future research is the use of multi-lingual and cross-lingual approaches for identifying Dhvani. Many works of Sanskrit poetry have been translated into other languages, and it would be interesting to investigate whether the same Dhvani can be identified across languages, and whether insights from one language can be used to improve the identification of Dhvani in another language.
Lastly, researchers could explore ways of integrating multiple sources of information to improve the accuracy of Dhvani identification. For example, incorporating biographical information about the composer, their other works, and the cultural and historical context of the composition could help to disambiguate multiple possible meanings and identify the intended Dhvani more accurately.

\noindent\textbf{Vakrokti school:} As the field of NLP advances, there is growing interest in exploring the use of oblique expressions in computational models. In particular, the study of oblique expressions can contribute to the development of natural language generation, sentiment analysis, and machine translation systems.
One potential area of research is the identification and classification of oblique expressions in large corpora of texts. Machine learning algorithms can be trained to recognize patterns of oblique expression usage and distinguish between different types of oblique expressions, such as the six categories identified in the present analysis of \'Sik\d{s}\={a}\d{s}\d{t}aka. This would require the creation of annotated datasets that can be used for training and testing purposes.
Another area of research is the development of computational models that can generate oblique expressions in natural language. Such models could be trained to generate oblique expressions based on the context of the text and the intended meaning. This would require a deep understanding of the different types of oblique expressions and their functions, as well as the ability to generate language that is both grammatically correct and semantically appropriate.
Despite the potential benefits of using oblique expressions in computational models, there are also several challenges that need to be addressed. One of the main challenges is the ambiguity of oblique expressions, which can make it difficult for computational models to accurately interpret their meaning. Another challenge is the variability of oblique expression usage across different languages and cultural contexts, which requires a more nuanced and context-specific approach.
Overall, the study of oblique expressions from a computational perspective holds great promise for advancing our understanding of language and communication, but also presents several challenges that must be addressed through interdisciplinary research and collaboration between linguists, computer scientists, and other experts.

\noindent\textbf{Aucitya school:} Moving forward, the development of a computational system for Aucitya analysis presents an exciting area of research. In terms of future directions, several questions can be explored. Firstly, the effectiveness of the proposed mutual compatibility approach between R\={\i}ti, Rasa, and Ala\.nk\=ara needs to be evaluated by analyzing the performance of the initial module. Furthermore, additional features could be incorporated to improve the accuracy of Aucitya analysis, such as word choice, meter, and rhyme patterns.
Moreover, there is a need to establish standardized guidelines for the annotation of corpora in different languages and cultural contexts. This would facilitate the creation of a large, diverse dataset that can be used to train the Aucitya analysis system. Additionally, research could focus on the development of data-driven metrics that automatically evaluate poetic appropriateness.
Another potential direction for research is to investigate the interplay between language, culture, and poetic appropriateness. This involves analyzing the extent to which poetic conventions vary across different languages and cultures and how these variations affect the evaluation of Aucitya. The impact of historical and temporal changes on poetic conventions can also be explored to understand the evolution of poetic conventions over time.
Finally, collaboration between experts in literature and computer science can help establish a consensus on what constitutes Aucitya. This could lead to the development of a standardized set of rules and models for evaluating poetic appropriateness, which could be applied across different languages, cultures, and time periods.

%% file: Appendices/AppendixA.tex

\chapter{SanskritShala: A Neural Sanskrit NLP Toolkit with Web-Based Interface for Pedagogical and Annotation Purposes} 

\label{AppendixA} 

\lhead{Appendix A. \emph{SanskritShala: A Neural Sanskrit NLP Toolkit}} 

We present  a  neural  Sanskrit  Natural Language Processing (NLP)  toolkit named SanskritShala\footnote{It means `a school of Sanskrit'.} to facilitate computational linguistic analyses for several tasks such as word segmentation, morphological tagging, dependency parsing, and compound type identification. Our systems currently report state-of-the-art performance on available benchmark datasets for all tasks. SanskritShala is deployed as a web-based application, which allows a user to get real-time analysis for the given input. It is built with easy-to-use interactive data annotation features that allow annotators to correct the system predictions when it makes mistakes. We publicly release the source codes of the 4 modules included in the toolkit, 7 word embedding models that have been trained on publicly available Sanskrit corpora and multiple annotated datasets such as word similarity, relatedness, categorization, analogy prediction to assess intrinsic properties of word embeddings.  So far as we know, this is the first neural-based Sanskrit NLP toolkit that has a web-based interface and a number of NLP modules. We are sure that the people who are willing to work with Sanskrit will find it useful for pedagogical and annotative purposes. SanskritShala is available at: \url{https://cnerg.iitkgp.ac.in/sanskritshala}. The demo video of our platform can be accessed at: \url{https://youtu.be/x0X31Y9k0mw4}.

\section{Motivation}
Sanskrit is a culture-bearing and knowledge-preserving language of ancient India. Digitization has come a long way, making it easy for people to access ancient Sanskrit manuscripts \cite{goyal-etal-2012-distributed,adiga-etal-2021-automatic}. 
However, we find that the utility of these digitized manuscripts is limited due to the  user's lack of language expertise and various linguistic phenomena exhibited by the language.
This motivates us to investigate how we can utilize natural language technologies to make Sanskrit texts more accessible.

The aim of this research is to create neural-based Sanskrit NLP systems that are accessible through a user-friendly web interface. The Sanskrit language presents a range of challenges for building deep learning solutions, such as the \textit{sandhi} phenomenon, a rich morphology, frequent compounding, flexible word order, and limited resources \cite{translist,Graph-Based,sandhan-etal-2021-little,sandhan-etal-2019-revisiting}. To overcome these challenges, 4 preliminary tasks were identified as essential for processing Sanskrit texts: word segmentation, morphological tagging, dependency parsing, and compound type identification. The word segmentation task is complicated by the \textit{sandhi} phenomenon, which transforms the word boundaries \cite{translist}. The lack of robust morphological analyzers makes it challenging to extract morphological information, which is crucial for dependency parsing. Similarly, dependency information is essential for several downstream tasks such as word order linearisation \cite{krishna-etal-2019-poetry} which helps to decode possible interpretation of the poetic composition. Additionally, the ubiquitous nature of compounding in Sanskrit is difficult due to the implicitly encoded semantic relationship between its constituents \cite{sandhan-etal-2022-novel}. 
These 4 tasks can be viewed as a preliminary requirement for developing robust NLP technology for Sanskrit.  
Thus, we develop novel neural-based linguistically informed architectures for all 4 tasks, reporting state-of-the-art performance on Sanskrit benchmark datasets \cite{sandhan-etal-2022-novel,translist,sandhan_systematic}. 
 We also illustrate the efficacy of our language agnostic proposed systems in multiple low-resource languages. 

In this work, we introduce  a  neural  Sanskrit  NLP toolkit named SanskritShala\footnote{Roughly, it can be translated as `a school of Sanskrit'.} to assist computational linguistic analyses involving multiple tasks such as word segmentation, morphological tagging, dependency parsing, and compound type identification.
SanskritShala is also deployed as a web application that enables users to input text and gain real-time linguistic analysis from our pretrained systems. It is also equipped with user-friendly interactive data annotation capabilities that allow annotators to rectify the system when it makes errors. It provides the following benefits: (1) A user with no prior experience with deep learning can utilise it for educational purposes. (2) It can function as a semi-supervised annotation tool that requires human oversight for erroneous corrections.   
We publicly release the source code of the 4 modules included in the toolkit, 7 word embedding models that have been trained on publicly available Sanskrit corpora and multiple annotated datasets such as word similarity, relatedness, categorization, analogy prediction to measure the word embeddings' quality.
 To the best of our knowledge, this is the first neural-based Sanskrit NLP toolkit that contains a variety of NLP modules integrated with a web-based interface.

Summarily, our key contributions are as follows:
\begin{itemize}
    \item We introduce the first neural Sanskrit NLP toolkit to facilitate automatic linguistic analyses for 4  downstream tasks (\S \ref{neural_toolkit}).
    \item We release 7 pretrained Sanskrit embeddings and suit of 4 intrinsic evaluation datasets to measure the word embeddings' quality.
    \item We integrate SanskritShala with a user-friendly web-based interface which is helpful for pedagogical purposes and in developing annotated datasets (\S \ref{webinterface}).
    \item We publicly release codebase and datasets of all the modules of SanskritShala which currently mark the state-of-the-art results.\footnote{\url{https://github.com/Jivnesh/SanskritShala}}
\end{itemize}

\begin{figure*}[!tbh]
    \centering
    \subfigure[]{\includegraphics[width=0.45\textwidth]{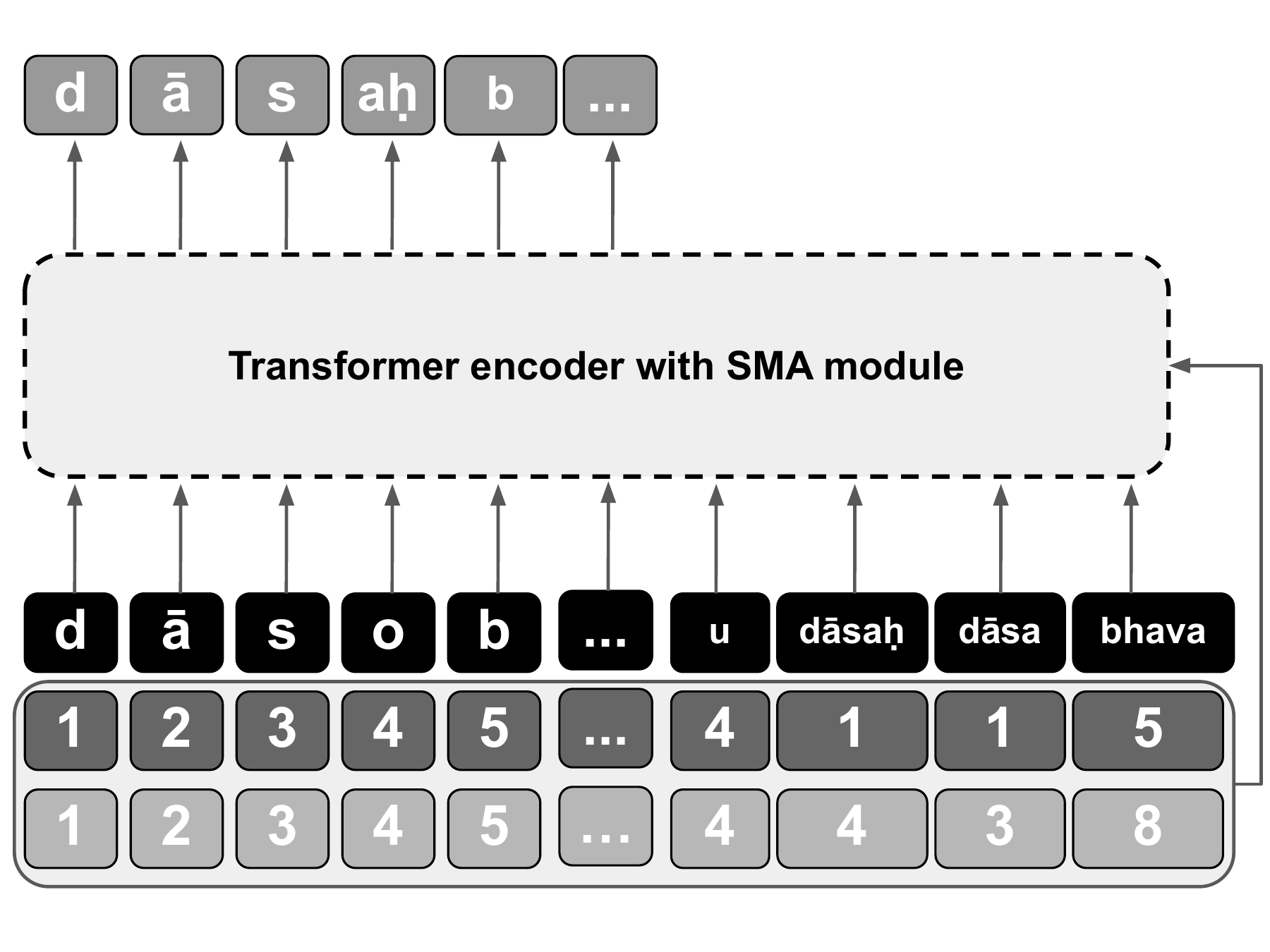}\label{fig:flat_architecture_ACL}}
    \subfigure[]
    {\includegraphics[width=0.45\textwidth]{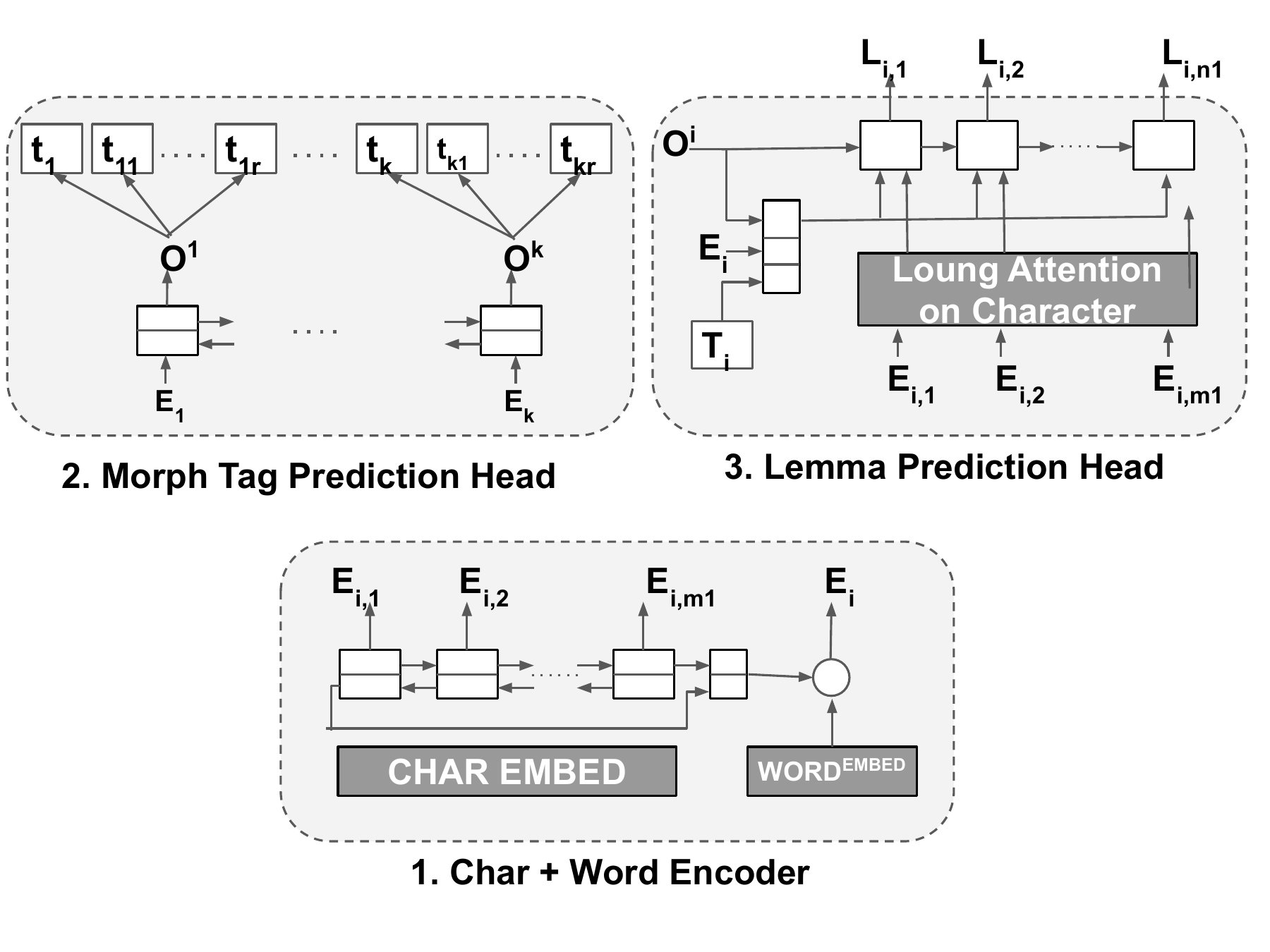}\label{fig:LemmaTag_ACL}}
    \subfigure[]
    {\includegraphics[width=0.45\textwidth]{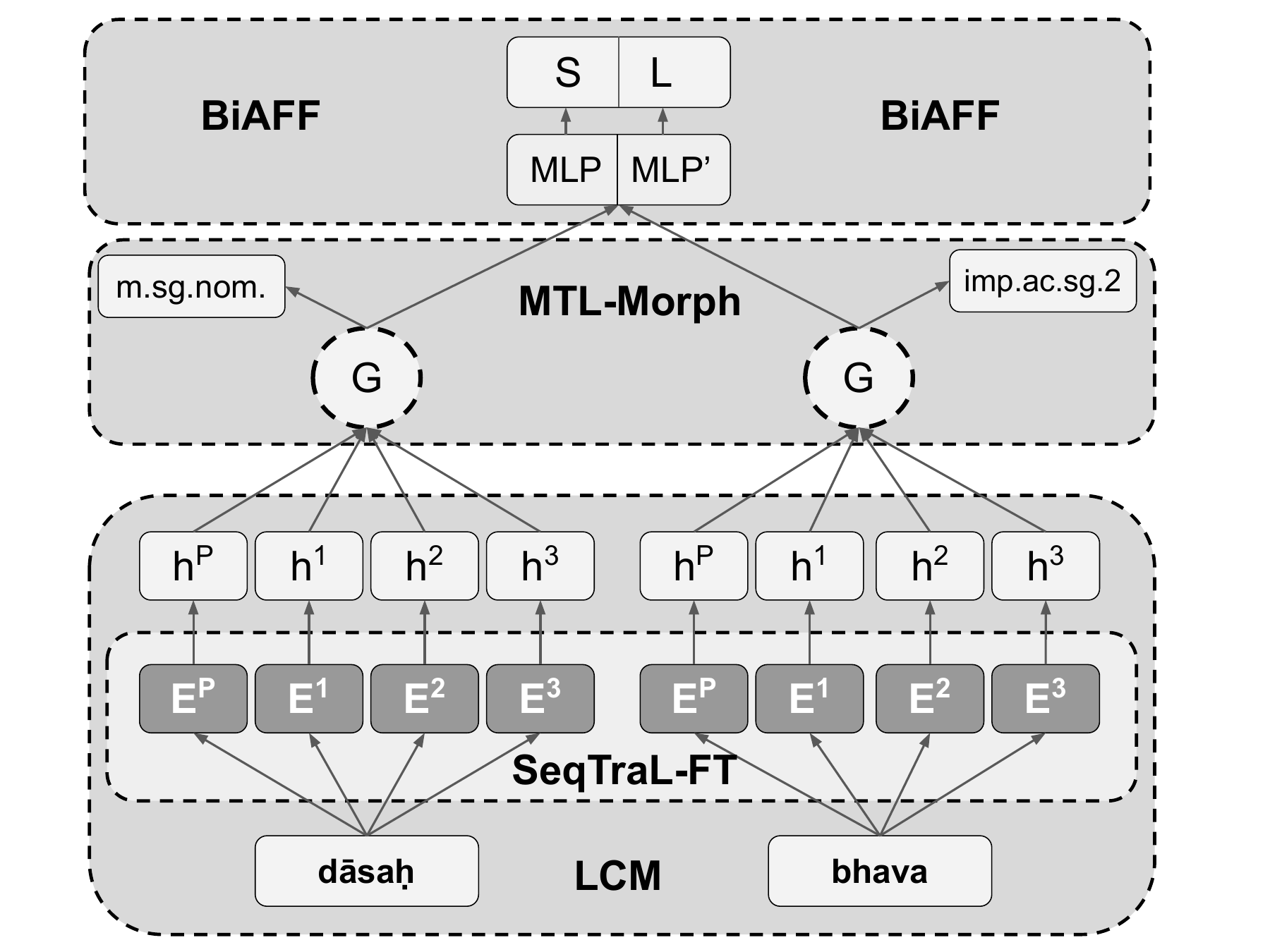}\label{fig:dep_architecture_ACL}}
    \subfigure[]
    {\includegraphics[width=0.45\textwidth]{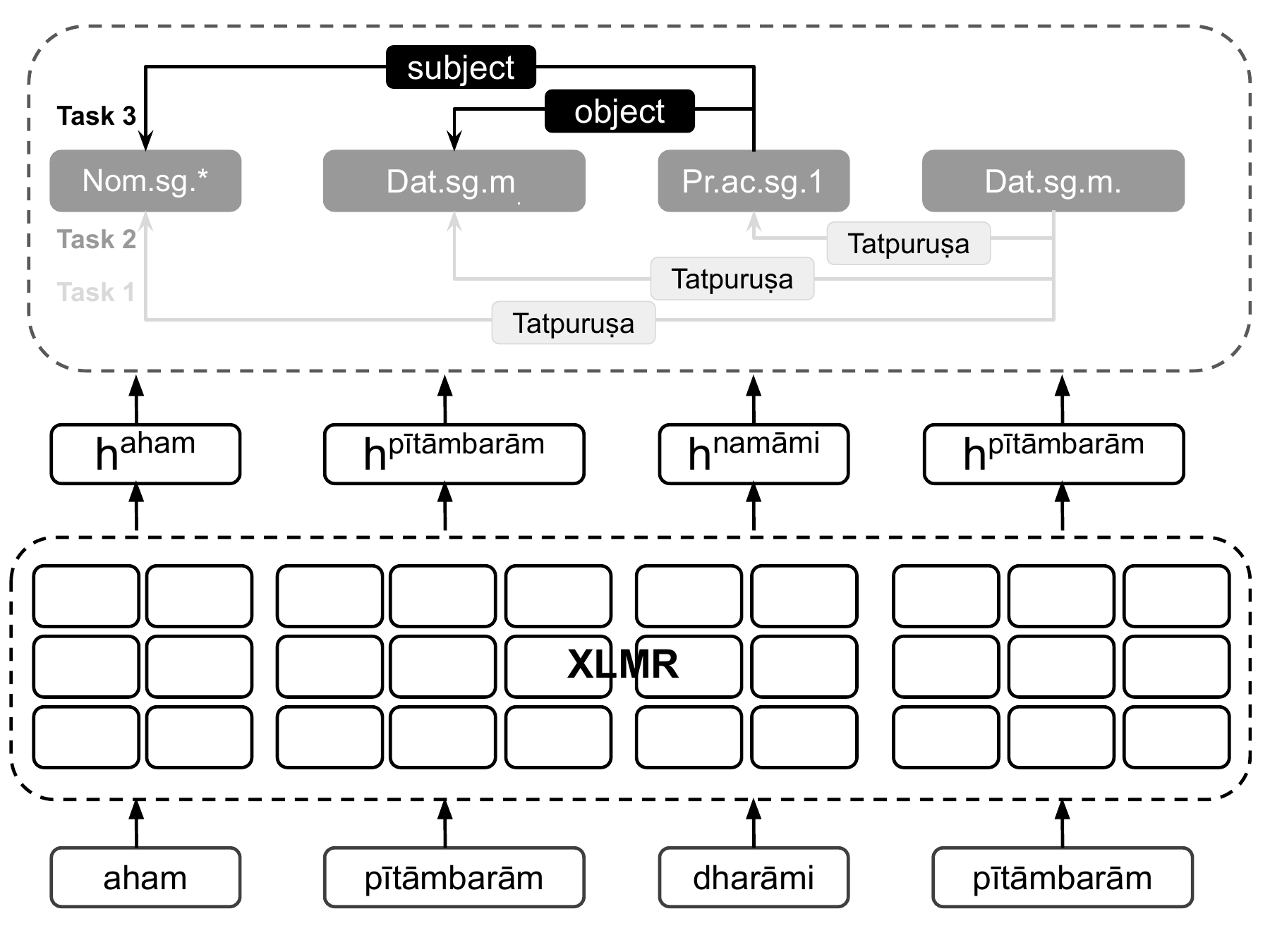}\label{fig:sacti_architecture_ACL}}
    \caption{(a) Toy illustration of the TransLIST system.``{\sl d\={a}sobhava}''. Translation: ``Become a servant.'' (b) LemmaTag architecture in which multi-task learning formulation is leveraged to predict morphological tags and lemmas by employing bidirectional RNNs with character-level and word-level representations. (c) Proposed ensembled architecture for dependency parsing   (d) Toy example illustrating the context-sensitive multi-task learning system: ``{\sl aham p\={\i}ta-ambaram dhar\={a}mi}'' (Translation: ``I wear a yellow cloth'') where `{\sl p\={\i}ta-ambaram}' is a compound having \textit{Tatpuru\d{s}a} semantic class according to the context presented.}
    \label{figure:complete_TransLIST}
\end{figure*}

\section{A Neural NLP Sanskrit Toolkit}
In this section, we describe SanskritShala, which is a neural Sanskrit NLP toolkit designed to aid computational linguistic analysis including various tasks, such as word segmentation, morphological tagging, dependency parsing, and compound type identification. It is also available as a web application that allows users to input text and obtain real-time linguistic analysis from our pretrained algorithms.
We elucidate SanskritShala by first elaborating on
its key modules.
\label{neural_toolkit}
\noindent\textbf{Word Tokenizer:}
Earlier \textit{lexicon-driven} systems for Sanskrit word segmentation (SWS) rely on Sanskrit Heritage Reader \cite[SHR]{goyaldesign16}, a rule-based system, to obtain the exhaustive solution space for segmentation, followed by diverse approaches to find the most valid solution. However, these systems are rendered moot while stumbling out-of-vocabulary words. Later, \textit{data-driven} systems for SWS are built using the most recent techniques in deep learning, but can not utilize the available candidate solution space.  To overcome the drawbacks of both lines of modelling, we build a \textbf{Tran}sformer-based \textbf{L}inguistically-\textbf{I}nformed \textbf{S}anskrit \textbf{T}okenizer (TransLIST) \cite{translist} containing (1) a component that encodes the character-level and word-level potential candidate solutions, which tackles \textit{sandhi} scenario typical to SWS and is compatible with partially available candidate solution space, (2) a novel soft-masked attention for prioritizing selected set of candidates and (3) a novel path ranking module to correct the mispredictions. 
Figure \ref{fig:flat_architecture_ACL}  illustrates the TransLIST architecture, where the candidate solutions obtained from SHR are used as auxiliary information. 
In terms of the perfect match (PM) evaluation metric, TransLIST surpasses the existing state-of-the-art \cite{hellwig-nehrdich-2018-sanskrit}  by 7.2 absolute points.
\noindent\textbf{Morphological Tagger:}
Sanskrit is a morphologically-rich fusional Indian language with 40,000 possible labels for inflectional morphology \cite{krishna-etal-2020-graph,gupta-etal-2020-evaluating}, where homonymy and syncretism are predominant \cite{krishna-etal-2018-free}. We train a neural-based architecture \cite[LemmaTag]{kondratyuk-etal-2018-lemmatag} on Sanskrit dataset \cite{hackathon}. Figure \ref{fig:LemmaTag_ACL} illustrates the system architecture in which multi-task learning formulation is leveraged to predict morphological tags and lemmas by employing bidirectional RNNs with character-level and word-level representations. We find that both tasks help by sharing the encoder, predicting label subcategories, and feeding the tagger output as input to the lemmatizer \cite{kondratyuk-etal-2018-lemmatag}.
Currently, our system trained on the Sanskrit dataset stands first on the Hackathon dataset \cite{hackathon} leaderboard. 
\noindent\textbf{Dependency Parser:}
We focus on low-resource Sanskrit dependency parsing. Numerous strategies are tailored to improve task-specific performance in low-resource scenarios.
 Although these strategies are well-known to the NLP community, it is not obvious to choose the best-performing ensemble of these methods for a low-resource language of interest, and not much effort has been given to gauging the usefulness of these methods. We investigate 5 low-resource strategies in our ensembled Sanskrit parser \cite{sandhan_systematic}: data augmentation, multi-task learning, sequential transfer learning, pretraining, cross/mono-lingual and self-training.
Figure \ref{fig:dep_architecture_ACL} shows our ensembled system, which supersedes the current state-of-the-art \cite{krishna-etal-2020-keep} for Sanskrit by 1.2 points absolute gain (Unlabelled Attached Score) and shows on par performance in terms of Labelled Attached Score. Our extensive multi-lingual experimentation on a variety of low-resource languages demonstrates significant improvements for languages that are not covered by pretrained language models.
\noindent\textbf{Sanskrit Compound Type Identifier:}
SaCTI is a multi-class classification task that identifies semantic relationships between the components of a compound. Prior methods only used the lexical information from the constituents and didn't take into account the most crucial syntactic and contextual information for SaCTI. However, the SaCTI task is difficult mainly due to the implicitly encrypted context-dependent semantic relationship between the compound's constituents. 
Thus, we introduce a novel multi-task learning approach \cite{sandhan-etal-2022-novel} (Figure \ref{fig:sacti_architecture_ACL}) which includes contextual information and enhances the complementary syntactic information employing morphological parsing and dependency parsing as two auxiliary tasks. SaCTI outperforms the state-of-the-art by  $7.7$ points (F1-score) absolute gain on the benchmark datasets. 
\begin{figure*}[!tbh]
    \centering
    \includegraphics[width=0.9\textwidth]{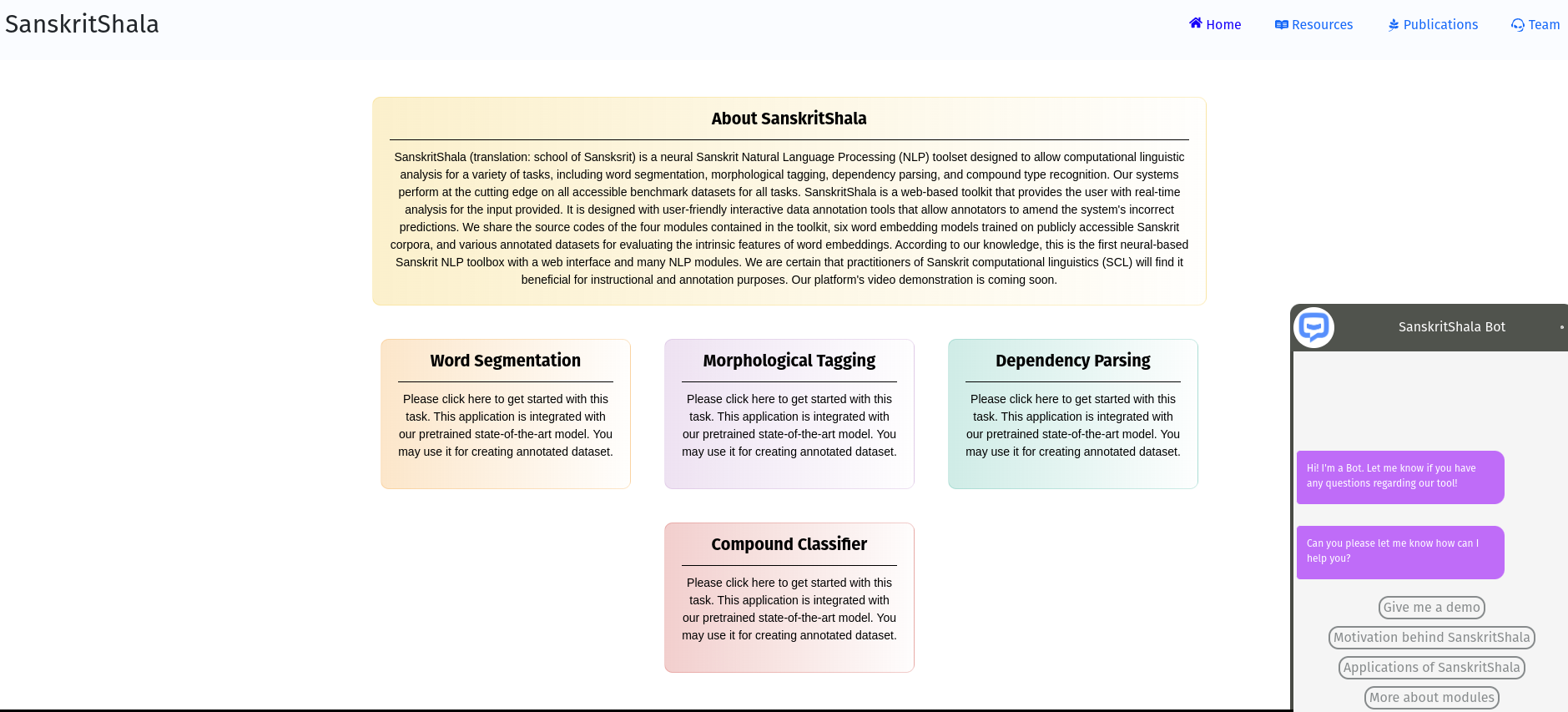}
    \caption{The web interface of the SanskritShala. 
    At the bottom right, a rule-based chatbot is added to navigate users on the platform to give users a user-friendly experience.}
    \label{figure:sanskritshala}
\end{figure*}
\begin{figure*}[!tbh]
    \centering
       \subfigure[]{\includegraphics[width=0.45\textwidth]{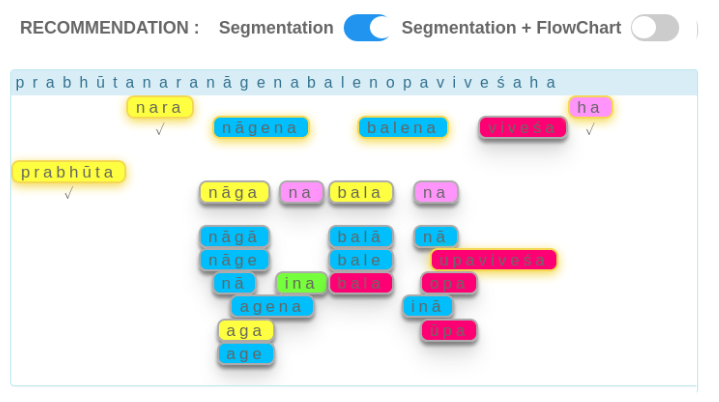}\label{fig:shr++}} 
    \subfigure[]{\includegraphics[width=0.45\textwidth]{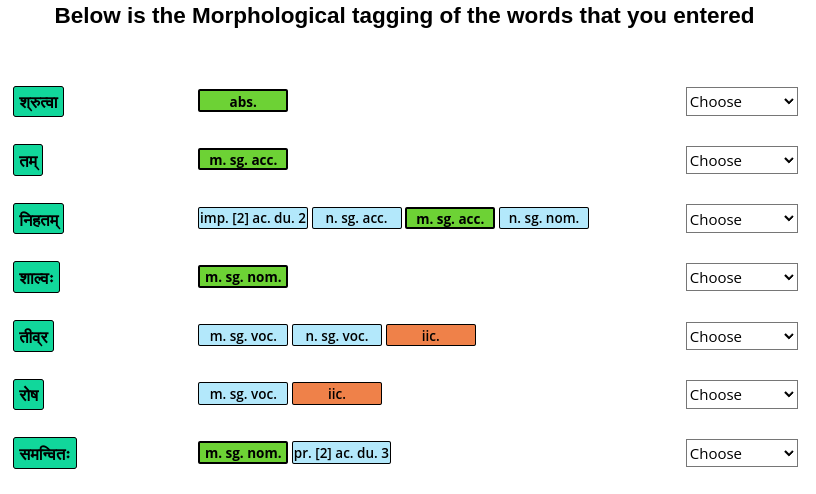}\label{fig:morph_tagger}}
    \caption{(a) The candidate solution space generated by SHR for the word segmentation task and the predicted solution by our pretrained model is recommended for the sequence ‘\textit{prabh\={u}tanaran\={a}gena balenopavive\'sa ha}’ using a yellow highlight. We expect a user to select a solution considering the recommendations shown by our system. The figure shows the stage at which the first two segments are selected manually. (b) Morphological Tagger: For each word, we show possible morphological analyses suggested by SHR as well as our system prediction in green if it falls in SHR's candidate space, otherwise in orange. Note that we prefer to use different examples in (a) and (b) for illustrating one of the features of a morphological tagger: If model prediction lies outside of SHR space, then it is highlighted with orange.}
    \label{figure:SHR}
\end{figure*}
\begin{figure*}[!tbh]
    \centering
      \subfigure[]{\includegraphics[width=0.45\textwidth]{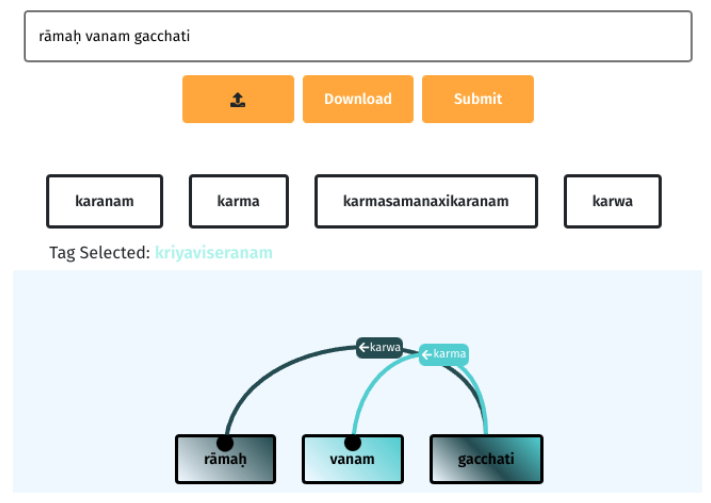}\label{fig:dep_parser}}
       \subfigure[]{\includegraphics[width=0.45\textwidth]{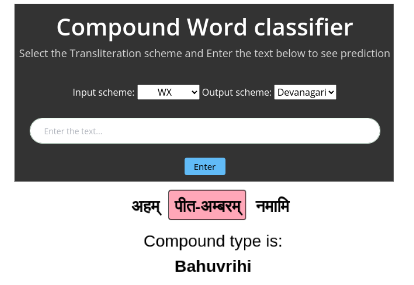}\label{fig:sacti}} 
  
\caption{ (a) Dependency parser: Interactive module for the dependency parsing task which directly loads predicted dependency trees from our pretrain model and allows user to correct mispredictions using our interactive interface. (b) Illustration of compound identifier}
    \label{figure:pipeline}
\end{figure*}
\section{Web-based Interface of SanskritShala}
\label{webinterface}
Figure \ref{figure:sanskritshala} shows our SanskritShala toolkit that offers interactive web-based predictions for various NLP tasks. The toolkit is built using React framework, which makes it user-friendly and easy to use. One of the tasks it handles is the word segmentation task, which is built on top of the web-based application called SHR++. The SHR++ demonstration is depicted in Figure \ref{fig:shr++}. The user inputs a Sanskrit string, which is then sent in real-time to SHR for potential word splits. The system prediction is then obtained from the pretrained word tokenizer. The human annotator is presented with the candidate solution space, with the system prediction highlighted in yellow.
The toolkit also features a flask-based application for morphological tagging, which takes user input and scrapes possible morphological tags for each word using SHR. As shown in Figure \ref{fig:morph_tagger}, the predictions of the pretrained morphological tagger are displayed in green or orange, depending on whether they are present in the candidate solution of SHR or not. The user can also add a new tag if the actual tag is missing in the SHR solution space or the system's prediction.
For the dependency parsing module, we have built a react-based front-end. The user input is passed to the pretrained model to generate a dependency structure. As illustrated in Figure \ref{fig:dep_parser}, the front-end automatically loads the predicted dependency tree and allows the user to make corrections if there are any mispredictions. Additionally, Figure \ref{fig:sacti} shows a flask-based application for the compound type identifier, where users can give input to the system through its web interface. The final annotations can be downloaded after each individual module. We plan to maintain the progress of Sanskrit NLP and offer an overview of available datasets and existing state-of-the-art via the leaderboard for various tasks.

\noindent\textbf{Interactive Chatbot:}
SanskritShala-bot is a rule-based chatbot that makes it easy to automate simple and repetitive user requests, like answering frequently asked questions and directing users to relevant resources. It is also easier to set up and maintain than AI-powered chatbots, which are more complicated, which makes them a good choice for us with limited resources. SanskritShala-bot is especially useful for helping many users quickly and effectively. It helps familiarize users with a platform by providing them with information and guidance on how to use it. It can answer questions about the platform's features, help users find their way around it, and explain step-by-step how to do certain tasks. This can make it easier for users to get started and  leading to a better experience.


\section{Resources and Dataset Contributions}
\noindent\textbf{Pretrained word embeddings for Sanskrit:}
There are two types of embedding methods: static and contextualized. Table \ref{tab:introTable} shows how they are categorized based on the smallest unit of input to the embedding model, such as character, subword, or token level. The paper focuses on two token-level word embeddings: \newcite[word2vec]{mikolov2013distributed} and \newcite[GloVe]{pennington-etal-2014-glove}. Word2vec is the foundation for all subsequent embeddings and works on the local context window, while GloVe considers the global context.
\begin{table}[!h]
\centering
\resizebox{0.45\textwidth}{!}{%
\begin{tabular}{|c|c|c|}
\hline
\textbf{Class}  & \textbf{Input type} & \textbf{Systems}    \\ \hline
Static   & character & charLM    \\ \hline 
   & subword & fastText    \\ \hline
   & token & word2vec, gloVe, LCM    \\ \hline\hline

Contextualized   & character & ELMo    \\ \hline
                & subword & ALBERT   \\ \hline 

\end{tabular}%
}
\caption{Overview of Sanskrit pretrained embeddings.}
\label{tab:introTable}
\end{table}
To address the OOV issue, subword \cite{wieting-etal-2016-charagram,bojanowski-etal-2017-enriching,heinzerling-strube-2018-bpemb} and character-level \cite{kim_charLM,jozefowicz2016exploring}  modeling have been proposed. 
We also explore two contextualized embeddings: ELMo \cite{elmo} and ALBERT \cite{ALBERT:}, a lighter version of BERT. 
We trained these 6 embedding methods on Sanskrit corpora and made the pretrained models publicly available \cite{sandhan-etal-2023-evaluating}.\footnote{\url{https://github.com/Jivnesh/SanskritShala/tree/master/EvalSan}}  The following section describes our proposed pretraining for low-resource settings.

\noindent\textbf{LCM Pretraining:}
 We propose a supervised pretraining, which automatically leverages morphological information using the pretrained encoders.
In a nutshell, LCM integrates word representations from multiple encoders trained on three independent auxiliary tasks into the encoder of the neural dependency parser. LCM follows a pipeline-based approach consisting of two steps: pretraining and integration. Pretraining uses a sequence labelling paradigm and trains encoders for three independent auxiliary tasks. Later, these pretrained encoders are combined with the encoder of the neural parser via a gating mechanism similar to ~\newcite{sato-etal-2017-adversarial}. The LCM consists of three sequence labelling-based auxiliary tasks, namely, predicting the dependency label between a modifier-modified pair (\textbf{LT}), the monolithic morphological label \textbf{(MT)}, and the case attribute of each nominal \textbf{(CT)}.
 We encourage readers to refer \newcite[LCM]{sandhan-etal-2021-little} for more details.

The quality of word embedding spaces is evaluated through intrinsic and extrinsic methods. This study focuses on intrinsic evaluation, which involves assessing semantic and syntactic information in the words without testing on NLP applications. It is based on works such as \newcite{mikolov2013distributed} and \newcite{baroni-etal-2014-dont}.
These evaluations require a query inventory containing a query word and a related target word. However, such query inventories are not readily available for Sanskrit. To address this, we annotated query inventories for 4 intrinsic tasks: analogy prediction, synonym detection, relatedness, and concept categorization. The inventories were constructed using resources such as Sanskrit WordNet \cite{Kulkarni2017}, Amarako\d{s}a  \cite{inproceedings_nair}, and Sanskrit Heritage Reader \cite{goyaldesign16,heut_13}. 

 \noindent\textbf{Intrinsic evaluation testbed for Sanskrit word embeddings:}
The quality of word embedding spaces is evaluated through intrinsic and extrinsic methods. This study focuses on intrinsic evaluation, which involves assessing semantic and syntactic information in the words without testing on NLP applications. It is based on works such as \newcite{mikolov2013distributed} and \newcite{baroni-etal-2014-dont}.
These evaluations require a query inventory containing a query word and a related target word. However, such query inventories are not readily available for Sanskrit. To address this, we annotated query inventories for 4 intrinsic tasks: analogy prediction, synonym detection, relatedness, and concept categorization. The inventories were constructed using resources such as Sanskrit WordNet \cite{Kulkarni2017}, Amarako\d{s}a  \cite{inproceedings_nair}, and Sanskrit Heritage Reader \cite{goyaldesign16,heut_13}. 


\noindent\textbf{Prabhupadavani Speech Translation Dataset:}
\label{intro}
Speech Translation (ST) is a task in which speech is simultaneously translated from source language to a different target language.\footnote{We refer to speech translation as a speech-to-text task.} It aids to overcome the language barriers across different communities for various applications such as social media, education, tourism, medical etc. 


Nowadays, most users prefer to communicate using a mixture of two or many languages on platforms such as social media, online blogs, chatbots, etc. Thus, code-mixing has become ubiquitous in all kinds of Natural Language Processing (NLP) resources/tasks \cite{khanuja2020new,chakravarthi2020corpus,singh2018twitter,singh2018named,dhar2018enabling}.  However, the existing NLP tools may not be robust enough to address this phenomenon of code-mixing for various downstream NLP applications \cite{srivastava2021challenges}. Therefore, there has been a surge in creating code-mixed datasets: (1) to understand reasons for the failure of existing models, and (2) to empower existing models for overcoming this phenomenon. Nevertheless, it is challenging to find natural resources that essentially capture different aspects of code-mixing for creating datasets for a wide range of NLP tasks.
Although there has been considerable research in generating copious data and novel architectures for the ST task, we find that not much attention has been given to address the code-mixing phenomenon on the ST task. Possibly, this can be justified due to the lack of a code-mixed ST dataset.
To the best of our knowledge, no such sufficiently large, multi-lingual, code-mixed dataset is available for the ST. 

Thus, in this work, we introduce \textbf{Prabhupadavani}, a multi-lingual, multi-domain, speech translation dataset for 25 languages containing 94 hours of speech by 130 speakers. The Prabhupadavani is about Vedic culture and heritage from Indic literature, where code-switching in the case of quotation from literature is important in the context of humanities teaching. The multiple domains cover utterances from public lectures, conversations, debates, and interviews on various social issues.
This is the first code-mixed data for speech translation to the best of our knowledge. It is code-mixed with English, Bengali and Sanskrit. From the typological point of view, the languages covered vary over ten language families. All the audios files have been manually aligned and translated. 
 We believe that our work will ignite research in this direction to understand- (1) How to make existing systems robust for handling this phenomenon effectively? (2) Can multi-lingual training help to improve performance on code-mixed speech translation? (3) Will the gap between the cascade and end-to-end systems be closed?

\section{Summary}
We present the first neural-based Sanskrit NLP toolkit, SanskritShala which facilitates diverse linguistic analysis for tasks such as word segmentation, morphological tagging, dependency parsing and compound type identification. It is set up as a web-based application to make the toolkit easier to use for teaching and annotating. All the codebase, datasets and web-based applications are publicly available. We also release word embedding models trained on publicly available Sanskrit corpora and various annotated datasets for 4 intrinsic evaluation tasks to access the intrinsic properties of word embeddings. We strongly believe that our toolkit will benefit people who are willing to work with Sanskrit and will eventually accelerate the Sanskrit NLP research.